\documentclass[]{worldbench}
\definecolor{w_blue}{RGB}{52,204,204}
\definecolor{w_yellow}{RGB}{255,192,0}

\usepackage{amsfonts}
\usepackage{amsmath}
\usepackage{amssymb}

\usepackage{colortbl}

\usepackage{makecell}
\usepackage{multicol}
\usepackage{multirow}
\usepackage{pifont}

\usepackage{wrapfig}

\usepackage{xspace}

\usepackage{pifont}

\definecolor{vla_purple}{RGB}{138,83,192}
\definecolor{vla_green}{RGB}{78,167,46}

\definecolor{tpami_blue}{RGB}{52,204,204}
\definecolor{tpami_gray}{RGB}{165,165,165}
\definecolor{tpami_red}{RGB}{192,0,0}
\definecolor{tpami_yellow}{RGB}{248,190,50}
\definecolor{link}{RGB}{248,190,50}

\newcommand{\crbx}[2]{\scalebox{0.75}{\fcolorbox{#1}{#1}{\makebox[1.5ex][c]{\rule{0pt}{1.5ex}#2}}}}

\usepackage{titletoc}

\definecolor{crArgoverse}{RGB}{220,100,180}

\definecolor{crBDD100}{RGB}{150,170,220}
\definecolor{crBDDX}{RGB}{110,180,200}
\definecolor{crBenchDrive}{RGB}{210,170,115}
\definecolor{crDriveBench}{RGB}{130,140,90}

\definecolor{crCarla}{RGB}{230,170,120}
\definecolor{crCarlaSC}{RGB}{150,140,110}
\definecolor{crCoVLA}{RGB}{240,210,100}
\definecolor{crCamOcc}{RGB}{220,140,120}
\definecolor{crOpenDV}{RGB}{235,140,120}

\definecolor{crDriveMLM}{RGB}{210,160,170}
\definecolor{crDriveLM}{RGB}{110,110,130}
\definecolor{crDriveCoT}{RGB}{180,140,120}
\definecolor{crDriveAction}{RGB}{140,150,110}

\definecolor{crHBD}{RGB}{200,220,80}

\definecolor{crImpromptuVLA}{RGB}{245,170,190}

\definecolor{crnuScenes}{RGB}{100,170,150}
\definecolor{crnuPlan}{RGB}{170,200,120}
\definecolor{crNoc}{RGB}{180,170,140}
\definecolor{crNAVSIM}{RGB}{140,160,180}
\definecolor{crNuInstruct}{RGB}{220,120,140}
\definecolor{crNuInteract}{RGB}{190,90,130}

\definecolor{crOpenOcc}{RGB}{170,140,190}
\definecolor{crOccThreeD}{RGB}{220,130,110}
\definecolor{crOmniDrive}{RGB}{130,180,130}
\definecolor{crOmniReasonN}{RGB}{100,150,110}
\definecolor{crOmniReasonB}{RGB}{160,190,220}

\definecolor{crPrivate}{RGB}{100,100,100}
\definecolor{crProcGen}{RGB}{90,130,130}
\definecolor{crPhysicalAI}{RGB}{100,120,32}

\definecolor{crRoboBEV}{RGB}{200,110,80}
\definecolor{crReasonDrive}{RGB}{100,150,110}

\definecolor{crSDN}{RGB}{90,140,180}
\definecolor{crSUPAD}{RGB}{190,140,210}

\definecolor{crTalkCar}{RGB}{140,90,170}

\definecolor{crVLAAD}{RGB}{240,180,80}

\definecolor{crWaymo}{RGB}{135,180,225}
\definecolor{crWOMDR}{RGB}{220,100,100}
\definecolor{crWODEE}{RGB}{100,220,150}

\definecolor{crLMDrive}{RGB}{160,180,100}
\definecolor{crLyft}{RGB}{110,160,120}

\definecolor{crMetaAD}{RGB}{190,120,120}

\newcommand{\Carla}{\crbx{crCarla}{\textcolor{white}{\textbf{\textsf{C}}}}}
\newcommand{\RoboBEV}{\crbx{crRoboBEV}{\textcolor{white}{\textbf{\textsf{R}}}}}
\newcommand{\nuScenes}{\crbx{crnuScenes}{\textcolor{white}{\textbf{\textsf{N}}}}}
\newcommand{\Waymo}{\crbx{crWaymo}{\textcolor{white}{\textbf{\textsf{W}}}}}
\newcommand{\BDDX}{\crbx{crBDDX}{\textcolor{white}{\textbf{\textsf{B}}}}}
\newcommand{\DriveMLM}{\crbx{crDriveMLM}{\textcolor{white}{\textbf{\textsf{D}}}}}
\newcommand{\nuPlan}{\crbx{crnuPlan}{\textcolor{white}{\textbf{\textsf{N}}}}}
\newcommand{\DriveLM}{\crbx{crDriveLM}{\textcolor{white}{\textbf{\textsf{D}}}}}
\newcommand{\BDD}{\crbx{crBDD100}{\textcolor{white}{\textbf{\textsf{B}}}}}
\newcommand{\TalkCar}{\crbx{crTalkCar}{\textcolor{white}{\textbf{\textsf{T}}}}}
\newcommand{\SDN}{\crbx{crSDN}{\textcolor{white}{\textbf{\textsf{S}}}}}
\newcommand{\Argoverse}{\crbx{crArgoverse}{\textcolor{white}{\textbf{\textsf{A}}}}}
\newcommand{\LMDrive}{\crbx{crLMDrive}{\textcolor{white}{\textbf{\textsf{L}}}}}
\newcommand{\OpenDV}{\crbx{crOpenDV}{\textcolor{white}{\textbf{\textsf{O}}}}}
\newcommand{\NAVSIM}{\crbx{crNAVSIM}{\textcolor{white}{\textbf{\textsf{N}}}}}

\newcommand{\OpenOcc}{\crbx{crOpenOcc}{\textcolor{white}{\textbf{\textsf{O}}}}}
\newcommand{\OmniDrive}{\crbx{crOmniDrive}{\textcolor{white}{\textbf{\textsf{O}}}}}
\newcommand{\BenchDrive}{\crbx{crBenchDrive}{\textcolor{white}{\textbf{\textsf{B}}}}}
\newcommand{\Private}{\crbx{crPrivate}{\textcolor{white}{\textbf{\textsf{P}}}}}
\newcommand{\Noc}{\crbx{crNoc}{\textcolor{white}{\textbf{\textsf{N}}}}}
\newcommand{\ProcGen}{\crbx{crProcGen}{\textcolor{white}{\textbf{\textsf{P}}}}}
\newcommand{\Lyft}{\crbx{crLyft}{\textcolor{white}{\textbf{\textsf{L}}}}}
\newcommand{\OccThreeD}{\crbx{crOccThreeD}{\textcolor{white}{\textbf{\textsf{O}}}}}
\newcommand{\HBD}{\crbx{crHBD}{\textcolor{white}{\textbf{\textsf{H}}}}}
\newcommand{\VLAAD}{\crbx{crVLAAD}{\textcolor{white}{\textbf{\textsf{V}}}}}
\newcommand{\SUPAD}{\crbx{crSUPAD}{\textcolor{white}{\textbf{\textsf{S}}}}}
\newcommand{\NuInstruct}{\crbx{crNuInstruct}{\textcolor{white}{\textbf{\textsf{N}}}}}
\newcommand{\WOMDR}{\crbx{crWOMDR}{\textcolor{white}{\textbf{\textsf{W}}}}}
\newcommand{\DriveCoT}{\crbx{crDriveCoT}{\textcolor{white}{\textbf{\textsf{D}}}}}
\newcommand{\ReasonDrive}{\crbx{crReasonDrive}{\textcolor{white}{\textbf{\textsf{R}}}}}
\newcommand{\DriveBench}{\crbx{crDriveBench}{\textcolor{white}{\textbf{\textsf{D}}}}}
\newcommand{\MetaAD}{\crbx{crMetaAD}{\textcolor{white}{\textbf{\textsf{M}}}}}
\newcommand{\CoVLA}{\crbx{crCoVLA}{\textcolor{white}{\textbf{\textsf{C}}}}}
\newcommand{\NuInteract}{\crbx{crNuInteract}{\textcolor{white}{\textbf{\textsf{N}}}}}
\newcommand{\DriveAction}{\crbx{crDriveAction}{\textcolor{white}{\textbf{\textsf{D}}}}}
\newcommand{\OmniReasonN}{\crbx{crOmniReasonN}{\textcolor{white}{\textbf{\textsf{O}}}}}
\newcommand{\ImpromptuVLA}{\crbx{crImpromptuVLA}{\textcolor{white}{\textbf{\textsf{I}}}}}
\newcommand{\OmniReasonB}{\crbx{crOmniReasonB}{\textcolor{white}{\textbf{\textsf{O}}}}}
\newcommand{\CamOcc}{\crbx{crCamOcc}{\textcolor{white}{\textbf{\textsf{C}}}}}
\newcommand{\WODEE}{\crbx{crWODEE}{\textcolor{white}{\textbf{\textsf{W}}}}}
\newcommand{\PhysicalAI}{\crbx{crPhysicalAI}{\textcolor{white}{\textbf{\textsf{P}}}}}

\usepackage{tabularx} 
\usepackage{enumitem}

\title{Vision-Language-Action Models for Autonomous Driving: Past, Present, and Future}

\author[]{Tianshuai~Hu$^{1,}$\raisebox{0.2em}{\includegraphics[width=0.019\linewidth]{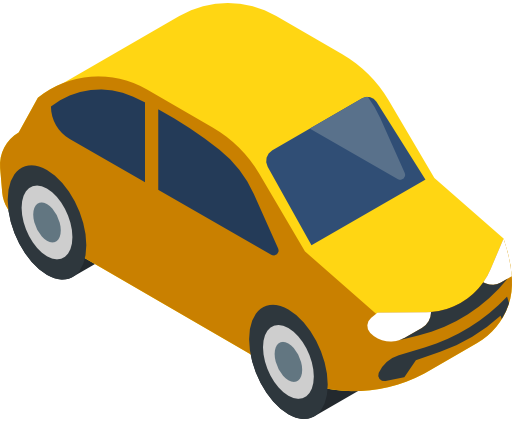}}}
\author[]{Xiaolu~Liu$^{2,}$\raisebox{0.2em}{\includegraphics[width=0.019\linewidth]{figures/icons/car1.png}}}
\author[]{Song~Wang$^{2,}$\raisebox{0.2em}{\includegraphics[width=0.019\linewidth]{figures/icons/car1.png}}}
\author[]{Yiyao~Zhu$^{1,}$\raisebox{0.2em}{\includegraphics[width=0.019\linewidth]{figures/icons/car1.png}}}
\author[]{Ao~Liang$^{3,}$\raisebox{0.2em}{\includegraphics[width=0.019\linewidth]{figures/icons/car1.png}}}
\author[]{Lingdong~Kong$^{3,}$\raisebox{0.2em}{\includegraphics[width=0.019\linewidth]{figures/icons/car1.png}}~\raisebox{0.2em}{\includegraphics[width=0.019\linewidth]{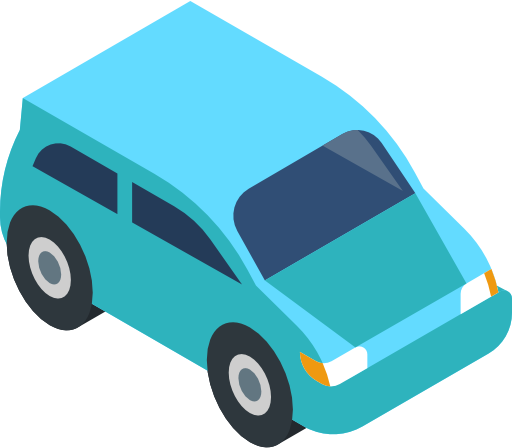}}}
\author[]{Guoyang~Zhao$^4$}
\author[]{Zeying~Gong$^4$}
\author[]{Jun~Cen$^5$}
\author[]{Zhiyu~Huang$^6$}
\author[]{Xiaoshuai~Hao$^7$}
\author[]{Linfeng~Li$^3$}
\author[]{Hang~Song$^8$}
\author[]{Xiangtai~Li$^9$}
\author[]{Jun~Ma$^{1,4}$}
\author[]{Shaojie~Shen$^1$}
\author[]{Jianke~Zhu$^2$}
\author[]{Dacheng~Tao$^9$}
\author[]{Ziwei~Liu$^{9,}$\raisebox{0.15em}{\includegraphics[width=0.017\linewidth]{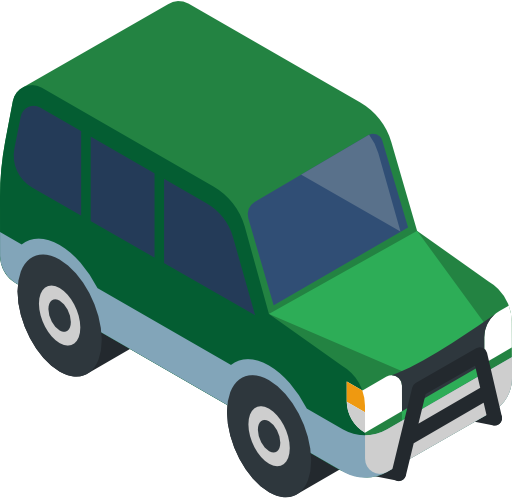}}}
\author[]{Junwei~Liang$^{1,4,}$\raisebox{0.15em}{\includegraphics[width=0.017\linewidth]{figures/icons/car4.png}}}

\affiliation[]{$^1$HKUST\quad
$^2$Zhejiang University\quad
$^3$National University of Singapore\quad
$^4$HKUST(GZ)\quad
$^5$DAMO~Academy,~Alibaba\quad
$^6$University of California, Los Angeles\quad
$^7$Xiaomi EV\quad
$^8$Xi'an~Jiaotong~University\quad
$^9$Nanyang Technological University, Singapore
\\[1.8ex]
\raisebox{-0.1em}{\includegraphics[width=0.029\linewidth]{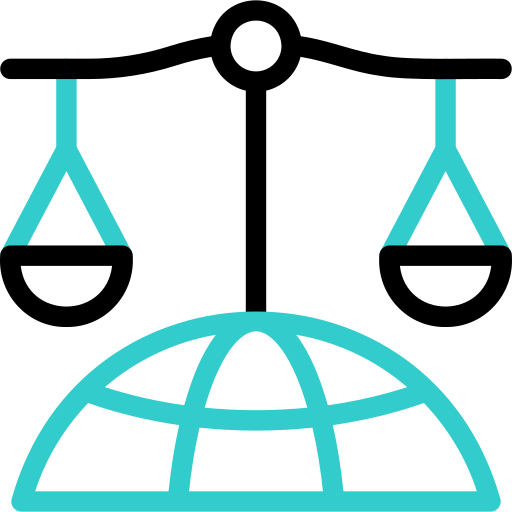}}~WorldBench Team
\\[2.2ex]
\raisebox{-0.2em}{\includegraphics[width=0.032\linewidth]{figures/icons/car1.png}}~{\small \textbf{Equal Contributions}}
\quad
\raisebox{-0.2em}{\includegraphics[width=0.031\linewidth]{figures/icons/car2.png}}~{\small \textbf{Project Lead}}
\quad
\raisebox{-0.2em}{\includegraphics[width=0.028\linewidth]{figures/icons/car4.png}}~{\small \textbf{Corresponding Authors}}
}

\abstract{
Autonomous driving has long relied on modular ``Perception-Decision-Action'' pipelines, where hand-crafted interfaces and rule-based components often break down in complex or long-tailed scenarios. Their cascaded design further propagates perception errors, degrading downstream planning and control. \textbf{Vision-Action (VA)} models address some limitations by learning direct mappings from visual inputs to actions, but they remain opaque, sensitive to distribution shifts, and lack structured reasoning or instruction-following capabilities. Recent progress in Large Language Models (LLMs) and multimodal learning has motivated the emergence of \textbf{Vision-Language-Action (VLA)} frameworks, which integrate perception with language-grounded decision making. By unifying visual understanding, linguistic reasoning, and actionable outputs, VLAs offer a more interpretable, generalizable, and human-aligned paradigm for driving policies. This work provides a structured characterization of the emerging VLA landscape for autonomous driving. We trace the evolution from early VA approaches to modern VLA frameworks and organize existing methods into two principal paradigms: \emph{End-to-End VLA}, which integrates perception, reasoning, and planning within a single model, and \emph{Dual-System VLA}, which separates slow deliberation (via VLMs) from fast, safety-critical execution (via planners). Within these paradigms, we further distinguish subclasses such as textual \emph{vs.} numerical action generators and explicit \emph{vs.} implicit guidance mechanisms. We also summarize representative datasets and benchmarks for evaluating VLA-based driving systems and highlight key challenges and open directions, including robustness, interpretability, and instruction fidelity. Overall, this work aims to establish a coherent foundation for advancing human-compatible autonomous driving systems.
}

\metadata[
\raisebox{-0.2em}{\includegraphics[width=0.025\linewidth]{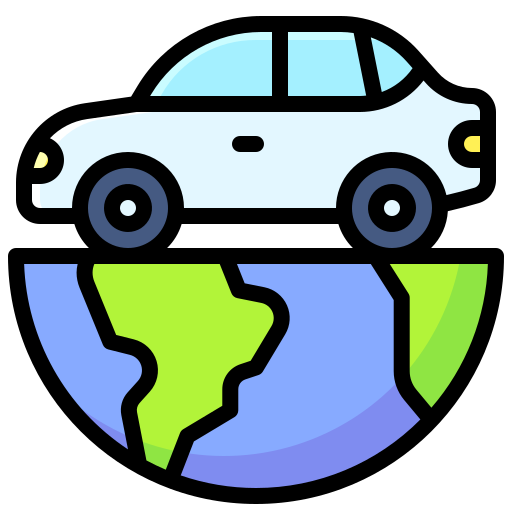}}~~Project Page]{\href{https://worldbench.github.io/vla4ad}{\texttt{https://worldbench.github.io/vla4ad}}
\\[-1.5ex]}

\metadata[
\raisebox{-0.2em}{\includegraphics[width=0.025\linewidth]{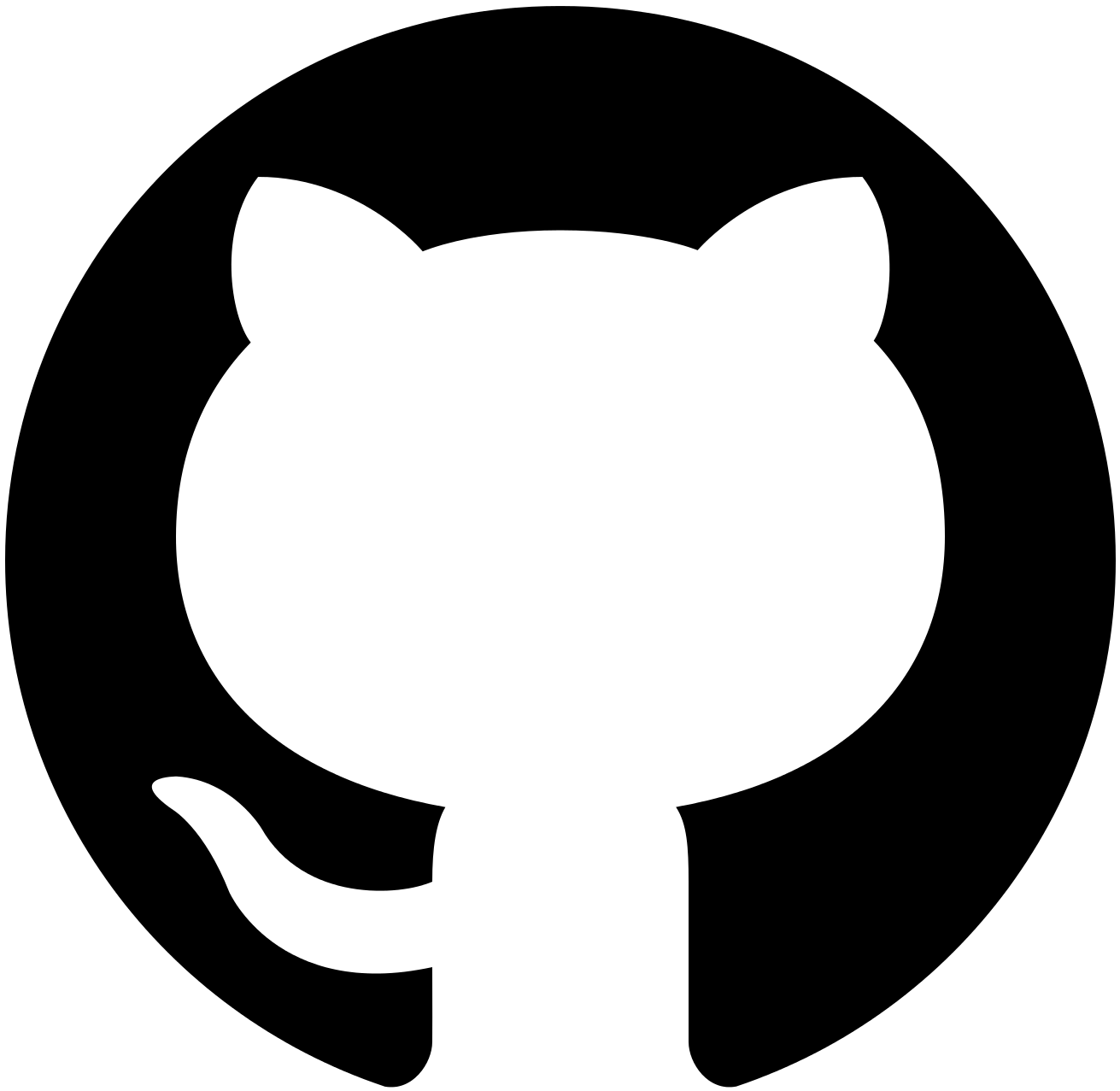}}~~GitHub Repo]{\href{https://github.com/worldbench/awesome-vla-for-ad}{\texttt{https://github.com/worldbench/awesome-vla-for-ad}}
\\[-1.5ex]}

\metadata[
\raisebox{-0.3em}{\includegraphics[width=0.026\linewidth]{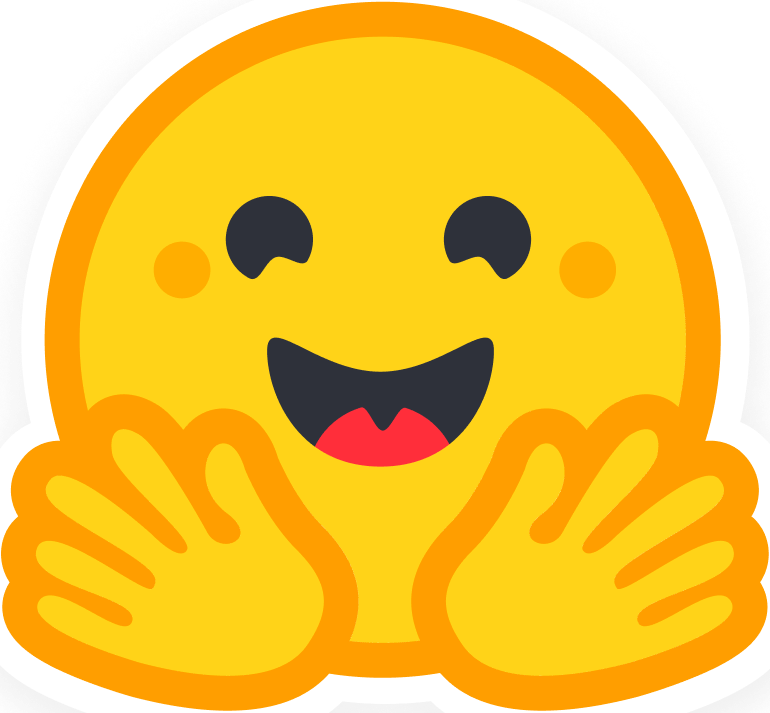}}~~HuggingFace Leaderboard]{\href{https://huggingface.co/spaces/worldbench/vla4ad}{\texttt{https://huggingface.co/spaces/worldbench/vla4ad}}}

\begin{document}

\maketitle

\begin{figure}[t]
    \centering
    \includegraphics[width=\linewidth]{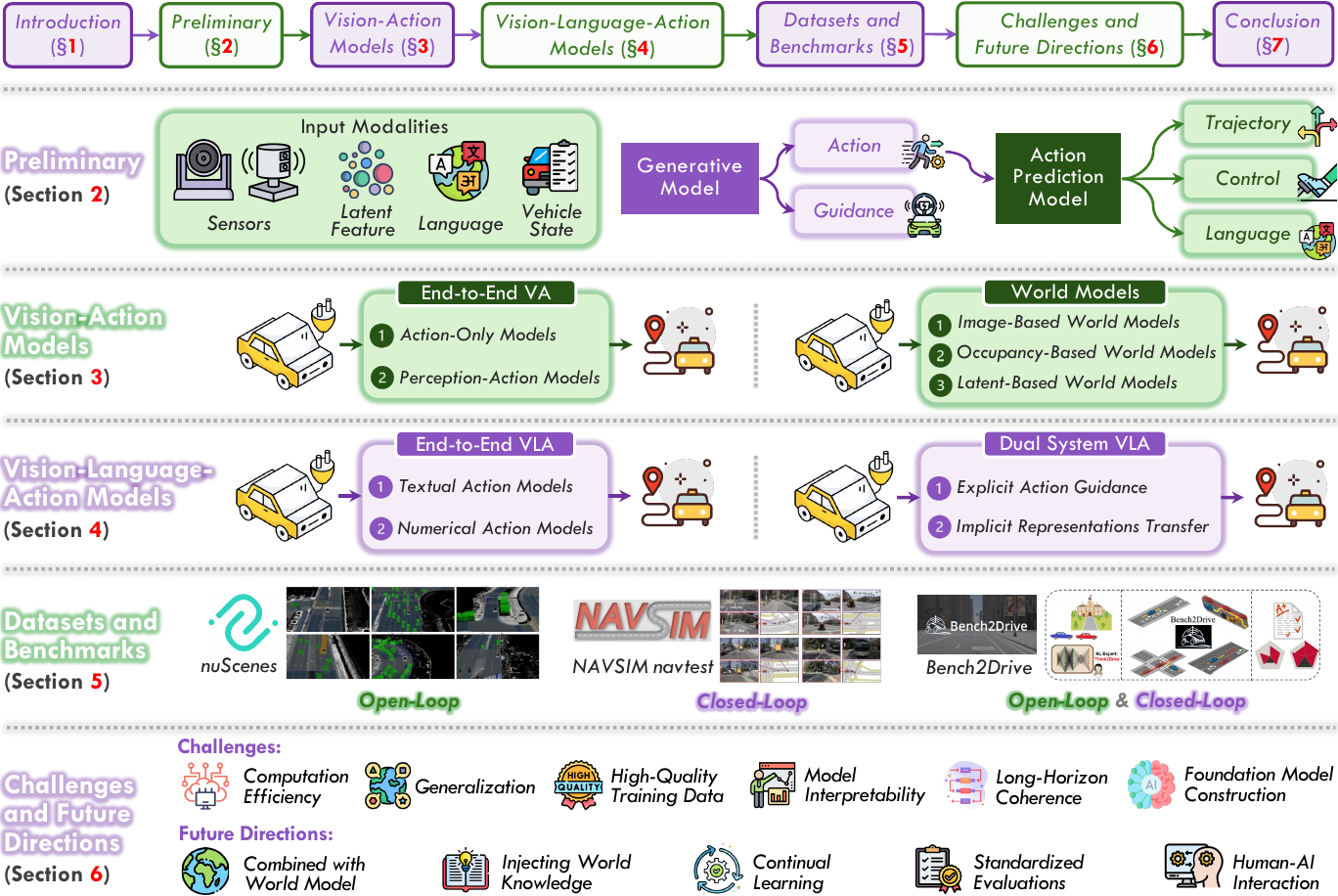}
    \vspace{-0.5cm}
    \caption{\textbf{Outline.} This work aims to provide a structured roadmap of the VLA paradigm for autonomous driving. We begin with \textbf{Preliminary Foundations} (Section~\ref{sec:preliminary}), which formalize the general formulation of VLA models and detail their three core components: the multi-modal input modalities, the VLM backbone, and the action prediction head. It then traces the evolution from \textbf{VA Models} (Section~\ref{sec:va}), which directly map perception to control, towards \textbf{VLA Models} (Section~\ref{sec:vla}), which incorporate language-grounded reasoning and interpretable decision-making. We further categorize VLA architectures into two major paradigms -- \textbf{End-to-End VLA} (Section~\ref{sec:e2evla}) and \textbf{Dual-System VLA} (Section~\ref{sec:dsvla}) -- that differ in their integration of vision, language, and action modules. Next, we review \textbf{Datasets \& Benchmarks} (Section~\ref{sec:datasets_benchmark_study}) that enable both open-loop and closed-loop evaluation of driving intelligence. Finally, we discuss \textbf{Challenges \& Future Directions} (Section~\ref{sec:challenges}), highlighting interpretability, reasoning, and human-AI interaction as central themes driving the next generation of VLA-based autonomous driving research.}
    \label{fig:structure}
\end{figure}

\section{Introduction}
\label{sec:intro}

The pursuit of fully autonomous driving (AD) has long been a central goal in AI and robotics~\cite{geiger2012we, chib2023recent,robodrive_challenge_2024}. Conventional AD systems typically adopt a modular ``Perception-Decision-Action'' pipeline, where mapping~\cite{hao2024your,hao2025mapfusion}, object detection~\cite{li2024bevformer,parktime, kong2023robo3d,liang2025pi3det}, motion prediction~\cite{yu2025combining, khandelwal2020if, djuric2020uncertainty}, and trajectory planning~\cite{li2025survey,yuan2024drama} are developed and optimized as separate components. While this design has achieved strong performance in structured environments, its reliance on hand-crafted interfaces and rules limits adaptability in complex~\cite{hao2025msc,hao2025safemap,shan2025stability}, dynamic \cite{xie2025benchmarking,xie2025vlms,kong2023rethinking}, and long-tailed scenarios~\cite{pan2024vlp, ghosh2024roadwork, yao2025lilodriver}. Moreover, the sequential cascade is prone to cross-stage error propagation, where perception noise is amplified by downstream reasoning and control, compromising stability and safety.

To mitigate these issues, research has increasingly moved toward end-to-end autonomous driving, where \textbf{Vision-Action (VA)} models directly map raw sensory inputs to control commands or trajectory waypoints using imitation~\cite{codevilla2018end, pan2017agile, popov2024mitigating} and reinforcement learning \cite{li2025survey, xu2025towards, gao2025rad}. 
Early systems such as ALVINN~\cite{pomerleau1988alvinn} and ChauffeurNet~\cite{bansal2018chauffeurnet} demonstrated the viability of behavior cloning at scale. 
Subsequent advances introduced more expressive architectures: TransFuser~\cite{chitta2022transfuser} exploited transformer-based multimodal fusion, UniAD~\cite{hu2023planning} unified perception and planning, VAD~\cite{jiang2023vad} leveraged vectorized scene representations, DriveTransformer~\cite{jia2025drivetransformer} explored scalable transformer backbones, and DiffusionDrive~\cite{liao2025diffusiondrive} applied generative modeling to multi-modal trajectory prediction. 
Collectively, these VA models show that complex driving policies can be learned directly from data, laying the foundation for modern end-to-end AD systems.

Despite these successes, VA models exhibit fundamental limitations. They largely behave as ``black boxes'', offering limited interpretability in safety-critical settings \cite{chen2024end,jing2022inaction,mirzaie2025interpretable,yuan2024rag,kong2025multi,kong2025largead}. Their generalization remains fragile under rare or long-tail scenarios that are underrepresented in training \cite{chen2024end,jiang2025survey,arai2025covla,yuan2024rag,gao2024vista,li2025recogdrive}. By directly mapping perception to low-level actions, they lack chain-of-thought (CoT) reasoning and contextual deliberation \cite{jiang2025survey,chen2024end,gao2025langcoop,xu2024vlm}, limiting their ability to resolve ambiguous or multi-stage interactions. Moreover, their focus on visual inputs prevents them from incorporating high-level plans or human instructions in natural language, leaving a gap in human-vehicle interaction~\cite{jiang2025survey,pan2024vlp,zhang2025safeauto,shao2024lmdrive}.

The emergence of Large Language Models (LLMs) and Large Multimodal Models (LMMs) has catalyzed a new paradigm: \textbf{Vision-Language-Action (VLA)} models \cite{adilkhanov2025survey, sapkota2025vision, zhong2025survey}. VLA models couple a Vision-Language Model (VLM) backbone with an action-prediction head, enabling direct mapping from multimodal inputs (vision + language) to executable driving actions. By jointly modeling perception, language understanding, and decision-making, VLA frameworks aspire to provide human-like reasoning, interpretability, and instruction-following~\cite{zhou2025autovla, fu2025orion, ge2025vla}. 
Initial explorations such as DriveMLM~\cite{wang2023drivemlm} and GPT-Driver~\cite{mao2023gpt} introduced language modules into driving pipelines for high-level decision understanding, paving the way for more integrated designs. Later systems advanced toward closed-loop and reasoning-centric VLA models: LMDrive~\cite{shao2024lmdrive} achieved language-guided closed-loop driving, DriveLM~\cite{sima2024drivelm} enabled structured reasoning via visual question answering, and DriveGPT4~\cite{xu2024drivegpt4} provided natural-language rationales for decisions. Recent works further investigate tightly coupled reasoning and control, including AutoVLA~\cite{zhou2025autovla} with fast/slow thinking and GRPO-based optimization~\cite{shao2024deepseekmath}, and SimLingo~\cite{renz2025simlingo}, which explicitly studies language-action alignment.

End-to-End VLA models, however, must simultaneously \emph{reason} and \emph{act} in real time, creating challenges for latency and safety. This has led to \emph{Dual-System} VLA designs, where high-level decision making is separated from low-level trajectory execution. 
DriveVLM~\cite{tian2024drivevlm} generates textual rationales or decisions with a VLM while relying on a classical planner for trajectories. 
VLP~\cite{pan2024vlp} tokenizes waypoints and value maps to produce planning-aware latent actions, and Diff-VLA~\cite{jiang2025diffvla} synthesizes language-guided trajectories refined by rule- or optimization-based controllers. 
InsightDrive~\cite{song2025insightdrive} integrates causal language reasoning with MPC, assigning \emph{why} to the VLM and \emph{how} to the planner.

Together, these developments signal a paradigm shift from perception-driven pipelines toward systems that jointly reason, understand, and act. Given the rapid evolution of this field, there is a need to consolidate its conceptual foundations, clarify architectural trends, and provide a structured analysis of emerging directions.

\noindent\textbf{Contributions.} 
This work provides a comprehensive characterization of VLA models for autonomous driving. Specifically:
\begin{itemize}
    \item We chart the evolution from precursor VA models (Section~\ref{sec:va}) to modern VLA frameworks (Section~\ref{sec:vla}), providing historical context and clarifying the motivations behind this paradigm shift. 
    
    \item We propose a taxonomy that categorizes VLA architectures into End-to-End (Section~\ref{sec:e2evla}) and Dual-System (Section~\ref{sec:dsvla}) designs, and compare their principles, advantages, and limitations.
    
    \item We present an organized synthesis of datasets and evaluation benchmarks relevant to VLA-based driving (Section~\ref{sec:datasets_benchmark_study}), facilitating consistent and meaningful comparisons.
    
    \item We identify key challenges in real-world VLA deployment and outline future research directions (Section~\ref{sec:challenges}) to guide progress toward safer and more reliable autonomous systems.
\end{itemize}

\begin{figure}[t]
    \centering
    \includegraphics[width=\linewidth]{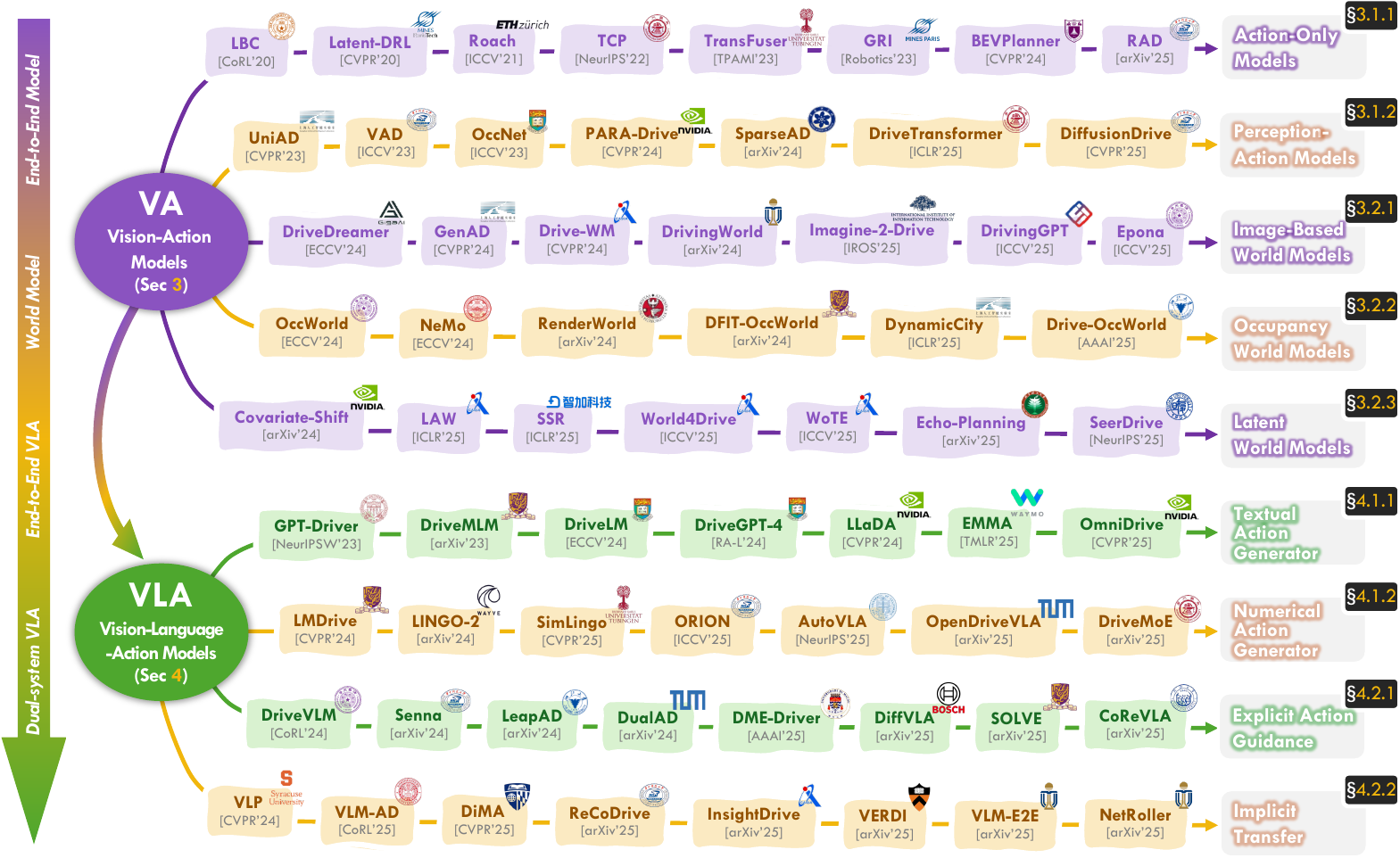}
    \vspace{-0.5cm}
    \caption{Summary of representative \textbf{VA} and \textbf{VLA} models from existing literature, spanning \emph{End-to-End Models}, \emph{World Models}, \emph{Dual-Systems}, etc. For the complete list of related approaches and the discussions on their specifications, configurations, and technical details, kindly refer to Section~\ref{sec:va} and Section~\ref{sec:vla}, respectively.}
    \label{fig:timeline}
\end{figure}

\noindent\textbf{Scope.} 
This work differs from prior studies on VLA models~\cite{adilkhanov2025survey, sapkota2025vision, zhong2025survey, jiang2025survey} through its domain-specific focus, historical framing, and architectural depth. 
$^1$\textbf{Domain-specific focus.} Unlike broader analyses that span robotics or embodied AI~\cite{ma2024survey, shao2025large}, our work \emph{focuses exclusively on autonomous driving}, allowing a fine-grained analysis of driving-specific challenges, dataset characteristics, and safety requirements. 
$^2$\textbf{Historical and conceptual continuity.} We adopt a ``Past-Present-Future'' narrative that traces the path from early VA models to modern VLA frameworks, emphasizing the motivations and technical lineage underlying the integration of language grounding into perception and control. 
$^3$\textbf{Fine-grained architectural taxonomy.} Unlike prior high-level overviews~\cite{zhou2024vision, yang2024llm4drive, cui2024survey, jiang2025survey}, we propose a hierarchical taxonomy that differentiates End-to-End and Dual-System VLA models and analyzes how they organize perception, reasoning, and control.

Through this combination of domain specificity, historical depth, and structured architectural analysis, we aim to provide a comprehensive and insightful reference for VLA research in autonomous driving.

\noindent\textbf{Organization.} 
The remainder of this paper is organized as follows. Section~\ref{sec:preliminary} introduces the preliminary foundations of VLA frameworks. Section~\ref{sec:va} outlines the evolution of VA models. Section~\ref{sec:vla} presents our taxonomy and analysis of VLA architectures. Section~\ref{sec:datasets_benchmark_study} summarizes datasets and benchmarks. Section~\ref{sec:challenges} discusses remaining challenges and future directions. Section~\ref{sec:conclusion} concludes this work.

\section{Preliminary Foundations}
\label{sec:preliminary}

Vision-Language-Action~(VLA) frameworks~\cite{ma2024survey, zhou2025autovla, arai2025covla, zhou2025opendrivevla} leverage large Vision-Language Models~(VLMs)~\cite{Qwen2-VL, li2024llava, chen2024internvl, chu2024mobilevlm} to interpret complex driving scenes and produce executable actions. A typical formulation can be expressed as:
\begin{equation}
\mathbf{a}_t = H(F(\mathbf{x}|\theta))~,
\end{equation}
where $\mathbf{x}$ denotes multimodal inputs at timestamp $t$, $F(\cdot)$ is a VLM backbone parameterized by $\theta$, and $H(\cdot)$ is an action-generation head. This section introduces these three components: the input modalities ($\mathbf{x}$), the VLM backbone ($F$), and the action prediction head ($H$).

\subsection{Input Modalities}
The input $\mathbf{x}$ aggregates heterogeneous signals that describe the external environment and the ego-vehicle state~\cite{caesar2020nuscenes, sun2020waymo, wilson2023argoverse}. These inputs can be grouped into four categories: sensor observations, latent scene representations, language instructions, and proprioceptive states.

\subsubsection{Sensor Inputs}
Sensor inputs include raw or preprocessed data directly obtained from vehicle-mounted sensors \cite{chen2020learning,chen2022learning,li2024place3d}.
\begin{itemize}
    \item \noindent \textbf{Visual Images.}  
    Surround-view RGB images that offer dense semantic information: $\mathbf{x}_{\mathrm{img}} \in \mathbb{R}^{N_c \times H \times W \times 3}$, where $N_c$ is the number of cameras (\emph{e.g.}, $6$ to $8$), and $H$, $W$ are the height and width of each image.

    \item \noindent \textbf{LiDAR Point Clouds.}  
    A sparse or dense set of 3D points representing the environment geometry: $\mathbf{x}_{\mathrm{lidar}} \in \mathbb{R}^{N_p \times D}, \quad D \geq 4$, where $N_p$ is the number of points, and $D$ includes dimensions such as $x, y, z$, velocity, and intensity.
\end{itemize}

\subsubsection{Latent Representations}
Multiple VLA systems operate on intermediate spatial representations that fuse multimodal sensor inputs.
\begin{itemize}
    \item \noindent \textbf{Bird's-Eye View (BEV) Features.}  
    Top-down view representation, often generated by fusing camera or LiDAR data \cite{li2024bevformer, liang2022bevfusion, liu2025dsdrive}: $\mathbf{x}_{\mathrm{bev}} \in \mathbb{R}^{C \times H_{\mathrm{bev}} \times W_{\mathrm{bev}}}$, where $C$ is the number of feature channels, and $H_{\mathrm{bev}}, W_{\mathrm{bev}}$ are the spatial dimensions of the BEV grid.

    \item \noindent \textbf{Occupancy Grids.}  
    3D volumetric representation predicting occupancy and semantics for each spatial location \cite{wang2024not, wei2024occllama, xu2025occ, liu2025occvla}: $\mathbf{x}_{\mathrm{occ}} \in \mathbb{R}^{C_{\mathrm{occ}} \times X \times Y \times Z}$, where $X, Y, Z$ are the spatial resolution of the 3D grid, and $C_{\mathrm{occ}}$ denotes the number of occupancy feature channels (\emph{e.g.}, occupancy, flow and semantics).
\end{itemize}

\subsubsection{Language Inputs}
To enable VLA capabilities, the model also receives high-level textual instructions or task descriptions~\cite{fu2024drive, xu2024drivegpt4, guo2025vdrive}. It is composed of a sequence of tokens representing the driving task or goal (\emph{e.g.}, \emph{``turn left at the next intersection''}): $\mathbf{x}_{\mathrm{lang}} \in \mathbb{Z}^{T}$ or $\mathbf{x}_{\mathrm{lang}} \in \mathbb{R}^{T \times D_{\mathrm{emb}}}$, where $T$ is the sequence length, and $D_{\mathrm{emb}}$ is the embedding dimension (if token embeddings are used).

\subsubsection{Vehicle State Information}
There is also proprioceptive information describing the current dynamic state of the ego-vehicle \cite{li2024ego, caesar2020nuscenes}: $\mathbf{x}_{\mathrm{state}} \in \mathbb{R}^{D_{\mathrm{state}}}$, where $D_{\mathrm{state}}$ is the dimension of the state vector, including speed, acceleration, steering angle, yaw rate, turn indicator status, etc.

\subsection{VLM Backbone ($F$)}
The VLM backbone $F(\cdot)$ is the core reasoning engine of the system. It is typically a large vision language model \cite{liu2023visual, Qwen2-VL, Qwen2.5-VL, chen2024internvl, zhu2025internvl3, genie3}. Its primary role is to fuse the diverse input modalities into a single, powerful latent representation. It consists of a vision encoder (\emph{e.g.}, a Vision Transformer, ViT) \cite{radford2021learning} to process visual inputs and an LLM decoder that conditions its generation on the visual features. 
A bridge network~\cite{liu2023visual,chen2024internvl} or unified multimodal token modelling mechanism~\cite{Qwen2-VL,Qwen2.5-VL} is used to align the vision features with the language embeddings. VLM can directly generate the actions or provide the guidance for another action expert to develop more robust results.
 
\subsubsection{VLM for Direct Action Generation (Single-System)}
In this paradigm, the VLM directly emits actions through its language head~\cite{wang2023drivemlm,mao2023gpt,sima2024drivelm,hwang2024emma} or a small attached head~\cite{xu2025drivegpt4,shao2024lmdrive,renz2025simlingo}. This fully end-to-end design exploits the VLM's reasoning capabilities to map from visual/language inputs to executable controls.

\subsubsection{VLM for Guidance Generation (Dual-System)}
Alternatively, the VLM functions as a high-level reasoning module that produces intermediate guidance—textual rationales~\cite{tian2024drivevlm,han2025dme,qian2024fasionad} or structured latent intents~\cite{pan2024vlp,li2025recogdrive}, which a downstream planner converts into low-level actions~\cite{wang2024dualad}. This ``slow thinking + fast execution'' architecture improves interpretability and enables planners to enforce physical feasibility and safety constraints.

\subsection{Action Prediction Head ($H$)}
The head $H(\cdot)$ converts the VLM latent representation into action outputs. Consistent with the taxonomy used in existing literature, we categorize action heads into four types based on their output formulation and generation mechanism: Language Head (LH), Regression (REG), Trajectory Selection (SEL), and Trajectory Generation (GEN).

\begin{itemize}
    \item \noindent \textbf{Language Head (LH).} 
    This design directly utilizes the VLM's inherent text-generation capabilities to produce actions in the language space. The head is typically the language modeling head of the VLM, trained to output either free-form textual commands (\emph{e.g.}, \emph{``turn left''})~\cite{xu2024drivegpt4} or a sequence of discretized action tokens~\cite{zhou2025autovla}. The model autoregressively predicts these tokens, which are subsequently parsed into executable signals.  This approach is widely adopted in textual action generators like DriveMLM~\cite{wang2023drivemlm} and DriveGPT4~\cite{xu2024drivegpt4}.

    \item \noindent \textbf{Regression (REG).} 
     This formulation employs a decoder structure followed by a regressor (typically a Multi-Layer Perceptron) to directly predict continuous values. Unlike language heads, it avoids discretization by mapping the latent features aggregated via Transformers or GRUs to specific numerical outputs such as steering angles, throttle/brake values, or trajectory waypoints. Representative methods using this deterministic approach include LMDrive~\cite{shao2024lmdrive} and DriveGPT4-V2 \cite{xu2025drivegpt4}.

    \item \noindent \textbf{Trajectory Selection (SEL).} 
     Instead of directly regressing a single path, this head evaluates a set of candidate trajectories and selects the optimal one based on a learned cost function or scoring mechanism. The model typically generates or samples a diverse set of dynamically feasible trajectories and uses the latent representation to predict the cost or probability for each candidate.  This approach, utilized by methods like WoTE~\cite{li2025end}  and SeerDrive~\cite{zhang2025future}, ensures that the final output adheres to kinematic constraints by selecting from pre-defined candidates.

    \item \noindent \textbf{Trajectory Generation (GEN).} 
     This generative formulation synthesizes actions through probabilistic modeling, most notably using diffusion models or variational autoencoder~\cite{liao2025diffusiondrive, zheng2024genad, jiang2025diffvla}. Starting from noise, the head iteratively refines the trajectory sample conditioned on the VLM latent state and optionally language instructions. This allows the model to capture the multi-modality and uncertainty of future distributions. Prominent examples include ORION~\cite{fu2025orion} and DiffVLA \cite{jiang2025diffvla}.
\end{itemize}

\subsection{Action in Driving}
In the context of autonomous driving, particularly for models like VLAs, the action space defines the set of possible outputs the model can generate to control the vehicle. The choice of action representation is a fundamental design decision that dictates how the model's reasoning is translated into physical motion. We outline three primary paradigms for action space representation below.

\subsubsection{Discrete Trajectory Representations}
This paradigm represents the vehicle's intended future path as a finite sequence of spatial waypoints \cite{liu2025dsdrive}. Each waypoint is a spatial coordinate that the vehicle is expected to reach at a specific future time step. This representation allows for explicit geometric path planning and trajectory optimization. The action, $\mathbf{a}_t$, formulated at the current time $t$, is a set of $\Phi$ future waypoints:
\begin{align}
    \mathbf{a}_t = \{ (x_i, y_i) \}_{i=1}^{\Phi}, \quad \text{where } (x_i, y_i) \in \mathbb{R}^2~.
\end{align}
Here, $\Phi$ is the prediction horizon (the total number of future steps), and each $(x_i, y_i)$ is a coordinate in a 2D Cartesian plane representing the target position at step $i$.

\subsubsection{Continuous Trajectory Representations}
Instead of discrete points, this approach parameterizes the vehicle's motion as a continuous function over a future time horizon \cite{liu2025vlm}. The trajectory is typically defined by functions that govern the vehicle's longitudinal and lateral motion, such as speed and turning radius. The action, $a_t$, is defined by these continuous functions over a time interval $[0, T]$:
\begin{align}
    \mathbf{a}_t = (v(t), \kappa(t)), \quad \text{for } t \in [0, T]~.
\end{align}
In this formulation, $v(t)$ represents the vehicle's speed profile, and $\kappa(t)$ represents its curvature profile over the future time horizon $T$. This inherently captures the continuous nature of vehicle dynamics.

\subsubsection{Direct Control Representations}
This paradigm involves the direct output of low-level vehicle control commands that are immediately sent to the vehicle's actuators \cite{xu2025drivegpt4}. These outputs typically consist of continuous signals for steering, acceleration, and braking control. The values are often normalized and constrained to lie within the vehicle's physical operational ranges. The action vector, $a_t$, represents control signals for a specific time step $t$:
\begin{align}
    \mathbf{a}_t = (\delta_t, \tau_t, \beta_t)~,
\end{align}
where $\delta_t$ is the steering angle, $\tau_t$ is the throttle input, and $\beta_t$ is the brake input at time step $t$. Each component is bound by the vehicle's hardware limits, \emph{e.g.}, $\delta_t \in [\delta_{\min}, \delta_{\max}]$.

\subsubsection{Language Representations}
This paradigm leverages the natural language capabilities of VLMs to express driving actions through textual descriptions \cite{cao2025fastdrivevla}. The action is represented as a sequence of discrete tokens from a predefined vocabulary:
\begin{align}
    \mathbf{a}_t = \{w_1, w_2, \ldots, w_T\}, \quad \text{where } w_i \in \mathcal{V}~.
\end{align}
Here, $\mathcal{V}$ represents the model's vocabulary, $T$ is the sequence length, and each token $w_i$ corresponds to an element in the vocabulary. The language-based action can range from high-level commands (\emph{e.g.}, \emph{``turn left at the intersection''}) to specific numerical trajectory representations encoded as text tokens.
\section{Vision-Action Models}
\label{sec:va}

Vision-Action (VA) models represent one of the earliest and most influential lines of research in autonomous driving. Their core idea is to directly map sensory observations -- typically camera inputs -- to driving actions, thereby avoiding explicit modular decomposition into perception, prediction, and planning. Enabled by deep neural networks, VA models have been explored through \textbf{two major training paradigms}: $^1$\emph{imitation learning}, which distills policies from expert demonstrations, and $^2$\emph{reinforcement learning}, which optimizes behavior through trial-and-error interaction. More recently, \emph{world models} have expanded this paradigm by enabling agents to simulate scene dynamics and reason about action consequences, improving robustness and scalability. Table~\ref{tab:va_methods} provides an overview of representative efforts.

From an architectural perspective, VA methodologies for autonomous driving can be broadly grouped into:
\begin{itemize}
    \item \textbf{End-to-End Models}, which directly predict control commands or planned trajectories from sensory inputs.
    
    \item \textbf{World Models}, which explicitly model action-conditioned future dynamics to support policy learning and decision-making.
\end{itemize}

\subsection{End-to-End Models for Autonomous Driving}
\label{sec:e2ead}
End-to-end (E2E) models learn a single neural network that maps raw or intermediate sensor observations to actions or planned trajectories \cite{chen2024end,chib2023recent,li2025end,bojarski2016end}. Unlike modular stacks, which isolate perception, prediction, and planning, E2E approaches implicitly couple these tasks within a unified representation \cite{sun2025sparsedrive,jiang2024senna,hu2022st}. Depending on whether perception supervision is employed, existing methods fall into two main categories: \emph{action-only models} and \emph{perception-action models}, as illustrated in Figure~\ref{fig:e2e-model}.

\subsubsection{Action-Only Model}
\label{sec:actiononly}
Action-only models adopt a streamlined one-stage formulation: sensory inputs are fed directly into a network that outputs low-level actions. These methods primarily differ in whether policies are learned from \textbf{demonstrations} or through \textbf{exploration}.

\noindent\textbf{Imitation Learning (IL)}, especially behavior cloning~\cite{bain1995framework}, learns a policy by matching expert actions, as visualized in the IL branch of Figure~\ref{fig:e2e-model}. Early works~\cite{pomerleau1988alvinn,bojarski2016end,pan2017agile,codevilla2018end} demonstrated that actions can be predicted directly from monocular or multi-view inputs, and subsequent designs refined backbone architectures \cite{prakash2021multi,chen2022learning,chitta2022transfuser}.

\begin{wraptable}{r}{0.639\textwidth}
    \centering
    \captionsetup{justification=raggedright, singlelinecheck=false}
    \caption{Summary of \textbf{Vision-Action} models in autonomous driving.
    \\
    $\bullet$ \textbf{Inputs}: \raisebox{-0.5ex}{\includegraphics[width=0.035\linewidth]{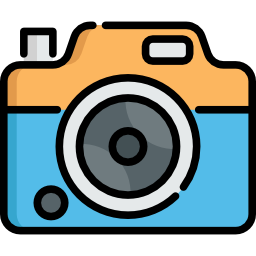}} Camera, \raisebox{-0.5ex}{\includegraphics[width=0.035\linewidth]{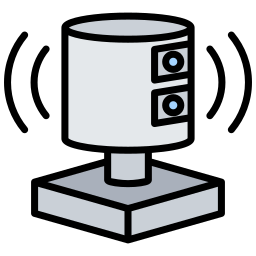}}: LiDAR, and 
    \raisebox{-0.5ex}{\includegraphics[width=0.035\linewidth]{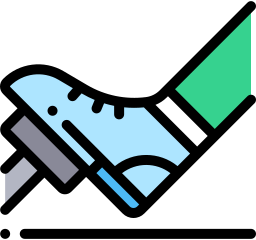}}: Ego-Status.
    \\
    $\bullet$ \textbf{Action Types}: \underline{RL}: Policy w/ Reinforcement Learning, \underline{REG}: Decoder + MLP, \underline{SEL}: Traj. Selection w/ Cost, and \underline{GEN}: Traj. Generation w/ Generative Model.
    \\
    $\bullet$ \textbf{Outputs}: \underline{Ctrl.}: Control Signal, \underline{Traj.}: Numerical Trajectory.
    \\
    $\bullet$ \textbf{Datasets:} 
    \Carla~CARLA~\cite{dosovitskiy2017carla},
    \Noc~NoCrash~\cite{codevilla2019exploring},
    \ProcGen~ProcGen~\cite{cobbe2020leveraging},
    \Lyft~Lyft~\cite{houston2021one}, 
    \nuScenes~nuScenes~\cite{caesar2020nuscenes}, 
    \BenchDrive~Bench2Drive~\cite{jia2024bench2drive},
    \NAVSIM~NAVSIM~\cite{dauner2024navsim},
    \OpenOcc~OpenOcc~\cite{tong2023scene},
    \OpenDV~OpenDV~\cite{yang2024generalized},
    \nuPlan~nuPlan~\cite{caesar2021nuplan},
    \OccThreeD~Occ3D~\cite{tian2023occ3d},
    \CamOcc~Cam4DOcc~\cite{ma2024cam4docc}, and
    \Private~Private Data.
}
\vspace{-0.2cm}
\renewcommand{\arraystretch}{1.0}
\resizebox{\linewidth}{!}{
\begin{tabular}{c|r|r|c|c|l|c|c}
    \toprule
    \textbf{\#} & \textbf{Model} & \textbf{Venue} & \textbf{Input} & \textbf{Dataset} & \textbf{Vision} & \textbf{Action} & \textbf{Output} 
    \\
    \midrule
    \rowcolor{tpami_yellow!10}\multicolumn{8}{l}{{$\bullet$~\textbf{Sec. \ref{sec:actiononly} Action-only Models}}} 
    \\
    1 & LBC \cite{chen2020learning} & {\scriptsize{CoRL'20}} & \raisebox{-0.5ex}{\includegraphics[width=0.035\linewidth]{figures/icons/rgb.png}} & \Carla~\Noc & ResNet \cite{he2016deep} & RL & Ctrl.+Traj.
    \\
    \rowcolor{gray!7}2 & Latent-DRL \cite{toromanoff2020end} & {\scriptsize{CVPR'20}} & \raisebox{-0.5ex}{\includegraphics[width=0.035\linewidth]{figures/icons/rgb.png}} & \Carla & ResNet \cite{he2016deep} & RL  & Ctrl.
    \\
    3 & NEAT \cite{chitta2021neat} & {\scriptsize{ICCV'21}} & \raisebox{-0.5ex}{\includegraphics[width=0.035\linewidth]{figures/icons/rgb.png}} & \Carla & ResNet \cite{he2016deep}  & REG  & Traj.
    \\
    \rowcolor{gray!7}4 & Roach \cite{zhang2021end} & {\scriptsize{ICCV'21}} & \raisebox{-0.5ex}{\includegraphics[width=0.035\linewidth]{figures/icons/rgb.png}} & \Carla~\Noc & ResNet \cite{he2016deep} & RL & Ctrl.
    \\
    5 & WoR \cite{chen2021learning} & {\scriptsize{ICCV'21}} & \raisebox{-0.5ex}{\includegraphics[width=0.035\linewidth]{figures/icons/rgb.png}} & \Carla~\Noc~\ProcGen & ResNet \cite{he2016deep} & REG & Ctrl.
    \\
    \rowcolor{gray!7}6 & TCP \cite{wu2022trajectoryguided} & {\scriptsize{NeurIPS'22}} & \raisebox{-0.5ex}{\includegraphics[width=0.035\linewidth]{figures/icons/rgb.png}} & \Carla & ResNet \cite{he2016deep} & REG & Ctrl.+Traj.
    \\
    7 & Urban-Driver \cite{scheel2022urban} & {\scriptsize{CoRL'22}} & \raisebox{-0.5ex}{\includegraphics[width=0.035\linewidth]{figures/icons/rgb.png}} & \Lyft & ResNet \cite{he2016deep}  & REG  & Traj.
    \\
    \rowcolor{gray!7}8 & LAV \cite{chen2022learning} & {\scriptsize{CVPR'22}} & \raisebox{-0.5ex}{\includegraphics[width=0.035\linewidth]{figures/icons/rgb.png}} \raisebox{-0.5ex}{\includegraphics[width=0.035\linewidth]{figures/icons/lidar.png}} & \Carla & ResNet \cite{he2016deep} & REG  & Ctrl.+Traj.
    \\
    9 & TransFuser \cite{chitta2022transfuser} & {\scriptsize{TPAMI'23}} & \raisebox{-0.5ex}{\includegraphics[width=0.035\linewidth]{figures/icons/rgb.png}} \raisebox{-0.5ex}{\includegraphics[width=0.035\linewidth]{figures/icons/lidar.png}} &  \Carla   & ResNet \cite{he2016deep} & REG & Traj.
    \\
    \rowcolor{gray!7}10 & GRI \cite{chekroun2023gri} & {\scriptsize{Robotics'23}} & \raisebox{-0.5ex}{\includegraphics[width=0.035\linewidth]{figures/icons/rgb.png}} & \Carla   & EfficientNet \cite{tan2019efficientnet}  & RL  & Ctrl.
    \\
    11 & BEVPlanner \cite{li2024ego} & {\scriptsize{CVPR'24}} & \raisebox{-0.5ex}{\includegraphics[width=0.035\linewidth]{figures/icons/rgb.png}} & \nuScenes & ResNet \cite{he2016deep} & REG & Traj.
    \\
    \rowcolor{gray!7}12 & Raw2Drive \cite{yang2025raw2drive} & {\scriptsize{NeurIPS'25}} & \raisebox{-0.5ex}{\includegraphics[width=0.035\linewidth]{figures/icons/rgb.png}} & \Carla~\BenchDrive   & ResNet \cite{he2016deep}  & RL  & Ctrl.
    \\
    13 & RAD \cite{gao2025rad} & {\scriptsize{NeurIPS'25}} & \raisebox{-0.5ex}{\includegraphics[width=0.035\linewidth]{figures/icons/rgb.png}} & \Private  & ResNet \cite{he2016deep} & RL  & Traj.
    \\
    \rowcolor{gray!7}14 & TrajDiff \cite{gui2025trajdiff} & {\scriptsize{arXiv'25}} & \raisebox{-0.5ex}{\includegraphics[width=0.035\linewidth]{figures/icons/rgb.png}} \raisebox{-0.5ex}{\includegraphics[width=0.035\linewidth]{figures/icons/lidar.png}} \raisebox{-0.5ex}{\includegraphics[width=0.035\linewidth]{figures/icons/status.png}} & \NAVSIM  & ResNet \cite{he2016deep}  & GEN  & Traj.
    \\
    \midrule
    \rowcolor{tpami_yellow!10}\multicolumn{8}{l}{{$\bullet$~\textbf{Sec. \ref{sec:perceptionaction} Perception-Action Models }}} 
    \\
    15 & ST-P3 \cite{hu2022st} & {\scriptsize{ECCV'22}} & \raisebox{-0.5ex}{\includegraphics[width=0.035\linewidth]{figures/icons/rgb.png}} & \nuScenes~\Carla &  EfficientNet \cite{tan2019efficientnet} & SEL  & Traj.
    \\
    \rowcolor{gray!7}16 & UniAD \cite{hu2023planning}  & {\scriptsize{CVPR'23}} & \raisebox{-0.5ex}{\includegraphics[width=0.035\linewidth]{figures/icons/rgb.png}} & \nuScenes & ResNet \cite{he2016deep} & REG  & Traj.
    \\
    17 & VAD \cite{jiang2023vad} & {\scriptsize{ICCV'23}} & \raisebox{-0.5ex}{\includegraphics[width=0.035\linewidth]{figures/icons/rgb.png}} & \nuScenes &  ResNet \cite{he2016deep} & REG  & Traj.
    \\
    \rowcolor{gray!7}18 & OccNet \cite{tong2023scene} & {\scriptsize{ICCV'23}} & \raisebox{-0.5ex}{\includegraphics[width=0.035\linewidth]{figures/icons/rgb.png}} & \nuScenes~\OpenOcc & ResNet \cite{he2016deep}  & SEL & Traj.
    \\
    19 & GenAD \cite{zheng2024genad} & {\scriptsize{ECCV'24}} & \raisebox{-0.5ex}{\includegraphics[width=0.035\linewidth]{figures/icons/rgb.png}} & \nuScenes & ResNet \cite{he2016deep} & 
    GEN & Traj.
    \\
    \rowcolor{gray!7}20 & PARA-Drive \cite{weng2024drive} & {\scriptsize{CVPR'24}} & \raisebox{-0.5ex}{\includegraphics[width=0.035\linewidth]{figures/icons/rgb.png}} & \nuScenes & ResNet \cite{he2016deep} & REG & Traj.
    \\
    21 & Hydra-MDP \cite{li2024hydra} & {\scriptsize{CVPRW'24}} & \raisebox{-0.5ex}{\includegraphics[width=0.035\linewidth]{figures/icons/rgb.png}} \raisebox{-0.5ex}{\includegraphics[width=0.035\linewidth]{figures/icons/lidar.png}} & \NAVSIM & ResNet \cite{he2016deep} & SEL & Traj.
    \\
    \rowcolor{gray!7}22 & SparseAD \cite{zhang2024sparsead} & {\scriptsize{arXiv'24}} & \raisebox{-0.5ex}{\includegraphics[width=0.035\linewidth]{figures/icons/rgb.png}} & \nuScenes & ResNet \cite{he2016deep}  & REG & Traj.
    \\
    23 & GaussianAD \cite{zheng2024gaussianad} & {\scriptsize{arXiv'24}} & \raisebox{-0.5ex}{\includegraphics[width=0.035\linewidth]{figures/icons/rgb.png}} & \nuScenes & ResNet \cite{he2016deep} & REG & Traj.
    \\
    \rowcolor{gray!7}24 &  DiFSD \cite{su2024difsd} & {\scriptsize{arXiv'24}} & \raisebox{-0.5ex}{\includegraphics[width=0.035\linewidth]{figures/icons/rgb.png}} & \nuScenes & ResNet \cite{he2016deep} & 
    GEN & Traj.
    \\
    25 &  DriveTransformer \cite{jia2025drivetransformer} & {\scriptsize{ICLR'25}} & \raisebox{-0.5ex}{\includegraphics[width=0.035\linewidth]{figures/icons/rgb.png}} & \nuScenes~\BenchDrive & ResNet \cite{he2016deep} & REG  & Traj. 
    \\
    \rowcolor{gray!7}26 & SparseDrive \cite{sun2025sparsedrive} & {\scriptsize{ICRA'25}} & \raisebox{-0.5ex}{\includegraphics[width=0.035\linewidth]{figures/icons/rgb.png}} & \nuScenes & ResNet \cite{he2016deep} & REG & Traj.
    \\
    27 & DiffusionDrive \cite{liao2025diffusiondrive} & {\scriptsize{CVPR'25}} & \raisebox{-0.5ex}{\includegraphics[width=0.035\linewidth]{figures/icons/rgb.png}} \raisebox{-0.5ex}{\includegraphics[width=0.035\linewidth]{figures/icons/lidar.png}} & \nuScenes~\NAVSIM  & ResNet \cite{he2016deep}  & 
    GEN & Traj.
    \\
    \rowcolor{gray!7}28 & GoalFlow \cite{xing2025goalflow} & {\scriptsize{CVPR'25}} & \raisebox{-0.5ex}{\includegraphics[width=0.035\linewidth]{figures/icons/rgb.png}} \raisebox{-0.5ex}{\includegraphics[width=0.035\linewidth]{figures/icons/lidar.png}} \raisebox{-0.5ex}{\includegraphics[width=0.035\linewidth]{figures/icons/status.png}} & \NAVSIM  & VoVNet \cite{lee2019energy}  & 
    GEN & Traj.
    \\
    29 & GuideFlow \cite{liu2025guideflow} & {\scriptsize{arXiv'25}} & \raisebox{-0.5ex}{\includegraphics[width=0.035\linewidth]{figures/icons/rgb.png}} & \nuScenes~\NAVSIM~\BenchDrive  & ResNet \cite{he2016deep}  & 
    GEN & Traj.
    \\
    \rowcolor{gray!7}30 & ETA \cite{hamdan2025eta} & {\scriptsize{arXiv'25}} & \raisebox{-0.5ex}{\includegraphics[width=0.035\linewidth]{figures/icons/rgb.png}}  & \BenchDrive& CLIP-ViT \cite{radford2021learning} & REG &Traj. 
    \\
    31 & Geo \cite{jia2025spatial} & {\scriptsize{arXiv'25}} & \raisebox{-0.5ex}{\includegraphics[width=0.035\linewidth]{figures/icons/rgb.png}}  & \nuScenes& ResNet \cite{he2016deep} & REG &Traj. 
    \\ 
    \rowcolor{gray!7}32 & DiffusionDriveV2 \cite{zou2025diffusiondrivev2} & {\scriptsize{arXiv'25}} & \raisebox{-0.5ex}{\includegraphics[width=0.035\linewidth]{figures/icons/rgb.png}} \raisebox{-0.5ex}{\includegraphics[width=0.035\linewidth]{figures/icons/lidar.png}}  & \NAVSIM& ResNet \cite{he2016deep} & GEN &Traj. 
    \\ 
    33 & NaviHydra \cite{wu2025navihydra} & {\scriptsize{arXiv'25}} & \raisebox{-0.5ex}{\includegraphics[width=0.035\linewidth]{figures/icons/rgb.png}} \raisebox{-0.5ex}{\includegraphics[width=0.035\linewidth]{figures/icons/lidar.png}}  & \NAVSIM& ResNet \cite{he2016deep} & SEL &Traj. 
    \\ 
    \rowcolor{gray!7}34 & Mimir \cite{xing2025mimir} & {\scriptsize{arXiv'25}} & \raisebox{-0.5ex}{\includegraphics[width=0.035\linewidth]{figures/icons/rgb.png}} \raisebox{-0.5ex}{\includegraphics[width=0.035\linewidth]{figures/icons/lidar.png}}  & \NAVSIM& ResNet \cite{he2016deep} & GEN &Traj. 
    \\ 
    \midrule
    \rowcolor{tpami_yellow!10}\multicolumn{8}{l}{{$\bullet$~\textbf{Sec. \ref{sec:imageworldmodel} Image-Based World Models }}} 
    \\
    35 & DriveDreamer \cite{wang2024drivedreamer} & {\scriptsize{ECCV'24}} & \raisebox{-0.5ex}{\includegraphics[width=0.035\linewidth]{figures/icons/rgb.png}} \raisebox{-0.5ex}{\includegraphics[width=0.035\linewidth]{figures/icons/status.png}} & \nuScenes & SD \cite{rombach2022high} & REG & Traj.  
    \\
    \rowcolor{gray!7}36 & GenAD \cite{yang2024generalized} & {\scriptsize{CVPR'24}} & \raisebox{-0.5ex}{\includegraphics[width=0.035\linewidth]{figures/icons/rgb.png}} & \OpenDV & SDXL \cite{podell2023sdxl} & REG & Traj.
    \\
    37 & Drive-WM \cite{wang2024driving} & {\scriptsize{CVPR'24}} & \raisebox{-0.5ex}{\includegraphics[width=0.035\linewidth]{figures/icons/rgb.png}} \raisebox{-0.5ex}{\includegraphics[width=0.035\linewidth]{figures/icons/status.png}} & \nuScenes & ConvNeXt \cite{liu2022convnet} & 
    SEL & Traj. 
    \\
    \rowcolor{gray!7}38 & DrivingWorld \cite{hu2024drivingworld} & {\scriptsize{arXiv'24}} & \raisebox{-0.5ex}{\includegraphics[width=0.035\linewidth]{figures/icons/rgb.png}} \raisebox{-0.5ex}{\includegraphics[width=0.035\linewidth]{figures/icons/status.png}} & \nuPlan & VQ-VAE \cite{van2017neural} & REG & Traj. 
    \\
    39 & Imagine-2-Drive \cite{garg2024imagine} & {\scriptsize{IROS'25}} & \raisebox{-0.5ex}{\includegraphics[width=0.035\linewidth]{figures/icons/rgb.png}} & \Carla & SVD \cite{blattmann2023stable} & 
    SEL & Traj. 
    \\
    \rowcolor{gray!7}40 & DrivingGPT \cite{chen2024drivinggpt} & {\scriptsize{ICCV'25}} & \raisebox{-0.5ex}{\includegraphics[width=0.035\linewidth]{figures/icons/rgb.png}} \raisebox{-0.5ex}{\includegraphics[width=0.035\linewidth]{figures/icons/status.png}} & \nuPlan~\NAVSIM & VQ-VAE \cite{van2017neural} & REG &  Traj. 
    \\
    41 & Epona \cite{zhang2025epona} & {\scriptsize{ICCV'25}} & \raisebox{-0.5ex}{\includegraphics[width=0.035\linewidth]{figures/icons/rgb.png}} \raisebox{-0.5ex}{\includegraphics[width=0.035\linewidth]{figures/icons/status.png}} & \nuScenes~\NAVSIM~\nuPlan & DC-AE \cite{chen2025dc} & REG & Traj.  
    \\
    \rowcolor{gray!7}42 & VaViM \cite{bartoccioni2025vavim} & {\scriptsize{arXiv'25}} & \raisebox{-0.5ex}{\includegraphics[width=0.035\linewidth]{figures/icons/rgb.png}}  & \OpenDV~\nuScenes~\nuPlan & LLaMAGen \cite{sun2024autoregressive} & GEN & Traj. 
    \\
    \midrule
    \rowcolor{tpami_yellow!10}\multicolumn{8}{l}{{$\bullet$~\textbf{Sec. \ref{sec:occworldmodel} Occupancy-Based World Models }}} 
    \\
    43 & OccWorld \cite{zheng2024occworld} & {\scriptsize{ECCV'24}} & \raisebox{-0.5ex}{\includegraphics[width=0.035\linewidth]{figures/icons/rgb.png}} \raisebox{-0.5ex}{\includegraphics[width=0.035\linewidth]{figures/icons/status.png}} & \nuScenes~\OccThreeD & ResNet \cite{he2016deep} & REG &  Traj. 
    \\
    \rowcolor{gray!7}44 & NeMo \cite{huang2024neural} & {\scriptsize{ECCV'24}} & \raisebox{-0.5ex}{\includegraphics[width=0.035\linewidth]{figures/icons/rgb.png}} & \nuScenes & ResNet \cite{he2016deep} & REG & Traj.  
    \\
    45 & OccVAR \cite{jinoccvar} & - & \raisebox{-0.5ex}{\includegraphics[width=0.035\linewidth]{figures/icons/rgb.png}} \raisebox{-0.5ex}{\includegraphics[width=0.035\linewidth]{figures/icons/status.png}} & \nuScenes~\OccThreeD & ResNet \cite{he2016deep} & REG & Traj.  
    \\
    \rowcolor{gray!7}46 & RenderWorld \cite{yan2024renderworld} & {\scriptsize{arXiv'24}} & \raisebox{-0.5ex}{\includegraphics[width=0.035\linewidth]{figures/icons/rgb.png}} & \nuScenes~\OccThreeD & Swin-T \cite{liu2021swin} & REG & Traj.  
    \\
    47 & DFIT-OccWorld \cite{zhang2024efficient} & {\scriptsize{arXiv'24}} & \raisebox{-0.5ex}{\includegraphics[width=0.035\linewidth]{figures/icons/rgb.png}} \raisebox{-0.5ex}{\includegraphics[width=0.035\linewidth]{figures/icons/status.png}} & \nuScenes~\OccThreeD & ResNet \cite{he2016deep} & REG & Traj.  
    \\
    \rowcolor{gray!7}48 & Drive-OccWorld \cite{yang2025driving} & {\scriptsize{AAAI'25}} & \raisebox{-0.5ex}{\includegraphics[width=0.035\linewidth]{figures/icons/rgb.png}} & \nuScenes~\CamOcc  & ResNet \cite{he2016deep} & REG &  Traj. 
    \\
    49 & $\text{T}^3$Former \cite{xu2025temporal} & {\scriptsize{arXiv'25}} & \raisebox{-0.5ex}{\includegraphics[width=0.035\linewidth]{figures/icons/rgb.png}} \raisebox{-0.5ex}{\includegraphics[width=0.035\linewidth]{figures/icons/status.png}} & \nuScenes~\OccThreeD & ResNet \cite{he2016deep} & REG & Traj.  
    \\
    \rowcolor{gray!7}50 & AD-R1 \cite{yan2025ad} & {\scriptsize{arXiv'25}} &  \raisebox{-0.5ex}{\includegraphics[width=0.035\linewidth]{figures/icons/rgb.png}} \raisebox{-0.5ex}{\includegraphics[width=0.035\linewidth]{figures/icons/lidar.png}} \raisebox{-0.5ex}{\includegraphics[width=0.035\linewidth]{figures/icons/status.png}} & \nuScenes~\NAVSIM & - & RL & Traj. 
    \\
    \midrule
    \rowcolor{tpami_yellow!10}\multicolumn{8}{l}{{$\bullet$~\textbf{Sec.  \ref{sec:latentworldmodel} Latent-Based World Models }}} 
    \\
    51 & Covariate-Shift \cite{popov2024mitigating} & {\scriptsize{arXiv'24}} & \raisebox{-0.5ex}{\includegraphics[width=0.035\linewidth]{figures/icons/rgb.png}} \raisebox{-0.5ex}{\includegraphics[width=0.035\linewidth]{figures/icons/status.png}} & ~\Carla & DINOv2 \cite{oquab2023dinov2} & REG & Traj.  
    \\
    \rowcolor{gray!7}52 & World4Drive \cite{zheng2025world4drive} & {\scriptsize{ICCV'25}} & \raisebox{-0.5ex}{\includegraphics[width=0.035\linewidth]{figures/icons/rgb.png}} & \nuScenes~\NAVSIM & ResNet \cite{he2016deep} & REG &  Traj. 
    \\
    53 & WoTE \cite{li2025end} & {\scriptsize{ICCV'25}} & \raisebox{-0.5ex}{\includegraphics[width=0.035\linewidth]{figures/icons/rgb.png}} \raisebox{-0.5ex}{\includegraphics[width=0.035\linewidth]{figures/icons/lidar.png}} & \NAVSIM~\BenchDrive & ResNet \cite{he2016deep} & 
    SEL &  Traj. 
    \\
    \rowcolor{gray!7}54 & LAW \cite{li2024enhancing} & {\scriptsize{ICLR'25}} & \raisebox{-0.5ex}{\includegraphics[width=0.035\linewidth]{figures/icons/rgb.png}} & \nuScenes~\NAVSIM~\Carla & Swin-T \cite{liu2021swin} & REG & Traj.  
    \\
    55 & SSR \cite{li2024navigation} & {\scriptsize{ICLR'25}} & \raisebox{-0.5ex}{\includegraphics[width=0.035\linewidth]{figures/icons/rgb.png}} & \nuScenes~\Carla & ResNet \cite{he2016deep} & REG & Traj.  
    \\
    \rowcolor{gray!7}56 & Echo-Planning \cite{sun2025echo} & {\scriptsize{arXiv'25}} & \raisebox{-0.5ex}{\includegraphics[width=0.035\linewidth]{figures/icons/rgb.png}} & \nuScenes & ResNet \cite{he2016deep} & REG & Traj. 
    \\
    57 & SeerDrive \cite{zhang2025future} & {\scriptsize{NeurIPS'25}} & \raisebox{-0.5ex}{\includegraphics[width=0.035\linewidth]{figures/icons/rgb.png}} \raisebox{-0.5ex}{\includegraphics[width=0.035\linewidth]{figures/icons/lidar.png}} & \nuScenes~\NAVSIM  & VoVNet \cite{lee2019energy}  & 
    SEL & Traj.
    \\
    \bottomrule
\end{tabular}
}
\label{tab:va_methods}
\vspace{-1cm}
\end{wraptable}
NEAT~\cite{chitta2021neat} highlights behaviorally relevant image regions via intermediate attention maps, while TCP~\cite{wu2022trajectoryguided} fuses a trajectory branch and control branch for complementary supervision. To better leverage scene geometry, BEV-Planner~\cite{li2024ego} predicts trajectories from BEV features enriched with ego states. Urban-Driver~\cite{scheel2022urban} moves beyond open-loop evaluation by training policies in a differentiable, data-driven simulator.

IL-based methods are simple, efficient, and require no reward engineering; however, they remain sensitive to distribution shift~\cite{ross2011reduction,dosovitskiy2017carla,prakash2020exploring} and causal confusion~\cite{muller2005off,de2019causal,park2021object}, which can impair reliability in long-tailed or rare-event scenarios.

\noindent\textbf{Reinforcement Learning (RL)} optimizes actions through interaction, offering greater flexibility than imitation-based approaches \cite{sutton1998reinforcement,kiran2021deep}. Several works address the sample inefficiency of RL by combining it with supervised pretraining: Latent-DRL~\cite{toromanoff2020end} and Gri~\cite{chekroun2023gri} pre-train visual encoders using semantic segmentation, while LSD~\cite{ohn2020learning} initializes policies via IL before performing RL fine-tuning. Privileged-information distillation has also proven effective: LBC~\cite{chen2020learning}, WoR~\cite{chen2021learning}, and Roach~\cite{zhang2021end} use simulator-only states to guide sensor-based agents.

Combined with the world model, Think2Drive \cite{li2024think2drive} trains the agent with the Model-Based RL (MBRL) method, paired with a compact latent world model learning the transitions of the environment. Raw2Drive~\cite{yang2025raw2drive} is a dual-stream MBRL approach, where the raw sensor world model is aligned with the privileged world model for camera-based action prediction. In contrast to studies in non-photorealistic CARLA \cite{dosovitskiy2017carla}, recent efforts have shifted toward photorealistic world modeling. RAD~\cite{gao2025rad} establishes a 3DGS-based~\cite{kerbl20233d} closed-loop RL training paradigm regulated by IL in a realistic 3DGS environment. The key challenges in RL-based models include sample inefficiency \cite{li2025survey}, reward function design \cite{knox2023reward}, and sim-to-real transfer \cite{chib2023recent}.

\subsubsection{Perception-Action Model}
\label{sec:perceptionaction}

\begin{wrapfigure}{r}{0.6\textwidth}
    \begin{minipage}{\linewidth}
        \centering
        \includegraphics[width=\linewidth]{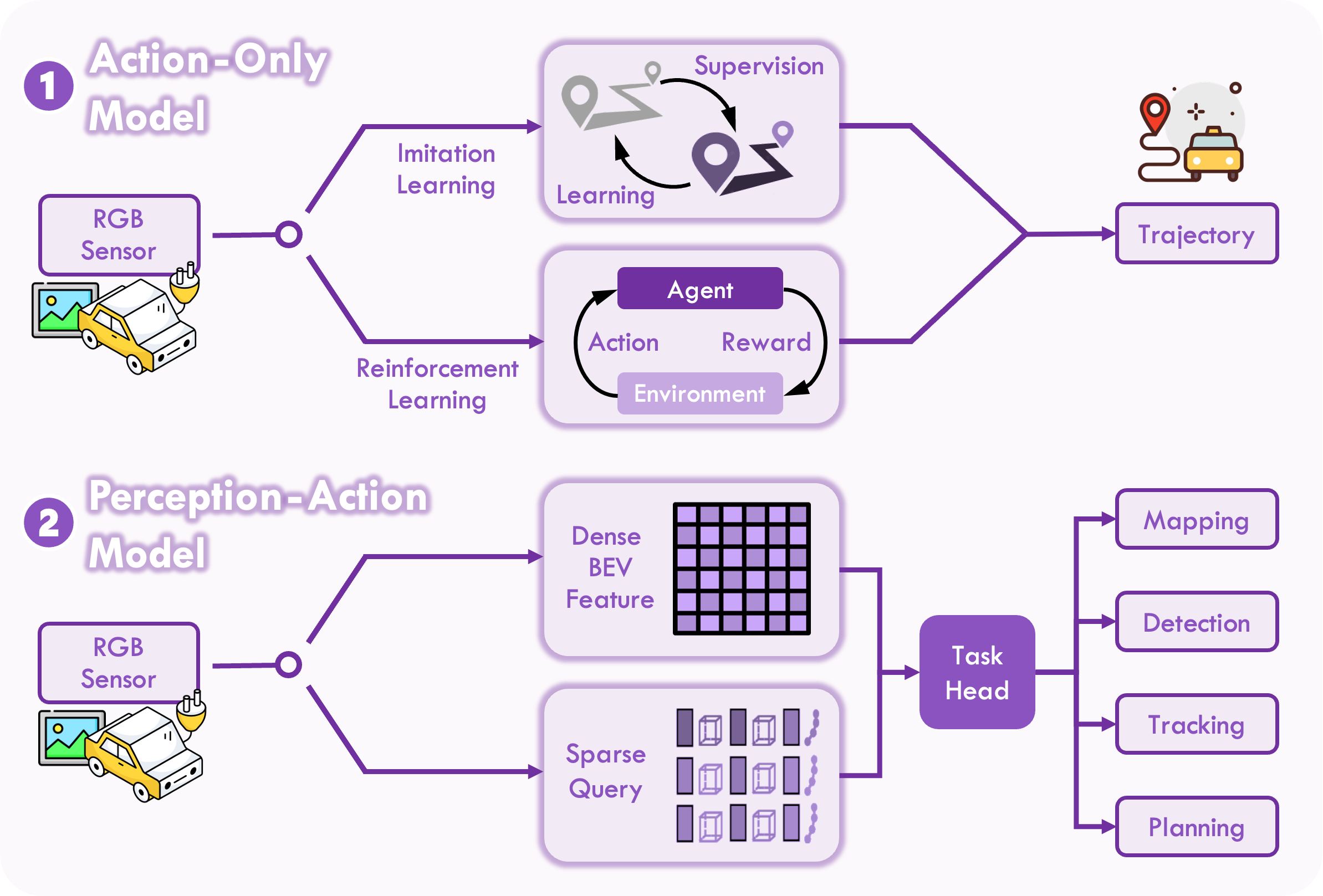}
        \vspace{-0.6cm}
        \caption{The categorization of \textbf{End-to-End VA models} based on model structures and outputs, including \emph{Action-Only Models} (Sec.~\ref{sec:actiononly}), and \emph{Perception-Action Models} (Sec.~\ref{sec:perceptionaction}).}
        \label{fig:e2e-model}
        \vspace{-0.4cm}
    \end{minipage}
\end{wrapfigure}

Perception-action models follow a two-stage paradigm in which perception tasks (\emph{e.g.}, mapping, tracking) supervise and constrain trajectory prediction. These methods generally adopt either dense BEV-based representations or sparse query-based representations, as shown in Figure~\ref{fig:e2e-model}.

\noindent\textbf{Dense BEV-Based Models} construct unified top-down features from multi-view cameras. ST-P3~\cite{hu2022st} jointly learns spatial-temporal features for perception and planning; UniAD~\cite{hu2023planning} integrates sequential task dependencies to support goal-directed planning. VAD~\cite{jiang2023vad} employs a vectorized scene representation to improve both planning safety and efficiency. OccNet~\cite{tong2023scene} incorporates occupancy embeddings to capture 3D scene geometry. Para-Drive~\cite{weng2024drive} proposes a fully parallel E2E architecture for real-time deployment.

Generative and sampling-based approaches have recently emerged: GenAD \cite{zheng2024genad} frames planning as sampling from learned distributions; DiffusionDrive \cite{liao2025diffusiondrive} introduces a truncated diffusion policy guided by multi-modal anchors; GuideFlow \cite{liu2025guideflow} incorporates explicit physical constraints into the generation process. While BEV representations naturally align with 2D trajectory planning, they require substantial computation due to their dense spatial structure.

\noindent\textbf{Sparse Query-Based Models} avoid explicit BEV grids by using latent queries to aggregate image features. SparseAD~\cite{zhang2024sparsead} and SparseDrive~\cite{sun2025sparsedrive} represent the entire scene using sparse perception queries and a parallel planner, achieving strong efficiency-accuracy trade-offs. DiFSD~\cite{su2024difsd} introduces an ego-centric sparse formulation and models uncertainty through trajectory denoising. DriveTransformer~\cite{jia2025drivetransformer} incorporates task parallelism, sparse attention, and streaming updates for improved stability. GaussianAD~\cite{zheng2024gaussianad} adopts 3D semantic Gaussians for fine-grained yet compact scene representation.

Sparse query methods significantly reduce inference latency, but the absence of a dense future-world representation can restrict long-horizon reasoning and planning safety.

\subsection{World Models for Autonomous Driving} 
\label{sec:wmad}
\begin{wrapfigure}{r}{0.6\textwidth}
    \begin{minipage}{\linewidth}
        \centering
        \vspace{-0.4cm}
        \includegraphics[width=\linewidth]{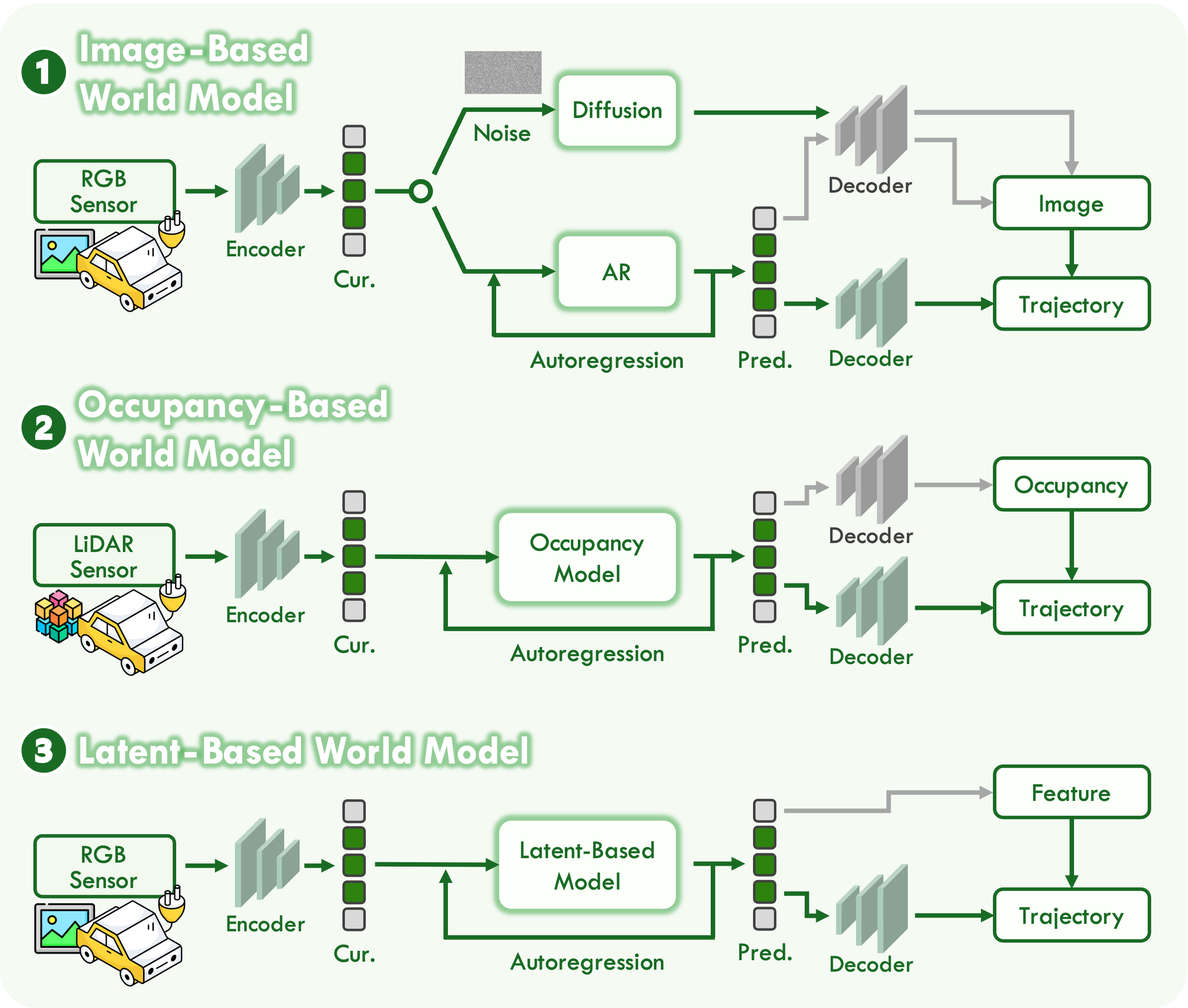}
        \vspace{-0.6cm}
        \caption{The categorization of \textbf{World Models} based on prediction modalities, including \emph{Image-Based Models} (Sec.~\ref{sec:imageworldmodel}), \emph{Occupancy-Based Models} (Sec.~\ref{sec:occworldmodel}), and \emph{Latent-Based Models} (Sec.~\ref{sec:latentworldmodel}).}
        \label{fig:world-model}
    \end{minipage}
    \vspace{-1cm}
\end{wrapfigure}

World models aim to predict how driving scenes evolve under different ego actions \cite{ha2018world,yan2024forging,survey_3d_4d_world_models}. By jointly modeling scene dynamics and ego motion, they provide a powerful mechanism for learning safe, long-horizon driving policies~\cite{ding2024understanding}. Their applications span immersive simulation~\cite{hu2023gaia,lu2024wovogen,ren2025cosmos,yan2025drivingsphere,bian2025dynamiccity,zhu2025spiral}, end-to-end planning~\cite{wang2024drivedreamer,gao2024vista,wang2024driving,hu2024drivingworld}, and feature learning for downstream tasks~\cite{min2024driveworld,yang2024visual,zhang2025visionpad,liu2023segment,chen2023clip2Scene,chen2023towards,li2025_3eed}. Here, we focus on world models designed for trajectory planning and categorize them by prediction modality and representation granularity into three groups: \textbf{image-based}, \textbf{occupancy-based}, and \textbf{latent-based} models (Figure~\ref{fig:world-model}).

\subsubsection{Image-Based World Model} 
\label{sec:imageworldmodel}
Image-based world models generate future frames conditioned on ego actions, enabling agents to ``dream'' scene evolution and evaluate the consequences of different trajectories. These methods leverage modern generative models to synthesize realistic, temporally coherent videos and are typically classified into diffusion-based and autoregressive architectures.

\noindent\textbf{Diffusion-Based World Models} use latent video diffusion~\cite{rombach2022high,blattmann2023align} to produce multi-step rollouts.
For front-view forecasting, GenAD~\cite{yang2024generalized} and Vista~\cite{gao2024vista} incorporate temporal reasoning modules to handle complex motion patterns. Imagine-2-Drive~\cite{garg2024imagine} integrates diffusion generation into a reinforcement-learning framework, training a policy actor inside the world model.
To support multi-view predictions, DriveDreamer~\cite{wang2024drivedreamer} employs a two-stage pipeline for video synthesis and policy learning. Drive-WM~\cite{wang2024driving} factors views within a spatiotemporal model and generates multiple plausible futures, selecting trajectories using image-based rewards.

\noindent\textbf{Autoregressive (AR) Models} tokenize images using VQ-VAE~\cite{van2017neural} and model scene evolution via next-token prediction~\cite{esser2021taming,lee2022autoregressive,yu2023language}.
DrivingWorld~\cite{hu2024drivingworld} builds a GPT-style architecture for high-fidelity long-horizon video generation. DrivingGPT~\cite{chen2024drivinggpt} interleaves image and action tokens, unifying simulation and trajectory planning. Epona~\cite{zhang2025epona} combines AR modeling with diffusion to produce high-resolution, long-duration rollouts.

Image-based world models provide photorealistic simulations crucial for training and evaluation. However, their reliance on 2D appearance limits explicit 3D reasoning, which can hinder safety-critical long-horizon planning.

\subsubsection{Occupancy-Based World Models}
\label{sec:occworldmodel}
Occupancy-based world models represent the driving scene as spatiotemporal occupancy grids and predict their evolution under different actions. Instead of synthesizing raw pixels, these models focus on the geometry and semantics of free space, obstacles, and agents \cite{liang2026lidarcrafter,liu2026lalalidar,zhu2025spiral}. As shown in the middle of Figure~\ref{fig:world-model}, AR prediction is commonly used in occupancy world models.

OccWorld~\cite{zheng2024occworld} first introduces occupancy forecasting for planning, using a scene tokenizer to discretize 3D occupancy before applying a GPT-style transformer to synthesize future scenes and ego trajectories. RenderWorld~\cite{yan2024renderworld} produces 3D occupancy through a self-supervised Gaussian module, while OccVAR~\cite{jinoccvar} performs coarse-to-fine 4D occupancy forecasting. $\text{T}^3$Former~\cite{xu2025temporal} encodes occupancy using compact triplanes and predicts future triplane updates from multi-scale history.

An alternative line employs single-stage feedforward prediction.
Drive-OccWorld~\cite{yang2025driving} uses predicted future BEV features for action-conditioned 3D forecasting. DFIT-OccWorld~\cite{zhang2024efficient} introduces a decoupled dynamic flow strategy to support efficient non-autoregressive prediction. NeMo~\cite{huang2024neural} improves vision-based occupancy forecasting by combining it with self-supervised image reconstruction signals.

Occupancy-based models offer strong geometric fidelity and explicit free-space reasoning but rely on costly 3D annotations, which can limit scalability across diverse environments.

\subsubsection{Latent-Based World Models} 
\label{sec:latentworldmodel}
Latent-based world models bypass explicit image or occupancy forecasting and instead predict future dynamics directly in a compressed latent space. By operating on high-level features, these models capture behavioral patterns and long-horizon dependencies while avoiding the computational overhead of pixel-level generation. Regarding the forecasting strategy, the latent world model utilizes single-frame or AR prediction presented at the bottom of Figure~\ref{fig:world-model}.

Early latent models~\cite{zhou2024dino,baldassarre2025back} learn feature-level dynamics for planning without generating visual frames. LAW~\cite{li2024enhancing} leverages self-supervised learning to predict future scene features from current features and planned ego trajectories, enabling end-to-end driving without perception labels.
World4Drive~\cite{zheng2025world4drive} employs vision foundation models to create latent representations from which diverse planning trajectories can be generated and evaluated. Echo-Planning~\cite{sun2025echo} introduces a bidirectional Current→Future→Current (CFC) cycle to enforce temporal consistency in latent BEV features.
For robustness in imitation learning, Covariate-Shift~\cite{popov2024mitigating} addresses distribution mismatch using latent rollouts. 
By injecting predicted BEV features, SeerDrive~\cite{zhang2025future} refines both latent prediction and trajectory generation in a closed-loop manner.

Latent world models offer efficient and semantically informed forecasting. However, achieving high-quality planning still requires auxiliary supervision from 2D/3D annotations, such as bounding boxes or HD maps.

\subsection{Limitations of VA Compared to VLA}
While VA models remain widely deployed, they face structural limitations that hinder performance in complex, ambiguous, or long-tailed scenarios: areas where VLA models excel.

\begin{itemize}
    \item \textbf{Limited Interpretability.} VA models provide little insight into their decision-making process. In contrast, VLA models can articulate reasoning steps or explanations through language.

    \item \textbf{Weak Generalization.} VA policies lack broad world knowledge and often require environment-specific retraining. VLA models leverage large-scale pretraining to generalize better under distribution shifts and long-tailed events.

    \item \textbf{No Chain-of-Thought Reasoning.} VA models directly map pixels to actions, making it difficult to perform explicit reasoning or contextual analysis. VLAs natively support step-wise reasoning.

    \item \textbf{No Language Understanding.} VA systems cannot incorporate human instructions or high-level goals expressed in texts. VLA models naturally integrate such inputs to guide planning and decision-making.
\end{itemize}
\section{Vision-Language-Action Models}
\label{sec:vla}

\begin{table}[t]
    \centering
    \caption{Categories of \textbf{natural language prompts} for Vision-Language-Action (VLA) models in autonomous driving.}
    \vspace{-0.2cm}
    \label{tab:prompt_category}
    \begin{footnotesize}
    \begin{tabularx}{0.77\textwidth}{lX}
    \toprule
    \textbf{Prompt Type} & \textbf{Explanations} 
    \\
    \midrule\midrule
    \raisebox{-.5\height}{\includegraphics[height=1.2em]{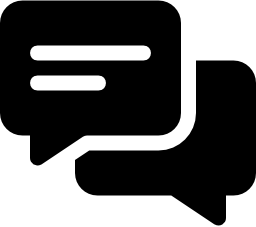}}~\textbf{System Prompt} & \cellcolor{vla_purple!7}Text templates or query formulations designed to interact with LLMs, guiding them to perform specific driving-related reasoning or trajectory prediction tasks. System prompts often define the task structure, provide role definitions, and shape the model’s reasoning behavior. 
    \\
    \midrule
    \raisebox{-.5\height}{\includegraphics[height=1.2em]{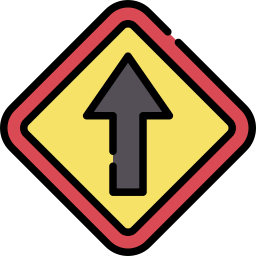}}~\textbf{Instructions} & \cellcolor{vla_green!7}Commands or instructions provided by humans or systems, typically describing the driving goal or required maneuver (\emph{e.g.}, ``turn left at the next intersection''). 
    \\
    \midrule
    \raisebox{-.5\height}{\includegraphics[height=1.2em]{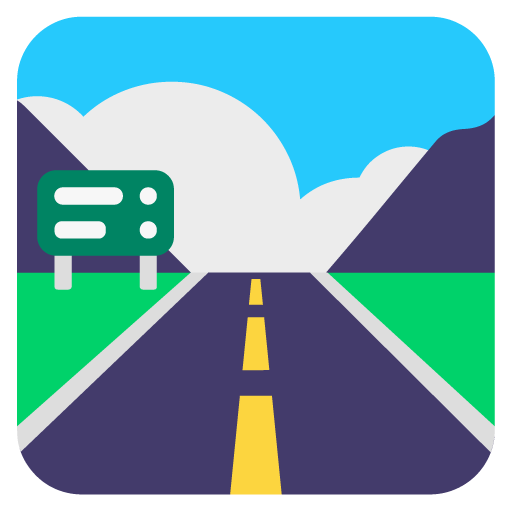}}~\textbf{Scene Description} & \cellcolor{tpami_yellow!7}Textual descriptions of surroundings, including perceived objects, road layout, and contextual factors (\emph{e.g.}, \emph{``a pedestrian is crossing on the right''}, or \emph{``a vehicle is 5 meters ahead on the left''}). 
    \\
    \midrule
    \raisebox{-.5\height}{\includegraphics[height=1.2em]{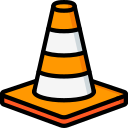}}~\textbf{Traffic Rules} & \cellcolor{vla_purple!7}Prompts encoding regulatory constraints or domain knowledge, such as traffic laws, traffic light status, right-of-way rules, or safety guidelines. 
    \\
    \midrule
    \raisebox{-.5\height}{\includegraphics[height=1.2em]{figures/icons/status.png}}~\textbf{Ego Status} & \cellcolor{vla_green!7}Information about the ego vehicle’s internal state, including speed, position, heading, or navigation intent. 
    \\
    \midrule
    \raisebox{-.5\height}{\includegraphics[height=1.3em]{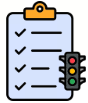}}~\textbf{Context Information~~} & \cellcolor{tpami_yellow!7}Demonstrations presented as paired examples of driving scenarios and corresponding actions, used to guide the model via in-context learning. 
    \\
    \bottomrule
\end{tabularx}
\end{footnotesize}
\end{table}

VLA models extend the Vision-Action paradigm by coupling visual perception with the multimodal reasoning capabilities of large vision-language models. Equipped with chain-of-thought style inference and broad world knowledge, these models are particularly promising for rare, ambiguous, and long-tailed driving scenarios. Table~\ref{tab:prompt_category} summarizes typical prompting strategies, and Table~\ref{tab:vla_methods} overviews representative VLA-based approaches.

\clearpage\clearpage
\begin{table}[!ht]
    \centering
    \vspace{-0.7cm}
    \caption{Summary of \textbf{Vision-Language-Action} models in autonomous driving.
    \\
    $\bullet$ \textbf{Input}: \raisebox{-0.5ex}{\includegraphics[width=0.02\linewidth]{figures/icons/rgb.png}}:Camera, \raisebox{-0.5ex}{\includegraphics[width=0.02\linewidth]{figures/icons/prompt.png}}: Sys. Prompt, \raisebox{-0.5ex}{\includegraphics[width=0.02\linewidth]{figures/icons/command.png}}: Instruct., \raisebox{-0.5ex}{\includegraphics[width=0.02\linewidth]{figures/icons/scene.png}}: Scene Descrip., \raisebox{-0.5ex}{\includegraphics[width=0.02\linewidth]{figures/icons/status.png}}: Status, \raisebox{-0.5ex}{\includegraphics[width=0.02\linewidth]{figures/icons/traffic_rule.png}}: Traffic Rule, \raisebox{-0.5ex}{\includegraphics[width=0.02\linewidth]{figures/icons/context.png}}: Context Info.
    \\
    $\bullet$ \textbf{Action}: \underline{LH}: Lang. Head, \underline{REG}: Decoder+MLP, \underline{GEN}: Traj. GEN with Generative Model.
    \\
    $\bullet$ \textbf{Output}: \underline{Desc.}: Linguistic Descriptions, \underline{Traj.}: Numerical Trajectory, \underline{Ctrl.}: Control Signal. \underline{Meta.}: Meta Action.
    \\
    $\bullet$ \textbf{Datasets:} \nuScenes: nuScenes~\cite{caesar2020nuscenes}, \BDDX: BDD-X~\cite{kim2018textual}, \DriveLM: DriveLM~\cite{sima2024drivelm}, \SDN: SDN~\cite{ma2022dorothie}, \VLAAD: VLAAD~\cite{park2024vlaad}, \BenchDrive: Bench2Drive~\cite{jia2024bench2drive}, \Waymo: Waymo~\cite{ettinger2021large}, \MetaAD: MetaAD~\cite{jiang2025alphadrive}, \Carla: Carla~\cite{dosovitskiy2017carla}, \ImpromptuVLA: ImpromptuVLA~\cite{chi2025impromptu}, \NAVSIM: NAVSIM~\cite{dauner2024navsim}, \OpenDV: OpenDV~\cite{yang2024generalized}, \nuPlan: nuPlan~\cite{caesar2021nuplan}, \TalkCar: Talk2Car~\cite{deruyttere2019talk2car}, \CoVLA: CoVLA~\cite{arai2025covla}, \PhysicalAI: PhysicalAI-AV~\cite{nvidia2025avdata} and \Private: Private Data. 
    \\
    }
    \vspace{-6mm}
    \resizebox{\linewidth}{!}{
\begin{tabular}{c|r|r|c|c|l|l|c|l}
    \toprule
    \textbf{\#} & \textbf{Model} & \textbf{Venue}  & \textbf{Input Modality} & \textbf{Dataset} & \textbf{Vision} & \textbf{Language} & \textbf{Action} & \textbf{Output} 
    \\
    \midrule\midrule
    \rowcolor{tpami_yellow!10}\multicolumn{9}{l}{{$\bullet$~\textbf{Sec.~\ref{sec:textualaction}: Textural Action Generator}}} 
    \\
    1 & DriveMLM  \cite{wang2023drivemlm} & {\scriptsize{arXiv'23}} & 
    \raisebox{-0.5ex}{\includegraphics[width=0.02\linewidth]{figures/icons/rgb.png}}~\raisebox{-0.5ex}{\includegraphics[width=0.02\linewidth]{figures/icons/prompt.png}}~\raisebox{-0.5ex}{\includegraphics[width=0.02\linewidth]{figures/icons/command.png}}~\raisebox{-0.5ex}{\includegraphics[width=0.02\linewidth]{figures/icons/traffic_rule.png}} & \Carla & EVA-CLIP~\cite{fang2023eva} & \raisebox{-0.5ex}{\includegraphics[width=0.016\linewidth]{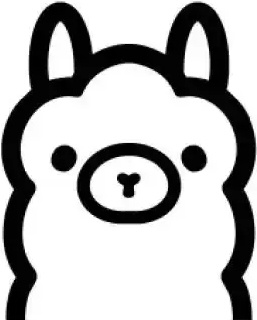}}~LLaMA \cite{touvron2023llama} & LH & Desc.+Meta. 
    \\
    \rowcolor{gray!7}2 & RAG-Driver  \cite{yuan2024rag} & {\scriptsize{RSS'24}} & \raisebox{-0.5ex}{\includegraphics[width=0.02\linewidth]{figures/icons/rgb.png}}~\raisebox{-0.5ex}{\includegraphics[width=0.02\linewidth]{figures/icons/prompt.png}}  \raisebox{-0.5ex}{\includegraphics[width=0.02\linewidth]{figures/icons/command.png}} \raisebox{-0.5ex}{\includegraphics[width=0.02\linewidth]{figures/icons/context.png}} \raisebox{-0.5ex}{\includegraphics[width=0.02\linewidth]{figures/icons/status.png}} & \BDDX & CLIP  \cite{radford2021learning} & Vicuna-1.5 \cite{chiang2023vicuna} & LH & Desc.+Ctrl. 
    \\
\bottomrule
\end{tabular}
  }
\label{tab:vla_methods}
\vspace{-0.4cm}
\end{table}

\clearpage\clearpage
From an architectural standpoint, current VLA methodologies for autonomous driving can be grouped into:
\begin{itemize}
    \item \textbf{End-to-End VLA}: a single model directly maps multimodal sensory inputs and language to actions.
    \item \textbf{Dual-System VLA}: a VLM provides high-level reasoning or guidance, while a specialized driving module executes fast, low-level action.
\end{itemize}

\subsection{End-to-End VLA for Autonomous Driving}
\label{sec:e2evla}
\begin{wrapfigure}{r}{0.6\textwidth}
    \begin{minipage}{\linewidth}
        \centering
        \vspace{-0.5cm}
        \includegraphics[width=\linewidth]{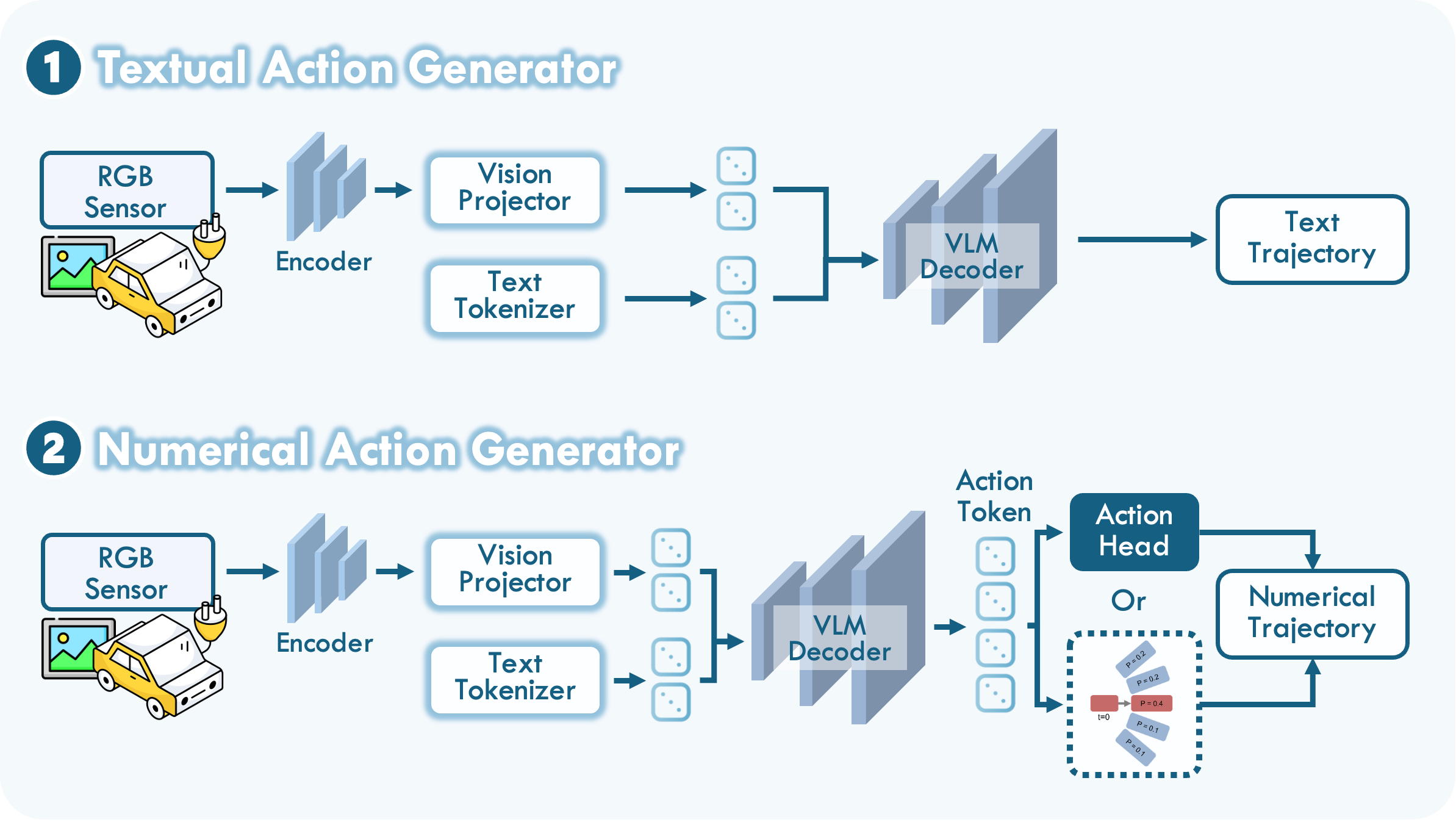}
        \vspace{-0.6cm}
        \caption{The categorization of \textbf{End-to-End VLA models} based on the form of model outputs, including \emph{Textual Action Models} (Sec.~\ref{sec:textualaction}), and \emph{Numerical Action Models} (Sec.~\ref{sec:numericalaction}).}
        \label{fig:e2e-vla}
    \end{minipage}
    \vspace{-0.3cm}
\end{wrapfigure}

End-to-end VLA frameworks aim to unify perception, reasoning, and planning within a single architecture. By leveraging the generalization ability of multimodal large language models (MLLMs), they directly transform multimodal observations into actions, reducing reliance on hand-crafted modules and task-specific heuristics.  
According to the form of their outputs, existing approaches can be broadly divided into two families, as illustrated in Figure~\ref{fig:e2e-vla}: \textbf{textual action generators}, which operate primarily in the language space, and \textbf{numerical action generators}, which predict trajectories or controls in a continuous or discretized numeric space.

\subsubsection{Textual Action Generator}
\label{sec:textualaction}
Textual action generators formulate driving as a reasoning problem in the language space. The model produces human-readable symbolic decisions, allowing it to ``think'' and justify its outputs in words. Depending on the abstraction level of these outputs, existing methods can be grouped into \emph{meta-actions} and \emph{trajectory waypoints}.

\textbf{Meta-Actions} are discrete, semantic driving decisions, such as ``accelerate'', ``stop'', or ``change lane''. They form an interpretable interface between high-level reasoning in VLMs and downstream controllers. Early works mainly used language models to output free-form text or conceptual descriptions, which are not directly executable. DriveMLM~\cite{wang2023drivemlm} narrows this gap by aligning LLM outputs with behavioral planning states in a modular stack, enabling language models to act as intermediate planners whose symbolic decisions can be converted into control commands.

Subsequent methods strengthen robustness and reasoning-planning alignment with reinforcement learning and chain-of-thought supervision~\cite{jiang2025alphadrive,shao2024deepseekmath,li2025drive}. AlphaDrive~\cite{jiang2025alphadrive} introduces Group Relative Policy Optimization (GRPO)~\cite{shao2024deepseekmath} to refine meta-actions using rewards that jointly consider trajectory quality, decision correctness, and format consistency. DriveAgent-R1~\cite{zheng2025driveagent} first fine-tunes on a curated CoT dataset to encourage step-wise visual reasoning, then applies RL with trajectory- and meta-action-based rewards to bias reasoning paths toward decisions that are practically useful for driving. Recognizing that single-frame front-view inputs limit temporal and spatial understanding, Sce2DriveX~\cite{zhao2025sce2drivex} further incorporates multi-view video streams and BEV representations, enabling context-aware meta-decisions that are consistent with road topology and spatiotemporal dynamics.


\textbf{Trajectory Waypoints}–based textual generators frame motion planning as the prediction of future coordinates expressed in natural language, thereby unifying reasoning and trajectory forecasting within a single linguistic sequence.
DriveLM~\cite{sima2024drivelm} is an early representative of this paradigm, modeling autonomous driving as graph-structured visual question answering and generating textualized trajectory waypoints conditioned on multi-stage perception, prediction, and planning. Building on this idea, subsequent works adopt end-to-end multimodal formulations. EMMA~\cite{hwang2024emma} integrates camera observations and navigation commands into a unified language-driven pipeline for joint perception, road-graph understanding, and trajectory prediction. To enhance robustness in challenging scenarios, ImpromptuVLA~\cite{chi2025impromptu} introduces an 80K-clip corner-case dataset, demonstrating that pretraining on diverse edge cases significantly improves trajectory accuracy and closed-loop stability. LightEMMA~\cite{qiao2025lightemma} further benchmarks 12 vision–language models, revealing clear trade-offs between interpretability and numerical precision.

A complementary research direction focuses on better aligning reasoning with decision-making. RDA-Driver~\cite{huang2024making} enforces consistency between chain-of-thought explanations and trajectory outputs through tailored constraints, while Drive-R1~\cite{li2025drive} leverages reinforcement learning to improve alignment between textual reasoning and waypoint prediction. Beyond alignment, efficiency and knowledge integration are explored by FastDriveVLA~\cite{cao2025fastdrivevla} via token pruning, WiseAD~\cite{zhang2024wisead} through explicit driving priors, and OmniDrive~\cite{wang2025omnidrive} using counterfactual reasoning. WKER~\cite{zhai2025world} further enhances robustness under occlusion by combining instruction-guided token selection with external knowledge sources.

Overall, textual action generators offer strong interpretability and rich reasoning but must bridge a fundamental gap between discrete language tokens and continuous control spaces. This mismatch can introduce precision limits and, in extreme cases, unstable or collapsed trajectories.

\subsubsection{Numerical Action Generator}
\label{sec:numericalaction}
Numerical action generators augment VLM backbones with mechanisms that produce directly usable numeric outputs. The model still leverages language-driven reasoning internally, but its final predictions are expressed as trajectories, waypoints, or control values that can be consumed by classical planners or low-level controllers. Two main realizations exist: \emph{additional action heads} attached to the backbone, and \emph{additional action tokens} that discretize continuous actions into a token space.

\textbf{Additional Action Head.} 
A common strategy is to attach specialized prediction heads to vision-language models. BEVDriver~\cite{winter2025bevdriver} couples a multimodal encoder with a GRU-based head over BEV features, linking language-grounded reasoning with spatial waypoint prediction. CoVLA-Agent~\cite{arai2025covla} uses a lightweight MLP head trained on the CoVLA dataset, demonstrating that joint supervision from trajectories and captions can simultaneously improve interpretability and numeric accuracy. DriveGPT4-V2~\cite{xu2025drivegpt4} augments token-based planning with an MLP that maps multimodal embeddings to continuous trajectories, enhancing sample efficiency while retaining GPT-style reasoning.

To specialize behaviors, DriveMoE~\cite{yang2025drivemoe} employs a Mixture-of-Experts design whose action head dynamically activates experts for skills such as lane following or overtaking. DSDrive~\cite{liu2025dsdrive} proposes a dual-head coordination module, with one head predicting waypoints and another generating reasoning outputs; distillation from larger VLMs keeps the model compact yet interpretable. LMDrive~\cite{shao2024lmdrive} integrates multimodal encoders with an MLP that directly outputs control signals in a closed loop, marking one of the first instruction-following, language-guided end-to-end systems.

Beyond simple MLPs, ORION~\cite{fu2025orion} replaces deterministic heads with a diffusion-based predictor, modeling multi-modal trajectory distributions under uncertainty. SimLingo~\cite{renz2025simlingo} decouples temporal speed waypoints from geometric path waypoints via a disentangled MLP head, enabling finer-grained control. 

\textbf{Additional Action Tokens.}
Instead of explicit heads, some works reuse the language token space to represent actions. AutoVLA~\cite{zhou2025autovla} discretizes continuous trajectories into a codebook of action tokens, which are autoregressively generated alongside reasoning tokens, thereby unifying semantic reasoning and planning within a single sequence. Reinforcement fine-tuning penalizes redundant reasoning and improves token efficiency. OpenDriveVLA~\cite{zhou2025opendrivevla} follows a similar token-based paradigm but grounds token generation in a hierarchical alignment between 2D/3D perception and the language model. Structured features are embedded into a unified semantic space, and interaction tokens for the ego vehicle, environment, and other agents are autoregressively decoded into driving actions.

Numerical action generators are well-suited for downstream control, as their outputs are natively compatible with planners and actuators. However, they typically sacrifice some interpretability and often require substantial supervised data for stable training. When discretized action tokens are used, quantization artifacts can further limit fine-grained control accuracy.

\subsection{Dual-System VLA}
\label{sec:dsvla}

Dual-system VLA frameworks draw inspiration from the dual-process theory popularized by \emph{Thinking, Fast and Slow}~\cite{kahneman2011thinking}. In this paradigm, a VLM serves as the \emph{slow}, deliberative system that performs high-level reasoning, situational assessment, and linguistic inference, while a specialized autonomous driving module acts as the \emph{fast} system responsible for real-time, low-latency trajectory generation and control. By combining these complementary strengths, dual-system frameworks aim to achieve both interpretability and safety-critical reactivity.

\begin{wrapfigure}{r}{0.6\textwidth}
    \begin{minipage}{\linewidth}
        \centering
        \vspace{-0.5cm}
        \includegraphics[width=\linewidth]{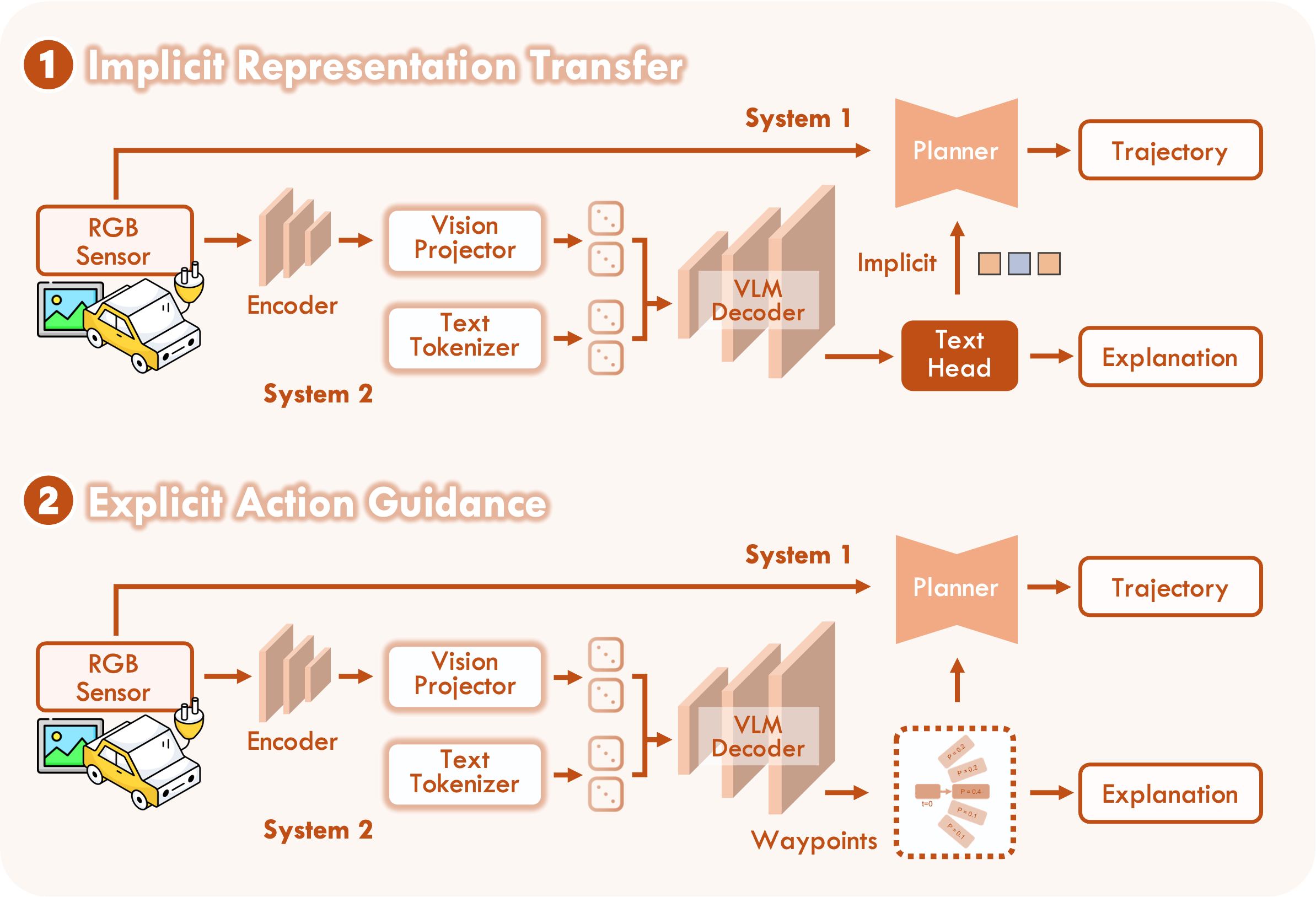}
        \vspace{-0.5cm}
        \caption{The categorization of \textbf{Dual-System VLA Models} based on how VLM interacts with E2E module, including explicit action guidance (Sec.~\ref{sec:explicitaction}), and implicit representations transfer models (Sec.~\ref{sec:implicitrepresentations}).}
        \label{fig:dual-system}
    \end{minipage}
    \vspace{-1cm}
\end{wrapfigure}
Depending on how VLM outputs interact with the specialized planner, existing methods can be categorized into two families: \textbf{explicit action guidance} and \textbf{implicit representation transfer}, as illustrated in Figure~\ref{fig:dual-system}.

\subsubsection{Explicit Action Guidance}
\label{sec:explicitaction}
Explicit action guidance frameworks use VLMs as structured action generators, whose high-level outputs are subsequently transformed or refined by the fast driving module. These approaches differ in their abstraction level and are mainly grouped into \textbf{meta-action guidance} and \textbf{waypoint supervision}.

\textbf{Meta-Action Guidance} resorts to VLMs to output symbolic driving intentions, such as ``slow down'', ``change lane'', or ``turn left'', which act as semantic priors for downstream planners. This design leverages the interpretability of linguistic actions while avoiding the precision challenges of directly generating continuous trajectories. Early work such as FasionAD~\cite{qian2024fasionad} embodies the dual-process design by pairing a fast, data-driven planner with a slow VLM that issues meta-actions; a learned switching mechanism selects the appropriate pathway based on confidence and scene context. LeapVAD~\cite{ma2025leapvad} refines this structure by combining an analytic branch that builds a memory bank with a heuristic branch that retrieves prior meta-actions for familiar situations.

More recent systems integrate high-level reasoning more tightly with planning. Senna~\cite{jiang2024senna} couples a commonsense VLM with an end-to-end planner: Senna-VLM produces natural-language decisions, which Senna-E2E converts into executable trajectories. DiffVLA~\cite{jiang2025diffvla} injects VLM-generated lateral and longitudinal decisions as one-hot priors into a diffusion-based planner, guiding multi-modal trajectory denoising.  

Hierarchical frameworks such as DME-Driver~\cite{han2025dme} further separate decision and execution: a VLM-based Decision-Maker supplies meta-decisions or visual attention priors, and a dedicated Executor translates them into fine-grained control. ReAL-AD~\cite{lu2025real} extends this to a full three-layer hierarchy: strategy, decision, and operation, where VLM-derived situational insights shape progressively refined planning commands.

\textbf{Waypoint Supervision} is an explicit guidance that uses VLMs to generate coarse trajectory waypoints, which the fast planning module refines into dense, executable trajectories. DriveVLM~\cite{tian2024drivevlm} adopts a hierarchical reasoning-to-planning pipeline: the VLM produces meta-actions and coarse waypoints through chain-of-thought reasoning, and conventional planners transform them into detailed trajectories.

SOLVE~\cite{chen2025solve} strengthens VLM-planner coordination through a shared vision encoder and a Trajectory Chain-of-Thought module that iteratively refines candidate waypoints before final selection by the E2E planner. These designs provide a tighter numerical interface between reasoning and control, enabling VLMs to influence planning while retaining stability through classical refinement.

Overall, explicit guidance approaches maintain strong interpretability and grant VLM a direct role in decision-making. However, they remain sensitive to the accuracy and consistency of VLM outputs; misaligned or ambiguous commands can propagate downstream and degrade planning safety.

\subsubsection{Implicit Representations Transfer}
\label{sec:implicitrepresentations}
Implicit feature constraint refers to methods where the VLM acts as a teacher or auxiliary module during training, transferring reasoning ability or cognitive priors as latent features to the compact E2E network. These approaches fall into two main groups: \textbf{knowledge distillation} and \textbf{multimodal feature fusion}.

\textbf{Knowledge Distillation}-based approaches transfer VLM-generated explanations, reasoning traces, or structured action semantics into the latent space of the E2E driving model. VLP~\cite{pan2024vlp} aligns BEV features and planning queries with pretrained language embeddings using contrastive and supervisory objectives, enabling planners to inherit commonsense scene understanding. VLM-AD~\cite{xu2024vlm} generates free-form textual justifications and structured behavior labels using a VLM, distilling them into the planner through an alignment head and an action classification head. This dual-supervision design helps the E2E module acquire richer semantic representations while remaining computationally light during deployment.

More comprehensive alignment is seen in VERDI~\cite{feng2025verdi}, which aligns perception, prediction, and planning outputs with VLM-generated chain-of-thought explanations, injecting structured reasoning across all stages of the pipeline. ALN-P3~\cite{ma2025aln} extends this principle with full-stack co-distillation: perception tokens, predicted motions, and planned trajectories are jointly aligned with VLM reasoning to unify cognition and execution.

\textbf{Multimodal Feature Fusion}-based approaches directly integrate VLM-derived features into the fast planner. InsightDrive~\cite{song2025insightdrive} introduces language-guided scene representations, where VLM-generated descriptions highlight critical regions and modulate BEV features via cross-attention. VLM-E2E~\cite{liu2025vlm} explicitly models driver attention by fusing textual attention cues with BEV features through a learnable gating mechanism. Beyond attention cues, NetRoller~\cite{netroller} extracts latent reasoning variables from VLMs and adapts them into compact features suitable for real-time planners. ReCogDrive~\cite{li2025recogdrive} aligns linguistic priors with a diffusion-based planner, refining trajectories through reinforcement learning to promote safety and human-like behavior. ETA~\cite{hamdan2025eta} focuses on efficiency: VLM reasoning is computed asynchronously in earlier frames and fused into current features using an action-mask mechanism, ensuring guidance without incurring high real-time costs.

Implicit transfer methods reduce inference cost and avoid dependence on large VLMs at runtime, but they may sacrifice interpretability, and excessive distillation can oversimplify reasoning signals. Their effectiveness also depends strongly on how well the distilled or fused features align with the capacity of the fast driving module.

\begin{table}[t]
\centering
\captionsetup{justification=raggedright, singlelinecheck=false}
\caption{Summary of existing \textbf{Datasets \& Benchmarks} for training and evaluating the VA and VLA models.
    \\
    $\bullet$ \textbf{Vision Sensor Inputs}: \raisebox{-0.5ex}{\includegraphics[width=0.02\linewidth]{figures/icons/rgb.png}} Camera, \raisebox{-0.5ex}{\includegraphics[width=0.02\linewidth]{figures/icons/lidar.png}} LiDAR point cloud, \raisebox{-0.5ex}{\includegraphics[width=0.02\linewidth]{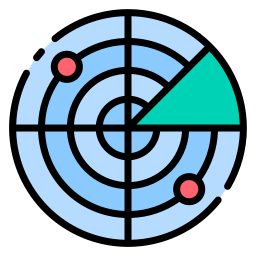}} RADAR point cloud, and \raisebox{-0.5ex}{\includegraphics[width=0.02\linewidth]{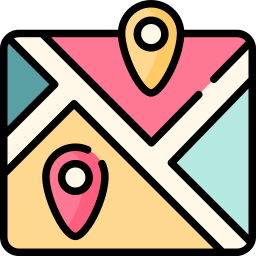}} Map.
    \\
    $\bullet$ \textbf{Vision Types}: \underline{Real}: Data collected from real driving scenes, and \underline{Sim}: Data collected from simulator. 
    \\
    $\bullet$ \textbf{Language Annotation Types}: \underline{A}: Automatic labeling process, and \underline{M}: Manual labeling process.
    \\
    $\bullet$ \textbf{Action Types}: \underline{Traj.}: Numerical trajectory output, and \underline{Ctrl}: Control signal output. 
    \\
    $\bullet$ \textbf{Action Metrics}: \underline{Open}: Open-loop Evaluation, \underline{CL}: Closed-loop Evaluation, and \underline{Lang.}: Language-based Evaluation.}
\vspace{-0.2cm}
\resizebox{\linewidth}{!}{
\begin{tabular}{llccccccccc}
    \toprule
    \multirow{2.5}{*}{\textbf{Dataset}} & \multirow{2.5}{*}{\textbf{Year}} & \multicolumn{3}{c}{\textbf{Vision} \raisebox{-0.25ex}{\includegraphics[width=0.022\linewidth]{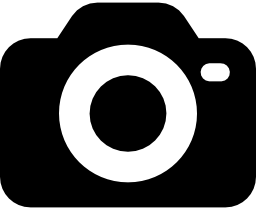}}} & \multicolumn{3}{c}{\textbf{Language} \raisebox{-0.5ex}{\includegraphics[width=0.021\linewidth]{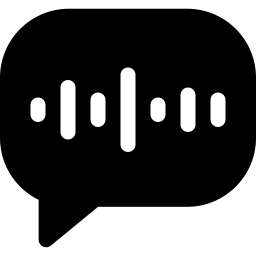}}} & \multicolumn{2}{c}{\textbf{Action \raisebox{-0.5ex}{\includegraphics[width=0.021\linewidth]{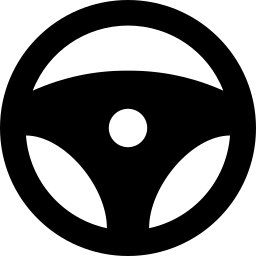}}}} & \multirow{2.5}{*}{\textbf{Other Tasks}} 
    \\
    \cmidrule(lr){3-5} \cmidrule(lr){6-8} \cmidrule(lr){9-10} & & \textbf{Sensor} & \textbf{Type} & \textbf{Scale} & \textbf{Category} & \textbf{Label} & \textbf{Scale} & \textbf{Type} & \textbf{Metric} & 
    \\
    \midrule
    \midrule
    \rowcolor{vla_purple!15}\multicolumn{11}{l}{\textcolor{vla_purple}{$\bullet$~\textbf{Vision-Action Datasets}}}
    \\ 
    \midrule
    \raisebox{0.2ex}{\BDD}~BDD100K \cite{yu2020bdd100k} & 2020 & \raisebox{-0.5ex}{\includegraphics[width=0.02\linewidth]{figures/icons/rgb.png}} & Real & 120M & - & - & - & Traj. & - & Percept.
    \\
    \rowcolor{gray!7}\raisebox{0.2ex}{\nuScenes}~nuScenes \cite{caesar2020nuscenes} & 2020 & \raisebox{-0.5ex}{\includegraphics[width=0.02\linewidth]{figures/icons/rgb.png}} \raisebox{-0.5ex}{\includegraphics[width=0.02\linewidth]{figures/icons/lidar.png}} \raisebox{-0.5ex}{\includegraphics[width=0.02\linewidth]{figures/icons/radar.png}} & Real & 1.4M & - & - & - & Traj. & Open & Percept.
    \\
    \raisebox{0.2ex}{\Waymo}~Waymo\cite{sun2020waymo} & 2020 & \raisebox{-0.5ex}{\includegraphics[width=0.02\linewidth]{figures/icons/rgb.png}} \raisebox{-0.5ex}{\includegraphics[width=0.02\linewidth]{figures/icons/lidar.png}} \raisebox{-0.5ex}{\includegraphics[width=0.02\linewidth]{figures/icons/radar.png}} & Real & 200M & - & - & - & Traj. & Open &  Percept., Forecast.
    \\
    \rowcolor{gray!7}\raisebox{0.2ex}{\nuPlan}~nuPlan \cite{caesar2021nuplan} & 2021 &\raisebox{-0.5ex}{\includegraphics[width=0.02\linewidth]{figures/icons/rgb.png}} \raisebox{-0.5ex}{\includegraphics[width=0.02\linewidth]{figures/icons/lidar.png}} \raisebox{-0.5ex}{\includegraphics[width=0.02\linewidth]{figures/icons/map.png}} & Real & 4.6M & - & - & - & Ctrl, Traj. & CL & Forecast.
    \\
    \raisebox{0.2ex}{\Argoverse}~Argoverse 2 \cite{wilson2023argoverse} & 2021 &\raisebox{-0.5ex}{\includegraphics[width=0.02\linewidth]{figures/icons/rgb.png}} \raisebox{-0.5ex}{\includegraphics[width=0.02\linewidth]{figures/icons/lidar.png}} \raisebox{-0.5ex}{\includegraphics[width=0.02\linewidth]{figures/icons/map.png}} & Real & 300K & - & - & - & Traj. & Open & Percept., Forecast.
    \\
    \rowcolor{gray!7}\raisebox{0.2ex}{\BenchDrive}~Bench2Drive \cite{jia2024bench2drive} & 2024 & \raisebox{-0.5ex}{\includegraphics[width=0.02\linewidth]{figures/icons/rgb.png}} \raisebox{-0.5ex}{\includegraphics[width=0.02\linewidth]{figures/icons/lidar.png}} \raisebox{-0.5ex}{\includegraphics[width=0.02\linewidth]{figures/icons/map.png}} & Sim & 2M & - & - & - & Traj. & CL & -
    \\
    \raisebox{0.2ex}{\RoboBEV}~RoboBEV \cite{xie2025benchmarking} & 2025 & \raisebox{-0.5ex}{\includegraphics[width=0.02\linewidth]{figures/icons/rgb.png}} \raisebox{-0.5ex}{\includegraphics[width=0.02\linewidth]{figures/icons/lidar.png}} \raisebox{-0.5ex}{\includegraphics[width=0.02\linewidth]{figures/icons/radar.png}} & Real &866K  & - & - & - & Traj. & Open & Percept.
    \\
    \rowcolor{gray!7}\raisebox{0.2ex}{\WODEE} WOD-E2E~\cite{xu2025wod}&2025&\raisebox{-0.5ex}{\includegraphics[width=0.02\linewidth]{figures/icons/rgb.png}} & Real & 800K & - & - & - & Traj. & Open & -
    \\
    \midrule
    \rowcolor{vla_green!15}\multicolumn{11}{l}{\textcolor{vla_green}{$\bullet$~\textbf{Vision-Language-Action Datasets}}} 
    \\
    \midrule
    \raisebox{0.2ex}{\BDDX}~BDD-X \cite{kim2018textual} & 2018 & \raisebox{-0.5ex}{\includegraphics[width=0.02\linewidth]{figures/icons/rgb.png}} & Real & 8.4M & Caption & M & 26K & Ctrl & Open & Reason.
    \\
    \rowcolor{gray!7}\raisebox{0.2ex}{\TalkCar}~Talk2Car \cite{deruyttere2022talk2car} & 2022 & \raisebox{-0.5ex}{\includegraphics[width=0.02\linewidth]{figures/icons/rgb.png}} \raisebox{-0.5ex}{\includegraphics[width=0.02\linewidth]{figures/icons/lidar.png}} \raisebox{-0.5ex}{\includegraphics[width=0.02\linewidth]{figures/icons/radar.png}} & Real & 400K & Caption & M & 12K & Ctrl, Traj.& Open & Ground.
    \\
    \raisebox{0.2ex}{\SDN}~SDN~\cite{ma2022dorothie} &2022&\raisebox{-0.5ex}{\includegraphics[width=0.02\linewidth]{figures/icons/rgb.png}}&Sim& -&Instruction, QA& A+M&8.4K&Ctrl, Traj.&CL&QA 
    \\
    \rowcolor{gray!7}\raisebox{0.2ex}{\DriveMLM}~DriveMLM \cite{wang2023drivemlm} & 2023 & \raisebox{-0.5ex}{\includegraphics[width=0.02\linewidth]{figures/icons/rgb.png}} \raisebox{-0.5ex}{\includegraphics[width=0.02\linewidth]{figures/icons/lidar.png}} & Sim & - & Reason., Deci. & A+M & - & Ctrl, Traj.& CL, Lang. & Reason., QA
    \\
    \raisebox{0.2ex}{\LMDrive}~LMDrive \cite{shao2024lmdrive} & 2024 & \raisebox{-0.5ex}{\includegraphics[width=0.02\linewidth]{figures/icons/rgb.png}} \raisebox{-0.5ex}{\includegraphics[width=0.02\linewidth]{figures/icons/lidar.png}} & Sim & 3M & Instruction & A+M & 64K & Traj. & CL & -
    \\
    \rowcolor{gray!7}\raisebox{0.2ex}{\BDD}~DriveLM-N \cite{sima2024drivelm} & 2024 & \raisebox{-0.5ex}{\includegraphics[width=0.02\linewidth]{figures/icons/rgb.png}} \raisebox{-0.5ex}{\includegraphics[width=0.02\linewidth]{figures/icons/lidar.png}} \raisebox{-0.5ex}{\includegraphics[width=0.02\linewidth]{figures/icons/radar.png}} & Real & 4.8K & QA & M & 445K & Ctrl, Traj. & Open & Reason., QA
    \\
    \raisebox{0.2ex}{\BDD}~DriveLM-C \cite{sima2024drivelm} & 2024 & \raisebox{-0.5ex}{\includegraphics[width=0.02\linewidth]{figures/icons/rgb.png}} \raisebox{-0.5ex}{\includegraphics[width=0.02\linewidth]{figures/icons/lidar.png}} \raisebox{-0.5ex}{\includegraphics[width=0.02\linewidth]{figures/icons/radar.png}} & Sim & 64K & QA & A & 3.76M & Ctrl, Traj. & Open & Reason., QA 
    \\
    \rowcolor{gray!7}\raisebox{0.2ex}{\HBD}~HBD~\cite{han2025dme}&2024&\raisebox{-0.5ex}{\includegraphics[width=0.02\linewidth]{figures/icons/rgb.png}} & Real,Sim & - & Deci., Descrip., QA&A+M&-&Traj.&Open&Descrip., QA
    \\
    \raisebox{0.2ex}{\VLAAD}~VLAAD~\cite{park2024vlaad}&2024& \raisebox{-0.5ex} {\includegraphics[width=0.02\linewidth]{figures/icons/rgb.png}} & Real & - & Reason., QA&A+M&64K&Ctrl.&Lang.&Caption, QA
    \\
    \rowcolor{gray!7}\raisebox{0.2ex}{\SUPAD}~SUP-AD~\cite{tian2024drivevlm} & 2024& \raisebox{-0.5ex}{\includegraphics[width=0.02\linewidth]{figures/icons/rgb.png}} &Real&-&Action, Reason., QA & A+M & -&Ctrl, Traj.& Open, Lang.&Reason, QA
    \\
    \raisebox{0.2ex}{\NuInstruct}~NuInstruct \cite{ding2024holistic} & 2024 & \raisebox{-0.5ex}{\includegraphics[width=0.02\linewidth]{figures/icons/rgb.png}} \raisebox{-0.5ex}{\includegraphics[width=0.02\linewidth]{figures/icons/lidar.png}} \raisebox{-0.5ex}{\includegraphics[width=0.02\linewidth]{figures/icons/radar.png}} & Real & 11.8K & Instruction & A & 91K & Ctrl & Lang. & Reason.
    \\
    \rowcolor{gray!7}\raisebox{0.2ex}{\WOMDR}~WOMD-Reason \cite{li2024womd} & 2024 & \raisebox{-0.5ex}{\includegraphics[width=0.02\linewidth]{figures/icons/rgb.png}} & Real & 63K & QA & A & 2940K & Plan. & Lang. & Reason., QA
    \\
    \raisebox{0.2ex}{\DriveCoT}~DriveCoT \cite{wang2024drivecot} & 2024 &\raisebox{-0.5ex}{\includegraphics[width=0.02\linewidth]{figures/icons/rgb.png}} \raisebox{-0.5ex}{\includegraphics[width=0.02\linewidth]{figures/icons/lidar.png}} & Sim & - & CoT, Deci. & A & 36K & Ctrl & Open & Reason. 
    \\
    \rowcolor{gray!7}\raisebox{0.2ex}{\ReasonDrive}~Reason2Drive \cite{nie2024reason2drive} & 2024 & \raisebox{-0.5ex}{\includegraphics[width=0.02\linewidth]{figures/icons/rgb.png}} \raisebox{-0.5ex}{\includegraphics[width=0.02\linewidth]{figures/icons/lidar.png}} \raisebox{-0.5ex}{\includegraphics[width=0.02\linewidth]{figures/icons/radar.png}} & Real & - & Reason., QA & A & 632K& Ctrl, Traj. & Open & Reason., QA 
    \\
    \raisebox{0.2ex}{\DriveBench}~DriveBench \cite{xie2025vlms} & 2025 & \raisebox{-0.5ex}{\includegraphics[width=0.02\linewidth]{figures/icons/rgb.png}} \raisebox{-0.5ex}{\includegraphics[width=0.02\linewidth]{figures/icons/lidar.png}} \raisebox{-0.5ex}{\includegraphics[width=0.02\linewidth]{figures/icons/radar.png}} & Real & 19.2K & QA & A+M & 20.5K & Ctrl & Lang. & QA
    \\
    \rowcolor{gray!7}\raisebox{0.2ex}{\MetaAD}~MetaAD~\cite{jiang2025alphadrive}&2025& \raisebox{-0.5ex}{\includegraphics[width=0.02\linewidth]{figures/icons/rgb.png}}&Real&120K& Reason., Plan, QA&-& 30K & Ctrl & Lang. & Reason. 
    \\
    \raisebox{0.2ex}{\OmniDrive}~OmniDrive \cite{wang2025omnidrive} & 2025 & \raisebox{-0.5ex}{\includegraphics[width=0.02\linewidth]{figures/icons/rgb.png}} \raisebox{-0.5ex}{\includegraphics[width=0.02\linewidth]{figures/icons/lidar.png}} \raisebox{-0.5ex}{\includegraphics[width=0.02\linewidth]{figures/icons/radar.png}} & Real & - & Reason., QA & A & - & Ctrl, Traj. & Open & Reason.
    \\
    \rowcolor{gray!7}\raisebox{0.2ex}{\NuInteract}~NuInteract \cite{zhao2025extending} & 2025 & \raisebox{-0.5ex}{\includegraphics[width=0.02\linewidth]{figures/icons/rgb.png}} \raisebox{-0.5ex}{\includegraphics[width=0.02\linewidth]{figures/icons/lidar.png}} \raisebox{-0.5ex}{\includegraphics[width=0.02\linewidth]{figures/icons/radar.png}} & Real & 34K & Caption, QA & A & 1.5M & Ctrl & Lang. & Percept., Ground. 
    \\
    \raisebox{0.2ex}{\DriveAction}~DriveAction \cite{hao2025driveaction} & 2025 & \raisebox{-0.5ex}{\includegraphics[width=0.02\linewidth]{figures/icons/rgb.png}} & Real & 2.6K & QA & A & 16.18K & Ctrl & Lang. & -
    \\
    \rowcolor{gray!7}\raisebox{0.2ex}{\ImpromptuVLA}~ImpromptuVLA~\cite{chi2025impromptu} & 2025 & \raisebox{-0.5ex}{\includegraphics[width=0.02\linewidth]{figures/icons/rgb.png}} & Real,Sim & 2M & Instruction, QA & A+M & 80K & Ctrl, Traj. & Open, CL & QA
    \\
    \raisebox{0.2ex}{\CoVLA}~CoVLA~\cite{arai2025covla} & 2025 & \raisebox{-0.5ex}{\includegraphics[width=0.02\linewidth]{figures/icons/rgb.png}} & Real & 6M & Caption & A & 6M & Traj. & Open & - 
    \\
    \rowcolor{gray!7}\raisebox{0.2ex}{\OmniReasonN}~OmniReason-N \cite{liu2025omnireason} & 2025 & \raisebox{-0.5ex}{\includegraphics[width=0.02\linewidth]{figures/icons/rgb.png}} \raisebox{-0.5ex}{\includegraphics[width=0.02\linewidth]{figures/icons/lidar.png}} \raisebox{-0.5ex}{\includegraphics[width=0.02\linewidth]{figures/icons/radar.png}} & Real & - & QA & A & - & Ctrl, Traj. & Open & Reason., QA
    \\
    \raisebox{0.2ex}{\OmniReasonB}~OmniReason-B2D \cite{liu2025omnireason} & 2025 & \raisebox{-0.5ex}{\includegraphics[width=0.02\linewidth]{figures/icons/rgb.png}} & Sim & - & QA & A & - & Ctrl, Traj. & Open & Reason., QA
    \\
    \bottomrule
\end{tabular}}
\label{tab:e2e_dataset}
\end{table}
\section{Datasets \& Benchmark}
\label{sec:datasets_benchmark_study}

Standardized datasets and benchmarks form the empirical foundation of VLA research, supporting model development, training, and evaluation. Since VLA driving systems integrate perception, language, and action, VLA datasets exhibit substantial diversity in modality composition, annotation granularity, and task definitions. Accordingly, evaluation protocols vary substantially, encompassing conventional trajectory-based metrics, language-centric assessments, and interactive closed-loop evaluations.

\subsection{Datasets for VLA in Autonomous Driving}
\label{sec:datasets}
Traditionally, VA datasets provide rich sensory observations (cameras, LiDAR, RADAR) paired with control actions, enabling end-to-end mapping from images to trajectories \cite{caesar2020nuscenes,caesar2021nuplan,yu2020bdd100k}. These datasets underpin the development of early IL/RL-based VA models.

As language becomes an increasingly important modality for reasoning, instruction following, and explainability, VLA datasets have emerged~\cite{wang2023drivemlm,sima2024drivelm,tian2024drivevlm,wang2024drivecot}. These datasets extend traditional driving logs with textual instructions, question-answer pairs, or rationales aligned with visual observations and expert actions \cite{chi2025impromptu,arai2025covla}.
In general, a dataset is considered VLA-compatible when it provides temporally or semantically aligned language annotations that connect visual observations with actions or trajectories, enabling tri-modal learning. The summarized collections are provided in Table~\ref{tab:e2e_dataset}.

\subsubsection{Vision-Action Datasets}
Originally, BDD100K~\cite{yu2020bdd100k} provides 100K diverse driving videos from across the United States, covering a wide spectrum of weather, lighting, and traffic conditions, making it a foundational dataset for behavioral cloning and end-to-end driving. Later, nuScenes~\cite{caesar2020nuscenes} offers 1,000 multi-sensor driving scenes with synchronized 6-camera surround views, LiDAR sweeps, radar, 3D boxes, and motion trajectories, supporting both perception tasks and multi-agent motion forecasting. Larger-scale datasets such as the Waymo Open Dataset~\cite{ettinger2021large} and Argoverse 2~\cite{wilson2023argoverse} further extend this paradigm with higher-resolution sensors, longer trajectories, and detailed HD maps, enabling robust training of perception-to-prediction pipelines in diverse urban settings. Complementing these efforts, nuPlan~\cite{caesar2021nuplan} incorporates long-horizon ego trajectories, dense map context, and simulation interfaces for closed-loop testing, providing comprehensive supervision for evaluating decision-making and planning under complex, real-world conditions.
While lacking explicit language supervision, these VA datasets establish the visual–action foundation for VLA development by providing structured supervision that links visual perception, temporal dynamics, and expert decision-making.

\subsubsection{Vision-Language-Action Datasets}
Building upon the visual–action foundation established by VA datasets, VLA datasets enrich driving logs with structured or free-form natural language to support joint perception-language-action learning.

\clearpage\clearpage
\begin{table}[ht]
\captionsetup{justification=raggedright, singlelinecheck=false}
\centering
\caption{
Comparisons of state-of-the-art models for \textbf{Open-Loop Planning} on the nuScenes \cite{caesar2020nuscenes} benchmark. 
\\
$\bullet$ \textbf{Input}: \raisebox{-0.5ex}{\includegraphics[width=0.02\linewidth]{figures/icons/rgb.png}}:Camera, \raisebox{-0.5ex}{\includegraphics[width=0.02\linewidth]{figures/icons/lidar.png}}:LiDAR, \raisebox{-0.5ex}{\includegraphics[width=0.02\linewidth]{figures/icons/prompt.png}}: Prompt, \raisebox{-0.5ex}{\includegraphics[width=0.02\linewidth]{figures/icons/command.png}}: Instruct., \raisebox{-0.5ex}{\includegraphics[width=0.02\linewidth]{figures/icons/scene.png}}: Scene Descrip., \raisebox{-0.5ex}{\includegraphics[width=0.02\linewidth]{figures/icons/status.png}}: Status, \raisebox{-0.5ex}{\includegraphics[width=0.02\linewidth]{figures/icons/traffic_rule.png}}: Rule, \raisebox{-0.5ex}{\includegraphics[width=0.02\linewidth]{figures/icons/context.png}}: Context.
\\
$\bullet$ \textbf{Action}: \underline{LH}: Language Head, \underline{RL}: Policy w/ Reinforcement Learning, \underline{REG}: Decoder + MLP, \underline{SEL}: Trajectory Selection w/ Cost, and \underline{GEN}: Trajectory Generation w/ Generative Model.
\\
$\bullet$ \textbf{Evaluation Metrics}: \underline{L2 ($\downarrow$)}: L2 Error in meters, and \underline{CR ($\downarrow$)}: Collision Rate.
}
\vspace{-0.2cm}
\footnotesize
\resizebox{\textwidth}{!}{
\begin{tabular}{rcclll|cccc|cccc}
    \toprule
    \multirow{2}{*}{\textbf{Model}} & \multirow{2}{*}{\textbf{Year}} & \multirow{2}{*}{\textbf{Input}} & \multirow{2}{*}{\textbf{Vision} \raisebox{-0.25ex}{\includegraphics[width=0.018\linewidth]{figures/icons/vision.png}}} & \multirow{2}{*}{\textbf{Language} \raisebox{-0.5ex}{\includegraphics[width=0.017\linewidth]{figures/icons/language.png}}} & \multirow{2}{*}{\textbf{Action} \raisebox{-0.5ex}{\includegraphics[width=0.017\linewidth]{figures/icons/action.png}}} & \multicolumn{4}{c|}{\textbf{L2 $\downarrow$}} & \multicolumn{4}{c}{\textbf{CR $\downarrow$}} 
    \\
    & & & & & & \textbf{1s} & \textbf{2s} & \textbf{3s} & \textbf{Avg.} & \textbf{1s} & \textbf{2s} & \textbf{3s} & \textbf{Avg.} 
    \\
    \midrule\midrule
    \rowcolor{vla_purple!15}\multicolumn{14}{l}{\textcolor{vla_purple}{$\bullet$~\textbf{Vision-Action Models}}} 
    \\
    \midrule
    ST-P3 \cite{hu2022st} & 2022 & \raisebox{-0.5ex}{\includegraphics[width=0.02\linewidth]{figures/icons/rgb.png}} & EfficientNet & - & REG & 1.33 & 2.11 & 2.90 & 2.11 & 0.23 & 0.62 & 1.27 & 0.71 
    \\
    \rowcolor{gray!7}UniAD \cite{hu2023planning} & 2022 & \raisebox{-0.5ex}{\includegraphics[width=0.02\linewidth]{figures/icons/rgb.png}} & ResNet & - & REG & 0.44 & 0.67 & 0.96 & 0.69 & 0.04 & 0.08 & 0.23 & 0.12
    \\
    VAD \cite{jiang2023vad} & 2023 & \raisebox{-0.5ex}{\includegraphics[width=0.02\linewidth]{figures/icons/rgb.png}} & ResNet & - & REG & 0.17 & 0.34& 0.60&0.37 & 0.07 & 0.10 & 0.24 &0.14
    \\
    \rowcolor{gray!7}OccNet \cite{tong2023scene} & 2023 & \raisebox{-0.5ex}{\includegraphics[width=0.02\linewidth]{figures/icons/rgb.png}} & ResNet & - & SEL & 1.29 & 2.13& 2.99&2.14 & 0.21 & 0.59 & 1.37 &0.72
    \\
\bottomrule
\end{tabular}}
\label{tab:nuscenes_compare}
\end{table}

\clearpage\clearpage
Representative examples include BDD-X~\cite{kim2018textual}, which extends BDD100K with time-aligned human rationales, where annotators describe why the driver performed a specific action. This dataset provides early grounding for language-based explanations. DriveLM~\cite{sima2024drivelm} constructs graph-structured question-answer pairs based on nuScenes and CARLA scenarios. These QA pairs target conditional reasoning, enabling models to infer high-level intent, spatial relations, and driving decisions. Impromptu VLA~\cite{chi2025impromptu} aggregates data from eight public driving datasets and supplements them with captions, instructions, and QA pairs aligned with expert trajectories. The focus is on corner cases and long-tailed events.
Other datasets, such as LingoQA~\cite{marcu2024lingoqa} and CoVLA~\cite{arai2025covla}, collect real-world driving videos paired with natural language QA or behavior descriptions, emphasizing spatiotemporal reasoning and human-understandable driving motivations.

Notably, QA-style annotations have emerged as a dominant paradigm for extending driving datasets, serving as a common foundation for training and evaluating reasoning and planning capabilities~\cite{sima2024drivelm, hao2025driveaction}. However, the scope and assumptions embedded in such annotations naturally influence model behavior, motivating further exploration of more diverse perspectives, planning horizons, and evaluation protocols for real-world deployment.

\subsection{Evaluation Metrics}
\label{sec:metrics}
Evaluation metrics differ according to the model’s output modality: \textbf{trajectory-based} metrics for continuous action prediction and \textbf{text-based} metrics for models producing linguistic commands or rationales.

\subsubsection{Trajectory-Based Action Evaluation}
Trajectory-based outputs are typically evaluated in open-loop and closed-loop settings. 

\noindent{\textbf{Open-Loop Evaluation.}} The predicted trajectory is directly compared against expert trajectories without executing in a simulator. Metrics such as L2 error and collision rate~\cite{hu2022st}, along with trajectory-based indicators including Average Displacement Error (ADE), Final Displacement Error (FDE), and Miss Rate (MR)~\cite{hu2023planning}, are widely used. These metrics regard human driving demonstrations as the ground truth and formulate planning essentially as an imitation learning task. By measuring the deviation between predicted and expert trajectories, they provide a straightforward way to assess the accuracy of motion prediction.

\noindent{\textbf{Closed-Loop Evaluation.}} Instead, it measures the model’s performance when interacting with a simulation environment (\emph{e.g.}, CARLA~\cite{dosovitskiy2017carla}). Representative metrics include route completion (RC), driving score (DS), and infraction distance (ID). Bench2Drive~\cite{jia2024bench2drive} further considers success rate, efficiency, and comfort. NAVSIM~\cite{dauner2024navsim}, built on nuPlan~\cite{caesar2021nuplan}, introduces the Predictive Driver Model Score (PDMS), which aggregates subscores for ego progress (EP), time-to-collision (TTC), and comfort (C), while applying penalties on collisions (NC) and driving admissibility (DAC). These metrics provide a holistic view of the safety, feasibility, and deployability of planning actions.

\subsubsection{Text-Based Action Evaluation}
For low-level vehicle control expressed in natural language, evaluation covers both linguistic quality and control effectiveness. Standard text metrics, such as BLEU, ROUGE, and CIDEr, are commonly used to assess the quality of generated language~\cite{zhao2025extending,sima2024drivelm}, which measures n-gram overlap with human-annotated reference commands. Beyond command accuracy, reasoning quality is assessed through rationale consistency~\cite{nie2024reason2drive} and human preference ratings of language explanations, particularly in datasets following the BDD-X~\cite{kim2018textual} format. To assess driving applications, execution-based metrics are introduced for behavior assessment. SimLingo~\cite{renz2025simlingo} introduces an action-dreaming benchmark. The corresponding actions are mapped from the input instruction, which is open-loop evaluated using the success rate. 

Regardless of output modality, these benchmarks emphasize key aspects of action quality, including accuracy, executability, safety, and intention alignment.

\begin{table}[t]
\captionsetup{justification=raggedright, singlelinecheck=false}
\caption{Comparisons of state-of-the-art models on the WOD-E2E \cite{xu2025wod} test split.
\\
$\bullet$ \textbf{Input}: \raisebox{-0.5ex}{\includegraphics[width=0.02\linewidth]{figures/icons/rgb.png}}:Camera, \raisebox{-0.5ex}{\includegraphics[width=0.02\linewidth]{figures/icons/prompt.png}}: Sys. Prompt, \raisebox{-0.5ex}{\includegraphics[width=0.02\linewidth]{figures/icons/command.png}}: Instruct., \raisebox{-0.5ex}{\includegraphics[width=0.02\linewidth]{figures/icons/scene.png}}: Scene Descrip., \raisebox{-0.5ex}{\includegraphics[width=0.02\linewidth]{figures/icons/status.png}}: Status, \raisebox{-0.5ex}{\includegraphics[width=0.02\linewidth]{figures/icons/traffic_rule.png}}: Traffic Rule, \raisebox{-0.5ex}{\includegraphics[width=0.02\linewidth]{figures/icons/context.png}}: Context Info.
\\
$\bullet$ \textbf{Action}: \underline{LH}: Language Head, \underline{RL}: Policy with Reinforcement Learning, \underline{REG}: Decoder + MLP, and \underline{GEN}: Trajectory Generation w/ Generative Model.
\\
$\bullet$ \textbf{Evaluation Metrics}: 
\underline{RFS (Overall/Spotlight)} ($\downarrow$): Rater Feedback Score,
\underline{ADE 5s/3s} ($\downarrow$): Average Displacement Error.
}
\label{tab:wod_e2e_results}
\vspace{-0.2cm}
\resizebox{\textwidth}{!}{
\begin{tabular}{rcclll|cccc}
    \toprule
    \textbf{Model} & \textbf{Year} & \textbf{Input} & \textbf{Vision \raisebox{-0.25ex}{\includegraphics[width=0.022\linewidth]{figures/icons/vision.png}}} & \textbf{Language \raisebox{-0.5ex}{\includegraphics[width=0.021\linewidth]{figures/icons/language.png}}} & \textbf{Action \raisebox{-0.5ex}{\includegraphics[width=0.021\linewidth]{figures/icons/action.png}}} & \textbf{RFS(Overall)($\uparrow$)} & \textbf{RFS(Spotlight)($\uparrow$)} & \textbf{ADE 5s($\downarrow$)} & \textbf{ADE 3s($\downarrow$)} 
    \\
    \midrule\midrule
    \rowcolor{vla_purple!15}\multicolumn{10}{l}{\textcolor{vla_purple}{$\bullet$~\textbf{Vision-Action Models}}} 
    \\
    \midrule

    Waymo Baseline & 2025 & \raisebox{-0.5ex}{\includegraphics[width=0.02\linewidth]{figures/icons/rgb.png}} & - & - & - & 7.53 & 6.60 & 3.02 & 1.32 
    \\

\rowcolor{gray!7}Swin-Trajectory~\cite{xu2025wod} & 2025 & \raisebox{-0.5ex}{\includegraphics[width=0.02\linewidth]{figures/icons/rgb.png}} & SwinT & - & REG & 7.54 & 6.68 & 2.81 & 1.21 \\
DiffusionDrive~\cite{liao2025diffusiondrive} & 2025 & \raisebox{-0.5ex}{\includegraphics[width=0.02\linewidth]{figures/icons/rgb.png}} &  ResNet & - & GEN & 7.69 & 6.65 & 2.99 & 1.31 \\


\rowcolor{gray!7}RAP-DINO~\cite{feng2025rap}& 2025 & \raisebox{-0.5ex}{\includegraphics[width=0.02\linewidth]{figures/icons/rgb.png}} & DINO & - & REG & 8.04&7.20&2.65&1.17\\
    \midrule
    \rowcolor{vla_green!15}\multicolumn{10}{l}{\textcolor{vla_green}{$\bullet$~\textbf{Vision-Language-Action Models}}} 
    \\
    \midrule
    OpenEMMA~\cite{xing2025openemma}&2025&\raisebox{-0.5ex}{\includegraphics[width=0.02\linewidth]{figures/icons/rgb.png}}~\raisebox{-0.5ex}{\includegraphics[width=0.02\linewidth]{figures/icons/prompt.png}} \raisebox{-0.5ex}{\includegraphics[width=0.02\linewidth]{figures/icons/status.png}} & \raisebox{-0.5ex}{\includegraphics[width=0.018\linewidth]{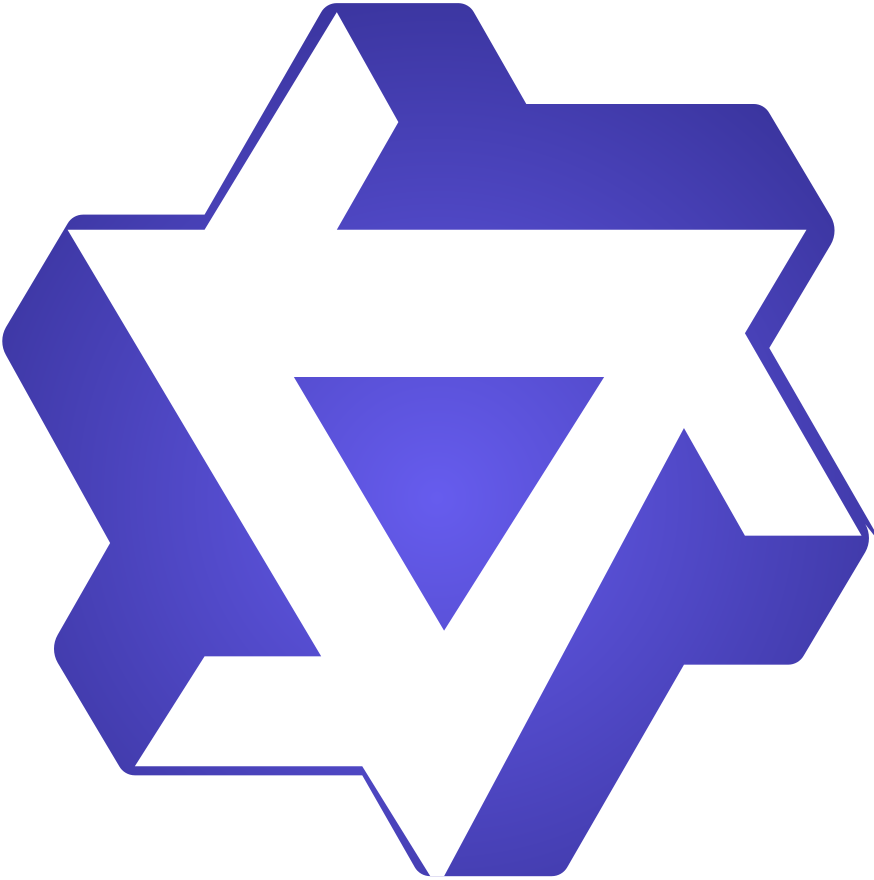}}~Qwen2-VL  & \raisebox{-0.5ex}{\includegraphics[width=0.018\linewidth]{figures/icons/qwen.png}}~Qwen2-VL & LH&5.16&4.71&12.74&6.68\\
    \rowcolor{gray!7}HMVLM~\cite{wang2025hmvlm} & 2025 & \raisebox{-0.5ex}{\includegraphics[width=0.02\linewidth]{figures/icons/rgb.png}} 
    \raisebox{-0.5ex}{\includegraphics[width=0.02\linewidth]{figures/icons/prompt.png}} 
    \raisebox{-0.5ex}{\includegraphics[width=0.02\linewidth]{figures/icons/command.png}} \raisebox{-0.5ex}{\includegraphics[width=0.02\linewidth]{figures/icons/status.png}}  & ViT & \raisebox{-0.5ex}{\includegraphics[width=0.018\linewidth]{figures/icons/qwen.png}}~Qwen2.5-VL & LH & 7.74 & 6.73 & 3.07 & 1.33 \\
    AutoVLA~\cite{zhou2025autovla} & 2025 & \raisebox{-0.5ex}{\includegraphics[width=0.02\linewidth]{figures/icons/rgb.png}} 
    \raisebox{-0.5ex}{\includegraphics[width=0.02\linewidth]{figures/icons/prompt.png}} 
    \raisebox{-0.5ex}{\includegraphics[width=0.02\linewidth]{figures/icons/command.png}} \raisebox{-0.5ex}{\includegraphics[width=0.02\linewidth]{figures/icons/status.png}}  & \raisebox{-0.5ex}{\includegraphics[width=0.018\linewidth]{figures/icons/qwen.png}}~Qwen2.5-VL & \raisebox{-0.5ex}{\includegraphics[width=0.018\linewidth]{figures/icons/qwen.png}}~Qwen2.5-VL & LH  & 7.56 & 6.94 & 2.96 & 1.35 \\
    \rowcolor{gray!7}Poutine~\cite{rowe2025poutine}& 2025 & \raisebox{-0.5ex}{\includegraphics[width=0.02\linewidth]{figures/icons/rgb.png}}~\raisebox{-0.5ex}{\includegraphics[width=0.02\linewidth]{figures/icons/prompt.png}}~\raisebox{-0.5ex}{\includegraphics[width=0.02\linewidth]{figures/icons/status.png}}~\raisebox{-0.5ex}{\includegraphics[width=0.02\linewidth]{figures/icons/command.png}}& ViT & \raisebox{-0.5ex}{\includegraphics[width=0.018\linewidth]{figures/icons/qwen.png}}~Qwen2.5-VL & LH & 7.99 & 6.89 & 2.74 & 1.21 \\
    LightEMMA~\cite{qiao2025lightemma}&2025&\raisebox{-0.5ex}{\includegraphics[width=0.02\linewidth]{figures/icons/rgb.png}}~\raisebox{-0.5ex}{\includegraphics[width=0.02\linewidth]{figures/icons/prompt.png}} &  \raisebox{-0.5ex}{\includegraphics[width=0.018\linewidth]{figures/icons/qwen.png}}~Qwen2.5-VL & \raisebox{-0.5ex}{\includegraphics[width=0.018\linewidth]{figures/icons/qwen.png}}~Qwen2.5-VL & LH&6.52&5.71&3.73&1.71\\
    \rowcolor{gray!7}dVLM-AD~\cite{ma2025dvlm}&2025&\raisebox{-0.5ex}{\includegraphics[width=0.02\linewidth]{figures/icons/rgb.png}}~\raisebox{-0.5ex}{\includegraphics[width=0.02\linewidth]{figures/icons/prompt.png}}~\raisebox{-0.5ex}{\includegraphics[width=0.02\linewidth]{figures/icons/status.png}}&SigLIP2&LLaDA-V&LH &7.63&-&3.02&1.29
    \\
    \bottomrule
\end{tabular}}
\end{table}
\begin{table}[t]
\captionsetup{justification=raggedright, singlelinecheck=false}
\caption{Comparisons of state-of-the-art models for \textbf{Open-Loop Planning} on the NAVSIM \cite{dauner2024navsim} \textit{navtest} benchmark.
\\
$\bullet$ \textbf{Input}: \raisebox{-0.5ex}{\includegraphics[width=0.02\linewidth]{figures/icons/rgb.png}}:Camera, \raisebox{-0.5ex}{\includegraphics[width=0.02\linewidth]{figures/icons/lidar.png}}: LiDAR, \raisebox{-0.5ex}{\includegraphics[width=0.02\linewidth]{figures/icons/prompt.png}}: Prompt, \raisebox{-0.5ex}{\includegraphics[width=0.02\linewidth]{figures/icons/command.png}}: Instruct., \raisebox{-0.5ex}{\includegraphics[width=0.02\linewidth]{figures/icons/scene.png}}: Scene Descrip., \raisebox{-0.5ex}{\includegraphics[width=0.02\linewidth]{figures/icons/status.png}}: Status, \raisebox{-0.5ex}{\includegraphics[width=0.02\linewidth]{figures/icons/traffic_rule.png}}: Rule, \raisebox{-0.5ex}{\includegraphics[width=0.02\linewidth]{figures/icons/context.png}}: Context.
\\
$\bullet$ \textbf{Action}: \underline{LH}: Language Head, \underline{RL}: Policy with Reinforcement Learning, \underline{REG}: Decoder + MLP, \underline{SEL}: Trajectory Selection w/ Cost, and \underline{GEN}: Trajectory Generation w/ Generative Model.
\\
$\bullet$ \textbf{Evaluation Metrics}: \underline{NC}: Navigation Completion,
\underline{DAC} ($\uparrow$): Driving Accuracy,
\underline{TTC} ($\uparrow$): Time-To-Collision,
\underline{Comf.} ($\uparrow$): Comfort,
\underline{EP} ($\uparrow$): Ego Progress, and
\underline{PDMS} ($\uparrow$): Perception Driving Metric Score.
}
\label{tab:navsim_compare}
\vspace{-0.2cm}
\resizebox{\textwidth}{!}{
\begin{tabular}{rcclll|cccccc}
    \toprule
    \textbf{Model} & \textbf{Year} & \textbf{Input} & \textbf{Vision \raisebox{-0.25ex}{\includegraphics[width=0.022\linewidth]{figures/icons/vision.png}}} & \textbf{Language \raisebox{-0.5ex}{\includegraphics[width=0.021\linewidth]{figures/icons/language.png}}} & \textbf{Action \raisebox{-0.5ex}{\includegraphics[width=0.021\linewidth]{figures/icons/action.png}}} & \textbf{NC($\uparrow$)} & \textbf{DAC($\uparrow$)} & \textbf{TTC($\uparrow$)} & \textbf{Comf($\uparrow$)} & \textbf{EP($\uparrow$)} & \textbf{PDMS($\uparrow$)} 
    \\
    \midrule\midrule
    \rowcolor{vla_purple!15}\multicolumn{12}{l}{\textcolor{vla_purple}{$\bullet$~\textbf{Vision-Action Models}}} 
    \\
    \midrule
    TransFuser \cite{chitta2022transfuser} & 2022 & \raisebox{-0.5ex}{\includegraphics[width=0.02\linewidth]{figures/icons/rgb.png}} \raisebox{-0.5ex}{\includegraphics[width=0.02\linewidth]{figures/icons/lidar.png}} & ResNet & - & REG & 97.7 & 92.8 & 92.8 & 100 & 79.2 & 84.0 
    \\
    \rowcolor{gray!7}UniAD~\cite{hu2023planning} & 2023 & \raisebox{-0.5ex}{\includegraphics[width=0.02\linewidth]{figures/icons/rgb.png}} & ResNet & - & REG & 97.8 & 91.9 & 92.9 & 100 & 78.8 & 83.4 
    \\
    VADv2 \cite{chen2024vadv2} & 2024 & \raisebox{-0.5ex}{\includegraphics[width=0.02\linewidth]{figures/icons/rgb.png}} & ResNet & - & REG &  97.2 & 89.1 & 91.6 & 100 & 76.0 & 80.9 
    \\
    \rowcolor{gray!7}PARA-Drive~\cite{weng2024drive} & 2024 & \raisebox{-0.5ex}{\includegraphics[width=0.02\linewidth]{figures/icons/rgb.png}} &ResNet&- & REG  & 97.9 & 92.4 & 93.0 & 99.8 & 79.3 & 84.0 
    \\
    LAW~\cite{li2024enhancing} & 2024 & \raisebox{-0.5ex}{\includegraphics[width=0.02\linewidth]{figures/icons/rgb.png}} &Swin-T&-&REG&96.4&95.4&88.7&99.9&81.7&84.6
    \\
    \rowcolor{gray!7}DRAMA~\cite{yuan2024drama} & 2024 & \raisebox{-0.5ex}{\includegraphics[width=0.02\linewidth]{figures/icons/rgb.png}} \raisebox{-0.5ex}{\includegraphics[width=0.02\linewidth]{figures/icons/lidar.png}} & ResNet & - & REG  & 98.0 & 93.1 & 94.8 & 100 & 80.1 & 85.5 
    \\
    DiffusionDrive~\cite{liao2025diffusiondrive}&2024& \raisebox{-0.5ex}{\includegraphics[width=0.02\linewidth]{figures/icons/rgb.png}} \raisebox{-0.5ex}{\includegraphics[width=0.02\linewidth]{figures/icons/lidar.png}} & ResNet & - & GEN & 98.2 & 96.2 & 94.7 & 100 & 82.2 & 88.1 
    \\
    \rowcolor{gray!7}WoTE~\cite{li2025end} & 2025 & \raisebox{-0.5ex}{\includegraphics[width=0.02\linewidth]{figures/icons/rgb.png}} \raisebox{-0.5ex}{\includegraphics[width=0.02\linewidth]{figures/icons/lidar.png}} &ResNet & - & SEL & 98.5 & 96.8 & 94.9 & 99.9 & 81.9 & 88.3 
    \\
    World4Drive~\cite{zheng2025world4drive}&2025& \raisebox{-0.5ex}{\includegraphics[width=0.02\linewidth]{figures/icons/rgb.png}} & ResNet & - & REG & 97.4 & 94.3 & 92.8 & 100 & 79.9 & 85.1
    \\
    \rowcolor{gray!7}DrivingGPT~\cite{chen2024drivinggpt} & 2025&\raisebox{-0.5ex}{\includegraphics[width=0.02\linewidth]{figures/icons/rgb.png}} & VQ-VAE & - & REG & 98.9 & 90.7 & 94.9 & 95.6 & 79.7 & 82.4
    \\
    AD-R1~\cite{yan2025ad}&2025&\raisebox{-0.5ex}{\includegraphics[width=0.02\linewidth]{figures/icons/rgb.png}} \raisebox{-0.5ex}{\includegraphics[width=0.02\linewidth]{figures/icons/lidar.png}} \raisebox{-0.5ex}{\includegraphics[width=0.02\linewidth]{figures/icons/status.png}}&-&-&RL&98.7&97.8&94.8&100&87.5&91.9
    \\
    \rowcolor{gray!7}SeerDrive~\cite{zhang2025future} & 2025&\raisebox{-0.5ex}{\includegraphics[width=0.02\linewidth]{figures/icons/rgb.png}} \raisebox{-0.5ex}{\includegraphics[width=0.02\linewidth]{figures/icons/lidar.png}} & VoVNet & - & SEL & 98.8 & 98.6 & 95.8 & 100 & 84.2 & 90.7
    \\
    Epona~\cite{zhang2025epona}&2025& \raisebox{-0.5ex}{\includegraphics[width=0.02\linewidth]{figures/icons/rgb.png}} \raisebox{-0.5ex}{\includegraphics[width=0.02\linewidth]{figures/icons/status.png}} & DC-AE & - & REG & 97.9 & 95.1 & 93.8 & 99.9 & 80.4 & 86.2
    \\
    \rowcolor{gray!7}GoalFlow~\cite{xing2025goalflow} & 2025&\raisebox{-0.5ex}{\includegraphics[width=0.02\linewidth]{figures/icons/rgb.png}} \raisebox{-0.5ex}{\includegraphics[width=0.02\linewidth]{figures/icons/lidar.png}} \raisebox{-0.5ex}{\includegraphics[width=0.02\linewidth]{figures/icons/status.png}} & VoVNet & - & GEN & 98.4 & 98.3 & 94.6 & 100 & 85.0 & 90.3
    \\
    TrajDiff~\cite{gui2025trajdiff}&2025& \raisebox{-0.5ex}{\includegraphics[width=0.02\linewidth]{figures/icons/rgb.png}} \raisebox{-0.5ex}{\includegraphics[width=0.02\linewidth]{figures/icons/lidar.png}} \raisebox{-0.5ex}{\includegraphics[width=0.02\linewidth]{figures/icons/status.png}} & ResNet & - & GEN & 98.1 & 97.0 & 94.3 & 100.0 & 82.7 & 88.5
    \\    
    \rowcolor{gray!7}DiffusionDriveV2~\cite{zou2025diffusiondrivev2}&2025& \raisebox{-0.5ex}{\includegraphics[width=0.02\linewidth]{figures/icons/rgb.png}} \raisebox{-0.5ex}{\includegraphics[width=0.02\linewidth]{figures/icons/lidar.png}} & ResNet & - & GEN & 98.3 & 97.9 & 94.8 & 99.9 & 87.5 & 91.2 
    \\    
    NaviHydra~\cite{wu2025navihydra}&2025& \raisebox{-0.5ex}{\includegraphics[width=0.02\linewidth]{figures/icons/rgb.png}} \raisebox{-0.5ex}{\includegraphics[width=0.02\linewidth]{figures/icons/lidar.png}} & ResNet & - & SEL & 98.7 & 98.6 & 88.7 & 96.2 & 100.0 & 92.7
    \\    
    \rowcolor{gray!7}Mimir~\cite{xing2025mimir}&2025& \raisebox{-0.5ex}{\includegraphics[width=0.02\linewidth]{figures/icons/rgb.png}} \raisebox{-0.5ex}{\includegraphics[width=0.02\linewidth]{figures/icons/lidar.png}} & ResNet & - & GEN & 98.2 & 97.5 & 94.6 & 100 & 83.6 & 89.3
    \\    
    \midrule
    \rowcolor{vla_green!15}\multicolumn{12}{l}{\textcolor{vla_green}{$\bullet$~\textbf{Vision-Language-Action Models}}} 
    \\
    \midrule
    ReCogDrive~\cite{li2025recogdrive} & 2025 &  \raisebox{-0.5ex}{\includegraphics[width=0.02\linewidth]{figures/icons/rgb.png}} \raisebox{-0.5ex}{\includegraphics[width=0.02\linewidth]{figures/icons/prompt.png}} \raisebox{-0.5ex}{\includegraphics[width=0.02\linewidth]{figures/icons/command.png}} & \raisebox{-0.5ex}{\includegraphics[width=0.019\linewidth]{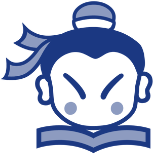}} InternViT  & \raisebox{-0.5ex}{\includegraphics[width=0.018\linewidth]{figures/icons/qwen.png}}~Qwen2.5-VL & GEN & 98.2 & 97.8 & 95.2 & 99.8 & 83.5 & 89.6
    \\
    \rowcolor{gray!7}AutoVLA ~\cite{zhou2025autovla} & 2025 & \raisebox{-0.5ex}{\includegraphics[width=0.02\linewidth]{figures/icons/rgb.png}} 
    \raisebox{-0.5ex}{\includegraphics[width=0.02\linewidth]{figures/icons/prompt.png}} 
    \raisebox{-0.5ex}{\includegraphics[width=0.02\linewidth]{figures/icons/command.png}} \raisebox{-0.5ex}{\includegraphics[width=0.02\linewidth]{figures/icons/status.png}} & \raisebox{-0.5ex}{\includegraphics[width=0.018\linewidth]{figures/icons/qwen.png}}~Qwen2.5-VL & \raisebox{-0.5ex}{\includegraphics[width=0.018\linewidth]{figures/icons/qwen.png}}~Qwen2.5-VL & LH & 99.1 & 97.1 & 97.1 & 99.9 & 87.6 & 92.1
    \\
    ReflectDrive \cite{li2025discrete} & 2025 & \raisebox{-0.5ex}{\includegraphics[width=0.02\linewidth]{figures/icons/rgb.png}} \raisebox{-0.5ex}{\includegraphics[width=0.02\linewidth]{figures/icons/prompt.png}} \raisebox{-0.5ex}{\includegraphics[width=0.018\linewidth]{figures/icons/scene.png}} \raisebox{-0.5ex}{\includegraphics[width=0.02\linewidth]{figures/icons/status.png}} & LLaDA-V & LLaDA-V & GEN & 99.7 & 99.5 & 99.1 & 99.9 & 88.9 & 94.7
    \\
    \rowcolor{gray!7}AdaThinkDrive \cite{luo2025adathinkdrive} & 2025 & \raisebox{-0.5ex}{\includegraphics[width=0.02\linewidth]{figures/icons/rgb.png}} \raisebox{-0.5ex}{\includegraphics[width=0.02\linewidth]{figures/icons/prompt.png}} \raisebox{-0.5ex}{\includegraphics[width=0.02\linewidth]{figures/icons/command.png}} \raisebox{-0.5ex}{\includegraphics[width=0.02\linewidth]{figures/icons/status.png}} & \raisebox{-0.5ex}{\includegraphics[width=0.019\linewidth]{figures/icons/internvl.png}}~InternVL3 & \raisebox{-0.5ex}{\includegraphics[width=0.019\linewidth]{figures/icons/internvl.png}}~InternVL3 & REG & 99.1 & 98.8 & 97.2 & 100.0 & 87.9 & 93.0
    \\
    Percept-WAM~\cite{han2025percept}&2025&\raisebox{-0.5ex}{\includegraphics[width=0.02\linewidth]{figures/icons/rgb.png}} \raisebox{-0.5ex}{\includegraphics[width=0.02\linewidth]{figures/icons/lidar.png}} \raisebox{-0.5ex}{\includegraphics[width=0.02\linewidth]{figures/icons/prompt.png}} \raisebox{-0.5ex}{\includegraphics[width=0.02\linewidth]{figures/icons/command.png}}&\raisebox{-0.5ex}{\includegraphics[width=0.019\linewidth]{figures/icons/internvl.png}} InternViT  \raisebox{-0.5ex}{\includegraphics[width=0.019\linewidth]{figures/icons/internvl.png}}&\raisebox{-0.5ex}{\includegraphics[width=0.019\linewidth]{figures/icons/internvl.png}}InternVL2&REG&98.8&98.6&94.4&99.5&84.8&90.2\\

    \rowcolor{gray!7}Reasoning-VLA~\cite{zhang2025reasoning}&2025&\raisebox{-0.5ex}{\includegraphics[width=0.02\linewidth]{figures/icons/rgb.png}} \raisebox{-0.5ex}{\includegraphics[width=0.02\linewidth]{figures/icons/prompt.png}} \raisebox{-0.5ex}{\includegraphics[width=0.02\linewidth]{figures/icons/status.png}}&\raisebox{-0.5ex}{\includegraphics[width=0.018\linewidth]{figures/icons/qwen.png}}~Qwen2.5-VL &\raisebox{-0.5ex}{\includegraphics[width=0.018\linewidth]{figures/icons/qwen.png}}~Qwen2.5-VL &REG&97.8&93.2&98.1&99.8&80.7&91.7
    \\
    \bottomrule
\end{tabular}}
\end{table}
\begin{table}[t]
\captionsetup{justification=raggedright, singlelinecheck=false}
\caption{Closed-loop and Open-loop performance comparison of E2E-AD Methods on the \textbf{Bench2Drive} benchmark.
\\ 
$\bullet$ \textbf{Input}: \raisebox{-0.5ex}{\includegraphics[width=0.02\linewidth]{figures/icons/rgb.png}}: Camera, \raisebox{-0.5ex}{\includegraphics[width=0.02\linewidth]{figures/icons/lidar.png}}: LiDAR, \raisebox{-0.5ex}{\includegraphics[width=0.02\linewidth]{figures/icons/prompt.png}}: Prompt, \raisebox{-0.5ex}{\includegraphics[width=0.02\linewidth]{figures/icons/command.png}}: Instruct., \raisebox{-0.5ex}{\includegraphics[width=0.02\linewidth]{figures/icons/scene.png}}: Scene Descrip., \raisebox{-0.5ex}{\includegraphics[width=0.02\linewidth]{figures/icons/status.png}}: Status, \raisebox{-0.5ex}{\includegraphics[width=0.02\linewidth]{figures/icons/traffic_rule.png}}: Rule, \raisebox{-0.5ex}{\includegraphics[width=0.02\linewidth]{figures/icons/context.png}}: Context.
\\
$\bullet$ \textbf{Action}: \underline{LH}: Language Head, \underline{RL}: Policy with Reinforcement Learning, \underline{REG}: Decoder + MLP, \underline{SEL}: Trajectory Selection w/ Cost, and \underline{GEN}: Trajectory Generation w/ Generative Model.
\\
$\bullet$ \textbf{Evaluation Metrics}: \underline{DS ($\uparrow$)}: Driving Score, \underline{SR ($\uparrow$)}: Success Rate. \underline{Avg. L2 ($\downarrow$)}: Averaged L2 distance of trajectory.}
\vspace{-0.2cm}
\resizebox{\linewidth}{!}{
\begin{tabular}{rcclllccccc}
    \toprule
    \multirow{2}{*}{\textbf{Method}} & \multirow{2}{*}{\textbf{Year}} & \multirow{2}{*}{\textbf{Input}}  & \multirow{2}{*}{\textbf{Vision \raisebox{-0.25ex}{\includegraphics[width=0.022\linewidth]{figures/icons/vision.png}}}} & \multirow{2}{*}{\textbf{Language \raisebox{-0.5ex}{\includegraphics[width=0.021\linewidth]{figures/icons/language.png}}}} & \multirow{2}{*}{\textbf{Action \raisebox{-0.5ex}{\includegraphics[width=0.021\linewidth]{figures/icons/action.png}}}} & \multicolumn{4}{c}{\textbf{Closed-Loop}} & \textbf{Open-Loop}
    \\
    \cmidrule(lr){7-10}\cmidrule(lr){10-11}
    & & & & & & DS$\uparrow$ & SR(\%)$\uparrow$ & Efficiency$\uparrow$ & Comfort$\uparrow$ & Avg. L2 $\downarrow$ 
    \\
    \midrule\midrule
    \rowcolor{vla_purple!15}\multicolumn{11}{l}{\textcolor{vla_purple}{$\bullet$~\textbf{Vision-Action Models}}}  
    \\
    \midrule
    TCP \cite{wu2022trajectoryguided} & 2022 & \raisebox{-0.5ex}{\includegraphics[width=0.02\linewidth]{figures/icons/rgb.png}} & ResNet & - & REG & 40.70 & 15.00 & 54.26 & 47.80 & 1.70   
    \\
    \rowcolor{gray!7}ThinkTwice \cite{jia2023thinktwice} & 2023 & \raisebox{-0.5ex}{\includegraphics[width=0.02\linewidth]{figures/icons/rgb.png}} & ResNet & - & REG & 62.44 & 31.23 & 69.33 & 16.22 & 0.95   
    \\
    DriveAdapter \cite{jia2023driveadapter} & 2023 & \raisebox{-0.5ex}{\includegraphics[width=0.02\linewidth]{figures/icons/rgb.png}} \raisebox{-0.5ex}{\includegraphics[width=0.02\linewidth]{figures/icons/lidar.png}} & ResNet & - & REG & 64.22 & 33.08 & 70.22 & 16.01 & 1.01  
    \\  
    \rowcolor{gray!7}UniAD-Base \cite{hu2023planning}&2023& \raisebox{-0.5ex}{\includegraphics[width=0.02\linewidth]{figures/icons/rgb.png}} & ResNet & - & REG & 45.81 & 16.36 & 129.21 & 43.58 & 0.73  
    \\
    VAD \cite{jiang2023vad}&2023 & \raisebox{-0.5ex}{\includegraphics[width=0.02\linewidth]{figures/icons/rgb.png}} & ResNet & - & REG & 42.35 & 15.00 & 157.94 & 46.01 & 0.91 
    \\ 
    \rowcolor{gray!7}GenAD \cite{zheng2024genad} & 2024 &\raisebox{-0.5ex}{\includegraphics[width=0.02\linewidth]{figures/icons/rgb.png}} & ResNet & - & GEN & 44.81 & 15.90 & - & - & -
    \\
    DriveTransformer \cite{jia2025drivetransformer} & 2025 & \raisebox{-0.5ex}{\includegraphics[width=0.02\linewidth]{figures/icons/rgb.png}} & ResNet & - & REG & 63.46 & 35.01 & 100.64 & 20.78 & 0.62  
    \\ 
    \rowcolor{gray!7}ETA \cite{hamdan2025eta}&2025&\raisebox{-0.5ex}{\includegraphics[width=0.02\linewidth]{figures/icons/rgb.png}} & CLIP & - & REG & 69.53 & 38.64 & 184.51 & 28.43 & -
    \\
    WoTE \cite{li2025end} & 2025 & \raisebox{-0.5ex}{\includegraphics[width=0.02\linewidth]{figures/icons/rgb.png}} \raisebox{-0.5ex}{\includegraphics[width=0.02\linewidth]{figures/icons/lidar.png}} & ResNet & - & SEL & 61.71 & 31.36 & - & - & -  
    \\
    \rowcolor{gray!7}GuideFlow \cite{liu2025guideflow} & 2025 & \raisebox{-0.5ex}{\includegraphics[width=0.02\linewidth]{figures/icons/rgb.png}} & ResNet & - & GEN & 75.21 & 51.36 & - & - & -  
    \\
    Raw2Drive \cite{yang2025raw2drive} & 2025 & \raisebox{-0.5ex}{\includegraphics[width=0.02\linewidth]{figures/icons/rgb.png}} & ResNet & - & RL & 71.36 & 50.24 & 214.17 & 22.42 & -  
    \\
    \midrule
    \rowcolor{vla_green!15}\multicolumn{11}{l}{\textcolor{vla_green}{$\bullet$~\textbf{Vision-Language-Action Models}}} 
    \\
    \midrule
    ORION  \cite{fu2025orion} & 2025 & \raisebox{-0.5ex}{\includegraphics[width=0.02\linewidth]{figures/icons/rgb.png}}  
    \raisebox{-0.5ex}{\includegraphics[width=0.02\linewidth]{figures/icons/prompt.png}} 
    \raisebox{-0.5ex}{\includegraphics[width=0.02\linewidth]{figures/icons/command.png}} & EVA-02 & Vicuna-1.5 & GEN & 77.74 & 54.62 & 151.48 & 17.38 & 0.68 
    \\
    \rowcolor{gray!7}AutoVLA \cite{zhou2025autovla} & 2025 &\raisebox{-0.5ex}{\includegraphics[width=0.02\linewidth]{figures/icons/rgb.png}} 
    \raisebox{-0.5ex}{\includegraphics[width=0.02\linewidth]{figures/icons/prompt.png}} 
    \raisebox{-0.5ex}{\includegraphics[width=0.02\linewidth]{figures/icons/command.png}} \raisebox{-0.5ex}{\includegraphics[width=0.02\linewidth]{figures/icons/status.png}}  & SigLIP & \raisebox{-0.5ex}{\includegraphics[width=0.018\linewidth]{figures/icons/qwen.png}}~Qwen2.5-VL & LH & 78.84 & 57.73 & 146.93 & 39.33 & -
    \\
    SimLingo-Base \cite{renz2025simlingo} & 2025 & \raisebox{-0.5ex}{\includegraphics[width=0.02\linewidth]{figures/icons/rgb.png}} 
    \raisebox{-0.5ex}{\includegraphics[width=0.02\linewidth]{figures/icons/prompt.png}} 
    \raisebox{-0.5ex}{\includegraphics[width=0.02\linewidth]{figures/icons/command.png}} & \raisebox{-0.5ex}{\includegraphics[width=0.019\linewidth]{figures/icons/internvl.png}} InternViT  & \raisebox{-0.5ex}{\includegraphics[width=0.018\linewidth]{figures/icons/qwen.png}}~Qwen2 & REG & 85.94 & 66.82 & 244.18 & 25.49 & -
    \\
    \rowcolor{gray!7}SimLingo \cite{renz2025simlingo} & 2025 & \raisebox{-0.5ex}{\includegraphics[width=0.02\linewidth]{figures/icons/rgb.png}} 
    \raisebox{-0.5ex}{\includegraphics[width=0.02\linewidth]{figures/icons/prompt.png}} 
    \raisebox{-0.5ex}{\includegraphics[width=0.02\linewidth]{figures/icons/command.png}} & \raisebox{-0.5ex}{\includegraphics[width=0.019\linewidth]{figures/icons/internvl.png}} InternViT  &  \raisebox{-0.5ex}{\includegraphics[width=0.018\linewidth]{figures/icons/qwen.png}}~Qwen2 & REG & 85.07 & 67.27 & 259.23 & 33.67 & -
    \\
    ReAL-AD \cite{lu2025real} & 2025 & \raisebox{-0.5ex}{\includegraphics[width=0.02\linewidth]{figures/icons/rgb.png}} \raisebox{-0.5ex}{\includegraphics[width=0.02\linewidth]{figures/icons/prompt.png}} \raisebox{-0.5ex}{\includegraphics[width=0.02\linewidth]{figures/icons/command.png}} & ResNet & \raisebox{-0.5ex}{\includegraphics[width=0.018\linewidth]{figures/icons/qwen.png}}~QwenVL & REG & 40.76 & 10.93 & - & - & 0.87
    \\
    \rowcolor{gray!7}ReasonPlan \cite{liu2025reasonplan} & 2025 & \raisebox{-0.5ex}{\includegraphics[width=0.02\linewidth]{figures/icons/rgb.png}} \raisebox{-0.5ex}{\includegraphics[width=0.02\linewidth]{figures/icons/prompt.png}} \raisebox{-0.5ex}{\includegraphics[width=0.02\linewidth]{figures/icons/command.png}}  \raisebox{-0.5ex}{\includegraphics[width=0.02\linewidth]{figures/icons/status.png}} \raisebox{-0.5ex}{\includegraphics[width=0.018\linewidth]{figures/icons/context.png}} & SigLIP & \raisebox{-0.5ex}{\includegraphics[width=0.018\linewidth]{figures/icons/qwen.png}}~Qwen & LH & 64.01 & 34.55 & 180.64 & 25.63 & 0.61
    \\
    DriveMoE~\cite{yang2025drivemoe} & 2025 & \raisebox{-0.5ex}{\includegraphics[width=0.02\linewidth]{figures/icons/rgb.png}} \raisebox{-0.5ex}{\includegraphics[width=0.02\linewidth]{figures/icons/prompt.png}} \raisebox{-0.5ex}{\includegraphics[width=0.02\linewidth]{figures/icons/status.png}} &  BEV Encoder & \raisebox{-0.5ex}{\includegraphics[width=0.016\linewidth]{figures/icons/llama.jpg}}~LLaMA  & REG  & 74.22 &48.64  & 175.96 & 15.31 &0.31
    \\
    \rowcolor{gray!7}VDRive\cite{guo2025vdrive} & 2025 &  \raisebox{-0.5ex}{\includegraphics[width=0.02\linewidth]{figures/icons/rgb.png}} \raisebox{-0.5ex}{\includegraphics[width=0.02\linewidth]{figures/icons/prompt.png}} \raisebox{-0.5ex}{\includegraphics[width=0.02\linewidth]{figures/icons/command.png}} \raisebox{-0.5ex}{\includegraphics[width=0.02\linewidth]{figures/icons/status.png}}&  \raisebox{-0.5ex}{\includegraphics[width=0.018\linewidth]{figures/icons/qwen.png}}~Qwen2.5-VL, CVQ& \raisebox{-0.5ex}{\includegraphics[width=0.019\linewidth]{figures/icons/internvl.png}}~InternVL3 & GEN & 66.15 & 50.51 & 110.23 & 22.90 & 0.55
    \\
    CoReVLA~\cite{fang2025corevla} & 2025 &  \raisebox{-0.5ex}{\includegraphics[width=0.02\linewidth]{figures/icons/rgb.png}} \raisebox{-0.5ex}{\includegraphics[width=0.02\linewidth]{figures/icons/prompt.png}} \raisebox{-0.5ex}{\includegraphics[width=0.02\linewidth]{figures/icons/command.png}} \raisebox{-0.5ex}{\includegraphics[width=0.02\linewidth]{figures/icons/context.png}} &  \raisebox{-0.5ex}{\includegraphics[width=0.018\linewidth]{figures/icons/qwen.png}}~Qwen2.5-VL & \raisebox{-0.5ex}{\includegraphics[width=0.018\linewidth]{figures/icons/qwen.png}}~Qwen2.5-VL & LH  & 72.18& 50.00&145.41 & 34.35 &-
    \\
    \bottomrule
\end{tabular}}
\label{tab: bench2drive_compare}
\end{table}

\subsection{Quantitative Experiments \& Analyses}
\label{sec:exp}
This section reviews quantitative benchmarks for evaluating VLA models across action prediction, planning accuracy, and closed-loop driving performance. Among them, nuScenes~\cite{caesar2020nuscenes}, NAVSIM~\cite{dauner2024navsim}, and Bench2Drive~\cite{jia2024bench2drive} are the most widely used. More recently, WOD-E2E~\cite{xu2025wod} introduces long-tail, safety-critical scenes with human-preference annotations, enabling more robust assessment of modern E2E and VLA systems.

\subsubsection{nuScenes Benchmark}
The nuScenes open-loop benchmark evaluates planning quality using trajectory-based metrics, including L2 displacement error and Collision Rate, as summarized in Table~\ref{tab:nuscenes_compare}. Basically, vision-action models such as UniAD~\cite{hu2023planning} reports 0.69m L2 and 0.12 collision rate. Incorporating language generally improves performance by providing semantic cues for safer planning. For instance, Drive-R1~\cite{li2025drive} combines supervised CoT alignment with RL finetuning to reach 0.31m L2 and 0.09 collision rate.

Beyond accuracy, recent studies explore the role of language in handling complex and long-tailed driving scenarios. While improvements are most evident in common cases, rare and highly complex situations remain an active area of investigation, motivating the integration of richer reasoning signals and data sources.

From a systems perspective, computational efficiency is an important consideration for practical deployment. Lightweight and efficiency-oriented designs, such as InsightDrive~\cite{song2025insightdrive} (16.3 FPS) and token-pruned architectures like FastDriveVLA~\cite{cao2025fastdrivevla}, illustrate ongoing efforts to balance model capacity with real-time feasibility.
For cross-domain evaluation, nuScenes highlights generalization to unseen cities and distribution shifts as a key benchmark dimension. Works such as VLP~\cite{pan2024vlp} and DiMA~\cite{hegde2025distilling} examine this setting and motivate complementary strategies including domain adaptation, distillation, and data augmentation.

\subsubsection{WOD-E2E Benchmark} 
The Waymo Open Dataset for End-to-End Driving (WOD-E2E)~\cite{xu2025wod} is a large-scale benchmark designed to evaluate end-to-end driving systems under long-tail, safety-critical scenarios that rarely appear in conventional datasets. It contains 4K segments with high-level routing commands, ego-status signals, and multi-camera views, enabling rigorous assessment of perception-planning coupling. A key contribution of WOD-E2E is the Rater Feedback Score (RFS), which measures trajectory quality based on alignment with human preference annotations rather than logged expert trajectories. As shown in Table~\ref{tab:wod_e2e_results}, RFS (Overall and Spotlight) complements conventional ADE metrics by providing a more human-aligned assessment of driving behavior.

\begin{figure}[t]
    \centering
    \includegraphics[width=\linewidth]{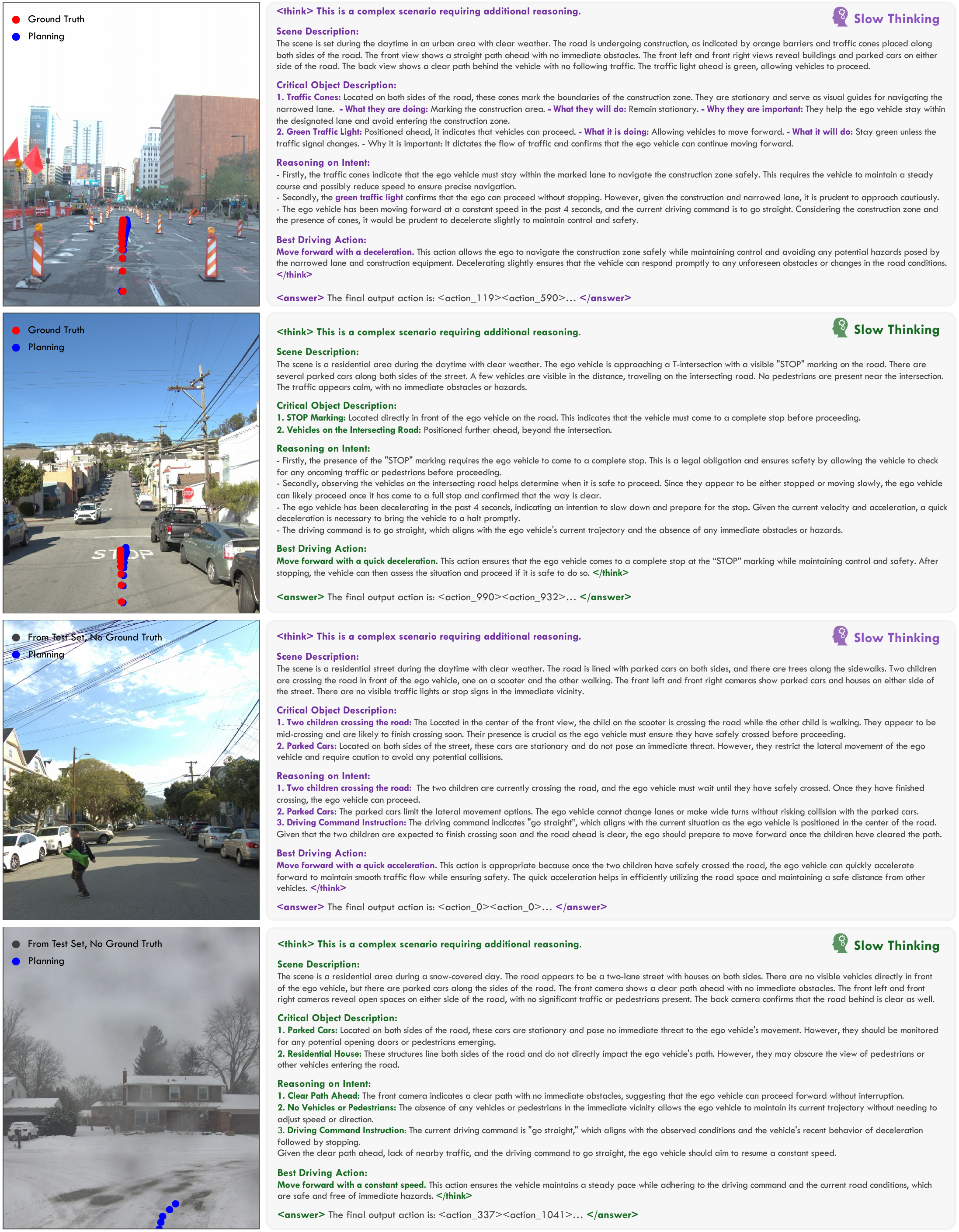}
    \caption{Visualization examples of the AutoVLA \cite{zhou2025autovla} reasoning/planning results on WOD-E2E~\cite{xu2025wod} dataset.}
    \label{fig:waymo_plot}
    \vspace{-0.4cm}
\end{figure}

Overall results indicate that while vision–action models achieve stable displacement accuracy, VLA models exhibit more diverse performance. Approaches such as  Poutine~\cite{rowe2025poutine}, and dVLM-AD~\cite{ma2025dvlm} achieve balanced RFS and ADE performance, highlighting the importance of effectively aligning language reasoning with trajectory generation. Figure~\ref{fig:waymo_plot}  presents the visualized performance of AutoVLA~\cite{zhou2025autovla} in the WOD-E2E dataset.

\subsubsection{NAVSIM Benchmark}
NAVSIM~\cite{dauner2024navsim} is built on OpenScene (a redistribution of nuPlan~\cite{caesar2021nuplan}), provides a closed-loop simulation environment designed to evaluate planning quality under realistic urban driving conditions. It adopts the PDMS metric, which aggregates multiple driving aspects, including No-Collision (NC), Driving Admissibility (DAC), Time-to-Collision (TTC), Ego Progress (EP), and Comfort (C), offering a holistic assessment of safety, efficiency, and driving smoothness. As shown in Table~\ref{tab:navsim_compare}, most methods achieve strong performance on safety-related metrics such as NC and DAC, while TTC and EP serve as more discriminative indicators of planning foresight and long-horizon decision quality. These metrics highlight differences in how models balance safety and progress when interacting with dynamic environments.

Representative vision–action approaches, such as WoTE~\cite{li2025end}, achieve 88.3 PDMS by integrating a BEV-based world model with reward-guided trajectory selection, demonstrating the effectiveness of structured world modeling for closed-loop planning. Building upon this foundation, language-conditioned methods further enhance decision-making. For instance, AutoVLA~\cite{zhou2025autovla} improves performance to 99.1 NC and 87.6 EP by leveraging language-driven decision priors and a Best-of-N oracle scoring strategy, illustrating how language supervision can guide trajectory selection and improve long-horizon planning behavior.

\subsubsection{Bench2Drive Benchmark}
Bench2Drive~\cite{jia2024bench2drive} provides a closed-loop evaluation protocol built on CARLA V2, focusing on success rate and composite driving scores to assess goal-directed driving behavior under interactive settings. Unlike open-loop benchmarks, Bench2Drive explicitly evaluates an agent’s ability to execute long-horizon tasks and respond to dynamic environmental feedback.

Recent VLA approaches demonstrate clear benefits from language grounding in this benchmark. For example, SimLingo~\cite{renz2025simlingo} introduces an action dreaming mechanism that aligns natural language instructions with control sequences, achieving a leading driving score of 85.94, as reported in Table~\ref{tab: bench2drive_compare}. These results indicate that language-guided reasoning can effectively influence closed-loop decision-making and improve planning performance in interactive driving scenarios.

Taken together with open-loop benchmarks, Bench2Drive highlights the growing importance of language–action alignment in VLA systems, particularly for interpreting high-level goals, guiding long-horizon behavior, and adapting actions under complex, human-centered instructions.
\section{Challenges \& Future Directions}
\label{sec:challenges}
VLA models mark a shift from modular stacks toward holistic, reasoning-driven driving agents. 
By leveraging large multimodal backbones, they promise richer environmental understanding, stronger generalization, and more interpretable decision-making. 
Yet, realizing their full potential in safety-critical autonomous driving requires addressing several fundamental challenges. In parallel, emerging research directions point toward next-generation systems that are more efficient, trustworthy, and capable of long-horizon reasoning.

\subsection{Current Challenges}

\subsubsection{Model Architecture and System Efficiency}
\label{subsec:model_architecture}

\textbf{Real-Time Processing and Latency.}
\label{para:latency}
VLA models inherit the substantial computational footprint of modern vision-language backbones. High-resolution, high-frame-rate camera inputs generate long visual-token sequences, and multi-view fusion amplifies memory and latency costs. Meeting the strict real-time constraints of autonomous vehicles, therefore, remains difficult~\cite{jihong2024edge, weng2024drive}. Recent advances in streaming token compression and adaptive visual encoders~\cite{li2025tokenpacker, cao2025fastdrivevla} offer promising directions, but achieving sub-50ms inference remains an unmet requirement for safety-critical deployment.

\noindent\textbf{Lack of Domain-Specific Foundation Models.}
\label{para:domain_models}
General-purpose VLMs~\cite{Qwen2.5-VL, chen2024internvl, zhu2025internvl3} provide strong priors but are not optimized for driving-specific perception, physics, or multi-sensor fusion. Autonomous driving requires precise spatial reasoning, adherence to traffic rules, and an understanding of rare, high-stakes edge cases -- abilities not fully captured by generic models. As highlighted in Section~\ref{sec:va}, dedicated driving foundation models~\cite{li2025march} remain a missing cornerstone for scalable and dependable VLA systems.

\clearpage\clearpage

\clearpage\clearpage
\subsubsection{Data and Generalization}
\label{subsec:data_generalization}

\noindent\textbf{Generalizing to Rare and Novel Scenarios.}
\label{para:long_tail}
One motivation for VLAs is their ability to leverage strong visual-language priors to interpret complex scenes. However, while VLM components may generalize well perceptually, aligning this understanding to the action space introduces new uncertainties. As noted in Section~\ref{sec:vla}, reasoning-rich representations do not automatically translate to robust action generation. Long-tailed scenarios -- misbehaving traffic agents, unusual road layouts, unpredictable weather -- remain failure points~\cite{ghosh2024roadwork, yao2025lilodriver, yurt2025ltda, tian2024tokenize, davidson2025refav}.

\noindent\textbf{Cost of High-Quality Data.}
\label{para:data_acquisition}
VLAs rely on diverse, high-quality multimodal datasets~\cite{liu2024surveyondata, li2023open}, yet collecting paired vision-action-language triplets at scale is expensive. Synthetic environments~\cite{song2023synthetic, chao2020survey, mutsch2023model} help, but face substantial sim-to-real gaps, with discrepancies in noise characteristics, lighting, and behavior of other agents~\cite{kong2023robo3d,xie2025benchmarking}. Improving data efficiency and mitigating distribution shifts remain long-standing challenges.

\subsubsection{Core Capabilities and Trustworthiness}
\label{subsec:capabilities_trust}

\noindent\textbf{Interpretability \& Hallucination.}
\label{para:interpretability}
While VLA models produce natural-language rationales via chain-of-thought prompting~\cite{fu2025orion, zhou2025autovla, wang2025pixelthink}, these explanations are generated artifacts -- not faithful reflections of the underlying causal reasoning. Language hallucination~\cite{rawte2023survey, liu2024survey, huang2025survey} presents new risks: the model may justify an incorrect decision with a confident but spurious narrative. Ensuring consistent grounding between perceptions, actions, and explanations is an open challenge.

\noindent \textbf{Long-Horizon Temporal Coherence.}
\label{para:temporal_coherence}
Driving depends on anticipating multi-stage interactions and maintaining situational awareness across extended time horizons~\cite{parktime, song2025don, xu2025beyond}. Current transformer-based VLA architectures remain constrained by limited context windows and short-term conditioning, inherited from standard VLM designs. Temporal fragmentation leads to inconsistent decisions, especially in multi-agent or highly dynamic traffic scenes.

\subsection{Future Directions}

\subsubsection{Next-Generation Model Paradigms}
\label{subsec:future_paradigms}

\noindent\textbf{Unified Vision-Language-World Models.} A promising evolution integrates VLA with predictive world models \cite{genie3, cen2025worldvla, li2025drivevlaw0, li2024enhancing}, extending the VA-based models in Section~\ref{sec:wmad}. Rather than reacting frame by frame, such systems simulate future scene evolution conditioned on candidate actions, enabling proactive planning and more reliable behavior under uncertainty. Building unified, end-to-end world models that jointly reason about perception, language, and dynamics may form the backbone of next-generation autonomous agents.

\noindent\textbf{Richer Multimodal Fusion.}
As sensor suites diversify, future architectures will incorporate early and tight fusion of LiDAR, Radar, event cameras, and high-definition maps~\cite{kong2023robo3d, xie2025benchmarking,shao2024lmdrive}. Language enhances semantic grounding, but robust 3D geometry is indispensable for safe decision-making~\cite{liu2025occvla, tian2023occ3d}. Holistic multimodal fusion can combine the interpretability of VLMs with the spatial precision of geometric sensors.

\subsubsection{Advancing Intelligence and Adaptation}
\label{subsec:advancing_intelligence}

\noindent\textbf{Socially Aware, Knowledge-Grounded Driving.} VLA models must acquire deeper commonsense reasoning -- understanding intent, conventions, and causal relationships beyond explicit annotations~\cite{li2023towards, zhai2025world}. Future efforts will draw from large-scale video-language corpora, leveraging external knowledge bases and structured reasoning modules to support socially compliant and anticipatory driving.

\noindent\textbf{Continual \& Onboard Learning.} Static, offline-trained models cannot capture evolving road infrastructures or regional driving customs~\cite{verwimp2023clad, zhuang2024online, cao2022autonomous}. 
Enabling safe, incremental learning from everyday driving, while avoiding catastrophic forgetting and ensuring safety guarantees, is essential for long-term deployment. This relates closely to addressing long-tail generalization gaps.

\clearpage\clearpage
\begin{table}[ht]
\renewcommand\arraystretch{1.35}
\centering
\vspace{-0.8cm}
\caption{Summary of the evaluation metrics used for evaluating the \textbf{trajectory-based} and \textbf{text-based} action outputs.}
\vspace{-0.2cm}
\scriptsize
\resizebox{\linewidth}{!}{
\begin{tabular}{|>{\centering\arraybackslash}m{1.2cm}|c|>{\centering\arraybackslash}m{2.4cm}|m{7.8cm}|c|}
    \hline
    \rowcolor{tpami_gray!15}\textbf{Abbr.} & \textbf{-} & \textbf{Full Name} & \textbf{Description} & \textbf{Ref.} 
    \\
    \hline\hline
    \rowcolor{vla_purple!15}\multicolumn{5}{|l|}{\textcolor{vla_purple}{\textbf{Action-Planning Open-Loop Evaluation}}}
    \\
    \hline
    L2 &  \textcolor{tpami_red}{$\downarrow$} & L2 Error & L2 distance error between the planned trajectory and the human driving trajectory in 3 seconds.& \cite{hu2022st} 
    \\
    \hline
    CR &  \textcolor{tpami_red}{$\downarrow$} & Collision Rate & How often the self-driving vehicle would collide with other agents on the road. & \cite{hu2022st}
    \\
    \hline
    ADE &  \textcolor{tpami_red}{$\downarrow$}&Average Displacement Error&Mean displacement error between predicted trajectories and expert waypoints across the horizon, reflecting overall trajectory accuracy.& \cite{hu2023planning}
    \\
    \hline
    FDE &  \textcolor{tpami_red}{$\downarrow$}&Final Displacement Error&Displacement error at the final predicted waypoint compared with expert trajectories, emphasizing long-term accuracy.& \cite{hu2023planning}
    \\
    
    \hline
    MR &  \textcolor{tpami_red}{$\downarrow$}&Miss Rate&Fraction of prediction time steps where displacement error exceeds horizon-specific thresholds, reflecting failure in trajectory coverage.& \cite{hu2023planning}
    \\
    \hline
    AHE &  \textcolor{tpami_red}{$\downarrow$} & Average Heading Error&Mean absolute angular deviation between predicted and expert heading over the trajectory horizon, measuring orientation accuracy. &  \cite{karnchanachari2024towards}
    \\
    \hline
    FHE &  \textcolor{tpami_red}{$\downarrow$}&Final Heading Error&Absolute angular deviation of predicted heading from expert at the final timestep, reflecting terminal orientation accuracy.&  \cite{karnchanachari2024towards}
    \\
    \hline
    SLE &  \textcolor{tpami_red}{$\downarrow$}&Speed L1 Error&Mean absolute error of predicted speed control signals.& \cite{jia2023adriver}
    \\
    \hline
    SALE &  \textcolor{tpami_red}{$\downarrow$}&Steer Angle L1 Error&Mean absolute error of predicted steering angle control signals.& \cite{jia2023adriver}
    \\
    \hline
    RFS&\textcolor{vla_green}{$\uparrow$}&Rater Feedback Score&Measure how well the predicted trajectory aligns with human driving preferences by checking whether it falls within trust regions.&\cite{xu2025wod} \\
   
    \hline\hline
    \rowcolor{vla_green!15}\multicolumn{5}{|l|}{\textcolor{vla_green}{\textbf{Trajectory-Based Closed-Loop Evaluation}}}
    \\
    \hline
    
    RC &  \textcolor{vla_green}{$\uparrow$} & Route Completion & The percentage of route distance completed.& \cite{prakash2021multi}
    \\
    \hline
    DS &  \textcolor{vla_green}{$\uparrow$} & Driving Score & RC weighted by a penalty factor that accounts for collisions with pedestrians, vehicles, etc.& \cite{prakash2021multi}
    \\
    \hline
    NC &  \textcolor{vla_green}{$\uparrow$}&No Collision&Fraction of scenarios without ego-fault collisions, focusing exclusively on responsibility-aware collision evaluation. & \cite{dauner2024navsim}
    \\
    \hline
    DAC&  \textcolor{vla_green}{$\uparrow$}&Driving Admissibility Check&Boolean evaluation that checks whether the ego vehicle remains inside drivable polygons throughout the rollout. & \cite{dauner2024navsim}
    \\
    \hline
    TTC&  \textcolor{vla_green}{$\uparrow$}&Time To Collision&Boolean verification that the time-to-collision value exceeds safety thresholds, preventing imminent crashes. & \cite{dauner2024navsim}
    \\
    \hline
    C&  \textcolor{vla_green}{$\uparrow$}&Driving Comfort&The comfort of driving.& \cite{dauner2024navsim}
    \\
    \hline
    EP&  \textcolor{vla_green}{$\uparrow$}&Ego Progress&Penalization of excessive jerk, acceleration, or yaw-rate, reflecting ride quality and passenger comfort.& \cite{dauner2024navsim}
    \\
    \hline
    PDMS&  \textcolor{vla_green}{$\uparrow$}&Predictive Driver Model Score&A flexible weighted evaluation score in autonomous driving that aggregates multiple safety, progress, and comfort subscores into a single metric.& \cite{dauner2024navsim}
    \\
    \hline
    SR &  \textcolor{vla_green}{$\uparrow$}&Success Rate&Percentage of navigation episodes that successfully reach the goal within a fixed time budget, indicating overall task completion. & \cite{dosovitskiy2017carla}
    \\
    \hline
    ID&  \textcolor{vla_green}{$\uparrow$}&Infraction Distance&Average driving distance between two infractions, with longer distances reflecting safer and more reliable policy behavior.& \cite{dosovitskiy2017carla}
    \\
    \hline\hline
    \rowcolor{vla_purple!15}\multicolumn{5}{|l|}{\textcolor{vla_purple}{\textbf{Text-Based Action Evaluation}}}
    \\
    \hline
    CIDEr &  \textcolor{vla_green}{$\uparrow$} & Consensus-based Image Description Evaluation&Measures similarity of generated captions to multiple human references using TF-IDF weighted n-grams.& \cite{ma2025aln}
    \\
    \hline
    BLEU &  \textcolor{vla_green}{$\uparrow$} & Bilingual Evaluation Understudy & Precision-based metric that compares n-grams of the generated text against reference texts.& \cite{ma2025aln}
    \\
    \hline
    METEOR &  \textcolor{vla_green}{$\uparrow$} & Metric for Evaluation of Translation with Explicit Ordering & Considers unigram precision and recall with stemming, synonym matching, and fragmentation penalty.& \cite{ma2025aln}
    \\
    \hline
    Rouge &  \textcolor{vla_green}{$\uparrow$}&Recall-Oriented Understudy for Gisting Evaluation& Recall-focused metric using overlapping n-grams, word sequences, or word pairs between generated and reference texts.& \cite{ma2025aln}
    \\
    \hline
    Top-1 Acc &  \textcolor{vla_green}{$\uparrow$} & Visual Question Answering Top-1
    Accuracy & Percentage of predictions where the most confident output matches the ground truth label.& \cite{ma2025aln}
    \\
    \hline
\end{tabular}
}
\label{tab:eval_metrics}
\end{table}

\clearpage\clearpage
\subsubsection{Ecosystem for Safe Deployment}
\label{subsec:safe_deployment}

\noindent\textbf{Standardized Evaluation \& Safety Guarantees.} 
Evaluation metrics from current benchmarks, \emph{e.g.}, NAVSIM~\cite{dauner2024navsim} and Bench2Drive~\cite{jia2024bench2drive}, assess safety and comfort but do not capture key VLA-specific risks such as reasoning failures, instruction-following errors, or cross-modal inconsistencies~\cite{liu2024survey, rawte2023survey}. 
Future benchmarks should evaluate multi-step instruction execution, robustness to ambiguous language, and resistance to hallucination. Beyond empirical testing, formal verification tools are needed to provide theoretical guarantees for safety-critical behaviors.

\noindent\textbf{Human-Centric Interaction \& Personalization.} 
VLA systems open the door to richer in-car interaction~\cite{yang2025human, schrum2024maveric}. Natural language enables drivers to specify goals, constraints, and preferences (``\textit{drive cautiously}'', ``\textit{avoid unprotected left turns}''). Personalization modules~\cite{hao2025styledrive} can adapt driving styles to different users, enhancing comfort and trust. The challenge lies in balancing personalization with strict safety and regulatory requirements.
\section{Conclusion}
\label{sec:conclusion}

Vision-Language-Action models are reshaping autonomous driving by coupling perception with high-level reasoning and natural language understanding. This work formalizes the VLA problem setting, outlines the progression from traditional VA pipelines, and organizes existing methods into coherent architectural families together with the datasets and benchmarks that support their development. VLA systems offer clear advantages in interpretability, generalization, and human interaction, but core challenges remain: aligning symbolic reasoning with continuous control, ensuring robustness in long-tail scenarios, and establishing evaluation protocols that faithfully measure instruction following and safety. Progress will depend on advances in efficient architectures, deeper multimodal fusion, world-model-driven planning, and more rigorous human-centered testing. Overall, VLA represents a promising direction for building autonomous agents that are not only competent drivers but also communicative, transparent, and responsive to human intent.

\bibliographystyle{plainnat}
\bibliography{main}

\begin{thebibliography}{345}
\providecommand{\natexlab}[1]{#1}
\providecommand{\url}[1]{\texttt{#1}}
\expandafter\ifx\csname urlstyle\endcsname\relax
  \providecommand{\doi}[1]{doi: #1}\else
  \providecommand{\doi}{doi: \begingroup \urlstyle{rm}\Url}\fi

\bibitem[Achiam et~al.(2023)Achiam, Adler, Agarwal, Ahmad, Akkaya, Aleman, Almeida, Altenschmidt, Altman, Anadkat, Avila, Babuschkin, Balaji, Balcom, Baltescu, Bao, Bavarian, Belgum, Bello, Berdine, Bernadett-Shapiro, Berner, Bogdonoff, Boiko, Boyd, Brakman, Brockman, Brooks, Brundage, Button, Cai, Campbell, Cann, Carey, Carlson, Carmichael, Chan, Chang, Chantzis, Chen, Chen, Chen, Chen, Chen, Chess, Cho, Chu, Chung, Cummings, Currier, Dai, Decareaux, Degry, Deutsch, Deville, Dhar, Dohan, Dowling, Dunning, Ecoffet, Eleti, Eloundou, Farhi, Fedus, Felix, Fishman, Forte, Fulford, Gao, Georges, Gibson, Goel, Gogineni, Goh, Gontijo-Lopes, Gordon, Grafstein, Gray, Greene, Gross, Gu, Guo, Hallacy, Han, Harris, He, Heaton, Heidecke, Hesse, Hickey, Hickey, Hoeschele, Houghton, Hsu, Hu, Hu, Huizinga, Jain, Jain, et~al.]{achiam2023gpt}
Josh Achiam, Steven Adler, Sandhini Agarwal, Lama Ahmad, Ilge Akkaya, Florencia~Leoni Aleman, Diogo Almeida, Janko Altenschmidt, Sam Altman, Shyamal Anadkat, Red Avila, Igor Babuschkin, Suchir Balaji, Valerie Balcom, Paul Baltescu, Haiming Bao, Mohammad Bavarian, Jeff Belgum, Irwan Bello, Jake Berdine, Gabriel Bernadett-Shapiro, Christopher Berner, Lenny Bogdonoff, Oleg Boiko, Madelaine Boyd, Anna-Luisa Brakman, Greg Brockman, Tim Brooks, Miles Brundage, Kevin Button, Trevor Cai, Rosie Campbell, Andrew Cann, Brittany Carey, Chelsea Carlson, Rory Carmichael, Brooke Chan, Che Chang, Fotis Chantzis, Derek Chen, Sully Chen, Ruby Chen, Jason Chen, Mark Chen, Ben Chess, Chester Cho, Casey Chu, Hyung~Won Chung, Dave Cummings, Jeremiah Currier, Yunxing Dai, Cory Decareaux, Thomas Degry, Noah Deutsch, Damien Deville, Arka Dhar, David Dohan, Steve Dowling, Sheila Dunning, Adrien Ecoffet, Atty Eleti, Tyna Eloundou, David Farhi, Liam Fedus, Niko Felix, Simón~Posada Fishman, Juston Forte, Isabella Fulford, Leo Gao, Elie
  Georges, Christian Gibson, Vik Goel, Tarun Gogineni, Gabriel Goh, Rapha Gontijo-Lopes, Jonathan Gordon, Morgan Grafstein, Scott Gray, Ryan Greene, Joshua Gross, Shixiang~Shane Gu, Yufei Guo, Chris Hallacy, Jesse Han, Jeff Harris, Yuchen He, Mike Heaton, Johannes Heidecke, Chris Hesse, Alan Hickey, Wade Hickey, Peter Hoeschele, Brandon Houghton, Kenny Hsu, Shengli Hu, Xin Hu, Joost Huizinga, Shantanu Jain, Shawn Jain, et~al.
\newblock {GPT-4} technical report.
\newblock \emph{arXiv preprint arXiv:2303.08774}, 2023.

\bibitem[Arai et~al.(2025)Arai, Miwa, Sasaki, Watanabe, Yamaguchi, Aoki, and Yamamoto]{arai2025covla}
Hidehisa Arai, Keita Miwa, Kento Sasaki, Kohei Watanabe, Yu~Yamaguchi, Shunsuke Aoki, and Issei Yamamoto.
\newblock {CoVLA}: Comprehensive vision-language-action dataset for autonomous driving.
\newblock In \emph{IEEE/CVF Winter Conf. Appl. Comput. Vis.}, pages 1933--1943, 2025.

\bibitem[Azzolini et~al.(2025)Azzolini, Bai, Brandon, Cao, Chattopadhyay, Chen, Chu, Cui, Diamond, Ding, Feng, Ferroni, Govindaraju, Gu, Gururani, Hanafi, Hao, Huffman, Jin, Johnson, Khan, Kurian, Lantz, Lee, Li, Li, Liao, Lin, Lin, Liu, Lu, Luo, Mathau, Ni, Pavao, Ping, Romero, Smelyanskiy, Song, Tchapmi, Wang, Wang, Wang, Wei, Xu, Xu, Yang, Yang, Yang, Zhang, Zeng, and Zhang]{azzolini2025cosmos}
Alisson Azzolini, Junjie Bai, Hannah Brandon, Jiaxin Cao, Prithvijit Chattopadhyay, Huayu Chen, Jinju Chu, Yin Cui, Jenna Diamond, Yifan Ding, Liang Feng, Francesco Ferroni, Rama Govindaraju, Jinwei Gu, Siddharth Gururani, Imad~El Hanafi, Zekun Hao, Jacob Huffman, Jingyi Jin, Brendan Johnson, Rizwan Khan, George Kurian, Elena Lantz, Nayeon Lee, Zhaoshuo Li, Xuan Li, Maosheng Liao, Tsung-Yi Lin, Yen-Chen Lin, Ming-Yu Liu, Xiangyu Lu, Alice Luo, Andrew Mathau, Yun Ni, Lindsey Pavao, Wei Ping, David~W. Romero, Misha Smelyanskiy, Shuran Song, Lyne Tchapmi, Andrew~Z. Wang, Boxin Wang, Haoxiang Wang, Fangyin Wei, Jiashu Xu, Yao Xu, Dinghao Yang, Xiaodong Yang, Zhuolin Yang, Jingxu Zhang, Xiaohui Zeng, and Zhe Zhang.
\newblock {Cosmos-Reason1}: From physical common sense to embodied reasoning.
\newblock \emph{arXiv preprint arXiv:2503.15558}, 2025.

\bibitem[Bai et~al.(2023)Bai, Bai, Chu, Cui, Dang, Deng, Fan, Ge, Han, Huang, Hui, Ji, Li, Lin, Lin, Liu, Liu, Lu, Lu, Ma, Men, Ren, Ren, Tan, Tan, Tu, Wang, Wang, Wang, Wu, Xu, Xu, Yang, Yang, Yang, Yang, Yao, Yu, Yuan, Yuan, Zhang, Zhang, Zhang, Zhang, Zhou, Zhou, Zhou, and Zhu]{bai2023qwen}
Jinze Bai, Shuai Bai, Yunfei Chu, Zeyu Cui, Kai Dang, Xiaodong Deng, Yang Fan, Wenbin Ge, Yu~Han, Fei Huang, Binyuan Hui, Luo Ji, Mei Li, Junyang Lin, Runji Lin, Dayiheng Liu, Gao Liu, Chengqiang Lu, Keming Lu, Jianxin Ma, Rui Men, Xingzhang Ren, Xuancheng Ren, Chuanqi Tan, Sinan Tan, Jianhong Tu, Peng Wang, Shijie Wang, Wei Wang, Shengguang Wu, Benfeng Xu, Jin Xu, An~Yang, Hao Yang, Jian Yang, Shusheng Yang, Yang Yao, Bowen Yu, Hongyi Yuan, Zheng Yuan, Jianwei Zhang, Xingxuan Zhang, Yichang Zhang, Zhenru Zhang, Chang Zhou, Jingren Zhou, Xiaohuan Zhou, and Tianhang Zhu.
\newblock Qwen technical report.
\newblock \emph{arXiv preprint arXiv:2309.16609}, 2023.

\bibitem[Bai et~al.(2025)Bai, Chen, Liu, Wang, Ge, Song, Dang, Wang, Wang, Tang, Zhong, Zhu, Yang, Li, Wan, Wang, Ding, Fu, Xu, Ye, Zhang, Xie, Cheng, Zhang, Yang, Xu, Lin, et~al.]{Qwen2.5-VL}
Shuai Bai, Keqin Chen, Xuejing Liu, Jialin Wang, Wenbin Ge, Sibo Song, Kai Dang, Peng Wang, Shijie Wang, Jun Tang, Humen Zhong, Yuanzhi Zhu, Mingkun Yang, Zhaohai Li, Jianqiang Wan, Pengfei Wang, Wei Ding, Zheren Fu, Yiheng Xu, Jiabo Ye, Xi~Zhang, Tianbao Xie, Zesen Cheng, Hang Zhang, Zhibo Yang, Haiyang Xu, Junyang Lin, et~al.
\newblock {Qwen2.5-VL} technical report.
\newblock \emph{arXiv preprint arXiv:2502.13923}, 2025.

\bibitem[Bain and Sammut(1995)]{bain1995framework}
Michael Bain and Claude Sammut.
\newblock A framework for behavioural cloning.
\newblock In \emph{Mach. intell.}, volume~15, pages 103--129, 1995.

\bibitem[Baldassarre et~al.(2025)Baldassarre, Szafraniec, Terver, Khalidov, Massa, LeCun, Labatut, Seitzer, and Bojanowski]{baldassarre2025back}
Federico Baldassarre, Marc Szafraniec, Basile Terver, Vasil Khalidov, Francisco Massa, Yann LeCun, Patrick Labatut, Maximilian Seitzer, and Piotr Bojanowski.
\newblock Back to the features: {DINO} as a foundation for video world models.
\newblock \emph{arXiv preprint arXiv:2507.19468}, 2025.

\bibitem[Ball et~al.(2025)Ball, Bauer, Belletti, Brownfield, Ephrat, Fruchter, Gupta, Holsheimer, Holynski, Hron, Kaplanis, Limont, McGill, Oliveira, Parker-Holder, Perbet, Scully, Shar, Spencer, Tov, Villegas, Wang, Yung, Baetu, Berbel, Bridson, Bruce, Buttimore, Chakera, Chandra, Collins, Cullum, Damoc, Dasagi, Gazeau, Gbadamosi, Han, Hirst, Kachra, Kerley, Kjems, Knoepfel, Koriakin, Lo, Lu, Mehring, Moufarek, Nandwani, Oliveira, Pardo, Park, Pierson, Poole, Ran, Salimans, Sanchez, Saprykin, Shen, Sidhwani, Smith, Stanton, Tomlinson, Vijaykumar, Wang, Wingfield, Wong, Xu, Yew, Young, Zubov, Eck, Erhan, Kavukcuoglu, Hassabis, Gharamani, Hadsell, van~den Oord, Mosseri, Bolton, Singh, and Rockt{\"a}schel]{genie3}
Philip~J. Ball, Jakob Bauer, Frank Belletti, Bethanie Brownfield, Ariel Ephrat, Shlomi Fruchter, Agrim Gupta, Kristian Holsheimer, Aleksander Holynski, Jiri Hron, Christos Kaplanis, Marjorie Limont, Matt McGill, Yanko Oliveira, Jack Parker-Holder, Frank Perbet, Guy Scully, Jeremy Shar, Stephen Spencer, Omer Tov, Ruben Villegas, Emma Wang, Jessica Yung, Cip Baetu, Jordi Berbel, David Bridson, Jake Bruce, Gavin Buttimore, Sarah Chakera, Bilva Chandra, Paul Collins, Alex Cullum, Bogdan Damoc, Vibha Dasagi, Maxime Gazeau, Charles Gbadamosi, Woohyun Han, Ed~Hirst, Ashyana Kachra, Lucie Kerley, Kristian Kjems, Eva Knoepfel, Vika Koriakin, Jessica Lo, Cong Lu, Zeb Mehring, Alex Moufarek, Henna Nandwani, Valeria Oliveira, Fabio Pardo, Jane Park, Andrew Pierson, Ben Poole, Helen Ran, Tim Salimans, Manuel Sanchez, Igor Saprykin, Amy Shen, Sailesh Sidhwani, Duncan Smith, Joe Stanton, Hamish Tomlinson, Dimple Vijaykumar, Luyu Wang, Piers Wingfield, Nat Wong, Keyang Xu, Christopher Yew, Nick Young, Vadim Zubov, Douglas
  Eck, Dumitru Erhan, Koray Kavukcuoglu, Demis Hassabis, Zoubin Gharamani, Raia Hadsell, A{\"a}ron van~den Oord, Inbar Mosseri, Adrian Bolton, Satinder Singh, and Tim Rockt{\"a}schel.
\newblock Genie 3: A new frontier for world models, 2025.
\newblock URL \url{https://deepmind.google/discover/blog/genie-3-a-new-frontier-for-world-models/}.

\bibitem[Bansal et~al.(2018)Bansal, Krizhevsky, and Ogale]{bansal2018chauffeurnet}
Mayank Bansal, Alex Krizhevsky, and Abhijit Ogale.
\newblock {ChauffeurNet}: Learning to drive by imitating the best and synthesizing the worst.
\newblock \emph{arXiv preprint arXiv:1812.03079}, 2018.

\bibitem[Bartoccioni et~al.(2025)Bartoccioni, Ramzi, Besnier, Venkataramanan, Vu, Xu, Chambon, Gidaris, Odabas, Hurych, Marlet, Boulch, Chen, Éloi Zablocki, Bursuc, Valle, and Cord]{bartoccioni2025vavim}
Florent Bartoccioni, Elias Ramzi, Victor Besnier, Shashanka Venkataramanan, Tuan-Hung Vu, Yihong Xu, Loick Chambon, Spyros Gidaris, Serkan Odabas, David Hurych, Renaud Marlet, Alexandre Boulch, Mickael Chen, Éloi Zablocki, Andrei Bursuc, Eduardo Valle, and Matthieu Cord.
\newblock {VaViM} and {VaVAM}: Autonomous driving through video generative modeling.
\newblock \emph{arXiv preprint arXiv:2502.15672}, 2025.

\bibitem[Beyer et~al.(2024)Beyer, Steiner, Pinto, Kolesnikov, Wang, Salz, Neumann, Alabdulmohsin, Tschannen, Bugliarello, Unterthiner, Keysers, Koppula, Liu, Grycner, Gritsenko, Houlsby, Kumar, Rong, Eisenschlos, Kabra, Bauer, Bošnjak, Chen, Minderer, Voigtlaender, Bica, Balazevic, Puigcerver, Papalampidi, Henaff, Xiong, Soricut, Harmsen, and Zhai]{beyer2024paligemma}
Lucas Beyer, Andreas Steiner, André~Susano Pinto, Alexander Kolesnikov, Xiao Wang, Daniel Salz, Maxim Neumann, Ibrahim Alabdulmohsin, Michael Tschannen, Emanuele Bugliarello, Thomas Unterthiner, Daniel Keysers, Skanda Koppula, Fangyu Liu, Adam Grycner, Alexey Gritsenko, Neil Houlsby, Manoj Kumar, Keran Rong, Julian Eisenschlos, Rishabh Kabra, Matthias Bauer, Matko Bošnjak, Xi~Chen, Matthias Minderer, Paul Voigtlaender, Ioana Bica, Ivana Balazevic, Joan Puigcerver, Pinelopi Papalampidi, Olivier Henaff, Xi~Xiong, Radu Soricut, Jeremiah Harmsen, and Xiaohua Zhai.
\newblock {PaliGemma}: A versatile {3B} {VLM} for transfer.
\newblock \emph{arXiv preprint arXiv:2407.07726}, 2024.

\bibitem[Bian et~al.(2025)Bian, Kong, Xie, Pan, Qiao, and Liu]{bian2025dynamiccity}
Hengwei Bian, Lingdong Kong, Haozhe Xie, Liang Pan, Yu~Qiao, and Ziwei Liu.
\newblock {DynamicCity}: Large-scale {4D} occupancy generation from dynamic scenes.
\newblock In \emph{Int. Conf. Learn. Represent.}, 2025.

\bibitem[Blattmann et~al.(2023{\natexlab{a}})Blattmann, Dockhorn, Kulal, Mendelevitch, Kilian, Lorenz, Levi, English, Voleti, Letts, Jampani, and Rombach]{blattmann2023stable}
Andreas Blattmann, Tim Dockhorn, Sumith Kulal, Daniel Mendelevitch, Maciej Kilian, Dominik Lorenz, Yam Levi, Zion English, Vikram Voleti, Adam Letts, Varun Jampani, and Robin Rombach.
\newblock Stable video diffusion: Scaling latent video diffusion models to large datasets.
\newblock \emph{arXiv preprint arXiv:2311.15127}, 2023{\natexlab{a}}.

\bibitem[Blattmann et~al.(2023{\natexlab{b}})Blattmann, Rombach, Ling, Dockhorn, Kim, Fidler, and Kreis]{blattmann2023align}
Andreas Blattmann, Robin Rombach, Huan Ling, Tim Dockhorn, Seung~Wook Kim, Sanja Fidler, and Karsten Kreis.
\newblock Align your latents: High-resolution video synthesis with latent diffusion models.
\newblock In \emph{IEEE/CVF Conf. Comput. Vis. Pattern Recog.}, pages 22563--22575, 2023{\natexlab{b}}.

\bibitem[Bojarski et~al.(2016)Bojarski, Testa, Dworakowski, Firner, Flepp, Goyal, Jackel, Monfort, Muller, Zhang, Zhang, Zhao, and Zieba]{bojarski2016end}
Mariusz Bojarski, Davide~Del Testa, Daniel Dworakowski, Bernhard Firner, Beat Flepp, Prasoon Goyal, Lawrence~D. Jackel, Mathew Monfort, Urs Muller, Jiakai Zhang, Xin Zhang, Jake Zhao, and Karol Zieba.
\newblock End-to-end learning for self-driving cars.
\newblock \emph{arXiv preprint arXiv:1604.07316}, 2016.

\bibitem[Caesar et~al.(2020)Caesar, Bankiti, Lang, Vora, Liong, Xu, Krishnan, Pan, Baldan, and Beijbom]{caesar2020nuscenes}
Holger Caesar, Varun Bankiti, Alex~H. Lang, Sourabh Vora, Venice~Erin Liong, Qiang Xu, Anush Krishnan, Yu~Pan, Giancarlo Baldan, and Oscar Beijbom.
\newblock {nuScenes}: A multimodal dataset for autonomous driving.
\newblock In \emph{IEEE/CVF Conf. Comput. Vis. Pattern Recog.}, pages 11621--11631, 2020.

\bibitem[Caesar et~al.(2021)Caesar, Kabzan, Tan, Fong, Wolff, Lang, Fletcher, Beijbom, and Omari]{caesar2021nuplan}
Holger Caesar, Juraj Kabzan, Kok~Seang Tan, Whye~Kit Fong, Eric Wolff, Alex Lang, Luke Fletcher, Oscar Beijbom, and Sammy Omari.
\newblock {nuPlan}: A closed-loop {ML}-based planning benchmark for autonomous vehicles.
\newblock \emph{arXiv preprint arXiv:2106.11810}, 2021.

\bibitem[Cao et~al.(2025)Cao, Zhang, Jia, Zhao, Lan, Zhang, Li, Wei, Chen, Li, Liu, Lu, Wang, and Zhang]{cao2025fastdrivevla}
Jiajun Cao, Qizhe Zhang, Peidong Jia, Xuhui Zhao, Bo~Lan, Xiaoan Zhang, Zhuo Li, Xiaobao Wei, Sixiang Chen, Liyun Li, Xianming Liu, Ming Lu, Yang Wang, and Shanghang Zhang.
\newblock {FastDriveVLA}: Efficient end-to-end driving via plug-and-play reconstruction-based token pruning.
\newblock \emph{arXiv preprint arXiv:2507.23318}, 2025.

\bibitem[Cao et~al.(2022)Cao, Li, Jiang, Zhou, Liu, Deng, and Yang]{cao2022autonomous}
Zhong Cao, Xiang Li, Kun Jiang, Weitao Zhou, Xiaoyu Liu, Nanshan Deng, and Diange Yang.
\newblock Autonomous driving policy continual learning with one-shot disengagement case.
\newblock \emph{IEEE Trans. Intell. Veh.}, 8\penalty0 (2):\penalty0 1380--1391, 2022.

\bibitem[Cen et~al.(2025)Cen, Yu, Yuan, Jiang, Huang, Guo, Li, Song, Luo, Wang, Zhao, and Chen]{cen2025worldvla}
Jun Cen, Chaohui Yu, Hangjie Yuan, Yuming Jiang, Siteng Huang, Jiayan Guo, Xin Li, Yibing Song, Hao Luo, Fan Wang, Deli Zhao, and Hao Chen.
\newblock {WorldVLA}: Towards autoregressive action world model.
\newblock \emph{arXiv preprint arXiv:2506.21539}, 2025.

\bibitem[Chao et~al.(2020)Chao, Bi, Li, Mao, Wang, Lin, and Deng]{chao2020survey}
Qianwen Chao, Huikun Bi, Weizi Li, Tianlu Mao, Zhaoqi Wang, Ming~C. Lin, and Zhigang Deng.
\newblock A survey on visual traffic simulation: Models, evaluations, and applications in autonomous driving.
\newblock In \emph{Computer Graphics Forum}, volume~39, pages 287--308. Wiley Online Library, 2020.

\bibitem[Chekroun et~al.(2023)Chekroun, Toromanoff, Hornauer, and Moutarde]{chekroun2023gri}
Raphael Chekroun, Marin Toromanoff, Sascha Hornauer, and Fabien Moutarde.
\newblock {GRI}: General reinforced imitation and its application to vision-based autonomous driving.
\newblock \emph{Robotics}, 12\penalty0 (5):\penalty0 127, 2023.

\bibitem[Chen and Kr{\"a}henb{\"u}hl(2022)]{chen2022learning}
Dian Chen and Philipp Kr{\"a}henb{\"u}hl.
\newblock Learning from all vehicles.
\newblock In \emph{IEEE/CVF Conf. Comput. Vis. Pattern Recog.}, pages 17222--17231, 2022.

\bibitem[Chen et~al.(2020)Chen, Zhou, Koltun, and Krähenbühl]{chen2020learning}
Dian Chen, Brady Zhou, Vladlen Koltun, and Philipp Krähenbühl.
\newblock Learning by cheating.
\newblock In \emph{Conf. Robot Learn.}, pages 66--75. PMLR, 2020.

\bibitem[Chen et~al.(2021)Chen, Koltun, and Kr\"ahenb\"uhl]{chen2021learning}
Dian Chen, Vladlen Koltun, and Philipp Kr\"ahenb\"uhl.
\newblock Learning to drive from a world on rails.
\newblock In \emph{IEEE/CVF Int. Conf. Comput. Vis.}, pages 15590--15599, 2021.

\bibitem[Chen et~al.(2025{\natexlab{a}})Chen, Zou, He, Chen, Xie, Han, and Cai]{chen2025dc}
Junyu Chen, Dongyun Zou, Wenkun He, Junsong Chen, Enze Xie, Song Han, and Han Cai.
\newblock {DC-AE 1.5}: Accelerating diffusion model convergence with structured latent space.
\newblock In \emph{IEEE/CVF Int. Conf. Comput. Vis.}, pages 19628--19637, 2025{\natexlab{a}}.

\bibitem[Chen et~al.(2024{\natexlab{a}})Chen, Wu, Chitta, Jaeger, Geiger, and Li]{chen2024end}
Li~Chen, Penghao Wu, Kashyap Chitta, Bernhard Jaeger, Andreas Geiger, and Hongyang Li.
\newblock End-to-end autonomous driving: Challenges and frontiers.
\newblock \emph{IEEE Trans. Pattern Anal. Mach. Intell.}, 46\penalty0 (12):\penalty0 10164--10183, 2024{\natexlab{a}}.

\bibitem[Chen et~al.(2023{\natexlab{a}})Chen, Liu, Kong, Chen, Zhu, Ma, Liu, and Wang]{chen2023towards}
Runnan Chen, Youquan Liu, Lingdong Kong, Nenglun Chen, Xinge Zhu, Yuexin Ma, Tongliang Liu, and Wenping Wang.
\newblock Towards label-free scene understanding by vision foundation models.
\newblock In \emph{Adv. Neural Inf. Process. Syst.}, volume~36, pages 75896--75910, 2023{\natexlab{a}}.

\bibitem[Chen et~al.(2023{\natexlab{b}})Chen, Liu, Kong, Zhu, Ma, Li, Hou, Qiao, and Wang]{chen2023clip2Scene}
Runnan Chen, Youquan Liu, Lingdong Kong, Xinge Zhu, Yuexin Ma, Yikang Li, Yuenan Hou, Yu~Qiao, and Wenping Wang.
\newblock {CLIP2Scene}: Towards label-efficient {3D} scene understanding by {CLIP}.
\newblock In \emph{IEEE/CVF Conf. Comput. Vis. Pattern Recog.}, pages 7020--7030, 2023{\natexlab{b}}.

\bibitem[Chen et~al.(2024{\natexlab{b}})Chen, Jiang, Gao, Liao, Xu, Zhang, Huang, Liu, and Wang]{chen2024vadv2}
Shaoyu Chen, Bo~Jiang, Hao Gao, Bencheng Liao, Qing Xu, Qian Zhang, Chang Huang, Wenyu Liu, and Xinggang Wang.
\newblock {VADv2}: End-to-end vectorized autonomous driving via probabilistic planning.
\newblock \emph{arXiv preprint arXiv:2402.13243}, 2024{\natexlab{b}}.

\bibitem[Chen et~al.(2023{\natexlab{c}})Chen, Wang, Beyer, Kolesnikov, Wu, Voigtlaender, Mustafa, Goodman, Alabdulmohsin, Padlewski, Salz, Xiong, Vlasic, Pavetic, Rong, Yu, Keysers, Zhai, and Soricut]{chen2023pali}
Xi~Chen, Xiao Wang, Lucas Beyer, Alexander Kolesnikov, Jialin Wu, Paul Voigtlaender, Basil Mustafa, Sebastian Goodman, Ibrahim Alabdulmohsin, Piotr Padlewski, Daniel Salz, Xi~Xiong, Daniel Vlasic, Filip Pavetic, Keran Rong, Tianli Yu, Daniel Keysers, Xiaohua Zhai, and Radu Soricut.
\newblock {PaLI-3} vision language models: Smaller, faster, stronger.
\newblock \emph{arXiv preprint arXiv:2310.09199}, 2023{\natexlab{c}}.

\bibitem[Chen et~al.(2025{\natexlab{b}})Chen, Huang, Ma, Fang, Shi, and Li]{chen2025solve}
Xuesong Chen, Linjiang Huang, Tao Ma, Rongyao Fang, Shaoshuai Shi, and Hongsheng Li.
\newblock {SOLVE}: Synergy of language-vision and end-to-end networks for autonomous driving.
\newblock In \emph{IEEE/CVF Conf. Comput. Vis. Pattern Recog.}, pages 12068--12077, 2025{\natexlab{b}}.

\bibitem[Chen et~al.(2024{\natexlab{c}})Chen, Wang, and Zhang]{chen2024drivinggpt}
Yuntao Chen, Yuqi Wang, and Zhaoxiang Zhang.
\newblock {DrivingGPT}: Unifying driving world modeling and planning with multi-modal autoregressive transformers.
\newblock \emph{arXiv preprint arXiv:2412.18607}, 2024{\natexlab{c}}.

\bibitem[Chen et~al.(2024{\natexlab{d}})Chen, Wu, Wang, Su, Chen, Xing, Zhong, Zhang, Zhu, Lu, Li, Luo, Lu, Qiao, and Dai]{chen2024internvl}
Zhe Chen, Jiannan Wu, Wenhai Wang, Weijie Su, Guo Chen, Sen Xing, Muyan Zhong, Qinglong Zhang, Xizhou Zhu, Lewei Lu, Bin Li, Ping Luo, Tong Lu, Yu~Qiao, and Jifeng Dai.
\newblock {InternVL}: Scaling up vision foundation models and aligning for generic visual-linguistic tasks.
\newblock In \emph{IEEE/CVF Conf. Comput. Vis. Pattern Recog.}, pages 24185--24198, 2024{\natexlab{d}}.

\bibitem[Chi et~al.(2025)Chi, ang Gao, Liu, Liu, Liu, Li, Yang, Yu, Wang, Li, Wang, Hu, Sun, Zhao, and Zhao]{chi2025impromptu}
Haohan Chi, Huan ang Gao, Ziming Liu, Jianing Liu, Chenyu Liu, Jinwei Li, Kaisen Yang, Yangcheng Yu, Zeda Wang, Wenyi Li, Leichen Wang, Xingtao Hu, Hao Sun, Hang Zhao, and Hao Zhao.
\newblock {Impromptu VLA}: Open weights and open data for driving vision-language-action models.
\newblock \emph{arXiv preprint arXiv:2505.23757}, 2025.

\bibitem[Chiang et~al.(2023)]{chiang2023vicuna}
Wei-Lin Chiang et~al.
\newblock Vicuna: An open-source chatbot impressing {GPT}-4 with 90\%* {ChatGPT} quality, 2023.
\newblock URL \url{https://vicuna.lmsys.org}.

\bibitem[Chib and Singh(2023)]{chib2023recent}
Pranav~Singh Chib and Pravendra Singh.
\newblock Recent advancements in end-to-end autonomous driving using deep learning: A survey.
\newblock \emph{IEEE Trans. Intell. Veh.}, 9\penalty0 (1):\penalty0 103--118, 2023.

\bibitem[Chitta et~al.(2021)Chitta, Prakash, and Geiger]{chitta2021neat}
Kashyap Chitta, Aditya Prakash, and Andreas Geiger.
\newblock {NEAT}: Neural attention fields for end-to-end autonomous driving.
\newblock In \emph{IEEE/CVF Int. Conf. Comput. Vis.}, pages 15793--15803, 2021.

\bibitem[Chitta et~al.(2022)Chitta, Prakash, Jaeger, Yu, Renz, and Geiger]{chitta2022transfuser}
Kashyap Chitta, Aditya Prakash, Bernhard Jaeger, Zehao Yu, Katrin Renz, and Andreas Geiger.
\newblock {TransFuser}: Imitation with transformer-based sensor fusion for autonomous driving.
\newblock \emph{IEEE Trans. Pattern Anal. Mach. Intell.}, 45\penalty0 (11):\penalty0 12878--12895, 2022.

\bibitem[Chu et~al.(2024)Chu, Qiao, Zhang, Xu, Wei, Yang, Sun, Hu, Lin, Zhang, and Shen]{chu2024mobilevlm}
Xiangxiang Chu, Limeng Qiao, Xinyu Zhang, Shuang Xu, Fei Wei, Yang Yang, Xiaofei Sun, Yiming Hu, Xinyang Lin, Bo~Zhang, and Chunhua Shen.
\newblock {MobileVLM} v2: Faster and stronger baseline for vision language model.
\newblock \emph{arXiv preprint arXiv:2402.03766}, 2024.

\bibitem[Cobbe et~al.(2020)Cobbe, Hesse, Hilton, and Schulman]{cobbe2020leveraging}
Karl Cobbe, Chris Hesse, Jacob Hilton, and John Schulman.
\newblock Leveraging procedural generation to benchmark reinforcement learning.
\newblock In \emph{Int. Conf. Mach. Learn.}, pages 2048--2056. PMLR, 2020.

\bibitem[Codevilla et~al.(2018)Codevilla, Müller, López, Koltun, and Dosovitskiy]{codevilla2018end}
Felipe Codevilla, Matthias Müller, Antonio López, Vladlen Koltun, and Alexey Dosovitskiy.
\newblock End-to-end driving via conditional imitation learning.
\newblock In \emph{IEEE Int. Conf. Robot. Autom.}, pages 4693--4700, 2018.

\bibitem[Codevilla et~al.(2019)Codevilla, Santana, Lopez, and Gaidon]{codevilla2019exploring}
Felipe Codevilla, Eder Santana, Antonio~M. Lopez, and Adrien Gaidon.
\newblock Exploring the limitations of behavior cloning for autonomous driving.
\newblock In \emph{IEEE/CVF Conf. Comput. Vis. Pattern Recog.}, pages 9329--9338, 2019.

\bibitem[Cui et~al.(2024)Cui, Ma, Cao, Ye, Zhou, Liang, Chen, Lu, Yang, Liao, Gao, Li, Tang, Cao, Zhou, Liu, Yan, Mei, Cao, Wang, and Zheng]{cui2024survey}
Can Cui, Yunsheng Ma, Xu~Cao, Wenqian Ye, Yang Zhou, Kaizhao Liang, Jintai Chen, Juanwu Lu, Zichong Yang, Kuei-Da Liao, Tianren Gao, Erlong Li, Kun Tang, Zhipeng Cao, Tong Zhou, Ao~Liu, Xinrui Yan, Shuqi Mei, Jianguo Cao, Ziran Wang, and Chao Zheng.
\newblock A survey on multimodal large language models for autonomous driving.
\newblock In \emph{IEEE/CVF Winter Conf. Appl. Comput. Vis.}, pages 958--979, 2024.

\bibitem[Cui et~al.(2025)Cui, Zhou, Peng, Park, Yang, Sankaranarayanan, Zhang, Zhang, and Wang]{cui2025vilad}
Can Cui, Yupeng Zhou, Juntong Peng, Sung-Yeon Park, Zichong Yang, Prashanth Sankaranarayanan, Jiaru Zhang, Ruqi Zhang, and Ziran Wang.
\newblock {ViLaD}: A large vision language diffusion framework for end-to-end autonomous driving.
\newblock \emph{arXiv preprint arXiv:2508.12603}, 2025.

\bibitem[Cui et~al.(2023)Cui, Wang, Li, Xie, Zou, Deng, Luo, Lu, Zhu, and Dai]{wang2023drivemlm}
Erfei Cui, Wenhai Wang, Zhiqi Li, Jiangwei Xie, Haoming Zou, Hanming Deng, Gen Luo, Lewei Lu, Xizhou Zhu, and Jifeng Dai.
\newblock {DriveMLM}: Aligning multi-modal large language models with behavioral planning states for autonomous driving.
\newblock \emph{arXiv preprint arXiv:2312.09245}, 2023.

\bibitem[Dauner et~al.(2024)Dauner, Hallgarten, Li, Weng, Huang, Yang, Li, Gilitschenski, Ivanovic, Pavone, Geiger, and Chitta]{dauner2024navsim}
Daniel Dauner, Marcel Hallgarten, Tianyu Li, Xinshuo Weng, Zhiyu Huang, Zetong Yang, Hongyang Li, Igor Gilitschenski, Boris Ivanovic, Marco Pavone, Andreas Geiger, and Kashyap Chitta.
\newblock {NAVSIM}: Data-driven non-reactive autonomous vehicle simulation and benchmarking.
\newblock In \emph{Adv. Neural Inf. Process. Syst.}, volume~37, pages 28706--28719, 2024.

\bibitem[Davidson et~al.(2025)Davidson, Ramanan, and Peri]{davidson2025refav}
Cainan Davidson, Deva Ramanan, and Neehar Peri.
\newblock {RefAV}: Towards planning-centric scenario mining.
\newblock \emph{arXiv preprint arXiv:2505.20981}, 2025.

\bibitem[De~Haan et~al.(2019)De~Haan, Jayaraman, and Levine]{de2019causal}
Pim De~Haan, Dinesh Jayaraman, and Sergey Levine.
\newblock Causal confusion in imitation learning.
\newblock In \emph{Adv. Neural Inf. Process. Syst.}, volume~32, pages 11698--11709, 2019.

\bibitem[Deruyttere et~al.(2019)Deruyttere, Vandenhende, Grujicic, Gool, and Moens]{deruyttere2019talk2car}
Thierry Deruyttere, Simon Vandenhende, Dusan Grujicic, Luc~Van Gool, and Marie-Francine Moens.
\newblock {Talk2Car}: Taking control of your self-driving car.
\newblock In \emph{Conf. Empirical Methods Natural Lang. Process.}, pages 2088--2098, 2019.

\bibitem[Deruyttere et~al.(2022)Deruyttere, Grujicic, Blaschko, and Moens]{deruyttere2022talk2car}
Thierry Deruyttere, Dusan Grujicic, Matthew~B. Blaschko, and Marie-Francine Moens.
\newblock {Talk2Car}: Predicting physical trajectories for natural language commands.
\newblock \emph{IEEE Access}, 10:\penalty0 123809--123834, 2022.

\bibitem[Devlin et~al.(2018)Devlin, Chang, Lee, and Toutanova]{devlin2018bert}
Jacob Devlin, Ming-Wei Chang, Kenton Lee, and Kristina Toutanova.
\newblock {BERT}: Pre-training of deep bidirectional transformers for language understanding.
\newblock \emph{arXiv preprint arXiv:1810.04805}, 2018.

\bibitem[Ding et~al.(2024{\natexlab{a}})Ding, Zhang, Shang, Feng, Zhang, Zong, Yuan, Su, Li, Piao, Deng, Sukiennik, Gao, Xu, and Li]{ding2024understanding}
Jingtao Ding, Yunke Zhang, Yu~Shang, Jie Feng, Yuheng Zhang, Zefang Zong, Yuan Yuan, Hongyuan Su, Nian Li, Jinghua Piao, Yucheng Deng, Nicholas Sukiennik, Chen Gao, Fengli Xu, and Yong Li.
\newblock Understanding world or predicting future? a comprehensive survey of world models.
\newblock \emph{ACM Comput. Surveys}, 2024{\natexlab{a}}.

\bibitem[Ding et~al.(2024{\natexlab{b}})Ding, Han, Xu, Liang, Zhang, and Li]{ding2024holistic}
Xinpeng Ding, Jianhua Han, Hang Xu, Xiaodan Liang, Wei Zhang, and Xiaomeng Li.
\newblock Holistic autonomous driving understanding by bird's-eye-view injected multi-modal large models.
\newblock In \emph{IEEE/CVF Conf. Comput. Vis. Pattern Recog.}, pages 13668--13677, 2024{\natexlab{b}}.

\bibitem[Djuric et~al.(2020)Djuric, Radosavljevic, Cui, Nguyen, Chou, Lin, SINGH, and Schneider]{djuric2020uncertainty}
Nemanja Djuric, Vladan Radosavljevic, Henggang Cui, Thi Nguyen, Fang-Chieh Chou, Tsung-Han Lin, NITIN SINGH, and Jeff Schneider.
\newblock Uncertainty-aware short-term motion prediction of traffic actors for autonomous driving.
\newblock In \emph{IEEE/CVF Winter Conf. Appl. Comput. Vis.}, pages 2095--2104, 2020.

\bibitem[Dosovitskiy et~al.(2017)Dosovitskiy, Ros, Codevilla, Lopez, and Koltun]{dosovitskiy2017carla}
Alexey Dosovitskiy, German Ros, Felipe Codevilla, Antonio Lopez, and Vladlen Koltun.
\newblock {CARLA}: An open urban driving simulator.
\newblock In \emph{Conf. Robot Learn.}, pages 1--16. PMLR, 2017.

\bibitem[Dosovitskiy et~al.(2021)Dosovitskiy, Beyer, Kolesnikov, Weissenborn, Zhai, Unterthiner, Dehghani, Minderer, Heigold, Gelly, Uszkoreit, and Houlsby]{dosovitskiy2020image}
Alexey Dosovitskiy, Lucas Beyer, Alexander Kolesnikov, Dirk Weissenborn, Xiaohua Zhai, Thomas Unterthiner, Mostafa Dehghani, Matthias Minderer, Georg Heigold, Sylvain Gelly, Jakob Uszkoreit, and Neil Houlsby.
\newblock An image is worth 16x16 words: Transformers for image recognition at scale.
\newblock In \emph{Int. Conf. Learn. Represent.}, 2021.

\bibitem[Esser et~al.(2021)Esser, Rombach, and Ommer]{esser2021taming}
Patrick Esser, Robin Rombach, and Bjorn Ommer.
\newblock Taming transformers for high-resolution image synthesis.
\newblock In \emph{IEEE/CVF Conf. Comput. Vis. Pattern Recog.}, pages 12873--12883, 2021.

\bibitem[Ettinger et~al.(2021)Ettinger, Cheng, Caine, Liu, Zhao, Pradhan, Chai, Sapp, Qi, Zhou, Yang, Chouard, Sun, Ngiam, Vasudevan, McCauley, Shlens, and Anguelov]{ettinger2021large}
Scott Ettinger, Shuyang Cheng, Benjamin Caine, Chenxi Liu, Hang Zhao, Sabeek Pradhan, Yuning Chai, Ben Sapp, Charles~R. Qi, Yin Zhou, Zoey Yang, Aur\'elien Chouard, Pei Sun, Jiquan Ngiam, Vijay Vasudevan, Alexander McCauley, Jonathon Shlens, and Dragomir Anguelov.
\newblock Large scale interactive motion forecasting for autonomous driving: The {Waymo Open Motion} dataset.
\newblock In \emph{IEEE/CVF Int. Conf. Comput. Vis.}, pages 9710--9719, 2021.

\bibitem[Fang et~al.(2025)Fang, Cui, Liang, Lv, Hang, and Sun]{fang2025corevla}
Shiyu Fang, Yiming Cui, Haoyang Liang, Chen Lv, Peng Hang, and Jian Sun.
\newblock {CoReVLA}: A dual-stage end-to-end autonomous driving framework for long-tail scenarios via collect-and-refine.
\newblock \emph{arXiv preprint arXiv:2509.15968}, 2025.

\bibitem[Fang et~al.(2023)Fang, Wang, Xie, Sun, Wu, Wang, Huang, Wang, and Cao]{fang2023eva}
Yuxin Fang, Wen Wang, Binhui Xie, Quan Sun, Ledell Wu, Xinggang Wang, Tiejun Huang, Xinlong Wang, and Yue Cao.
\newblock {EVA}: Exploring the limits of masked visual representation learning at scale.
\newblock In \emph{IEEE/CVF Conf. Comput. Vis. Pattern Recog.}, pages 19358--19369, 2023.

\bibitem[Fang et~al.(2024)Fang, Sun, Wang, Huang, Wang, and Cao]{fang2024eva}
Yuxin Fang, Quan Sun, Xinggang Wang, Tiejun Huang, Xinlong Wang, and Yue Cao.
\newblock {EVA-02}: A visual representation for neon genesis.
\newblock \emph{Image and Vision Computing}, 149:\penalty0 105171, 2024.

\bibitem[Feng et~al.(2025{\natexlab{a}})Feng, Mei, Li, Ost, Ghilotti, Girgis, Majumdar, and Heide]{feng2025verdi}
Bowen Feng, Zhiting Mei, Baiang Li, Julian Ost, Filippo Ghilotti, Roger Girgis, Anirudha Majumdar, and Felix Heide.
\newblock {VERDI}: {VLM}-embedded reasoning for autonomous driving.
\newblock \emph{arXiv preprint arXiv:2505.15925}, 2025{\natexlab{a}}.

\bibitem[Feng et~al.(2025{\natexlab{b}})Feng, Gao, Zablocki, Li, Li, Liu, Cord, and Alahi]{feng2025rap}
Lan Feng, Yang Gao, Eloi Zablocki, Quanyi Li, Wuyang Li, Sichao Liu, Matthieu Cord, and Alexandre Alahi.
\newblock {RAP}: {3D} rasterization augmented end-to-end planning.
\newblock \emph{arXiv preprint arXiv:2510.04333}, 2025{\natexlab{b}}.

\bibitem[Fu et~al.(2024)Fu, Li, Wen, Dou, Cai, Shi, and Qiao]{fu2024drive}
Daocheng Fu, Xin Li, Licheng Wen, Min Dou, Pinlong Cai, Botian Shi, and Yu~Qiao.
\newblock Drive like a human: Rethinking autonomous driving with large language models.
\newblock In \emph{IEEE/CVF Winter Conf. Appl. Comput. Vis. Worksh.}, pages 910--919, 2024.

\bibitem[Fu et~al.(2025)Fu, Zhang, Zhao, Cui, Liang, Zhang, Zhang, Xie, Wang, and Bai]{fu2025orion}
Haoyu Fu, Diankun Zhang, Zongchuang Zhao, Jianfeng Cui, Dingkang Liang, Chong Zhang, Dingyuan Zhang, Hongwei Xie, Bing Wang, and Xiang Bai.
\newblock {ORION}: A holistic end-to-end autonomous driving framework by vision-language instructed action generation.
\newblock \emph{arXiv preprint arXiv:2503.19755}, 2025.

\bibitem[Gao et~al.(2025{\natexlab{a}})Gao, Chen, Jiang, Liao, Shi, Guo, Pu, Yin, Li, Zhang, Zhang, Liu, Zhang, and Wang]{gao2025rad}
Hao Gao, Shaoyu Chen, Bo~Jiang, Bencheng Liao, Yiang Shi, Xiaoyang Guo, Yuechuan Pu, Haoran Yin, Xiangyu Li, Xinbang Zhang, Ying Zhang, Wenyu Liu, Qian Zhang, and Xinggang Wang.
\newblock {RAD}: Training an end-to-end driving policy via large-scale {3DGS}-based reinforcement learning.
\newblock \emph{arXiv preprint arXiv:2502.13144}, 2025{\natexlab{a}}.

\bibitem[Gao et~al.(2024)Gao, Yang, Chen, Chitta, Qiu, Geiger, Zhang, and Li]{gao2024vista}
Shenyuan Gao, Jiazhi Yang, Li~Chen, Kashyap Chitta, Yihang Qiu, Andreas Geiger, Jun Zhang, and Hongyang Li.
\newblock Vista: A generalizable driving world model with high fidelity and versatile controllability.
\newblock \emph{Adv. Neural Inf. Process. Syst.}, 37:\penalty0 91560--91596, 2024.

\bibitem[Gao et~al.(2025{\natexlab{b}})Gao, Wu, Wang, Liu, Zhou, and Tu]{gao2025langcoop}
Xiangbo Gao, Yuheng Wu, Rujia Wang, Chenxi Liu, Yang Zhou, and Zhengzhong Tu.
\newblock {LangCoop}: Collaborative driving with language.
\newblock In \emph{IEEE/CVF Conf. Comput. Vis. Pattern Recog.}, pages 4226--4237, 2025{\natexlab{b}}.

\bibitem[Garg and Madhava~Krishna(2024)]{garg2024imagine}
Anant Garg and K~Madhava~Krishna.
\newblock {Imagine-2-Drive}: Leveraging high-fidelity world models via multi-modal diffusion policies.
\newblock \emph{arXiv preprint arXiv:2411.10171}, 2024.

\bibitem[Ge et~al.(2025)Ge, Ohtani, Niu, Zhang, and Takeda]{ge2025vla}
Maoning Ge, Kento Ohtani, Yingjie Niu, Yuxiao Zhang, and Kazuya Takeda.
\newblock {VLA-MP}: A vision-language-action framework for multimodal perception and physics-constrained action generation in autonomous driving.
\newblock \emph{Sensors}, 25\penalty0 (19):\penalty0 6163, 2025.

\bibitem[Geiger et~al.(2012)Geiger, Lenz, and Urtasun]{geiger2012we}
Andreas Geiger, Philip Lenz, and Raquel Urtasun.
\newblock Are we ready for autonomous driving? the {KITTI} vision benchmark suite.
\newblock In \emph{IEEE/CVF Conf. Comput. Vis. Pattern Recog.}, pages 3354--3361, 2012.

\bibitem[Ghosh et~al.(2024)Ghosh, Zheng, Tamburo, Vuong, Alvarez-Padilla, Zhu, Cardei, Dunn, Mertz, and Narasimhan]{ghosh2024roadwork}
Anurag Ghosh, Shen Zheng, Robert Tamburo, Khiem Vuong, Juan Alvarez-Padilla, Hailiang Zhu, Michael Cardei, Nicholas Dunn, Christoph Mertz, and Srinivasa~G. Narasimhan.
\newblock {ROADWork} dataset: Learning to recognize, observe, analyze and drive through work zones.
\newblock \emph{arXiv preprint arXiv:2406.07661}, 2024.

\bibitem[Grattafiori et~al.(2024)Grattafiori, Dubey, Jauhri, Pandey, Kadian, Al-Dahle, Letman, Mathur, Schelten, Vaughan, Yang, Fan, Goyal, Hartshorn, Yang, Mitra, Sravankumar, Korenev, Hinsvark, Rao, Zhang, Rodriguez, Gregerson, Spataru, Roziere, Biron, Tang, Chern, Caucheteux, Nayak, Bi, Marra, McConnell, Keller, Touret, Wu, Wong, Ferrer, Nikolaidis, Allonsius, Song, Pintz, Livshits, Wyatt, Esiobu, Choudhary, Mahajan, Garcia-Olano, Perino, Hupkes, Lakomkin, AlBadawy, Lobanova, Dinan, Smith, Radenovic, Guzmán, Zhang, Synnaeve, Lee, Anderson, Thattai, Nail, Mialon, Pang, Cucurell, Nguyen, Korevaar, Xu, Touvron, Zarov, Ibarra, Kloumann, Misra, Evtimov, Zhang, Copet, Lee, Geffert, Vranes, Park, Mahadeokar, Shah, van~der Linde, Billock, Hong, Lee, Fu, Chi, Huang, Liu, Wang, Yu, Bitton, Spisak, Park, Rocca, et~al.]{dubey2024llama}
Aaron Grattafiori, Abhimanyu Dubey, Abhinav Jauhri, Abhinav Pandey, Abhishek Kadian, Ahmad Al-Dahle, Aiesha Letman, Akhil Mathur, Alan Schelten, Alex Vaughan, Amy Yang, Angela Fan, Anirudh Goyal, Anthony Hartshorn, Aobo Yang, Archi Mitra, Archie Sravankumar, Artem Korenev, Arthur Hinsvark, Arun Rao, Aston Zhang, Aurelien Rodriguez, Austen Gregerson, Ava Spataru, Baptiste Roziere, Bethany Biron, Binh Tang, Bobbie Chern, Charlotte Caucheteux, Chaya Nayak, Chloe Bi, Chris Marra, Chris McConnell, Christian Keller, Christophe Touret, Chunyang Wu, Corinne Wong, Cristian~Canton Ferrer, Cyrus Nikolaidis, Damien Allonsius, Daniel Song, Danielle Pintz, Danny Livshits, Danny Wyatt, David Esiobu, Dhruv Choudhary, Dhruv Mahajan, Diego Garcia-Olano, Diego Perino, Dieuwke Hupkes, Egor Lakomkin, Ehab AlBadawy, Elina Lobanova, Emily Dinan, Eric~Michael Smith, Filip Radenovic, Francisco Guzmán, Frank Zhang, Gabriel Synnaeve, Gabrielle Lee, Georgia~Lewis Anderson, Govind Thattai, Graeme Nail, Gregoire Mialon, Guan Pang,
  Guillem Cucurell, Hailey Nguyen, Hannah Korevaar, Hu~Xu, Hugo Touvron, Iliyan Zarov, Imanol~Arrieta Ibarra, Isabel Kloumann, Ishan Misra, Ivan Evtimov, Jack Zhang, Jade Copet, Jaewon Lee, Jan Geffert, Jana Vranes, Jason Park, Jay Mahadeokar, Jeet Shah, Jelmer van~der Linde, Jennifer Billock, Jenny Hong, Jenya Lee, Jeremy Fu, Jianfeng Chi, Jianyu Huang, Jiawen Liu, Jie Wang, Jiecao Yu, Joanna Bitton, Joe Spisak, Jongsoo Park, Joseph Rocca, et~al.
\newblock The {LLaMA} 3 herd of models.
\newblock \emph{arXiv preprint arXiv:2407.21783}, 2024.

\bibitem[Gui et~al.(2025)Gui, Zhao, Han, Wang, Gong, Tan, zhong Xu, and Shen]{gui2025trajdiff}
Xingtai Gui, Jianbo Zhao, Wencheng Han, Jikai Wang, Jiahao Gong, Feiyang Tan, Cheng zhong Xu, and Jianbing Shen.
\newblock {TrajDiff}: End-to-end autonomous driving without perception annotation.
\newblock \emph{arXiv preprint arXiv:2512.00723}, 2025.

\bibitem[Guo and Zhang(2025)]{guo2025vdrive}
Ziang Guo and Zufeng Zhang.
\newblock {VDRive}: Leveraging reinforced {VLA} and diffusion policy for end-to-end autonomous driving.
\newblock \emph{arXiv preprint arXiv:2510.15446}, 2025.

\bibitem[Ha and Schmidhuber(2018)]{ha2018world}
David Ha and J{\"u}rgen Schmidhuber.
\newblock World models.
\newblock \emph{arXiv preprint arXiv:1803.10122}, 2018.

\bibitem[Hamdan et~al.(2025)Hamdan, Sima, Yang, Li, and Guney]{hamdan2025eta}
Shadi Hamdan, Chonghao Sima, Zetong Yang, Hongyang Li, and Fatma Guney.
\newblock {ETA}: Efficiency through thinking ahead, a dual approach to self-driving with large models.
\newblock In \emph{IEEE/CVF Int. Conf. Comput. Vis.}, pages 26529--26538, 2025.

\bibitem[Han et~al.(2025{\natexlab{a}})Han, Tian, Zhu, He, Zhang, Guo, Zhu, Tang, Xu, Guo, Niu, Zhu, Dong, Yan, Dong, Hou, Huang, Jia, and Xu]{han2025percept}
Jianhua Han, Meng Tian, Jiangtong Zhu, Fan He, Huixin Zhang, Sitong Guo, Dechang Zhu, Hao Tang, Pei Xu, Yuze Guo, Minzhe Niu, Haojie Zhu, Qichao Dong, Xuechao Yan, Siyuan Dong, Lu~Hou, Qingqiu Huang, Xiaosong Jia, and Hang Xu.
\newblock {Percept-WAM}: Perception-enhanced world-awareness-action model for robust end-to-end autonomous driving.
\newblock \emph{arXiv preprint arXiv:2511.19221}, 2025{\natexlab{a}}.

\bibitem[Han et~al.(2025{\natexlab{b}})Han, Guo, Xu, and Shen]{han2025dme}
Wencheng Han, Dongqian Guo, Cheng-Zhong Xu, and Jianbing Shen.
\newblock {DME-Driver}: Integrating human decision logic and {3D} scene perception in autonomous driving.
\newblock In \emph{AAAI Conf. Artifi. Intell.}, volume~39, pages 3347--3355, 2025{\natexlab{b}}.

\bibitem[Hao et~al.(2025{\natexlab{a}})Hao, Jing, Yu, and Nie]{hao2025styledrive}
Ruiyang Hao, Bowen Jing, Haibao Yu, and Zaiqing Nie.
\newblock {StyleDrive}: Towards driving-style aware benchmarking of end-to-end autonomous driving.
\newblock \emph{arXiv preprint arXiv:2506.23982}, 2025{\natexlab{a}}.

\bibitem[Hao et~al.(2024)Hao, Wei, Yang, Zhao, Zhang, Zhou, Wang, Li, Kong, and Zhang]{hao2024your}
Xiaoshuai Hao, Mengchuan Wei, Yifan Yang, Haimei Zhao, Hui Zhang, Yi~Zhou, Qiang Wang, Weiming Li, Lingdong Kong, and Jing Zhang.
\newblock Is your {HD} map constructor reliable under sensor corruptions?
\newblock \emph{Adv. Neural Inf. Process. Syst.}, 37:\penalty0 22441--22482, 2024.

\bibitem[Hao et~al.(2025{\natexlab{b}})Hao, Diao, Wei, Yang, Hao, Yin, Zhang, Li, Zhao, and Liu]{hao2025mapfusion}
Xiaoshuai Hao, Yunfeng Diao, Mengchuan Wei, Yifan Yang, Peng Hao, Rong Yin, Hui Zhang, Weiming Li, Shu Zhao, and Yu~Liu.
\newblock {MapFusion}: A novel {BEV} feature fusion network for multi-modal map construction.
\newblock \emph{Information Fusion}, 119:\penalty0 103018, 2025{\natexlab{b}}.

\bibitem[Hao et~al.(2025{\natexlab{c}})Hao, Kong, Yin, Wang, Zhang, Diao, and Zhao]{hao2025safemap}
Xiaoshuai Hao, Lingdong Kong, Rong Yin, Pengwei Wang, Jing Zhang, Yunfeng Diao, and Shu Zhao.
\newblock {SafeMap}: Robust {HD} map construction from incomplete observations.
\newblock In \emph{Int. Conf. Mach. Learn.}, pages 22091--22102. PMLR, 2025{\natexlab{c}}.

\bibitem[Hao et~al.(2025{\natexlab{d}})Hao, Liu, Zhao, Ji, Wei, Zhao, Kong, Yin, and Liu]{hao2025msc}
Xiaoshuai Hao, Guanqun Liu, Yuting Zhao, Yuheng Ji, Mengchuan Wei, Haimei Zhao, Lingdong Kong, Rong Yin, and Yu~Liu.
\newblock {MSC-Bench}: Benchmarking and analyzing multi-sensor corruption for driving perception.
\newblock \emph{arXiv preprint arXiv:2501.01037}, 2025{\natexlab{d}}.

\bibitem[Hao et~al.(2025{\natexlab{e}})Hao, Li, Sun, Wang, Yi, Song, Qin, Zhou, Zhan, and Lang]{hao2025driveaction}
Yuhan Hao, Zhengning Li, Lei Sun, Weilong Wang, Naixin Yi, Sheng Song, Caihong Qin, Mofan Zhou, Yifei Zhan, and Xianpeng Lang.
\newblock {DriveAction}: A benchmark for exploring human-like driving decisions in {VLA} models.
\newblock \emph{arXiv preprint arXiv:2506.05667}, 2025{\natexlab{e}}.

\bibitem[He et~al.(2016)He, Zhang, Ren, and Sun]{he2016deep}
Kaiming He, Xiangyu Zhang, Shaoqing Ren, and Jian Sun.
\newblock Deep residual learning for image recognition.
\newblock In \emph{IEEE/CVF Conf. Comput. Vis. Pattern Recog.}, pages 770--778, 2016.

\bibitem[Hegde et~al.(2025)Hegde, Yasarla, Cai, Han, Bhattacharyya, Mahajan, Liu, Garrepalli, Patel, and Porikli]{hegde2025distilling}
Deepti Hegde, Rajeev Yasarla, Hong Cai, Shizhong Han, Apratim Bhattacharyya, Shweta Mahajan, Litian Liu, Risheek Garrepalli, Vishal~M. Patel, and Fatih Porikli.
\newblock Distilling multi-modal large language models for autonomous driving.
\newblock In \emph{IEEE/CVF Conf. Comput. Vis. Pattern Recog.}, pages 27575--27585, 2025.

\bibitem[Houston et~al.(2021)Houston, Zuidhof, Bergamini, Ye, Chen, Jain, Omari, Iglovikov, and Ondruska]{houston2021one}
John Houston, Guido Zuidhof, Luca Bergamini, Yawei Ye, Long Chen, Ashesh Jain, Sammy Omari, Vladimir Iglovikov, and Peter Ondruska.
\newblock One thousand and one hours: Self-driving motion prediction dataset.
\newblock In \emph{Conf. Robot Learn.}, pages 409--418. PMLR, 2021.

\bibitem[Hu et~al.(2023{\natexlab{a}})Hu, Russell, Yeo, Murez, Fedoseev, Kendall, Shotton, and Corrado]{hu2023gaia}
Anthony Hu, Lloyd Russell, Hudson Yeo, Zak Murez, George Fedoseev, Alex Kendall, Jamie Shotton, and Gianluca Corrado.
\newblock {GAIA-1}: A generative world model for autonomous driving.
\newblock \emph{arXiv preprint arXiv:2309.17080}, 2023{\natexlab{a}}.

\bibitem[Hu et~al.(2022)Hu, Chen, Wu, Li, Yan, and Tao]{hu2022st}
Shengchao Hu, Li~Chen, Penghao Wu, Hongyang Li, Junchi Yan, and Dacheng Tao.
\newblock {ST-P3}: End-to-end vision-based autonomous driving via spatial-temporal feature learning.
\newblock In \emph{Eur. Conf. Comput. Vis.}, pages 533--549. Springer, 2022.

\bibitem[Hu et~al.(2024)Hu, Yin, Jia, Deng, Guo, Zhang, Long, and Tan]{hu2024drivingworld}
Xiaotao Hu, Wei Yin, Mingkai Jia, Junyuan Deng, Xiaoyang Guo, Qian Zhang, Xiaoxiao Long, and Ping Tan.
\newblock {DrivingWorld}: Constructing world model for autonomous driving via video {GPT}.
\newblock \emph{arXiv preprint arXiv:2412.19505}, 2024.

\bibitem[Hu et~al.(2023{\natexlab{b}})Hu, Yang, Chen, Li, Sima, Zhu, Chai, Du, Lin, Wang, Lu, Jia, Liu, Dai, Qiao, and Li]{hu2023planning}
Yihan Hu, Jiazhi Yang, Li~Chen, Keyu Li, Chonghao Sima, Xizhou Zhu, Siqi Chai, Senyao Du, Tianwei Lin, Wenhai Wang, Lewei Lu, Xiaosong Jia, Qiang Liu, Jifeng Dai, Yu~Qiao, and Hongyang Li.
\newblock Planning-oriented autonomous driving.
\newblock In \emph{IEEE/CVF Conf. Comput. Vis. Pattern Recog.}, pages 17853--17862, 2023{\natexlab{b}}.

\bibitem[Huang et~al.(2025)Huang, Yu, Ma, Zhong, Feng, Wang, Chen, Peng, Feng, Qin, and Liu]{huang2025survey}
Lei Huang, Weijiang Yu, Weitao Ma, Weihong Zhong, Zhangyin Feng, Haotian Wang, Qianglong Chen, Weihua Peng, Xiaocheng Feng, Bing Qin, and Ting Liu.
\newblock A survey on hallucination in large language models: Principles, taxonomy, challenges, and open questions.
\newblock \emph{ACM Trans. Info. Syst.}, 43\penalty0 (2):\penalty0 1--55, 2025.

\bibitem[Huang et~al.(2024{\natexlab{a}})Huang, Sansom, Ma, Gervits, and Chai]{huang2024drivlme}
Yidong Huang, Jacob Sansom, Ziqiao Ma, Felix Gervits, and Joyce Chai.
\newblock {DriVLMe}: Enhancing llm-based autonomous driving agents with embodied and social experiences.
\newblock In \emph{IEEE/RSJ Int. Conf. Intell. Robots Syst.}, pages 3153--3160, 2024{\natexlab{a}}.

\bibitem[Huang et~al.(2024{\natexlab{b}})Huang, Zhang, and Ohn-Bar]{huang2024neural}
Zanming Huang, Jimuyang Zhang, and Eshed Ohn-Bar.
\newblock Neural volumetric world models for autonomous driving.
\newblock In \emph{Eur. Conf. Comput. Vis.}, pages 195--213. Springer, 2024{\natexlab{b}}.

\bibitem[Huang et~al.(2024{\natexlab{c}})Huang, Tang, Chen, Lin, Jie, Ma, Wang, , and Liang]{huang2024making}
Zhijian Huang, Tao Tang, Shaoxiang Chen, Sihao Lin, Zequn Jie, Lin Ma, Guangrun Wang, , and Xiaodan Liang.
\newblock Making large language models better planners with reasoning-decision alignment.
\newblock In \emph{Eur. Conf. Comput. Vis.}, pages 73--90. Springer, 2024{\natexlab{c}}.

\bibitem[Hwang et~al.(2024)Hwang, Xu, Lin, Hung, Ji, Choi, Huang, He, Covington, Sapp, Zhou, Guo, Anguelov, and Tan]{hwang2024emma}
Jyh-Jing Hwang, Runsheng Xu, Hubert Lin, Wei-Chih Hung, Jingwei Ji, Kristy Choi, Di~Huang, Tong He, Paul Covington, Benjamin Sapp, Yin Zhou, James Guo, Dragomir Anguelov, and Mingxing Tan.
\newblock {EMMA}: End-to-end multimodal model for autonomous driving.
\newblock \emph{arXiv preprint arXiv:2410.23262}, 2024.

\bibitem[Ilharco et~al.(2021)Ilharco, Wortsman, Carlini, Taori, Dave, Shankar, Namkoong, et~al.]{ilharco2021openclip}
Gabriel Ilharco, Mitchell Wortsman, Nicholas Carlini, Rohan Taori, Achal Dave, Vaishaal Shankar, Hongseok Namkoong, et~al.
\newblock {OpenCLIP}.
\newblock \emph{Zenodo}, 2021.

\bibitem[Jia et~al.(2023{\natexlab{a}})Jia, Mao, Liu, Zhao, Wen, Zhang, Zhang, and Wang]{jia2023adriver}
Fan Jia, Weixin Mao, Yingfei Liu, Yucheng Zhao, Yuqing Wen, Chi Zhang, Xiangyu Zhang, and Tiancai Wang.
\newblock {ADriver-I}: A general world model for autonomous driving.
\newblock \emph{arXiv preprint arXiv:2311.13549}, 2023{\natexlab{a}}.

\bibitem[Jia et~al.(2023{\natexlab{b}})Jia, Gao, Chen, Yan, Liu, and Li]{jia2023driveadapter}
Xiaosong Jia, Yulu Gao, Li~Chen, Junchi Yan, Patrick~Langechuan Liu, and Hongyang Li.
\newblock {DriveAdapter}: Breaking the coupling barrier of perception and planning in end-to-end autonomous driving.
\newblock In \emph{IEEE/CVF Int. Conf. Comput. Vis.}, pages 7953--7963, 2023{\natexlab{b}}.

\bibitem[Jia et~al.(2023{\natexlab{c}})Jia, Wu, Chen, Xie, He, Yan, and Li]{jia2023thinktwice}
Xiaosong Jia, Penghao Wu, Li~Chen, Jiangwei Xie, Conghui He, Junchi Yan, and Hongyang Li.
\newblock Think twice before driving: Towards scalable decoders for end-to-end autonomous driving.
\newblock In \emph{IEEE/CVF Conf. Comput. Vis. Pattern Recog.}, 2023{\natexlab{c}}.

\bibitem[Jia et~al.(2024)Jia, Yang, Li, Zhang, and Yan]{jia2024bench2drive}
Xiaosong Jia, Zhenjie Yang, Qifeng Li, Zhiyuan Zhang, and Junchi Yan.
\newblock {Bench2Drive}: Towards multi-ability benchmarking of closed-loop end-to-end autonomous driving.
\newblock In \emph{Adv. Neural Inf. Process. Syst.}, volume~37, pages 819--844, 2024.

\bibitem[Jia et~al.(2025{\natexlab{a}})Jia, You, Zhang, and Yan]{jia2025drivetransformer}
Xiaosong Jia, Junqi You, Zhiyuan Zhang, and Junchi Yan.
\newblock {DriveTransformer}: Unified transformer for scalable end-to-end autonomous driving.
\newblock In \emph{Int. Conf. Learn. Represent.}, 2025{\natexlab{a}}.

\bibitem[Jia et~al.(2025{\natexlab{b}})Jia, Zhang, Jiang, Wong, Zhang, Chen, Zhang, Zhou, Yang, Yan, and Jiang]{jia2025spatial}
Xiaosong Jia, Chenhe Zhang, Yule Jiang, Songbur Wong, Zhiyuan Zhang, Chen Chen, Shaofeng Zhang, Xuanhe Zhou, Xue Yang, Junchi Yan, and Yu-Gang Jiang.
\newblock Spatial retrieval augmented autonomous driving.
\newblock \emph{arXiv preprint arXiv:2512.06865}, 2025{\natexlab{b}}.

\bibitem[Jiang et~al.(2025{\natexlab{a}})Jiang, Gao, Sun, Wang, Wang, Chai, Cao, Heng, Jiang, Dong, Zhang, Guo, Sun, and Zhao]{jiang2025diffvla}
Anqing Jiang, Yu~Gao, Zhigang Sun, Yiru Wang, Jijun Wang, Jinghao Chai, Qian Cao, Yuweng Heng, Hao Jiang, Yunda Dong, Zongzheng Zhang, Xianda Guo, Hao Sun, and Hao Zhao.
\newblock {DiffVLA}: Vision-language guided diffusion planning for autonomous driving.
\newblock \emph{arXiv preprint arXiv:2505.19381}, 2025{\natexlab{a}}.

\bibitem[Jiang et~al.(2023)Jiang, Chen, Xu, Liao, Chen, Zhou, Zhang, Liu, Huang, and Wang]{jiang2023vad}
Bo~Jiang, Shaoyu Chen, Qing Xu, Bencheng Liao, Jiajie Chen, Helong Zhou, Qian Zhang, Wenyu Liu, Chang Huang, and Xinggang Wang.
\newblock {VAD}: Vectorized scene representation for efficient autonomous driving.
\newblock In \emph{IEEE/CVF Conf. Comput. Vis. Pattern Recog.}, pages 8340--8350, 2023.

\bibitem[Jiang et~al.(2024)Jiang, Chen, Liao, Zhang, Yin, Zhang, Huang, Liu, and Wang]{jiang2024senna}
Bo~Jiang, Shaoyu Chen, Bencheng Liao, Xingyu Zhang, Wei Yin, Qian Zhang, Chang Huang, Wenyu Liu, and Xinggang Wang.
\newblock Senna: Bridging large vision-language models and end-to-end autonomous driving.
\newblock \emph{arXiv preprint arXiv:2410.22313}, 2024.

\bibitem[Jiang et~al.(2025{\natexlab{b}})Jiang, Chen, Zhang, Liu, and Wang]{jiang2025alphadrive}
Bo~Jiang, Shaoyu Chen, Qian Zhang, Wenyu Liu, and Xinggang Wang.
\newblock {AlphaDrive}: Unleashing the power of {VLMs} in autonomous driving via reinforcement learning and reasoning.
\newblock \emph{arXiv preprint arXiv:2503.07608}, 2025{\natexlab{b}}.

\bibitem[Jiang et~al.(2025{\natexlab{c}})Jiang, Huang, Qian, Luo, Zhu, Zhong, Tang, Kong, Wang, Jiao, Ye, Sheng, Zhao, Wen, Fu, Chen, Jiang, Yang, Choi, and Sun]{jiang2025survey}
Sicong Jiang, Zilin Huang, Kangan Qian, Ziang Luo, Tianze Zhu, Yang Zhong, Yihong Tang, Menglin Kong, Yunlong Wang, Siwen Jiao, Hao Ye, Zihao Sheng, Xin Zhao, Tuopu Wen, Zheng Fu, Sikai Chen, Kun Jiang, Diange Yang, Seongjin Choi, and Lijun Sun.
\newblock A survey on vision-language-action models for autonomous driving.
\newblock \emph{arXiv preprint arXiv:2506.24044}, 2025{\natexlab{c}}.

\bibitem[Jin et~al.(2025)Jin, Hu, Gu, Zheng, Guo, et~al.]{jinoccvar}
Bu~Jin, Xiaotao Hu, Songen Gu, Yupeng Zheng, Xiaoyang Guo, et~al.
\newblock {OccVAR}: Scalable {4D} occupancy prediction via next-scale prediction.
\newblock https://openreview.net/forum?id=X2HnTFsFm8, 2025.

\bibitem[Jing et~al.(2022)Jing, Xia, Tian, Ding, Luo, Domeyer, Sherony, and Ding]{jing2022inaction}
Taotao Jing, Haifeng Xia, Renran Tian, Haoran Ding, Xiao Luo, Joshua Domeyer, Rini Sherony, and Zhengming Ding.
\newblock Inaction: Interpretable action decision making for autonomous driving.
\newblock In \emph{European Conference on Computer Vision}, pages 370--387. Springer, 2022.

\bibitem[Kahneman(2011)]{kahneman2011thinking}
Daniel Kahneman.
\newblock Thinking, fast and slow.
\newblock \emph{Farrar, Straus and Giroux}, 2011.

\bibitem[Karnchanachari et~al.(2024)Karnchanachari, Geromichalos, Tan, Li, Eriksen, Yaghoubi, Mehdipour, Bernasconi, Fong, Guo, and Caesar]{karnchanachari2024towards}
Napat Karnchanachari, Dimitris Geromichalos, Kok~Seang Tan, Nanxiang Li, Christopher Eriksen, Shakiba Yaghoubi, Noushin Mehdipour, Gianmarco Bernasconi, Whye~Kit Fong, Yiluan Guo, and Holger Caesar.
\newblock Towards learning-based planning: The {nuPlan} benchmark for real-world autonomous driving.
\newblock In \emph{IEEE Int. Conf. Robot. Autom.}, pages 629--636, 2024.

\bibitem[Kerbl et~al.(2023)Kerbl, Kopanas, Leimkühler, and Drettakis]{kerbl20233d}
Bernhard Kerbl, Georgios Kopanas, Thomas Leimkühler, and George Drettakis.
\newblock {3D} gaussian splatting for real-time radiance field rendering.
\newblock \emph{ACM Trans. Graph.}, 42\penalty0 (4):\penalty0 139--1, 2023.

\bibitem[Khandelwal et~al.(2020)Khandelwal, Qi, Singh, Hartnett, and Ramanan]{khandelwal2020if}
Siddhesh Khandelwal, William Qi, Jagjeet Singh, Andrew Hartnett, and Deva Ramanan.
\newblock What-if motion prediction for autonomous driving.
\newblock \emph{arXiv preprint arXiv:2008.10587}, 2020.

\bibitem[Kim et~al.(2018)Kim, Rohrbach, Darrell, Canny, and Akata]{kim2018textual}
Jinkyu Kim, Anna Rohrbach, Trevor Darrell, John Canny, and Zeynep Akata.
\newblock Textual explanations for self-driving vehicles.
\newblock In \emph{Eur. Conf. Comput. Vis.}, pages 563--578. Springer, 2018.

\bibitem[Kiran et~al.(2021)Kiran, Sobh, Talpaert, Mannion, Sallab, Yogamani, and Pérez]{kiran2021deep}
B~Ravi Kiran, Ibrahim Sobh, Victor Talpaert, Patrick Mannion, Ahmad A.~Al Sallab, Senthil Yogamani, and Patrick Pérez.
\newblock Deep reinforcement learning for autonomous driving: A survey.
\newblock \emph{IEEE Int. Conf. Intell. Transport. Syst.}, 23\penalty0 (6):\penalty0 4909--4926, 2021.

\bibitem[Knox et~al.(2023)Knox, Allievi, Banzhaf, Schmitt, and Stone]{knox2023reward}
W~Bradley Knox, Alessandro Allievi, Holger Banzhaf, Felix Schmitt, and Peter Stone.
\newblock Reward (mis)design for autonomous driving.
\newblock \emph{Artifi. Intell.}, 316:\penalty0 103829, 2023.

\bibitem[Kong et~al.(2023{\natexlab{a}})Kong, Liu, Chen, Ma, Zhu, Li, Hou, Qiao, and Liu]{kong2023rethinking}
Lingdong Kong, Youquan Liu, Runnan Chen, Yuexin Ma, Xinge Zhu, Yikang Li, Yuenan Hou, Yu~Qiao, and Ziwei Liu.
\newblock Rethinking range view representation for {LiDAR} segmentation.
\newblock In \emph{IEEE/CVF Int. Conf. Comput. Vis.}, pages 228--240, 2023{\natexlab{a}}.

\bibitem[Kong et~al.(2023{\natexlab{b}})Kong, Liu, Li, Chen, Zhang, Ren, Pan, Chen, and Liu]{kong2023robo3d}
Lingdong Kong, Youquan Liu, Xin Li, Runnan Chen, Wenwei Zhang, Jiawei Ren, Liang Pan, Kai Chen, and Ziwei Liu.
\newblock {Robo3D}: Towards robust and reliable {3D} perception against corruptions.
\newblock In \emph{IEEE/CVF Int. Conf. Comput. Vis.}, pages 19994--20006, 2023{\natexlab{b}}.

\bibitem[Kong et~al.(2024)Kong, Xie, Hu, Niu, Ooi, Cottereau, Ng, Ma, Zhang, Pan, Chen, Liu, Qiu, Zhang, Cao, Lu, Chen, Kang, Zhou, Ying, Shang, Wei, Dong, Yang, Jiang, Ma, Ji, Li, Huang, Tian, Kou, Jia, Liu, Wang, Li, Hao, Yang, Zhang, Wei, Zhou, Zhao, Zhang, Li, He, Cheng, Zhang, Zhao, Ding, Liu, Yan, Wang, Ye, Luo, Tian, Zuo, Cao, Ren, Li, Liu, Wu, Mao, Li, Liu, Liu, Qin, Chu, Xu, Zhao, Jiang, Liu, Wang, Li, Li, Yuan, Yang, Liu, Chen, Zhou, Wang, Zhang, Sun, Chen, Yang, Wang, Fu, Lin, Yang, Li, Luo, Cheng, and Xu]{robodrive_challenge_2024}
Lingdong Kong, Shaoyuan Xie, Hanjiang Hu, Yaru Niu, Wei~Tsang Ooi, Benoit~R. Cottereau, Lai~Xing Ng, Yuexin Ma, Wenwei Zhang, Liang Pan, Kai Chen, Ziwei Liu, Weichao Qiu, Wei Zhang, Xu~Cao, Hao Lu, Ying-Cong Chen, Caixin Kang, Xinning Zhou, Chengyang Ying, Wentao Shang, Xingxing Wei, Yinpeng Dong, Bo~Yang, Shengyin Jiang, Zeliang Ma, Dengyi Ji, Haiwen Li, Xingliang Huang, Yu~Tian, Genghua Kou, Fan Jia, Yingfei Liu, Tiancai Wang, Ying Li, Xiaoshuai Hao, Yifan Yang, Hui Zhang, Mengchuan Wei, Yi~Zhou, Haimei Zhao, Jing Zhang, Jinke Li, Xiao He, Xiaoqiang Cheng, Bingyang Zhang, Lirong Zhao, Dianlei Ding, Fangsheng Liu, Yixiang Yan, Hongming Wang, Nanfei Ye, Lun Luo, Yubo Tian, Yiwei Zuo, Zhe Cao, Yi~Ren, Yunfan Li, Wenjie Liu, Xun Wu, Yifan Mao, Ming Li, Jian Liu, Jiayang Liu, Zihan Qin, Cunxi Chu, Jialei Xu, Wenbo Zhao, Junjun Jiang, Xianming Liu, Ziyan Wang, Chiwei Li, Shilong Li, Chendong Yuan, Songyue Yang, Wentao Liu, Peng Chen, Bin Zhou, Yubo Wang, Chi Zhang, Jianhang Sun, Hai Chen, Xiao Yang, Lizhong Wang,
  Dongyi Fu, Yongchun Lin, Huitong Yang, Haoang Li, Yadan Luo, Xianjing Cheng, and Yong Xu.
\newblock The {RoboDrive} challenge: Drive anytime anywhere in any condition.
\newblock \emph{arXiv preprint arXiv:2405.08816}, 2024.

\bibitem[Kong et~al.(2025{\natexlab{a}})Kong, Xu, Liu, Cen, Chen, Zhang, Pan, Chen, and Liu]{kong2025largead}
Lingdong Kong, Xiang Xu, Youquan Liu, Jun Cen, Runnan Chen, Wenwei Zhang, Liang Pan, Kai Chen, and Ziwei Liu.
\newblock {LargeAD}: Large-scale cross-sensor data pretraining for autonomous driving.
\newblock \emph{IEEE Trans. Pattern Anal. Mach. Intell.}, 2025{\natexlab{a}}.

\bibitem[Kong et~al.(2025{\natexlab{b}})Kong, Xu, Ren, Zhang, Pan, Chen, Ooi, and Liu]{kong2025multi}
Lingdong Kong, Xiang Xu, Jiawei Ren, Wenwei Zhang, Liang Pan, Kai Chen, Wei~Tsang Ooi, and Ziwei Liu.
\newblock Multi-modal data-efficient {3D} scene understanding for autonomous driving.
\newblock \emph{IEEE Trans. Pattern Anal. Mach. Intell.}, 47\penalty0 (5):\penalty0 3748--3765, 2025{\natexlab{b}}.

\bibitem[Kong et~al.(2025{\natexlab{c}})Kong, Yang, Mei, Liu, Liang, Zhu, Lu, Yin, Hu, Jia, Deng, Zhang, Wu, Yan, Gao, Wang, Li, Pan, Liu, Zhu, Ooi, Hoi, and Liu]{survey_3d_4d_world_models}
Lingdong Kong, Wesley Yang, Jianbiao Mei, Youquan Liu, Ao~Liang, Dekai Zhu, Dongyue Lu, Wei Yin, Xiaotao Hu, Mingkai Jia, Junyuan Deng, Kaiwen Zhang, Yang Wu, Tianyi Yan, Shenyuan Gao, Song Wang, Linfeng Li, Liang Pan, Yong Liu, Jianke Zhu, Wei~Tsang Ooi, Steven C.~H. Hoi, and Ziwei Liu.
\newblock {3D} and {4D} world modeling: A survey.
\newblock \emph{arXiv preprint arXiv:2509.07996}, 2025{\natexlab{c}}.

\bibitem[Lang et~al.(2019)Lang, Vora, Caesar, Zhou, Yang, and Beijbom]{lang2019pointpillars}
Alex~H. Lang, Sourabh Vora, Holger Caesar, Lubing Zhou, Jiong Yang, and Oscar Beijbom.
\newblock {PointPillars}: Fast encoders for object detection from point clouds.
\newblock In \emph{IEEE/CVF Conf. Comput. Vis. Pattern Recog.}, pages 12697--12705, 2019.

\bibitem[Lee et~al.(2022)Lee, Kim, Kim, Cho, and Han]{lee2022autoregressive}
Doyup Lee, Chiheon Kim, Saehoon Kim, Minsu Cho, and Wook-Shin Han.
\newblock Autoregressive image generation using residual quantization.
\newblock In \emph{IEEE/CVF Conf. Comput. Vis. Pattern Recog.}, pages 11523--11532, 2022.

\bibitem[Lee et~al.(2019)Lee, Hwang, Lee, Bae, and Park]{lee2019energy}
Youngwan Lee, Joong-won Hwang, Sangrok Lee, Yuseok Bae, and Jongyoul Park.
\newblock An energy and {GPU}-computation efficient backbone network for real-time object detection.
\newblock In \emph{IEEE/CVF Conf. Comput. Vis. Pattern Recog. Worksh.}, pages 1--9, 2019.

\bibitem[Li et~al.(2024{\natexlab{a}})Li, Zhang, Guo, Zhang, Li, Zhang, Zhang, Zhang, Li, Liu, and Li]{li2024llava}
Bo~Li, Yuanhan Zhang, Dong Guo, Renrui Zhang, Feng Li, Hao Zhang, Kaichen Zhang, Peiyuan Zhang, Yanwei Li, Ziwei Liu, and Chunyuan Li.
\newblock {LlaVA-OneVision}: Easy visual task transfer.
\newblock \emph{arXiv preprint arXiv:2408.03326}, 2024{\natexlab{a}}.

\bibitem[Li et~al.(2024{\natexlab{b}})Li, Wang, Mao, Ivanovic, Veer, Leung, and Pavone]{li2024driving}
Boyi Li, Yue Wang, Jiageng Mao, Boris Ivanovic, Sushant Veer, Karen Leung, and Marco Pavone.
\newblock Driving everywhere with large language model policy adaptation.
\newblock In \emph{IEEE/CVF Conf. Comput. Vis. Pattern Recog.}, pages 14948--14957, 2024{\natexlab{b}}.

\bibitem[Li et~al.(2023{\natexlab{a}})Li, Li, Wang, Zeng, Xu, Cai, Chen, Yan, Xu, Xiong, Wang, Zhu, Xu, Wang, Xia, Mu, Peng, Lin, and Qiao]{li2023open}
Hongyang Li, Yang Li, Huijie Wang, Jia Zeng, Huilin Xu, Pinlong Cai, Li~Chen, Junchi Yan, Feng Xu, Lu~Xiong, Jingdong Wang, Futang Zhu, Chunjing Xu, Tiancai Wang, Fei Xia, Beipeng Mu, Zhihui Peng, Dahua Lin, and Yu~Qiao.
\newblock Open-sourced data ecosystem in autonomous driving: The present and future.
\newblock \emph{arXiv preprint arXiv:2312.03408}, 2023{\natexlab{a}}.

\bibitem[Li et~al.(2023{\natexlab{b}})Li, Li, Savarese, and Hoi]{li2023blip}
Junnan Li, Dongxu Li, Silvio Savarese, and Steven Hoi.
\newblock {BLIP-2}: Bootstrapping language-image pre-training with frozen image encoders and large language models.
\newblock In \emph{Int. Conf. Mach. Learn.}, pages 19730--19742. PMLR, 2023{\natexlab{b}}.

\bibitem[Li and Cui(2024)]{li2024navigation}
Peidong Li and Dixiao Cui.
\newblock Navigation-guided sparse scene representation for end-to-end autonomous driving.
\newblock \emph{arXiv preprint arXiv:2409.18341}, 2024.

\bibitem[Li et~al.(2025{\natexlab{a}})Li, Zhang, Holtz, Yu, Yang, Lai, Song, Geiger, and Zell]{li2025spacedrive}
Peizheng Li, Zhenghao Zhang, David Holtz, Hang Yu, Yutong Yang, Yuzhi Lai, Rui Song, Andreas Geiger, and Andreas Zell.
\newblock {SpaceDrive}: Infusing spatial awareness into {VLM}-based autonomous driving.
\newblock \emph{arXiv preprint arXiv:2512.10719}, 2025{\natexlab{a}}.

\bibitem[Li et~al.(2025{\natexlab{b}})Li, Zheng, Wang, Wang, Zhao, Liu, Zhan, Zhan, and Lang]{li2025discrete}
Pengxiang Li, Yinan Zheng, Yue Wang, Huimin Wang, Hang Zhao, Jingjing Liu, Xianyuan Zhan, Kun Zhan, and Xianpeng Lang.
\newblock Discrete diffusion for reflective vision-language-action models in autonomous driving.
\newblock \emph{arXiv preprint arXiv:2509.20109}, 2025{\natexlab{b}}.

\bibitem[Li et~al.(2024{\natexlab{c}})Li, Jia, Wang, and Yan]{li2024think2drive}
Qifeng Li, Xiaosong Jia, Shaobo Wang, and Junchi Yan.
\newblock {Think2Drive}: Efficient reinforcement learning by thinking with latent world model for autonomous driving (in {CARLA-V2}).
\newblock In \emph{Eur. Conf. Comput. Vis.}, pages 142--158. Springer, 2024{\natexlab{c}}.

\bibitem[Li et~al.(2025{\natexlab{c}})Li, Dong, Hu, Liang, Liu, Lu, Pan, Kong, Liang, and Liu]{li2025_3eed}
Rong Li, Yuhao Dong, Tianshuai Hu, Ao~Liang, Youquan Liu, Dongyue Lu, Liang Pan, Lingdong Kong, Junwei Liang, and Ziwei Liu.
\newblock {3EED}: Ground everything everywhere in {3D}.
\newblock In \emph{Adv. Neural Inf. Process. Syst.}, volume~38, 2025{\natexlab{c}}.

\bibitem[Li et~al.(2025{\natexlab{d}})Li, Yuan, Liu, Tang, Wang, Qin, Zhu, and Zhang]{li2025tokenpacker}
Wentong Li, Yuqian Yuan, Jian Liu, Dongqi Tang, Song Wang, Jie Qin, Jianke Zhu, and Lei Zhang.
\newblock {TokenPacker}: Efficient visual projector for multimodal {LLM}.
\newblock \emph{Int. J. Comput. Vis.}, 133:\penalty0 6794–6812, 2025{\natexlab{d}}.

\bibitem[Li et~al.(2023{\natexlab{c}})Li, Bai, Cai, Wen, Fu, Zhang, Yang, Cai, Ma, Guo, Gao, Dou, Li, Shi, Liu, He, and Qiao]{li2023towards}
Xin Li, Yeqi Bai, Pinlong Cai, Licheng Wen, Daocheng Fu, Bo~Zhang, Xuemeng Yang, Xinyu Cai, Tao Ma, Jianfei Guo, Xing Gao, Min Dou, Yikang Li, Botian Shi, Yong Liu, Liang He, and Yu~Qiao.
\newblock Towards knowledge-driven autonomous driving.
\newblock \emph{arXiv preprint arXiv:2312.04316}, 2023{\natexlab{c}}.

\bibitem[Li et~al.(2024{\natexlab{d}})Li, Kong, Hu, Xu, and Huang]{li2024place3d}
Ye~Li, Lingdong Kong, Hanjiang Hu, Xiaohao Xu, and Xiaonan Huang.
\newblock Is your {LiDAR} placement optimized for {3D} scene understanding?
\newblock In \emph{Adv. Neural Inf. Process. Syst.}, volume~37, pages 34980--35017, 2024{\natexlab{d}}.

\bibitem[Li et~al.(2024{\natexlab{e}})Li, Fan, Ge, Zhao, Li, Xu, Yao, Tomizuka, Zhou, Tang, Ding, and Zhan]{li2024womd}
Yiheng Li, Cunxin Fan, Chongjian Ge, Zhihao Zhao, Chenran Li, Chenfeng Xu, Huaxiu Yao, Masayoshi Tomizuka, Bolei Zhou, Chen Tang, Mingyu Ding, and Wei Zhan.
\newblock {WOMD-Reasoning}: A large-scale dataset for interaction reasoning in driving.
\newblock \emph{arXiv preprint arXiv:2407.04281}, 2024{\natexlab{e}}.

\bibitem[Li et~al.(2024{\natexlab{f}})Li, Fan, He, Wang, Chen, Zhang, and Tan]{li2024enhancing}
Yingyan Li, Lue Fan, Jiawei He, Yuqi Wang, Yuntao Chen, Zhaoxiang Zhang, and Tieniu Tan.
\newblock Enhancing end-to-end autonomous driving with latent world model.
\newblock \emph{arXiv preprint arXiv:2406.08481}, 2024{\natexlab{f}}.

\bibitem[Li et~al.(2025{\natexlab{e}})Li, Shang, Liu, Zhan, Wang, Wang, Chen, Wang, An, Tang, Hou, Fan, and Zhang]{li2025drivevlaw0}
Yingyan Li, Shuyao Shang, Weisong Liu, Bing Zhan, Haochen Wang, Yuqi Wang, Yuntao Chen, Xiaoman Wang, Yasong An, Chufeng Tang, Lu~Hou, Lue Fan, and Zhaoxiang Zhang.
\newblock {DriveVLA-W0}: World models amplify data scaling law in autonomous driving.
\newblock \emph{arXiv preprint arXiv:2510.12796}, 2025{\natexlab{e}}.

\bibitem[Li et~al.(2025{\natexlab{f}})Li, Wang, Liu, He, Fan, and Zhang]{li2025end}
Yingyan Li, Yuqi Wang, Yang Liu, Jiawei He, Lue Fan, and Zhaoxiang Zhang.
\newblock End-to-end driving with online trajectory evaluation via bev world model.
\newblock \emph{arXiv preprint arXiv:2504.01941}, 2025{\natexlab{f}}.

\bibitem[Li et~al.(2025{\natexlab{g}})Li, Xiong, Guo, Li, Yan, Xu, Zhou, Chen, Sun, Wang, Ma, Chen, Ye, Liu, and Wang]{li2025recogdrive}
Yongkang Li, Kaixin Xiong, Xiangyu Guo, Fang Li, Sixu Yan, Gangwei Xu, Lijun Zhou, Long Chen, Haiyang Sun, Bing Wang, Kun Ma, Guang Chen, Hangjun Ye, Wenyu Liu, and Xinggang Wang.
\newblock {ReCogDrive}: A reinforced cognitive framework for end-to-end autonomous driving.
\newblock \emph{arXiv preprint arXiv:2506.08052}, 2025{\natexlab{g}}.

\bibitem[Li et~al.(2025{\natexlab{h}})Li, Tian, Zhu, Zhu, Lin, Xiong, and Zhao]{li2025drive}
Yue Li, Meng Tian, Dechang Zhu, Jiangtong Zhu, Zhenyu Lin, Zhiwei Xiong, and Xinhai Zhao.
\newblock {Drive-R1}: Bridging reasoning and planning in {VLMs} for autonomous driving with reinforcement learning.
\newblock \emph{arXiv preprint arXiv:2506.18234}, 2025{\natexlab{h}}.

\bibitem[Li et~al.(2024{\natexlab{g}})Li, Li, Wang, Lan, Yu, Ji, Li, Zhu, Kautz, Wu, Jiang, and Alvarez]{li2024hydra}
Zhenxin Li, Kailin Li, Shihao Wang, Shiyi Lan, Zhiding Yu, Yishen Ji, Zhiqi Li, Ziyue Zhu, Jan Kautz, Zuxuan Wu, Yu-Gang Jiang, and Jose~M. Alvarez.
\newblock {Hydra-MDP}: End-to-end multimodal planning with multi-target hydra-distillation.
\newblock \emph{arXiv preprint arXiv:2406.06978}, 2024{\natexlab{g}}.

\bibitem[Li et~al.(2024{\natexlab{h}})Li, Wang, Li, Xie, Sima, Lu, Yu, and Dai]{li2024bevformer}
Zhiqi Li, Wenhai Wang, Hongyang Li, Enze Xie, Chonghao Sima, Tong Lu, Qiao Yu, and Jifeng Dai.
\newblock {BEVFormer}: learning bird's-eye-view representation from {LiDAR}-camera via spatiotemporal transformers.
\newblock \emph{IEEE Trans. Pattern Anal. Mach. Intell.}, 47\penalty0 (3):\penalty0 2020--2036, 2024{\natexlab{h}}.

\bibitem[Li et~al.(2024{\natexlab{i}})Li, Yu, Lan, Li, Kautz, Lu, and Alvarez]{li2024ego}
Zhiqi Li, Zhiding Yu, Shiyi Lan, Jiahan Li, Jan Kautz, Tong Lu, and Jose~M Alvarez.
\newblock Is ego status all you need for open-loop end-to-end autonomous driving?
\newblock In \emph{IEEE/CVF Conf. Comput. Vis. Pattern Recog.}, pages 14864--14873, 2024{\natexlab{i}}.

\bibitem[Li et~al.(2025{\natexlab{i}})Li, Jin, Yu, Chen, Li, Han, Xiong, Leng, Hu, Kolmanovsky, and Filev]{li2025survey}
Zhuoren Li, Guizhe Jin, Ran Yu, Zhiwen Chen, Nan Li, Wei Han, Lu~Xiong, Bo~Leng, Jia Hu, Ilya Kolmanovsky, and Dimitar Filev.
\newblock A survey of reinforcement learning-based motion planning for autonomous driving: Lessons learned from a driving task perspective.
\newblock \emph{arXiv preprint arXiv:2503.23650}, 2025{\natexlab{i}}.

\bibitem[{Li Auto Inc.}(2025)]{li2025march}
{Li Auto Inc.}
\newblock {MindVLA}, 2025.
\newblock URL \url{https://ir.lixiang.com/news-releases/news-release-details/li-auto-inc-march-2025-delivery-update}.

\bibitem[Liang et~al.(2025)Liang, Kong, Lu, Liu, Fang, Zhao, and Ooi]{liang2025pi3det}
Ao~Liang, Lingdong Kong, Dongyue Lu, Youquan Liu, Jian Fang, Huaici Zhao, and Wei~Tsang Ooi.
\newblock Perspective-invariant {3D} object detection.
\newblock In \emph{IEEE/CVF Int. Conf. Comput. Vis.}, pages 27725--27738, 2025.

\bibitem[Liang et~al.(2026)Liang, Liu, Yang, Lu, Li, Kong, Zhao, and Ooi]{liang2026lidarcrafter}
Ao~Liang, Youquan Liu, Yu~Yang, Dongyue Lu, Linfeng Li, Lingdong Kong, Huaici Zhao, and Wei~Tsang Ooi.
\newblock {LiDARCrafter}: Dynamic {4D} world modeling from {LiDAR} sequences.
\newblock In \emph{AAAI Conf. Artifi. Intell.}, volume~40, 2026.

\bibitem[Liang et~al.(2022)Liang, Xie, Yu, Xia, Lin, Wang, Tang, Wang, and Tang]{liang2022bevfusion}
Tingting Liang, Hongwei Xie, Kaicheng Yu, Zhongyu Xia, Zhiwei Lin, Yongtao Wang, Tao Tang, Bing Wang, and Zhi Tang.
\newblock {BEVFusion}: A simple and robust {LiDAR}-camera fusion framework.
\newblock \emph{Adv. Neural Inf. Process. Syst.}, 35:\penalty0 10421--10434, 2022.

\bibitem[Liao et~al.(2025)Liao, Chen, Yin, Jiang, Wang, Yan, Zhang, Li, Zhang, Zhang, and Wang]{liao2025diffusiondrive}
Bencheng Liao, Shaoyu Chen, Haoran Yin, Bo~Jiang, Cheng Wang, Sixu Yan, Xinbang Zhang, Xiangyu Li, Ying Zhang, Qian Zhang, and Xinggang Wang.
\newblock {DiffusionDrive}: Truncated diffusion model for end-to-end autonomous driving.
\newblock In \emph{IEEE/CVF Conf. Comput. Vis. Pattern Recog.}, pages 12037--12047, 2025.

\bibitem[Lin et~al.(2023)Lin, Ye, Zhu, Cui, Ning, Jin, and Yuan]{lin2023video}
Bin Lin, Yang Ye, Bin Zhu, Jiaxi Cui, Munan Ning, Peng Jin, and Li~Yuan.
\newblock {Video-LLaVA}: Learning united visual representation by alignment before projection.
\newblock \emph{arXiv preprint arXiv:2311.10122}, 2023.

\bibitem[Liu et~al.(2024{\natexlab{a}})Liu, Zhao, Zhuo, Lin, Xin, Li, Qin, Qiao, Li, and Gao]{liu2024lumina}
Dongyang Liu, Shitian Zhao, Le~Zhuo, Weifeng Lin, Yi~Xin, Xinyue Li, Qi~Qin, Yu~Qiao, Hongsheng Li, and Peng Gao.
\newblock {Lumina-MGPT}: Illuminate flexible photorealistic text-to-image generation with multimodal generative pretraining.
\newblock \emph{arXiv preprint arXiv:2408.02657}, 2024{\natexlab{a}}.

\bibitem[Liu et~al.(2024{\natexlab{b}})Liu, Xue, Chen, Chen, Zhao, Wang, Hou, Li, and Peng]{liu2024survey}
Hanchao Liu, Wenyuan Xue, Yifei Chen, Dapeng Chen, Xiutian Zhao, Ke~Wang, Liping Hou, Rongjun Li, and Wei Peng.
\newblock A survey on hallucination in large vision-language models.
\newblock \emph{arXiv preprint arXiv:2402.00253}, 2024{\natexlab{b}}.

\bibitem[Liu et~al.(2023{\natexlab{a}})Liu, Li, Wu, and Lee]{liu2023visual}
Haotian Liu, Chunyuan Li, Qingyang Wu, and Yong~Jae Lee.
\newblock Visual instruction tuning.
\newblock In \emph{Adv. Neural Inf. Process. Syst.}, volume~36, pages 34892--34916, 2023{\natexlab{a}}.

\bibitem[Liu et~al.(2024{\natexlab{c}})Liu, Li, Li, and Lee]{liu2024improved}
Haotian Liu, Chunyuan Li, Yuheng Li, and Yong~Jae Lee.
\newblock Improved baselines with visual instruction tuning.
\newblock In \emph{IEEE/CVF Conf. Comput. Vis. Pattern Recog.}, pages 26296--26306, 2024{\natexlab{c}}.

\bibitem[Liu et~al.(2025{\natexlab{a}})Liu, Jia, Yu, Song, Li, Jia, Wu, Hao, and Luo]{liu2025guideflow}
Lin Liu, Caiyan Jia, Guanyi Yu, Ziying Song, JunQiao Li, Feiyang Jia, Peiliang Wu, Xiaoshuai Hao, and Yandan Luo.
\newblock {GuideFlow}: Constraint-guided flow matching for planning in end-to-end autonomous driving.
\newblock \emph{arXiv preprint arXiv:2511.18729}, 2025{\natexlab{a}}.

\bibitem[Liu et~al.(2024{\natexlab{d}})Liu, Yurtsever, Fossaert, Zhou, Zimmer, Cui, Zagar, and Knoll]{liu2024surveyondata}
Mingyu Liu, Ekim Yurtsever, Jonathan Fossaert, Xingcheng Zhou, Walter Zimmer, Yuning Cui, Bare~Luka Zagar, and Alois~C. Knoll.
\newblock A survey on autonomous driving datasets: Statistics, annotation quality, and a future outlook.
\newblock \emph{IEEE Trans. Intell. Veh.}, 9\penalty0 (11):\penalty0 7138--7164, 2024{\natexlab{d}}.

\bibitem[Liu et~al.(2025{\natexlab{b}})Liu, Liu, Liu, Liu, Ni, and Ma]{liu2025vlm}
Pei Liu, Haipeng Liu, Haichao Liu, Xin Liu, Jinxin Ni, and Jun Ma.
\newblock {VLM-E2E}: Enhancing end-to-end autonomous driving with multimodal driver attention fusion.
\newblock \emph{arXiv preprint arXiv:2502.18042}, 2025{\natexlab{b}}.

\bibitem[Liu et~al.(2025{\natexlab{c}})Liu, Lu, Liu, Liu, Liu, Yao, Li, and Ma]{liu2025omniscene}
Pei Liu, Hongliang Lu, Haichao Liu, Haipeng Liu, Xin Liu, Ruoyu Yao, Shengbo~Eben Li, and Jun Ma.
\newblock {OmniScene}: Attention-augmented multimodal {4D} scene understanding for autonomous driving.
\newblock \emph{arXiv preprint arXiv:2509.19973}, 2025{\natexlab{c}}.

\bibitem[Liu et~al.(2025{\natexlab{d}})Liu, Ning, Lu, Liu, Ma, She, Jia, Lang, and Ma]{liu2025omnireason}
Pei Liu, Qingtian Ning, Xinyan Lu, Haipeng Liu, Weiliang Ma, Dangen She, Peng Jia, Xianpeng Lang, and Jun Ma.
\newblock {OmniReason}: A temporal-guided vision-language-action framework for autonomous driving.
\newblock \emph{arXiv preprint arXiv:2509.00789}, 2025{\natexlab{d}}.

\bibitem[Liu et~al.(2025{\natexlab{e}})Liu, Kong, Li, and Zhao]{liu2025occvla}
Ruixun Liu, Lingyu Kong, Derun Li, and Hang Zhao.
\newblock {OccVLA}: Vision-language-action model with implicit {3D} occupancy supervision.
\newblock \emph{arXiv preprint arXiv:2509.05578}, 2025{\natexlab{e}}.

\bibitem[Liu et~al.(2025{\natexlab{f}})Liu, Liu, and Ma]{liu2025dsdrive}
Wenru Liu, Pei Liu, and Jun Ma.
\newblock {DSDrive}: Distilling large language model for lightweight end-to-end autonomous driving with unified reasoning and planning.
\newblock \emph{arXiv preprint arXiv:2505.05360}, 2025{\natexlab{f}}.

\bibitem[Liu et~al.(2025{\natexlab{g}})Liu, Zhong, Guo, Liu, Su, Zhang, Wang, Gao, Zheng, Lin, Chen, and Zhao]{liu2025reasonplan}
Xueyi Liu, Zuodong Zhong, Yuxin Guo, Yun-Fu Liu, Zhiguo Su, Qichao Zhang, Junli Wang, Yinfeng Gao, Yupeng Zheng, Qiao Lin, Huiyong Chen, and Dongbin Zhao.
\newblock {ReasonPlan}: Unified scene prediction and decision reasoning for closed-loop autonomous driving.
\newblock \emph{arXiv preprint arXiv:2505.20024}, 2025{\natexlab{g}}.

\bibitem[Liu et~al.(2023{\natexlab{b}})Liu, Kong, Cen, Chen, Zhang, Pan, Chen, and Liu]{liu2023segment}
Youquan Liu, Lingdong Kong, Jun Cen, Runnan Chen, Wenwei Zhang, Liang Pan, Kai Chen, and Ziwei Liu.
\newblock Segment any point cloud sequences by distilling vision foundation models.
\newblock In \emph{Adv. Neural Inf. Process. Syst.}, volume~36, pages 37193--37229, 2023{\natexlab{b}}.

\bibitem[Liu et~al.(2026)Liu, Kong, Yang, Li, Liang, Chen, Fei, and Liu]{liu2026lalalidar}
Youquan Liu, Lingdong Kong, Weidong Yang, Xin Li, Ao~Liang, Runnan Chen, Ben Fei, and Tongliang Liu.
\newblock {La La LiDAR}: Large-scale layout generation from {LiDAR} data.
\newblock In \emph{AAAI Conf. Artifi. Intell.}, volume~40, 2026.

\bibitem[Liu et~al.(2021)Liu, Lin, Cao, Hu, Wei, Zhang, Lin, and Guo]{liu2021swin}
Ze~Liu, Yutong Lin, Yue Cao, Han Hu, Yixuan Wei, Zheng Zhang, Stephen Lin, and Baining Guo.
\newblock Swin transformer: Hierarchical vision transformer using shifted windows.
\newblock In \emph{IEEE/CVF Int. Conf. Comput. Vis.}, pages 10012--10022, 2021.

\bibitem[Liu et~al.(2022)Liu, Mao, Wu, Feichtenhofer, Darrell, and Xie]{liu2022convnet}
Zhuang Liu, Hanzi Mao, Chao-Yuan Wu, Christoph Feichtenhofer, Trevor Darrell, and Saining Xie.
\newblock A {ConvNet} for the 2020s.
\newblock In \emph{IEEE/CVF Conf. Comput. Vis. Pattern Recog.}, pages 11976--11986, 2022.

\bibitem[Lu et~al.(2025{\natexlab{a}})Lu, Liu, Jiang, Luo, Chen, Zhang, and Chen]{lu2025uniugp}
Hao Lu, Ziyang Liu, Guangfeng Jiang, Yuanfei Luo, Sheng Chen, Yangang Zhang, and Ying-Cong Chen.
\newblock {UniUGP}: Unifying understanding, generation, and planing for end-to-end autonomous driving.
\newblock \emph{arXiv preprint arXiv:2512.09864}, 2025{\natexlab{a}}.

\bibitem[Lu et~al.(2024)Lu, Huang, Yang, Zhang, and Zhang]{lu2024wovogen}
Jiachen Lu, Ze~Huang, Zeyu Yang, Jiahui Zhang, and Li~Zhang.
\newblock {WoVoGen}: World volume-aware diffusion for controllable multi-camera driving scene generation.
\newblock In \emph{Eur. Conf. Comput. Vis.}, pages 329--345. Springer, 2024.

\bibitem[Lu et~al.(2025{\natexlab{b}})Lu, Tu, Ma, and Zhu]{lu2025real}
Yuhang Lu, Jiadong Tu, Yuexin Ma, and Xinge Zhu.
\newblock {ReAL-AD}: Towards human-like reasoning in end-to-end autonomous driving.
\newblock In \emph{IEEE/CVF Int. Conf. Comput. Vis.}, pages 27783--27793, 2025{\natexlab{b}}.

\bibitem[Luo et~al.(2025)Luo, Li, Xu, Lai, Yang, Chen, Luo, Xie, Jiang, Liu, Chen, Wang, and xin Yang]{luo2025adathinkdrive}
Yuechen Luo, Fang Li, Shaoqing Xu, Zhiyi Lai, Lei Yang, Qimao Chen, Ziang Luo, Zixun Xie, Shengyin Jiang, Jiaxin Liu, Long Chen, Bing Wang, and Zhi xin Yang.
\newblock {AdaThinkDrive}: Adaptive thinking via reinforcement learning for autonomous driving.
\newblock \emph{arXiv preprint arXiv:2509.13769}, 2025.

\bibitem[Ma et~al.(2024{\natexlab{a}})Ma, Chen, Huang, Xu, Luo, Xu, Gu, Ai, and Wang]{ma2024cam4docc}
Junyi Ma, Xieyuanli Chen, Jiawei Huang, Jingyi Xu, Zhen Luo, Jintao Xu, Weihao Gu, Rui Ai, and Hesheng Wang.
\newblock {Cam4DOcc}: Benchmark for camera-only {4D} occupancy forecasting in autonomous driving applications.
\newblock In \emph{IEEE/CVF Conf. Comput. Vis. Pattern Recog.}, pages 21486--21495, 2024{\natexlab{a}}.

\bibitem[Ma et~al.(2025{\natexlab{a}})Ma, Cao, Ding, Zhang, Wang, Ivanovic, Jiang, Pavone, and Xiao]{ma2025dvlm}
Yingzi Ma, Yulong Cao, Wenhao Ding, Shuibai Zhang, Yan Wang, Boris Ivanovic, Ming Jiang, Marco Pavone, and Chaowei Xiao.
\newblock {dVLM-AD}: Enhance diffusion vision-language-model for driving via controllable reasoning.
\newblock \emph{arXiv preprint arXiv:2512.04459}, 2025{\natexlab{a}}.

\bibitem[Ma et~al.(2024{\natexlab{b}})Ma, Song, Zhuang, Hao, and King]{ma2024survey}
Yueen Ma, Zixing Song, Yuzheng Zhuang, Jianye Hao, and Irwin King.
\newblock A survey on vision-language-action models for embodied {AI}.
\newblock \emph{arXiv preprint arXiv:2405.14093}, 2024{\natexlab{b}}.

\bibitem[Ma et~al.(2025{\natexlab{b}})Ma, Song, Zhuang, Hao, and King]{adilkhanov2025survey}
Yueen Ma, Zixing Song, Yuzheng Zhuang, Jianye Hao, and Irwin King.
\newblock Survey on vision-language-action models.
\newblock \emph{arXiv preprint arXiv:2502.06851}, 2025{\natexlab{b}}.

\bibitem[Ma et~al.(2025{\natexlab{c}})Ma, Wei, Zhong, Mei, Hu, Wen, Yang, Shi, and Liu]{ma2025leapvad}
Yukai Ma, Tiantian Wei, Naiting Zhong, Jianbiao Mei, Tao Hu, Licheng Wen, Xuemeng Yang, Botian Shi, and Yong Liu.
\newblock {LeapVAD}: A leap in autonomous driving via cognitive perception and dual-process thinking.
\newblock \emph{arXiv preprint arXiv:2501.08168}, 2025{\natexlab{c}}.

\bibitem[Ma et~al.(2025{\natexlab{d}})Ma, Yaman, Ye, Yurt, Luo, Mallik, Wang, and Ren]{ma2025aln}
Yunsheng Ma, Burhaneddin Yaman, Xin Ye, Mahmut Yurt, Jingru Luo, Abhirup Mallik, Ziran Wang, and Liu Ren.
\newblock {ALN-P3}: Unified language alignment for perception, prediction, and planning in autonomous driving.
\newblock \emph{arXiv preprint arXiv:2505.15158}, 2025{\natexlab{d}}.

\bibitem[Ma et~al.(2022)Ma, VanDerPloeg, Bara, Huang, Kim, Gervits, Marge, and Chai]{ma2022dorothie}
Ziqiao Ma, Benjamin VanDerPloeg, Cristian-Paul Bara, Yidong Huang, Eui-In Kim, Felix Gervits, Matthew Marge, and Joyce Chai.
\newblock {DOROTHIE}: Spoken dialogue for handling unexpected situations in interactive autonomous driving agents.
\newblock \emph{arXiv preprint arXiv:2210.12511}, 2022.

\bibitem[Mao et~al.(2023)Mao, Qian, Ye, Zhao, and Wang]{mao2023gpt}
Jiageng Mao, Yuxi Qian, Junjie Ye, Hang Zhao, and Yue Wang.
\newblock {GPT-Driver}: Learning to drive with {GPT}.
\newblock \emph{arXiv preprint arXiv:2310.01415}, 2023.

\bibitem[Mao et~al.(2025)Mao, Ye, Qian, Pavone, and Wang]{mao2023language}
Jiageng Mao, Junjie Ye, Yuxi Qian, Marco Pavone, and Yue Wang.
\newblock A language agent for autonomous driving.
\newblock In \emph{Conf. Lang. Model.}, 2025.

\bibitem[Marcu et~al.(2024)Marcu, Chen, Hünermann, Karnsund, Hanotte, Chidananda, Nair, Badrinarayanan, Kendall, Shotton, Arani, and Sinavski]{marcu2024lingoqa}
Ana-Maria Marcu, Long Chen, Jan Hünermann, Alice Karnsund, Benoit Hanotte, Prajwal Chidananda, Saurabh Nair, Vijay Badrinarayanan, Alex Kendall, Jamie Shotton, Elahe Arani, and Oleg Sinavski.
\newblock {LingoQA}: Visual question answering for autonomous driving.
\newblock In \emph{Eur. Conf. Comput. Vis.}, pages 252--269. Springer, 2024.

\bibitem[Mei et~al.(2024)Mei, Ma, Yang, Wen, Cai, Li, Fu, Zhang, Cai, Dou, Shi, He, Liu, and Qiao]{mei2024continuously}
Jianbiao Mei, Yukai Ma, Xuemeng Yang, Licheng Wen, Xinyu Cai, Xin Li, Daocheng Fu, Bo~Zhang, Pinlong Cai, Min Dou, Botian Shi, Liang He, Yong Liu, and Yu~Qiao.
\newblock Continuously learning, adapting, and improving: A dual-process approach to autonomous driving.
\newblock In \emph{Adv. Neural Inf. Process. Syst.}, volume~37, pages 123261--123290, 2024.

\bibitem[Min et~al.(2024)Min, Zhao, Xiao, Zhao, Xu, Zhu, Jin, Li, Guo, Xing, Jing, Nie, and Dai]{min2024driveworld}
Chen Min, Dawei Zhao, Liang Xiao, Jian Zhao, Xinli Xu, Zheng Zhu, Lei Jin, Jianshu Li, Yulan Guo, Junliang Xing, Liping Jing, Yiming Nie, and Bin Dai.
\newblock {DriveWorld}: {4D} pre-trained scene understanding via world models for autonomous driving.
\newblock In \emph{IEEE/CVF Conf. Comput. Vis. Pattern Recog.}, pages 15522--15533, 2024.

\bibitem[Mirzaie and Rosenhahn(2025)]{mirzaie2025interpretable}
Mona Mirzaie and Bodo Rosenhahn.
\newblock Interpretable decision-making for end-to-end autonomous driving.
\newblock \emph{arXiv preprint arXiv:2508.18898}, 2025.

\bibitem[Muller et~al.(2005)Muller, Ben, Cosatto, Flepp, and Cun]{muller2005off}
Urs Muller, Jan Ben, Eric Cosatto, Beat Flepp, and Yann Cun.
\newblock Off-road obstacle avoidance through end-to-end learning.
\newblock In \emph{Adv. Neural Inf. Process. Syst.}, volume~18, pages 739--746, 2005.

\bibitem[Mütsch et~al.(2023)Mütsch, Gremmelmaier, Becker, Bogdoll, Zofka, and Zöllner]{mutsch2023model}
Ferdinand Mütsch, Helen Gremmelmaier, Nicolas Becker, Daniel Bogdoll, Marc~René Zofka, and J.~Marius Zöllner.
\newblock From model-based to data-driven simulation: Challenges and trends in autonomous driving.
\newblock \emph{arXiv preprint arXiv:2305.13960}, 2023.

\bibitem[Nie et~al.(2024)Nie, Peng, Wang, Cai, Han, Xu, and Zhang]{nie2024reason2drive}
Ming Nie, Renyuan Peng, Chunwei Wang, Xinyue Cai, Jianhua Han, Hang Xu, and Li~Zhang.
\newblock {Reason2Drive}: Towards interpretable and chain-based reasoning for autonomous driving.
\newblock In \emph{Eur. Conf. Comput. Vis.}, pages 292--308. Springer, 2024.

\bibitem[NVIDIA(2025)]{nvidia2025avdata}
NVIDIA.
\newblock Physical {AI} autonomous vehicles dataset.
\newblock \url{https://huggingface.co/datasets/nvidia/PhysicalAI-Autonomous-Vehicles}, October 2025.

\bibitem[Ohn-Bar et~al.(2020)Ohn-Bar, Prakash, Behl, Chitta, and Geiger]{ohn2020learning}
Eshed Ohn-Bar, Aditya Prakash, Aseem Behl, Kashyap Chitta, and Andreas Geiger.
\newblock Learning situational driving.
\newblock In \emph{IEEE/CVF Conf. Comput. Vis. Pattern Recog.}, pages 11296--11305, 2020.

\bibitem[OpenAI(2024)]{gpt4o}
OpenAI.
\newblock Hello {GPT4-o}.
\newblock \url{https://openai.com/index/hello-gpt-4o}, 2024.

\bibitem[Oquab et~al.(2023)Oquab, Darcet, Moutakanni, Vo, Szafraniec, Khalidov, Fernandez, Haziza, Massa, El-Nouby, Assran, Ballas, Galuba, Howes, Huang, Li, Misra, Rabbat, Sharma, Synnaeve, Xu, Jegou, Mairal, Labatut, Joulin, and Bojanowski]{oquab2023dinov2}
Maxime Oquab, Timothée Darcet, Théo Moutakanni, Huy Vo, Marc Szafraniec, Vasil Khalidov, Pierre Fernandez, Daniel Haziza, Francisco Massa, Alaaeldin El-Nouby, Mahmoud Assran, Nicolas Ballas, Wojciech Galuba, Russell Howes, Po-Yao Huang, Shang-Wen Li, Ishan Misra, Michael Rabbat, Vasu Sharma, Gabriel Synnaeve, Hu~Xu, Hervé Jegou, Julien Mairal, Patrick Labatut, Armand Joulin, and Piotr Bojanowski.
\newblock {DINOv2}: Learning robust visual features without supervision.
\newblock \emph{arXiv preprint arXiv:2304.07193}, 2023.

\bibitem[Pan et~al.(2024)Pan, Yaman, Nesti, Mallik, Allievi, Velipasalar, and Ren]{pan2024vlp}
Chenbin Pan, Burhaneddin Yaman, Tommaso Nesti, Abhirup Mallik, Alessandro~G. Allievi, Senem Velipasalar, and Liu Ren.
\newblock {VLP}: Vision language planning for autonomous driving.
\newblock In \emph{IEEE/CVF Conf. Comput. Vis. Pattern Recog.}, pages 14760--14769, 2024.

\bibitem[Pan et~al.(2018)Pan, Cheng, Saigol, Lee, Yan, Theodorou, and Boots]{pan2017agile}
Yunpeng Pan, Ching-An Cheng, Kamil Saigol, Keuntaek Lee, Xinyan Yan, Evangelos Theodorou, and Byron Boots.
\newblock Agile autonomous driving using end-to-end deep imitation learning.
\newblock In \emph{Robot. Sci. Syst.}, 2018.

\bibitem[Park et~al.(2023)Park, Xu, Yang, Keutzer, Kitani, Tomizuka, and Zhan]{parktime}
Jinhyung Park, Chenfeng Xu, Shijia Yang, Kurt Keutzer, Kris Kitani, Masayoshi Tomizuka, and Wei Zhan.
\newblock {Time Will Tell}: New outlooks and a baseline for temporal multi-view {3D} object detection.
\newblock In \emph{Int. Conf. Learn. Represent.}, 2023.

\bibitem[Park et~al.(2021)Park, Seo, Liu, Zhao, Qin, Shin, and Liu]{park2021object}
Jongjin Park, Younggyo Seo, Chang Liu, Li~Zhao, Tao Qin, Jinwoo Shin, and Tie-Yan Liu.
\newblock Object-aware regularization for addressing causal confusion in imitation learning.
\newblock In \emph{Adv. Neural Inf. Process. Syst.}, volume~34, pages 3029--3042, 2021.

\bibitem[Park et~al.(2024)Park, Lee, Kang, Choi, Park, Cho, Lee, and Kim]{park2024vlaad}
SungYeon Park, MinJae Lee, JiHyuk Kang, Hahyeon Choi, Yoonah Park, Juhwan Cho, Adam Lee, and DongKyu Kim.
\newblock {VLAAD}: Vision and language assistant for autonomous driving.
\newblock In \emph{IEEE/CVF Winter Conf. Appl. Comput. Vis. Worksh.}, pages 980--987, 2024.

\bibitem[Podell et~al.(2023)Podell, English, Lacey, Blattmann, Dockhorn, Müller, Penna, and Rombach]{podell2023sdxl}
Dustin Podell, Zion English, Kyle Lacey, Andreas Blattmann, Tim Dockhorn, Jonas Müller, Joe Penna, and Robin Rombach.
\newblock {SDXL}: Improving latent diffusion models for high-resolution image synthesis.
\newblock In \emph{Int. Conf. Learn. Represent.}, 2023.

\bibitem[Pomerleau(1988)]{pomerleau1988alvinn}
Dean~A Pomerleau.
\newblock {ALVINN}: An autonomous land vehicle in a neural network.
\newblock In \emph{Adv. Neural Inf. Process. Syst.}, volume~1, pages 305--313, 1988.

\bibitem[Popov et~al.(2024)Popov, Degirmenci, Wehr, Hegde, Oldja, Kamenev, Douillard, Nistér, Muller, Bhargava, Birchfield, and Smolyanskiy]{popov2024mitigating}
Alexander Popov, Alperen Degirmenci, David Wehr, Shashank Hegde, Ryan Oldja, Alexey Kamenev, Bertrand Douillard, David Nistér, Urs Muller, Ruchi Bhargava, Stan Birchfield, and Nikolai Smolyanskiy.
\newblock Mitigating covariate shift in imitation learning for autonomous vehicles using latent space generative world models.
\newblock \emph{arXiv preprint arXiv:2409.16663}, 2024.

\bibitem[Prakash et~al.(2020)Prakash, Behl, Ohn-Bar, Chitta, and Geiger]{prakash2020exploring}
Aditya Prakash, Aseem Behl, Eshed Ohn-Bar, Kashyap Chitta, and Andreas Geiger.
\newblock Exploring data aggregation in policy learning for vision-based urban autonomous driving.
\newblock In \emph{IEEE/CVF Conf. Comput. Vis. Pattern Recog.}, pages 11763--11773, 2020.

\bibitem[Prakash et~al.(2021)Prakash, Chitta, and Geiger]{prakash2021multi}
Aditya Prakash, Kashyap Chitta, and Andreas Geiger.
\newblock Multi-modal fusion transformer for end-to-end autonomous driving.
\newblock In \emph{IEEE/CVF Conf. Comput. Vis. Pattern Recog.}, pages 7077--7087, 2021.

\bibitem[Qian et~al.(2024)Qian, Ma, He, Luo, Shi, Zhu, Li, Wang, Chen, He, Shi, Fu, Jiao, Jiang, Yang, and Matsumaru]{qian2024fasionad}
Kangan Qian, Zhikun Ma, Yangfan He, Ziang Luo, Tianyu Shi, Tianze Zhu, Jiayin Li, Jianhui Wang, Ziyu Chen, Xiao He, Yining Shi, Zheng Fu, Xinyu Jiao, Kun Jiang, Diange Yang, and Takafumi Matsumaru.
\newblock {FasionAD}: Fast and slow fusion thinking systems for human-like autonomous driving with adaptive feedback.
\newblock \emph{arXiv preprint arXiv:2411.18013}, 2024.

\bibitem[Qian et~al.(2025)Qian, Luo, Jiang, Huang, Miao, Ma, Zhu, Li, He, Fu, Shi, Wang, Lin, Chen, Yu, Jiao, Yang, Jiang, and Yang]{qian2025fasionad++}
Kangan Qian, Ziang Luo, Sicong Jiang, Zilin Huang, Jinyu Miao, Zhikun Ma, Tianze Zhu, Jiayin Li, Yangfan He, Zheng Fu, Yining Shi, Boyue Wang, Hezhe Lin, Ziyu Chen, Jiangbo Yu, Xinyu Jiao, Mengmeng Yang, Kun Jiang, and Diange Yang.
\newblock {FasionAD++}: Integrating high-level instruction and information bottleneck in fat-slow fusion systems for enhanced safety in autonomous driving with adaptive feedback.
\newblock \emph{arXiv preprint arXiv:2503.08162}, 2025.

\bibitem[Qiao et~al.(2025)Qiao, Li, Cao, and Liu]{qiao2025lightemma}
Zhijie Qiao, Haowei Li, Zhong Cao, and Henry~X. Liu.
\newblock {LightEMMA}: Lightweight end-to-end multimodal model for autonomous driving.
\newblock \emph{arXiv preprint arXiv:2505.00284}, 2025.

\bibitem[Radford et~al.(2021)Radford, Kim, Hallacy, Ramesh, Goh, Agarwal, Sastry, Askell, Mishkin, Clark, Krueger, and Sutskever]{radford2021learning}
Alec Radford, Jong~Wook Kim, Chris Hallacy, Aditya Ramesh, Gabriel Goh, Sandhini Agarwal, Girish Sastry, Amanda Askell, Pamela Mishkin, Jack Clark, Gretchen Krueger, and Ilya Sutskever.
\newblock Learning transferable visual models from natural language supervision.
\newblock In \emph{Int. Conf. Mach. Learn.}, pages 8748--8763. PMLR, 2021.

\bibitem[Rawte et~al.(2023)Rawte, Sheth, and Das]{rawte2023survey}
Vipula Rawte, Amit Sheth, and Amitava Das.
\newblock A survey of hallucination in large foundation models.
\newblock \emph{arXiv preprint arXiv:2309.05922}, 2023.

\bibitem[Ren et~al.(2025)Ren, Lu, Cao, Gao, Huang, Sabour, Shen, Pfaff, Wu, Chen, Kim, Gao, Leal-Taixe, Chen, Fidler, and Ling]{ren2025cosmos}
Xuanchi Ren, Yifan Lu, Tianshi Cao, Ruiyuan Gao, Shengyu Huang, Amirmojtaba Sabour, Tianchang Shen, Tobias Pfaff, Jay~Zhangjie Wu, Runjian Chen, Seung~Wook Kim, Jun Gao, Laura Leal-Taixe, Mike Chen, Sanja Fidler, and Huan Ling.
\newblock {Cosmos-Drive-Dreams}: Scalable synthetic driving data generation with world foundation models.
\newblock \emph{arXiv preprint arXiv:2506.09042}, 2025.

\bibitem[Renz et~al.(2025)Renz, Chen, Arani, and Sinavski]{renz2025simlingo}
Katrin Renz, Long Chen, Elahe Arani, and Oleg Sinavski.
\newblock {SimLingo}: Vision-only closed-loop autonomous driving with language-action alignment.
\newblock In \emph{IEEE/CVF Conf. Comput. Vis. Pattern Recog.}, pages 11993--12003, 2025.

\bibitem[Rombach et~al.(2022)Rombach, Blattmann, Lorenz, Esser, and Ommer]{rombach2022high}
Robin Rombach, Andreas Blattmann, Dominik Lorenz, Patrick Esser, and Bj\"orn Ommer.
\newblock High-resolution image synthesis with latent diffusion models.
\newblock In \emph{IEEE/CVF Conf. Comput. Vis. Pattern Recog.}, pages 10684--10695, 2022.

\bibitem[Ross et~al.(2011)Ross, Gordon, and Bagnell]{ross2011reduction}
St{\'e}phane Ross, Geoffrey Gordon, and Drew Bagnell.
\newblock A reduction of imitation learning and structured prediction to no-regret online learning.
\newblock In \emph{Int. Conf. Artifi. Intell. Stat.}, pages 627--635, 2011.

\bibitem[Rowe et~al.(2025)Rowe, de~Schaetzen, Girgis, Pal, and Paull]{rowe2025poutine}
Luke Rowe, Rodrigue de~Schaetzen, Roger Girgis, Christopher Pal, and Liam Paull.
\newblock Poutine: Vision-language-trajectory pre-training and reinforcement learning post-training enable robust end-to-end autonomous driving.
\newblock \emph{arXiv preprint arXiv:2506.11234}, 2025.

\bibitem[Sapkota et~al.(2025)Sapkota, Cao, Roumeliotis, and Karkee]{sapkota2025vision}
Ranjan Sapkota, Yang Cao, Konstantinos~I. Roumeliotis, and Manoj Karkee.
\newblock Vision-language-action models: Concepts, progress, applications and challenges.
\newblock \emph{arXiv preprint arXiv:2505.04769}, 2025.

\bibitem[Scheel et~al.(2022)Scheel, Bergamini, Wołczyk, Osiński, and Ondruska]{scheel2022urban}
Oliver Scheel, Luca Bergamini, Maciej Wołczyk, Błażej Osiński, and Peter Ondruska.
\newblock Urban driver: Learning to drive from real-world demonstrations using policy gradients.
\newblock In \emph{Conf. Robot Learn.}, pages 718--728. PMLR, 2022.

\bibitem[Schrum et~al.(2024)Schrum, Sumner, Gombolay, and Best]{schrum2024maveric}
Mariah~L Schrum, Emily Sumner, Matthew~C Gombolay, and Andrew Best.
\newblock Maveric: A data-driven approach to personalized autonomous driving.
\newblock \emph{IEEE Trans. Robot.}, 40:\penalty0 1952--1965, 2024.

\bibitem[Shan et~al.(2025)Shan, Li, Jiang, Fan, Yan, Li, Hao, Zhao, Cui, Ren, and Yu]{shan2025stability}
Hao Shan, Ruikai Li, Han Jiang, Yizhe Fan, Ziyang Yan, Bohan Li, Xiaoshuai Hao, Hao Zhao, Zhiyong Cui, Yilong Ren, and Haiyang Yu.
\newblock Stability under scrutiny: Benchmarking representation paradigms for online {HD} mapping.
\newblock \emph{arXiv preprint arXiv:2510.10660}, 2025.

\bibitem[Shao et~al.(2023)Shao, Wang, Chen, Li, and Liu]{shao2023safety}
Hao Shao, Letian Wang, Ruobing Chen, Hongsheng Li, and Yu~Liu.
\newblock Safety-enhanced autonomous driving using interpretable sensor fusion transformer.
\newblock In \emph{Conf. Robot Learn.}, pages 726--737. PMLR, 2023.

\bibitem[Shao et~al.(2024{\natexlab{a}})Shao, Hu, Wang, Song, Waslander, Liu, and Li]{shao2024lmdrive}
Hao Shao, Yuxuan Hu, Letian Wang, Guanglu Song, Steven~L. Waslander, Yu~Liu, and Hongsheng Li.
\newblock {LMDrive}: Closed-loop end-to-end driving with large language models.
\newblock In \emph{IEEE/CVF Conf. Comput. Vis. Pattern Recog.}, pages 15120--15130, 2024{\natexlab{a}}.

\bibitem[Shao et~al.(2025)Shao, Li, Zhang, Zhang, Liu, Chen, and Nie]{shao2025large}
Rui Shao, Wei Li, Lingsen Zhang, Renshan Zhang, Zhiyang Liu, Ran Chen, and Liqiang Nie.
\newblock Large {VLM}-based vision-language-action models for robotic manipulation: A survey.
\newblock \emph{arXiv preprint arXiv:2508.13073}, 2025.

\bibitem[Shao et~al.(2024{\natexlab{b}})Shao, Wang, Zhu, Xu, Song, Bi, Zhang, Zhang, Li, Wu, and Guo]{shao2024deepseekmath}
Zhihong Shao, Peiyi Wang, Qihao Zhu, Runxin Xu, Junxiao Song, Xiao Bi, Haowei Zhang, Mingchuan Zhang, Y.K. Li, Y.~Wu, and Daya Guo.
\newblock {DeepSeekMath}: Pushing the limits of mathematical reasoning in open language models.
\newblock \emph{arXiv preprint arXiv:2402.03300}, 2024{\natexlab{b}}.

\bibitem[Sima et~al.(2024)Sima, Renz, Chitta, Chen, Zhang, Xie, Beißwenger, Luo, Geiger, , and Li]{sima2024drivelm}
Chonghao Sima, Katrin Renz, Kashyap Chitta, Li~Chen, Hanxue Zhang, Chengen Xie, Jens Beißwenger, Ping Luo, Andreas Geiger, , and Hongyang Li.
\newblock {DriveLM}: Driving with graph visual question answering.
\newblock In \emph{Eur. Conf. Comput. Vis.}, pages 256--274. Springer, 2024.

\bibitem[Song et~al.(2025{\natexlab{a}})Song, Zhang, Zhu, Deng, and Zhang]{song2025lmad}
Nan Song, Bozhou Zhang, Xiatian Zhu, Jiankang Deng, and Li~Zhang.
\newblock {LMAD}: Integrated end-to-end vision-language model for explainable autonomous driving.
\newblock \emph{arXiv preprint arXiv:2508.12404}, 2025{\natexlab{a}}.

\bibitem[Song et~al.(2025{\natexlab{b}})Song, Guo, Peng, Wei, Wu, and Chen]{song2025insightdrive}
Ruiqi Song, Xianda Guo, Yanlun Peng, Qinggong Wei, Hangbin Wu, and Long Chen.
\newblock {InsightDrive}: Insight scene representation for end-to-end autonomous driving.
\newblock \emph{arXiv preprint arXiv:2503.13047}, 2025{\natexlab{b}}.

\bibitem[Song et~al.(2023)Song, He, Li, Ma, Ming, Mao, Pei, Peng, Hu, Yao, and Zhang]{song2023synthetic}
Zhihang Song, Zimin He, Xingyu Li, Qiming Ma, Ruibo Ming, Zhiqi Mao, Huaxin Pei, Lihui Peng, Jianming Hu, Danya Yao, and Yi~Zhang.
\newblock Synthetic datasets for autonomous driving: A survey.
\newblock \emph{IEEE Trans. Intell. Veh.}, 9\penalty0 (1):\penalty0 1847--1864, 2023.

\bibitem[Song et~al.(2025{\natexlab{c}})Song, Jia, Liu, Pan, Zhang, Wang, Zhang, Xu, Yang, and Luo]{song2025don}
Ziying Song, Caiyan Jia, Lin Liu, Hongyu Pan, Yongchang Zhang, Junming Wang, Xingyu Zhang, Shaoqing Xu, Lei Yang, and Yadan Luo.
\newblock Don't shake the wheel: Momentum-aware planning in end-to-end autonomous driving.
\newblock In \emph{IEEE/CVF Conf. Comput. Vis. Pattern Recog.}, pages 22432--22441, 2025{\natexlab{c}}.

\bibitem[Steiner et~al.(2024)Steiner, Pinto, Tschannen, Keysers, Wang, Bitton, Gritsenko, Minderer, Sherbondy, Long, Qin, Ingle, Bugliarello, Kazemzadeh, Mesnard, Alabdulmohsin, Beyer, and Zhai]{steiner2024paligemma}
Andreas Steiner, André~Susano Pinto, Michael Tschannen, Daniel Keysers, Xiao Wang, Yonatan Bitton, Alexey Gritsenko, Matthias Minderer, Anthony Sherbondy, Shangbang Long, Siyang Qin, Reeve Ingle, Emanuele Bugliarello, Sahar Kazemzadeh, Thomas Mesnard, Ibrahim Alabdulmohsin, Lucas Beyer, and Xiaohua Zhai.
\newblock {PaliGemma 2}: A family of versatile {VLMs} for transfer.
\newblock \emph{arXiv preprint arXiv:2412.03555}, 2024.

\bibitem[Su et~al.(2024)Su, Wu, and Yan]{su2024difsd}
Haisheng Su, Wei Wu, and Junchi Yan.
\newblock {DiFSD}: Ego-centric fully sparse paradigm with uncertainty denoising and iterative refinement for efficient end-to-end self-driving.
\newblock \emph{arXiv preprint arXiv:2409.09777}, 2024.

\bibitem[Sun et~al.(2025{\natexlab{a}})Sun, Cao, Wang, Wang, Shang, Feng, Lu, Shi, Yang, Yan, and Song]{suna2025minddrive}
Bin Sun, Yaoguang Cao, Yan Wang, Rui Wang, Jiachen Shang, Xiejie Feng, Jiayi Lu, Jia Shi, Shichun Yang, Xiaoyu Yan, and Ziying Song.
\newblock {MindDrive}: An all-in-one framework bridging world models and vision-language model for end-to-end autonomous driving.
\newblock \emph{arXiv preprint arXiv:2512.04441}, 2025{\natexlab{a}}.

\bibitem[Sun et~al.(2025{\natexlab{b}})Sun, Zhang, Ding, and Zheng]{sun2025echo}
Jintao Sun, Hu~Zhang, Gangyi Ding, and Zhedong Zheng.
\newblock Echo planning for autonomous driving: From current observations to future trajectories and back.
\newblock \emph{arXiv preprint arXiv:2505.18945}, 2025{\natexlab{b}}.

\bibitem[Sun et~al.(2020)Sun, Kretzschmar, Dotiwalla, Chouard, Patnaik, Tsui, Guo, Zhou, Chai, Caine, Vasudevan, Han, Ngiam, Zhao, Timofeev, Ettinger, Krivokon, Gao, Joshi, Zhang, Shlens, Chen, and Anguelov]{sun2020waymo}
Pei Sun, Henrik Kretzschmar, Xerxes Dotiwalla, Aurelien Chouard, Vijaysai Patnaik, Paul Tsui, James Guo, Yin Zhou, Yuning Chai, Benjamin Caine, Vijay Vasudevan, Wei Han, Jiquan Ngiam, Hang Zhao, Aleksei Timofeev, Scott Ettinger, Maxim Krivokon, Amy Gao, Aditya Joshi, Yu~Zhang, Jonathon Shlens, Zhifeng Chen, and Dragomir Anguelov.
\newblock Scalability in perception for autonomous driving: Waymo open dataset.
\newblock In \emph{IEEE/CVF Conf. Comput. Vis. Pattern Recog.}, pages 2446--2454, 2020.

\bibitem[Sun et~al.(2024)Sun, Jiang, Chen, Zhang, Peng, Luo, and Yuan]{sun2024autoregressive}
Peize Sun, Yi~Jiang, Shoufa Chen, Shilong Zhang, Bingyue Peng, Ping Luo, and Zehuan Yuan.
\newblock Autoregressive model beats diffusion: Llama for scalable image generation.
\newblock \emph{arXiv preprint arXiv:2406.06525}, 2024.

\bibitem[Sun et~al.(2025{\natexlab{c}})Sun, Lin, Shi, Zhang, Wu, and Zheng]{sun2025sparsedrive}
Wenchao Sun, Xuewu Lin, Yining Shi, Chuang Zhang, Haoran Wu, and Sifa Zheng.
\newblock {SparseDrive}: End-to-end autonomous driving via sparse scene representation.
\newblock In \emph{IEEE Int. Conf. Robot. Autom.}, pages 8795--8801, 2025{\natexlab{c}}.

\bibitem[Sutton et~al.(1998)Sutton, Barto, et~al.]{sutton1998reinforcement}
Richard~S Sutton, Andrew~G Barto, et~al.
\newblock \emph{Reinforcement Learning: An Introduction}, volume~1.
\newblock MIT Press, Cambridge, 1998.

\bibitem[Tan and Le(2019)]{tan2019efficientnet}
Mingxing Tan and Quoc Le.
\newblock {EfficientNet}: Rethinking model scaling for convolutional neural networks.
\newblock In \emph{Int. Conf. Mach. Learn.}, pages 6105--6114. PMLR, 2019.

\bibitem[Tan et~al.(2025)Tan, Chitta, Chen, Tian, You, Wang, Luo, Cao, Krahenbuhl, Pavone, and Ivanovic]{tan2025latent}
Shuhan Tan, Kashyap Chitta, Yuxiao Chen, Ran Tian, Yurong You, Yan Wang, Wenjie Luo, Yulong Cao, Philipp Krahenbuhl, Marco Pavone, and Boris Ivanovic.
\newblock Latent chain-of-thought world modeling for end-to-end driving.
\newblock \emph{arXiv preprint arXiv:2512.10226}, 2025.

\bibitem[Tang et~al.(2025)Tang, Liao, Nie, He, Qu, Chen, Ma, Li, Sun, and Xu]{tang2025e3ad}
Yihong Tang, Haicheng Liao, Tong Nie, Junlin He, Ao~Qu, Kehua Chen, Wei Ma, Zhenning Li, Lijun Sun, and Chengzhong Xu.
\newblock {E3AD}: An emotion-aware vision-language-action model for human-centric end-to-end autonomous driving.
\newblock \emph{arXiv preprint arXiv:2512.04733}, 2025.

\bibitem[Team(2024{\natexlab{a}})]{team2024chameleon}
Chameleon Team.
\newblock Chameleon: Mixed-modal early-fusion foundation models.
\newblock \emph{arXiv preprint arXiv:2405.09818}, 2024{\natexlab{a}}.

\bibitem[Team et~al.(2023)Team, Anil, Borgeaud, Alayrac, Yu, Soricut, Schalkwyk, Dai, Hauth, Millican, Silver, Johnson, Antonoglou, Schrittwieser, Glaese, Chen, Pitler, Lillicrap, Lazaridou, Firat, Molloy, Isard, Barham, Hennigan, Lee, Viola, Reynolds, Xu, Doherty, Collins, Meyer, Rutherford, Moreira, Ayoub, Goel, Krawczyk, Du, Chi, Cheng, Ni, Shah, Kane, Chan, Faruqui, Severyn, Lin, Li, Cheng, Ittycheriah, Mahdieh, Chen, Sun, Tran, Bagri, Lakshminarayanan, Liu, Orban, Güra, Zhou, Song, Boffy, Ganapathy, Zheng, Choe, Ágoston Weisz, Zhu, Lu, Gopal, Kahn, Kula, Pitman, Shah, Taropa, Merey, Baeuml, Chen, Shafey, Zhang, Sercinoglu, Tucker, Piqueras, Krikun, Barr, Savinov, Danihelka, Roelofs, White, Andreassen, von Glehn, Yagati, Kazemi, Gonzalez, Khalman, Sygnowski, Frechette, Smith, Culp, et~al.]{team2023gemini}
Gemini Team, Rohan Anil, Sebastian Borgeaud, Jean-Baptiste Alayrac, Jiahui Yu, Radu Soricut, Johan Schalkwyk, Andrew~M. Dai, Anja Hauth, Katie Millican, David Silver, Melvin Johnson, Ioannis Antonoglou, Julian Schrittwieser, Amelia Glaese, Jilin Chen, Emily Pitler, Timothy Lillicrap, Angeliki Lazaridou, Orhan Firat, James Molloy, Michael Isard, Paul~R. Barham, Tom Hennigan, Benjamin Lee, Fabio Viola, Malcolm Reynolds, Yuanzhong Xu, Ryan Doherty, Eli Collins, Clemens Meyer, Eliza Rutherford, Erica Moreira, Kareem Ayoub, Megha Goel, Jack Krawczyk, Cosmo Du, Ed~Chi, Heng-Tze Cheng, Eric Ni, Purvi Shah, Patrick Kane, Betty Chan, Manaal Faruqui, Aliaksei Severyn, Hanzhao Lin, YaGuang Li, Yong Cheng, Abe Ittycheriah, Mahdis Mahdieh, Mia Chen, Pei Sun, Dustin Tran, Sumit Bagri, Balaji Lakshminarayanan, Jeremiah Liu, Andras Orban, Fabian Güra, Hao Zhou, Xinying Song, Aurelien Boffy, Harish Ganapathy, Steven Zheng, HyunJeong Choe, Ágoston Weisz, Tao Zhu, Yifeng Lu, Siddharth Gopal, Jarrod Kahn, Maciej Kula, Jeff
  Pitman, Rushin Shah, Emanuel Taropa, Majd~Al Merey, Martin Baeuml, Zhifeng Chen, Laurent~El Shafey, Yujing Zhang, Olcan Sercinoglu, George Tucker, Enrique Piqueras, Maxim Krikun, Iain Barr, Nikolay Savinov, Ivo Danihelka, Becca Roelofs, Anaïs White, Anders Andreassen, Tamara von Glehn, Lakshman Yagati, Mehran Kazemi, Lucas Gonzalez, Misha Khalman, Jakub Sygnowski, Alexandre Frechette, Charlotte Smith, Laura Culp, et~al.
\newblock Gemini: A family of highly capable multimodal models.
\newblock \emph{arXiv preprint arXiv:2312.11805}, 2023.

\bibitem[Team et~al.(2025)Team, Abeyruwan, Ainslie, Alayrac, Arenas, Armstrong, Balakrishna, Baruch, Bauza, Blokzijl, Bohez, Bousmalis, Brohan, Buschmann, Byravan, Cabi, Caluwaerts, Casarini, Chang, Chen, Chen, Chiang, Choromanski, D'Ambrosio, Dasari, Davchev, Devin, Palo, Ding, Dostmohamed, Driess, Du, Dwibedi, Elabd, Fantacci, Fong, Frey, Fu, Giustina, Gopalakrishnan, Graesser, Hasenclever, Heess, Hernaez, Herzog, Hofer, Humplik, Iscen, Jacob, Jain, Julian, Kalashnikov, Karagozler, Karp, Kew, Kirkland, Kirmani, Kuang, Lampe, Laurens, Leal, Lee, Lee, Liang, Lin, Maddineni, Majumdar, Michaely, Moreno, Neunert, Nori, Parada, Parisotto, Pastor, Pooley, Rao, Reymann, Sadigh, Saliceti, Sanketi, Sermanet, Shah, Sharma, Shea, Shu, Sindhwani, Singh, Soricut, Springenberg, Sterneck, Surdulescu, Tan, Tompson, Vanhoucke, Varley, Vesom, Vezzani, Vinyals, Wahid, Welker, et~al.]{team2025gemini}
Gemini~Robotics Team, Saminda Abeyruwan, Joshua Ainslie, Jean-Baptiste Alayrac, Montserrat~Gonzalez Arenas, Travis Armstrong, Ashwin Balakrishna, Robert Baruch, Maria Bauza, Michiel Blokzijl, Steven Bohez, Konstantinos Bousmalis, Anthony Brohan, Thomas Buschmann, Arunkumar Byravan, Serkan Cabi, Ken Caluwaerts, Federico Casarini, Oscar Chang, Jose~Enrique Chen, Xi~Chen, Hao-Tien~Lewis Chiang, Krzysztof Choromanski, David D'Ambrosio, Sudeep Dasari, Todor Davchev, Coline Devin, Norman~Di Palo, Tianli Ding, Adil Dostmohamed, Danny Driess, Yilun Du, Debidatta Dwibedi, Michael Elabd, Claudio Fantacci, Cody Fong, Erik Frey, Chuyuan Fu, Marissa Giustina, Keerthana Gopalakrishnan, Laura Graesser, Leonard Hasenclever, Nicolas Heess, Brandon Hernaez, Alexander Herzog, R.~Alex Hofer, Jan Humplik, Atil Iscen, Mithun~George Jacob, Deepali Jain, Ryan Julian, Dmitry Kalashnikov, M.~Emre Karagozler, Stefani Karp, Chase Kew, Jerad Kirkland, Sean Kirmani, Yuheng Kuang, Thomas Lampe, Antoine Laurens, Isabel Leal, Alex~X. Lee,
  Tsang-Wei~Edward Lee, Jacky Liang, Yixin Lin, Sharath Maddineni, Anirudha Majumdar, Assaf~Hurwitz Michaely, Robert Moreno, Michael Neunert, Francesco Nori, Carolina Parada, Emilio Parisotto, Peter Pastor, Acorn Pooley, Kanishka Rao, Krista Reymann, Dorsa Sadigh, Stefano Saliceti, Pannag Sanketi, Pierre Sermanet, Dhruv Shah, Mohit Sharma, Kathryn Shea, Charles Shu, Vikas Sindhwani, Sumeet Singh, Radu Soricut, Jost~Tobias Springenberg, Rachel Sterneck, Razvan Surdulescu, Jie Tan, Jonathan Tompson, Vincent Vanhoucke, Jake Varley, Grace Vesom, Giulia Vezzani, Oriol Vinyals, Ayzaan Wahid, Stefan Welker, et~al.
\newblock Gemini robotics: Bringing {AI} into the physical world.
\newblock \emph{arXiv preprint arXiv:2503.20020}, 2025.

\bibitem[Team(2024{\natexlab{b}})]{qwen1.5}
Qwen Team.
\newblock Introducing {Qwen1.5}, February 2024{\natexlab{b}}.
\newblock URL \url{https://qwenlm.github.io/blog/qwen1.5}.

\bibitem[Team(2024{\natexlab{c}})]{wayve2024lingo}
Wayve Team.
\newblock {LINGO-2}: Driving with natural language, 2024{\natexlab{c}}.
\newblock URL \url{https://wayve.ai/thinking/lingo-2-driving-with-language}.

\bibitem[Team et~al.(2024)]{waywe2024lingo}
Waywe~Research Team et~al.
\newblock {LINGO-2}: Driving with natural language, 2024.

\bibitem[Tian et~al.(2025{\natexlab{a}})Tian, Li, Weng, Chen, Schmerling, Wang, Ivanovic, and Pavone]{tian2024tokenize}
Ran Tian, Boyi Li, Xinshuo Weng, Yuxiao Chen, Edward Schmerling, Yue Wang, Boris Ivanovic, and Marco Pavone.
\newblock Tokenize the world into object-level knowledge to address long-tail events in autonomous driving.
\newblock In \emph{Conf. Robot Learn.}, pages 3656--3673. PMLR, 2025{\natexlab{a}}.

\bibitem[Tian et~al.(2023)Tian, Jiang, Yun, Mao, Yang, Wang, Wang, and Zhao]{tian2023occ3d}
Xiaoyu Tian, Tao Jiang, Longfei Yun, Yucheng Mao, Huitong Yang, Yue Wang, Yilun Wang, and Hang Zhao.
\newblock {Occ3D}: A large-scale {3D} occupancy prediction benchmark for autonomous driving.
\newblock In \emph{Adv. Neural Inf. Process. Syst.}, volume~36, pages 64318--64330, 2023.

\bibitem[Tian et~al.(2025{\natexlab{b}})Tian, Gu, Li, Liu, Wang, Zhao, Zhan, Jia, Lang, and Zhao]{tian2024drivevlm}
Xiaoyu Tian, Junru Gu, Bailin Li, Yicheng Liu, Yang Wang, Zhiyong Zhao, Kun Zhan, Peng Jia, Xianpeng Lang, and Hang Zhao.
\newblock {DriveVLM}: The convergence of autonomous driving and large vision-language models.
\newblock In \emph{Conf. Robot Learn.}, pages 4698--4726. PMLR, 2025{\natexlab{b}}.

\bibitem[Tong et~al.(2023)Tong, Sima, Wang, Chen, Wu, Deng, Gu, Lu, Luo, Lin, and Li]{tong2023scene}
Wenwen Tong, Chonghao Sima, Tai Wang, Li~Chen, Silei Wu, Hanming Deng, Yi~Gu, Lewei Lu, Ping Luo, Dahua Lin, and Hongyang Li.
\newblock Scene as occupancy.
\newblock In \emph{IEEE/CVF Int. Conf. Comput. Vis.}, pages 8406--8415, 2023.

\bibitem[Toromanoff et~al.(2020)Toromanoff, Wirbel, and Moutarde]{toromanoff2020end}
Marin Toromanoff, Emilie Wirbel, and Fabien Moutarde.
\newblock End-to-end model-free reinforcement learning for urban driving using implicit affordances.
\newblock In \emph{IEEE/CVF Conf. Comput. Vis. Pattern Recog.}, pages 7153--7162, 2020.

\bibitem[Touvron et~al.(2023)Touvron, Martin, Stone, Albert, Almahairi, Babaei, Bashlykov, Batra, Bhargava, Bhosale, Bikel, Blecher, Ferrer, Chen, Cucurull, Esiobu, Fernandes, Fu, Fu, Fuller, Gao, Goswami, Goyal, Hartshorn, Hosseini, Hou, Inan, Kardas, Kerkez, Khabsa, Kloumann, Korenev, Koura, Lachaux, Lavril, Lee, Liskovich, Lu, Mao, Martinet, Mihaylov, Mishra, Molybog, Nie, Poulton, Reizenstein, Rungta, Saladi, Schelten, Silva, Smith, Subramanian, Tan, Tang, Taylor, Williams, Kuan, Xu, Yan, Zarov, Zhang, Fan, Kambadur, Narang, Rodriguez, Stojnic, Edunov, and Scialom]{touvron2023llama}
Hugo Touvron, Louis Martin, Kevin Stone, Peter Albert, Amjad Almahairi, Yasmine Babaei, Nikolay Bashlykov, Soumya Batra, Prajjwal Bhargava, Shruti Bhosale, Dan Bikel, Lukas Blecher, Cristian~Canton Ferrer, Moya Chen, Guillem Cucurull, David Esiobu, Jude Fernandes, Jeremy Fu, Wenyin Fu, Brian Fuller, Cynthia Gao, Vedanuj Goswami, Naman Goyal, Anthony Hartshorn, Saghar Hosseini, Rui Hou, Hakan Inan, Marcin Kardas, Viktor Kerkez, Madian Khabsa, Isabel Kloumann, Artem Korenev, Punit~Singh Koura, Marie-Anne Lachaux, Thibaut Lavril, Jenya Lee, Diana Liskovich, Yinghai Lu, Yuning Mao, Xavier Martinet, Todor Mihaylov, Pushkar Mishra, Igor Molybog, Yixin Nie, Andrew Poulton, Jeremy Reizenstein, Rashi Rungta, Kalyan Saladi, Alan Schelten, Ruan Silva, Eric~Michael Smith, Ranjan Subramanian, Xiaoqing~Ellen Tan, Binh Tang, Ross Taylor, Adina Williams, Jian~Xiang Kuan, Puxin Xu, Zheng Yan, Iliyan Zarov, Yuchen Zhang, Angela Fan, Melanie Kambadur, Sharan Narang, Aurelien Rodriguez, Robert Stojnic, Sergey Edunov, and Thomas
  Scialom.
\newblock {LLaMA} 2: Open foundation and fine-tuned chat models.
\newblock \emph{arXiv preprint arXiv:2307.09288}, 2023.

\bibitem[Tschannen et~al.(2025)Tschannen, Gritsenko, Wang, Naeem, Alabdulmohsin, Parthasarathy, Evans, Beyer, Xia, Mustafa, Hénaff, Harmsen, Steiner, and Zhai]{tschannen2025siglip}
Michael Tschannen, Alexey Gritsenko, Xiao Wang, Muhammad~Ferjad Naeem, Ibrahim Alabdulmohsin, Nikhil Parthasarathy, Talfan Evans, Lucas Beyer, Ye~Xia, Basil Mustafa, Olivier Hénaff, Jeremiah Harmsen, Andreas Steiner, and Xiaohua Zhai.
\newblock {SigLIP 2}: Multilingual vision-language encoders with improved semantic understanding, localization, and dense features.
\newblock \emph{arXiv preprint arXiv:2502.14786}, 2025.

\bibitem[Van Den~Oord et~al.(2017)Van Den~Oord, Vinyals, et~al.]{van2017neural}
Aaron Van Den~Oord, Oriol Vinyals, et~al.
\newblock Neural discrete representation learning.
\newblock In \emph{Adv. Neural Inf. Process. Syst.}, volume~30, pages 6309--6318, 2017.

\bibitem[Verwimp et~al.(2023)Verwimp, Yang, Parisot, Hong, McDonagh, Pérez-Pellitero, Lange, and Tuytelaars]{verwimp2023clad}
Eli Verwimp, Kuo Yang, Sarah Parisot, Lanqing Hong, Steven McDonagh, Eduardo Pérez-Pellitero, Matthias~De Lange, and Tinne Tuytelaars.
\newblock {CLAD}: A realistic continual learning benchmark for autonomous driving.
\newblock \emph{Neural Net.}, 161:\penalty0 659--669, 2023.

\bibitem[Wang et~al.(2025{\natexlab{a}})Wang, Song, He, Chen, Pan, Deng, and Gu]{wang2025hmvlm}
Daming Wang, Yuhao Song, Zijian He, Kangliang Chen, Xing Pan, Lu~Deng, and Weihao Gu.
\newblock {HMVLM}: Multistage reasoning-enhanced vision-language model for long-tailed driving scenarios.
\newblock \emph{arXiv preprint arXiv:2506.05883}, 2025{\natexlab{a}}.

\bibitem[Wang et~al.(2025{\natexlab{b}})Wang, Kaufeld, and Betz]{wang2024dualad}
Dingrui Wang, Marc Kaufeld, and Johannes Betz.
\newblock {DualAD}: Dual-layer planning for reasoning in autonomous driving.
\newblock In \emph{IEEE/RSJ Int. Conf. Intell. Robots Syst.}, 2025{\natexlab{b}}.

\bibitem[Wang et~al.(2024{\natexlab{a}})Wang, Bai, Tan, Wang, Fan, Bai, Chen, Liu, Wang, Ge, Fan, Dang, Du, Ren, Men, Liu, Zhou, Zhou, and Lin]{Qwen2-VL}
Peng Wang, Shuai Bai, Sinan Tan, Shijie Wang, Zhihao Fan, Jinze Bai, Keqin Chen, Xuejing Liu, Jialin Wang, Wenbin Ge, Yang Fan, Kai Dang, Mengfei Du, Xuancheng Ren, Rui Men, Dayiheng Liu, Chang Zhou, Jingren Zhou, and Junyang Lin.
\newblock {Qwen2-VL}: Enhancing vision-language model's perception of the world at any resolution.
\newblock \emph{arXiv preprint arXiv:2409.12191}, 2024{\natexlab{a}}.

\bibitem[Wang et~al.(2025{\natexlab{c}})Wang, Yu, Jiang, Lan, Shi, Chang, Kautz, Li, and Alvarez]{wang2025omnidrive}
Shihao Wang, Zhiding Yu, Xiaohui Jiang, Shiyi Lan, Min Shi, Nadine Chang, Jan Kautz, Ying Li, and Jose~M. Alvarez.
\newblock {OmniDrive}: A holistic vision-language dataset for autonomous driving with counterfactual reasoning.
\newblock In \emph{IEEE/CVF Conf. Comput. Vis. Pattern Recog.}, pages 22442--22452, 2025{\natexlab{c}}.

\bibitem[Wang et~al.(2024{\natexlab{b}})Wang, Yu, Li, Liu, Liu, Chen, and Zhu]{wang2024not}
Song Wang, Jiawei Yu, Wentong Li, Wenyu Liu, Xiaolu Liu, Junbo Chen, and Jianke Zhu.
\newblock Not all voxels are equal: Hardness-aware semantic scene completion with self-distillation.
\newblock In \emph{IEEE/CVF Conf. Comput. Vis. Pattern Recog.}, pages 14792--14801, 2024{\natexlab{b}}.

\bibitem[Wang et~al.(2025{\natexlab{d}})Wang, Fang, Kong, Li, Xu, Yang, Li, Zhu, and Wang]{wang2025pixelthink}
Song Wang, Gongfan Fang, Lingdong Kong, Xiangtai Li, Jianyun Xu, Sheng Yang, Qiang Li, Jianke Zhu, and Xinchao Wang.
\newblock {PixelThink}: Towards efficient chain-of-pixel reasoning.
\newblock \emph{arXiv preprint arXiv:2505.23727}, 2025{\natexlab{d}}.

\bibitem[Wang et~al.(2024{\natexlab{c}})Wang, Xie, Chu, Li, and Luo]{wang2024drivecot}
Tianqi Wang, Enze Xie, Ruihang Chu, Zhenguo Li, and Ping Luo.
\newblock {DriveCoT}: Integrating chain-of-thought reasoning with end-to-end driving.
\newblock \emph{arXiv preprint arXiv:2403.16996}, 2024{\natexlab{c}}.

\bibitem[Wang et~al.(2024{\natexlab{d}})Wang, Chen, Wang, Cao, Liu, Gao, Zhu, Zhu, Lu, Qiao, and Dai]{wang2024enhancing}
Weiyun Wang, Zhe Chen, Wenhai Wang, Yue Cao, Yangzhou Liu, Zhangwei Gao, Jinguo Zhu, Xizhou Zhu, Lewei Lu, Yu~Qiao, and Jifeng Dai.
\newblock Enhancing the reasoning ability of multimodal large language models via mixed preference optimization.
\newblock \emph{arXiv preprint arXiv:2411.10442}, 2024{\natexlab{d}}.

\bibitem[Wang et~al.(2024{\natexlab{e}})Wang, Zhu, Huang, Chen, Zhu, and Lu]{wang2024drivedreamer}
Xiaofeng Wang, Zheng Zhu, Guan Huang, Xinze Chen, Jiagang Zhu, and Jiwen Lu.
\newblock {DriveDreamer}: Towards real-world-drive world models for autonomous driving.
\newblock In \emph{Eur. Conf. Comput. Vis.}, pages 55--72. Springer, 2024{\natexlab{e}}.

\bibitem[Wang et~al.(2025{\natexlab{e}})Wang, Luo, Bai, Cao, Che, Chen, Chen, Diamond, Ding, Ding, Feng, Heinrich, Huang, Karkus, Li, Li, Lin, Liu, Liu, Liu, Liu, Lu, Mao, Molchanov, Pavao, Peng, Ranzinger, Schmerling, Shen, Shi, Tariq, Tian, Wekel, Weng, Xiao, Yang, Yang, You, Zeng, Zhang, Ivanovic, and Pavone]{wang2025alpamayo}
Yan Wang, Wenjie Luo, Junjie Bai, Yulong Cao, Tong Che, Ke~Chen, Yuxiao Chen, Jenna Diamond, Yifan Ding, Wenhao Ding, Liang Feng, Greg Heinrich, Jack Huang, Peter Karkus, Boyi Li, Pinyi Li, Tsung-Yi Lin, Dongran Liu, Ming-Yu Liu, Langechuan Liu, Zhijian Liu, Jason Lu, Yunxiang Mao, Pavlo Molchanov, Lindsey Pavao, Zhenghao Peng, Mike Ranzinger, Ed~Schmerling, Shida Shen, Yunfei Shi, Sarah Tariq, Ran Tian, Tilman Wekel, Xinshuo Weng, Tianjun Xiao, Eric Yang, Xiaodong Yang, Yurong You, Xiaohui Zeng, Wenyuan Zhang, Boris Ivanovic, and Marco Pavone.
\newblock {Alpamayo-R1}: Bridging reasoning and action prediction for generalizable autonomous driving in the long tail.
\newblock \emph{arXiv preprint arXiv:2511.00088}, 2025{\natexlab{e}}.

\bibitem[Wang et~al.(2024{\natexlab{f}})Wang, He, Fan, Li, Chen, and Zhang.]{wang2024driving}
Yuqi Wang, Jiawei He, Lue Fan, Hongxin Li, Yuntao Chen, and Zhaoxiang Zhang.
\newblock Driving into the future: Multiview visual forecasting and planning with world model for autonomous driving.
\newblock In \emph{IEEE/CVF Conf. Comput. Vis. Pattern Recog.}, pages 14749--14759, 2024{\natexlab{f}}.

\bibitem[Wei et~al.(2024)Wei, Yuan, Li, et~al.]{wei2024occllama}
Julong Wei, Shanshuai Yuan, Pengfei Li, et~al.
\newblock {OccLLaMA}: An occupancy-language-action generative world model for autonomous driving.
\newblock \emph{arXiv preprint arXiv:2409.03272}, 2024.

\bibitem[Weng et~al.(2024)Weng, Ivanovic, Wang, Wang, and Pavone]{weng2024drive}
Xinshuo Weng, Boris Ivanovic, Yan Wang, Yue Wang, and Marco Pavone.
\newblock {PARA-Drive}: Parallelized architecture for real-time autonomous driving.
\newblock In \emph{IEEE/CVF Conf. Comput. Vis. Pattern Recog.}, pages 15449--15458, 2024.

\bibitem[Wilson et~al.(2023)Wilson, Qi, Agarwal, Lambert, Singh, Khandelwal, Pan, Kumar, Hartnett, Pontes, Ramanan, Carr, and Hays]{wilson2023argoverse}
Benjamin Wilson, William Qi, Tanmay Agarwal, John Lambert, Jagjeet Singh, Siddhesh Khandelwal, Bowen Pan, Ratnesh Kumar, Andrew Hartnett, Jhony~Kaesemodel Pontes, Deva Ramanan, Peter Carr, and James Hays.
\newblock Argoverse 2: Next generation datasets for self-driving perception and forecasting.
\newblock \emph{arXiv preprint arXiv:2301.00493}, 2023.

\bibitem[Winter et~al.(2025)Winter, Azer, and Flohr]{winter2025bevdriver}
Katharina Winter, Mark Azer, and Fabian~B. Flohr.
\newblock {BEVDriver}: Leveraging {BEV} maps in {LLMs} for robust closed-loop driving.
\newblock \emph{arXiv preprint arXiv:2503.03074}, 2025.

\bibitem[Wu et~al.(2025)Wu, Steiner, Schmidt, Marcos-Ramiro, and Stiller]{wu2025navihydra}
Hanfeng Wu, Marlon Steiner, Michael Schmidt, Alvaro Marcos-Ramiro, and Christoph Stiller.
\newblock {NaviHydra}: Controllable navigation-guided end-to-end autonomous driving with hydra-distillation.
\newblock \emph{arXiv preprint arXiv:2512.10660}, 2025.

\bibitem[Wu et~al.(2022)Wu, Jia, Chen, Yan, Li, and Qiao]{wu2022trajectoryguided}
Penghao Wu, Xiaosong Jia, Li~Chen, Junchi Yan, Hongyang Li, and Yu~Qiao.
\newblock Trajectory-guided control prediction for end-to-end autonomous driving: A simple yet strong baseline.
\newblock In \emph{Adv. Neural Inf. Process. Syst.}, volume~35, pages 6119--6132, 2022.

\bibitem[Xie et~al.(2024)Xie, Zhou, and Cheng]{jihong2024edge}
Jihong Xie, Xiang Zhou, and Lu~Cheng.
\newblock Edge computing for real-time decision making in autonomous driving: Review of challenges, solutions, and future trends.
\newblock \emph{Int. J. Adv. Comput. Sci. \& Appl.}, 15\penalty0 (7), 2024.

\bibitem[Xie et~al.(2025{\natexlab{a}})Xie, Kong, Dong, Sima, Zhang, Chen, Liu, and Pan]{xie2025vlms}
Shaoyuan Xie, Lingdong Kong, Yuhao Dong, Chonghao Sima, Wenwei Zhang, Qi~Alfred Chen, Ziwei Liu, and Liang Pan.
\newblock Are {VLMs} ready for autonomous driving? an empirical study from the reliability, data, and metric perspectives.
\newblock In \emph{IEEE/CVF Int. Conf. Comput. Vis.}, pages 6585--6597, 2025{\natexlab{a}}.

\bibitem[Xie et~al.(2025{\natexlab{b}})Xie, Kong, Zhang, Ren, Pan, Chen, and Liu]{xie2025benchmarking}
Shaoyuan Xie, Lingdong Kong, Wenwei Zhang, Jiawei Ren, Liang Pan, Kai Chen, and Ziwei Liu.
\newblock Benchmarking and improving bird's eye view perception robustness in autonomous driving.
\newblock \emph{IEEE Trans. Pattern Anal. Mach. Intell.}, 47\penalty0 (5):\penalty0 3878--3894, 2025{\natexlab{b}}.

\bibitem[Xie et~al.(2025{\natexlab{c}})Xie, Xu, He, Hwang, Luo, Ji, Lin, Chen, Lu, Leng, Anguelov, and Tan]{xie2025s4}
Yichen Xie, Runsheng Xu, Tong He, Jyh-Jing Hwang, Katie Luo, Jingwei Ji, Hubert Lin, Letian Chen, Yiren Lu, Zhaoqi Leng, Dragomir Anguelov, and Mingxing Tan.
\newblock {S4-Driver}: Scalable self-supervised driving multimodal large language model with spatio-temporal visual representation.
\newblock In \emph{IEEE/CVF Conf. Comput. Vis. Pattern Recog.}, pages 1622--1632, 2025{\natexlab{c}}.

\bibitem[Xin et~al.(2025)Xin, Liu, Mei, Liu, Ye, Chen, and Ma]{netroller}
Ren Xin, Hongji Liu, Xiaodong Mei, Wenru Liu, Maosheng Ye, Zhili Chen, and Jun Ma.
\newblock {NetRoller}: Interfacing general and specialized models for end-to-end autonomous driving.
\newblock \emph{arXiv preprint arXiv:2506.14589}, 2025.

\bibitem[Xing et~al.(2025{\natexlab{a}})Xing, Qian, Wang, Hua, Tian, Zhou, and Tu]{xing2025openemma}
Shuo Xing, Chengyuan Qian, Yuping Wang, Hongyuan Hua, Kexin Tian, Yang Zhou, and Zhengzhong Tu.
\newblock {OpenEMMA}: Open-source multimodal model for end-to-end autonomous driving.
\newblock In \emph{IEEE/CVF Winter Conf. Appl. Comput. Vis.}, pages 1001--1009, 2025{\natexlab{a}}.

\bibitem[Xing et~al.(2025{\natexlab{b}})Xing, Zhang, Hu, Jiang, He, Zhang, Long, and Yin]{xing2025goalflow}
Zebin Xing, Xingyu Zhang, Yang Hu, Bo~Jiang, Tong He, Qian Zhang, Xiaoxiao Long, and Wei Yin.
\newblock {GoalFlow}: Goal-driven flow matching for multimodal trajectories generation in end-to-end autonomous driving.
\newblock In \emph{IEEE/CVF Conf. Comput. Vis. Pattern Recog.}, pages 1602--1611, 2025{\natexlab{b}}.

\bibitem[Xing et~al.(2025{\natexlab{c}})Xing, Zheng, Zhang, Ding, Yang, Gu, Xia, and Zhao]{xing2025mimir}
Zebin Xing, Yupeng Zheng, Qichao Zhang, Zhixing Ding, Pengxuan Yang, Songen Gu, Zhongpu Xia, and Dongbin Zhao.
\newblock Mimir: Hierarchical goal-driven diffusion with uncertainty propagation for end-to-end autonomous driving.
\newblock \emph{arXiv preprint arXiv:2512.07130}, 2025{\natexlab{c}}.

\bibitem[Xu et~al.(2025{\natexlab{a}})Xu, Liu, Guo, Zhang, Hang, and Sun]{xu2025towards}
Chengkai Xu, Jiaqi Liu, Yicheng Guo, Yuhang Zhang, Peng Hang, and Jian Sun.
\newblock Towards human-centric autonomous driving: A fast-slow architecture integrating large language model guidance with reinforcement learning.
\newblock \emph{arXiv preprint arXiv:2505.06875}, 2025{\natexlab{a}}.

\bibitem[Xu et~al.(2025{\natexlab{b}})Xu, Peng, Tan, Chang, Zhao, and Tian]{xu2025temporal}
Haoran Xu, Peixi Peng, Guang Tan, Yiqian Chang, Yisen Zhao, and Yonghong Tian.
\newblock Temporal triplane transformers as occupancy world models.
\newblock \emph{arXiv preprint arXiv:2503.07338}, 2025{\natexlab{b}}.

\bibitem[Xu et~al.(2025{\natexlab{c}})Xu, Lin, Jeon, Feng, Zou, Sun, Gorman, Tolstaya, Tang, White, Sapp, Tan, Hwang, and Anguelov]{xu2025wod}
Runsheng Xu, Hubert Lin, Wonseok Jeon, Hao Feng, Yuliang Zou, Liting Sun, John Gorman, Ekaterina Tolstaya, Sarah Tang, Brandyn White, Ben Sapp, Mingxing Tan, Jyh-Jing Hwang, and Dragomir Anguelov.
\newblock {WOD-E2E}: Waymo open dataset for end-to-end driving in challenging long-tail scenarios.
\newblock \emph{arXiv preprint arXiv:2510.26125}, 2025{\natexlab{c}}.

\bibitem[Xu et~al.(2025{\natexlab{d}})Xu, Lu, Yan, Cai, Liu, and Chen]{xu2025occ}
Tianshuo Xu, Hao Lu, Xu~Yan, Yingjie Cai, Bingbing Liu, and Yingcong Chen.
\newblock {Occ-LLM}: Enhancing autonomous driving with occupancy-based large language models.
\newblock In \emph{IEEE Int. Conf. Robot. Autom.}, 2025{\natexlab{d}}.

\bibitem[Xu et~al.(2025{\natexlab{e}})Xu, Kong, Wang, Zhou, and Liu]{xu2025beyond}
Xiang Xu, Lingdong Kong, Song Wang, Chuanwei Zhou, and Qingshan Liu.
\newblock Beyond one shot, beyond one perspective: Cross-view and long-horizon distillation for better {LiDAR} representations.
\newblock In \emph{IEEE/CVF Int. Conf. Comput. Vis.}, pages 25506--25518, 2025{\natexlab{e}}.

\bibitem[Xu et~al.(2024{\natexlab{a}})Xu, Hu, Zhang, Meyer, Mustikovela, Srinivasa, Wolff, and Huang]{xu2024vlm}
Yi~Xu, Yuxin Hu, Zaiwei Zhang, Gregory~P. Meyer, Siva~Karthik Mustikovela, Siddhartha Srinivasa, Eric~M. Wolff, and Xin Huang.
\newblock {VLM-AD}: End-to-end autonomous driving through vision-language model supervision.
\newblock \emph{arXiv preprint arXiv:2412.14446}, 2024{\natexlab{a}}.

\bibitem[Xu et~al.(2024{\natexlab{b}})Xu, Zhang, Xie, Zhao, Guo, Wong, Li, and Zhao]{xu2024drivegpt4}
Zhenhua Xu, Yujia Zhang, Enze Xie, Zhen Zhao, Yong Guo, Kwan-Yee~K. Wong, Zhenguo Li, and Hengshuang Zhao.
\newblock {DriveGPT4}: Interpretable end-to-end autonomous driving via large language model.
\newblock \emph{IEEE Robot. Autom. Lett.}, 9\penalty0 (10):\penalty0 8186--8193, 2024{\natexlab{b}}.

\bibitem[Xu et~al.(2025{\natexlab{f}})Xu, Bai, Zhang, Li, Xia, Wong, Wang, and Zhao]{xu2025drivegpt4}
Zhenhua Xu, Yan Bai, Yujia Zhang, Zhuoling Li, Fei Xia, Kwan-Yee~K. Wong, Jianqiang Wang, and Hengshuang Zhao.
\newblock {DriveGPT4-V2}: Harnessing large language model capabilities for enhanced closed-loop autonomous driving.
\newblock In \emph{IEEE/CVF Conf. Comput. Vis. Pattern Recog.}, pages 17261--17270, 2025{\natexlab{f}}.

\bibitem[Yan et~al.(2025{\natexlab{a}})Yan, Tang, Gui, Li, Zhesng, Huang, Kong, Han, Zhou, Zhang, Zhan, Zhan, zhong Xu, and Shen]{yan2025ad}
Tianyi Yan, Tao Tang, Xingtai Gui, Yongkang Li, Jiasen Zhesng, Weiyao Huang, Lingdong Kong, Wencheng Han, Xia Zhou, Xueyang Zhang, Yifei Zhan, Kun Zhan, Cheng zhong Xu, and Jianbing Shen.
\newblock {AD-R1}: Closed-loop reinforcement learning for end-to-end autonomous driving with impartial world models.
\newblock \emph{arXiv preprint arXiv:2511.20325}, 2025{\natexlab{a}}.

\bibitem[Yan et~al.(2025{\natexlab{b}})Yan, Wu, Han, Jiang, Zhou, Zhan, Xu, and Shen]{yan2025drivingsphere}
Tianyi Yan, Dongming Wu, Wencheng Han, Junpeng Jiang, Xia Zhou, Kun Zhan, Cheng-zhong Xu, and Jianbing Shen.
\newblock {DrivingSphere}: Building a high-fidelity {4D} world for closed-loop simulation.
\newblock In \emph{IEEE/CVF Conf. Comput. Vis. Pattern Recog.}, pages 27531--27541, 2025{\natexlab{b}}.

\bibitem[Yan et~al.(2024)Yan, Zhang, Cai, Guo, Qiu, Gao, Zhou, Zhao, Jin, Gao, Li, Jiang, Zhang, Zhang, Dai, and Liu]{yan2024forging}
Xu~Yan, Haiming Zhang, Yingjie Cai, Jingming Guo, Weichao Qiu, Bin Gao, Kaiqiang Zhou, Yue Zhao, Huan Jin, Jiantao Gao, Zhen Li, Lihui Jiang, Wei Zhang, Hongbo Zhang, Dengxin Dai, and Bingbing Liu.
\newblock Forging vision foundation models for autonomous driving: Challenges, methodologies, and opportunities.
\newblock \emph{arXiv preprint arXiv:2401.08045}, 2024.

\bibitem[Yan et~al.(2025{\natexlab{c}})Yan, Dong, Shao, Lu, Liu, Liu, Wang, Wang, Wang, Remondino, and Ma]{yan2024renderworld}
Ziyang Yan, Wenzhen Dong, Yihua Shao, Yuhang Lu, Haiyang Liu, Jingwen Liu, Haozhe Wang, Zhe Wang, Yan Wang, Fabio Remondino, and Yuexin Ma.
\newblock {RenderWorld}: World model with self-supervised {3D} label.
\newblock In \emph{IEEE Int. Conf. Robot. Autom.}, pages 6063--6070, 2025{\natexlab{c}}.

\bibitem[Yang et~al.(2025{\natexlab{a}})Yang, Li, Yang, Zhang, Hui, Zheng, Yu, Gao, Huang, Lv, Zheng, Liu, Zhou, Huang, Hu, Ge, Wei, Lin, Tang, Yang, Tu, Zhang, Yang, Yang, Zhou, Zhou, Lin, Dang, Bao, Yang, Yu, Deng, Li, Xue, Li, Zhang, Wang, Zhu, Men, Gao, Liu, Luo, Li, Tang, Yin, Ren, Wang, Zhang, Ren, Fan, Su, Zhang, Zhang, Wan, Liu, Wang, Cui, Zhang, Zhou, and Qiu]{yang2025qwen3}
An~Yang, Anfeng Li, Baosong Yang, Beichen Zhang, Binyuan Hui, Bo~Zheng, Bowen Yu, Chang Gao, Chengen Huang, Chenxu Lv, Chujie Zheng, Dayiheng Liu, Fan Zhou, Fei Huang, Feng Hu, Hao Ge, Haoran Wei, Huan Lin, Jialong Tang, Jian Yang, Jianhong Tu, Jianwei Zhang, Jianxin Yang, Jiaxi Yang, Jing Zhou, Jingren Zhou, Junyang Lin, Kai Dang, Keqin Bao, Kexin Yang, Le~Yu, Lianghao Deng, Mei Li, Mingfeng Xue, Mingze Li, Pei Zhang, Peng Wang, Qin Zhu, Rui Men, Ruize Gao, Shixuan Liu, Shuang Luo, Tianhao Li, Tianyi Tang, Wenbiao Yin, Xingzhang Ren, Xinyu Wang, Xinyu Zhang, Xuancheng Ren, Yang Fan, Yang Su, Yichang Zhang, Yinger Zhang, Yu~Wan, Yuqiong Liu, Zekun Wang, Zeyu Cui, Zhenru Zhang, Zhipeng Zhou, and Zihan Qiu.
\newblock Qwen3 technical report.
\newblock \emph{arXiv preprint arXiv:2505.09388}, 2025{\natexlab{a}}.

\bibitem[Yang et~al.(2024{\natexlab{a}})]{yang2024qwen25}
An~Yang et~al.
\newblock Qwen2.5 technical report.
\newblock \emph{arXiv preprint arXiv:2412.15115}, 2024{\natexlab{a}}.

\bibitem[Yang et~al.(2025{\natexlab{b}})Yang, Zhou, Wu, Liu, Yang, and Lv]{yang2025human}
Haohan Yang, Yanxin Zhou, Jingda Wu, Haochen Liu, Lie Yang, and Chen Lv.
\newblock Human-guided continual learning for personalized decision-making of autonomous driving.
\newblock \emph{IEEE Int. Conf. Intell. Transport. Syst.}, 26\penalty0 (4):\penalty0 5435--5447, 2025{\natexlab{b}}.

\bibitem[Yang et~al.(2024{\natexlab{b}})Yang, Gao, Qiu, Chen, Li, Dai, Chitta, Wu, Zeng, Luo, Zhang, Geiger, Qiao, and Li]{yang2024generalized}
Jiazhi Yang, Shenyuan Gao, Yihang Qiu, Li~Chen, Tianyu Li, Bo~Dai, Kashyap Chitta, Penghao Wu, Jia Zeng, Ping Luo, Jun Zhang, Andreas Geiger, Yu~Qiao, and Hongyang Li.
\newblock Generalized predictive model for autonomous driving.
\newblock In \emph{IEEE/CVF Conf. Comput. Vis. Pattern Recog.}, pages 14662--14672, 2024{\natexlab{b}}.

\bibitem[Yang et~al.(2025{\natexlab{c}})Yang, Mei, Ma, Du, Chen, Qian, Feng, and Liu]{yang2025driving}
Yu~Yang, Jianbiao Mei, Yukai Ma, Siliang Du, Wenqing Chen, Yijie Qian, Yuxiang Feng, and Yong Liu.
\newblock Driving in the occupancy world: Vision-centric {4D} occupancy forecasting and planning via world models for autonomous driving.
\newblock In \emph{AAAI Conf. Artifi. Intell.}, volume~39, pages 9327--9335, 2025{\natexlab{c}}.

\bibitem[Yang et~al.(2024{\natexlab{c}})Yang, Chen, Sun, and Li]{yang2024visual}
Zetong Yang, Li~Chen, Yanan Sun, and Hongyang Li.
\newblock Visual point cloud forecasting enables scalable autonomous driving.
\newblock In \emph{IEEE/CVF Conf. Comput. Vis. Pattern Recog.}, pages 14673--14684, 2024{\natexlab{c}}.

\bibitem[Yang et~al.(2025{\natexlab{d}})Yang, Chai, Jia, Li, Shao, Zhu, Su, and Yan]{yang2025drivemoe}
Zhenjie Yang, Yilin Chai, Xiaosong Jia, Qifeng Li, Yuqian Shao, Xuekai Zhu, Haisheng Su, and Junchi Yan.
\newblock {DriveMoE}: Mixture-of-experts for vision-language-action model in end-to-end autonomous driving.
\newblock \emph{arXiv preprint arXiv:2505.16278}, 2025{\natexlab{d}}.

\bibitem[Yang et~al.(2025{\natexlab{e}})Yang, Jia, Li, Yang, Yao, and Yan]{yang2025raw2drive}
Zhenjie Yang, Xiaosong Jia, Qifeng Li, Xue Yang, Maoqing Yao, and Junchi Yan.
\newblock {Raw2Drive}: Reinforcement learning with aligned world models for end-to-end autonomous driving (in {CARLA-V2}).
\newblock \emph{arXiv preprint arXiv:2505.16394}, 2025{\natexlab{e}}.

\bibitem[Yang et~al.(2024{\natexlab{d}})]{yang2024llm4drive}
Zhenjie Yang et~al.
\newblock {LLM4Drive}: A survey of large language models for autonomous driving.
\newblock In \emph{Adv. Neural Inf. Process. Syst. Worksh.}, 2024{\natexlab{d}}.

\bibitem[Yao et~al.(2025)Yao, Li, Jin, Zheng, Liu, Mu, Su, Zhang, Chen, and Li]{yao2025lilodriver}
Huaiyuan Yao, Pengfei Li, Bu~Jin, Yupeng Zheng, An~Liu, Lisen Mu, Qing Su, Qian Zhang, Yilun Chen, and Peng Li.
\newblock {LiloDriver}: A lifelong learning framework for closed-loop motion planning in long-tail autonomous driving scenarios.
\newblock \emph{arXiv preprint arXiv:2505.17209}, 2025.

\bibitem[You et~al.(2025)You, Nie, Zhang, Hu, Zhou, Lu, Wen, and Li]{you2025llada}
Zebin You, Shen Nie, Xiaolu Zhang, Jun Hu, Jun Zhou, Zhiwu Lu, Ji-Rong Wen, and Chongxuan Li.
\newblock {LLaDA-V}: Large language diffusion models with visual instruction tuning.
\newblock \emph{arXiv preprint arXiv:2505.16933}, 2025.

\bibitem[Yu et~al.(2020)Yu, Chen, Wang, Xian, Chen, Liu, Madhavan, and Darrell]{yu2020bdd100k}
Fisher Yu, Haofeng Chen, Xin Wang, Wenqi Xian, Yingying Chen, Fangchen Liu, Vashisht Madhavan, and Trevor Darrell.
\newblock {BDD100K}: A diverse driving dataset for heterogeneous multitask learning.
\newblock In \emph{IEEE/CVF Conf. Comput. Vis. Pattern Recog.}, pages 2636--2645, 2020.

\bibitem[Yu et~al.(2023)Yu, Lezama, Gundavarapu, Versari, Sohn, Minnen, Cheng, Birodkar, Gupta, Gu, Hauptmann, Gong, Yang, Essa, Ross, and Jiang]{yu2023language}
Lijun Yu, José Lezama, Nitesh~B. Gundavarapu, Luca Versari, Kihyuk Sohn, David Minnen, Yong Cheng, Vighnesh Birodkar, Agrim Gupta, Xiuye Gu, Alexander~G. Hauptmann, Boqing Gong, Ming-Hsuan Yang, Irfan Essa, David~A. Ross, and Lu~Jiang.
\newblock Language model beats diffusion-tokenizer is key to visual generation.
\newblock \emph{arXiv preprint arXiv:2310.05737}, 2023.

\bibitem[Yu et~al.(2024)Yu, Zhang, Li, Xu, Yao, Chen, Lu, Cui, Dang, He, Feng, Song, Zheng, Liu, Chua, and Sun]{yu2024rlaif}
Tianyu Yu, Haoye Zhang, Qiming Li, Qixin Xu, Yuan Yao, Da~Chen, Xiaoman Lu, Ganqu Cui, Yunkai Dang, Taiwen He, Xiaocheng Feng, Jun Song, Bo~Zheng, Zhiyuan Liu, Tat-Seng Chua, and Maosong Sun.
\newblock {RLAIF-V}: Open-source {AI} feedback leads to super {GPT-4V} trustworthiness.
\newblock \emph{arXiv preprint arXiv:2405.17220}, 2024.

\bibitem[Yu et~al.(2025)Yu, Li, Wei, Lyu, and Tan]{yu2025combining}
Ze~Yu, Jun Li, Yuzhen Wei, Yuandong Lyu, and Xiaojun Tan.
\newblock Combining camera-{LiDAR} fusion and motion planning using bird’s-eye view representation for end-to-end autonomous driving.
\newblock \emph{Drones}, 9\penalty0 (4):\penalty0 281, 2025.

\bibitem[Yuan et~al.(2024{\natexlab{a}})Yuan, Zhang, Sun, Sun, Huang, Lee, Li, Han, Wong, Tee, and Jr]{yuan2024drama}
Chengran Yuan, Zhanqi Zhang, Jiawei Sun, Shuo Sun, Zefan Huang, Christina Dao~Wen Lee, Dongen Li, Yuhang Han, Anthony Wong, Keng~Peng Tee, and Marcelo H.~Ang Jr.
\newblock {DRAMA}: An efficient end-to-end motion planner for autonomous driving with mamba.
\newblock \emph{arXiv preprint arXiv:2408.03601}, 2024{\natexlab{a}}.

\bibitem[Yuan et~al.(2024{\natexlab{b}})Yuan, Sun, Omeiza, Zhao, Newman, Kunze, and Gadd]{yuan2024rag}
Jianhao Yuan, Shuyang Sun, Daniel Omeiza, Bo~Zhao, Paul Newman, Lars Kunze, and Matthew Gadd.
\newblock {RAG-Driver}: Generalisable driving explanations with retrieval-augmented in-context learning in multi-modal large language model.
\newblock \emph{arXiv preprint arXiv:2402.10828}, 2024{\natexlab{b}}.

\bibitem[Yuan et~al.(2025)Yuan, Qian, Tang, Chen, Song, Sun, Chu, Cai, Zhang, and Li]{yuan2025autodrive}
Zhenlong Yuan, Chengxuan Qian, Jing Tang, Rui Chen, Zijian Song, Lei Sun, Xiangxiang Chu, Yujun Cai, Dapeng Zhang, and Shuo Li.
\newblock {AutoDrive-R2}: Incentivizing reasoning and self-reflection capacity for {VLA} model in autonomous driving.
\newblock \emph{arXiv preprint arXiv:2509.01944}, 2025.

\bibitem[Yurt et~al.(2025)Yurt, Ye, Ma, Luo, Mallik, Pauly, Yaman, and Ren]{yurt2025ltda}
Mahmut Yurt, Xin Ye, Yunsheng Ma, Jingru Luo, Abhirup Mallik, John Pauly, Burhaneddin Yaman, and Liu Ren.
\newblock {LTDA-Drive}: {LLMs}-guided generative models based long-tail data augmentation for autonomous driving.
\newblock \emph{arXiv preprint arXiv:2505.18198}, 2025.

\bibitem[Zeng et~al.(2025)Zeng, Chang, Xie, Liu, Bai, Pan, Xu, Wei, and Guo]{zeng2025futuresightdrive}
Shuang Zeng, Xinyuan Chang, Mengwei Xie, Xinran Liu, Yifan Bai, Zheng Pan, Mu~Xu, Xing Wei, and Ning Guo.
\newblock {FutureSightDrive}: Thinking visually with spatio-temporal {CoT} for autonomous driving.
\newblock \emph{arXiv preprint arXiv:2505.17685}, 2025.

\bibitem[Zhai et~al.(2025)Zhai, Li, Guo, Yang, Qin, Zhao, Han, Tao, Wu, , and Jia]{zhai2025world}
Mingliang Zhai, Cheng Li, Zengyuan Guo, Ningrui Yang, Xiameng Qin, Sanyuan Zhao, Junyu Han, Ji~Tao, Yuwei Wu, , and Yunde Jia.
\newblock World knowledge-enhanced reasoning using instruction-guided interactor in autonomous driving.
\newblock In \emph{AAAI Conf. Artifi. Intell.}, volume~39, pages 9842--9850, 2025.

\bibitem[Zhai et~al.(2023)Zhai, Mustafa, Kolesnikov, and Beyer]{zhai2023sigmoid}
Xiaohua Zhai, Basil Mustafa, Alexander Kolesnikov, and Lucas Beyer.
\newblock Sigmoid loss for language image pre-training.
\newblock In \emph{IEEE/CVF Int. Conf. Comput. Vis.}, pages 11975--11986, 2023.

\bibitem[Zhang et~al.(2025{\natexlab{a}})Zhang, Song, Li, Zhu, Deng, and Zhang]{zhang2025future}
Bozhou Zhang, Nan Song, Jingyu Li, Xiatian Zhu, Jiankang Deng, and Li~Zhang.
\newblock Future-aware end-to-end driving: Bidirectional modeling of trajectory planning and scene evolution.
\newblock \emph{arXiv preprint arXiv:2510.11092}, 2025{\natexlab{a}}.

\bibitem[Zhang et~al.(2025{\natexlab{b}})Zhang, Yuan, Chen, Liao, Chen, Shen, Zhou, and Chua]{zhang2025reasoning}
Dapeng Zhang, Zhenlong Yuan, Zhangquan Chen, Chih-Ting Liao, Yinda Chen, Fei Shen, Qingguo Zhou, and Tat-Seng Chua.
\newblock {Reasoning-VLA}: A fast and general vision-language-action reasoning model for autonomous driving.
\newblock \emph{arXiv preprint arXiv:2511.19912}, 2025{\natexlab{b}}.

\bibitem[Zhang et~al.(2024{\natexlab{a}})Zhang, Wang, Zhu, Zhao, Chen, Zhang, Gong, Zhou, Zhang, Wang, Tan, Zhou, Xu, Yao, Zhang, Liu, Di, and Li]{zhang2024sparsead}
Diankun Zhang, Guoan Wang, Runwen Zhu, Jianbo Zhao, Xiwu Chen, Siyu Zhang, Jiahao Gong, Qibin Zhou, Wenyuan Zhang, Ningzi Wang, Feiyang Tan, Hangning Zhou, Ziyao Xu, Haotian Yao, Chi Zhang, Xiaojun Liu, Xiaoguang Di, and Bin Li.
\newblock {SparseAD}: Sparse query-centric paradigm for efficient end-to-end autonomous driving.
\newblock \emph{arXiv preprint arXiv:2404.06892}, 2024{\natexlab{a}}.

\bibitem[Zhang et~al.(2024{\natexlab{b}})Zhang, Xue, Yan, Zhang, Qiu, Bai, Liu, Cui, and Li]{zhang2024efficient}
Haiming Zhang, Ying Xue, Xu~Yan, Jiacheng Zhang, Weichao Qiu, Dongfeng Bai, Bingbing Liu, Shuguang Cui, and Zhen Li.
\newblock An efficient occupancy world model via decoupled dynamic flow and image-assisted training.
\newblock \emph{arXiv preprint arXiv:2412.13772}, 2024{\natexlab{b}}.

\bibitem[Zhang et~al.(2025{\natexlab{c}})Zhang, Zhou, Zhu, Yan, Gao, Bai, Cai, Liu, Cui, and Li]{zhang2025visionpad}
Haiming Zhang, Wending Zhou, Yiyao Zhu, Xu~Yan, Jiantao Gao, Dongfeng Bai, Yingjie Cai, Bingbing Liu, Shuguang Cui, and Zhen Li.
\newblock {VisionPAD}: A vision-centric pre-training paradigm for autonomous driving.
\newblock In \emph{IEEE/CVF Conf. Comput. Vis. Pattern Recog.}, pages 17165--17175, 2025{\natexlab{c}}.

\bibitem[Zhang et~al.(2025{\natexlab{d}})Zhang, Yang, Wang, Yao, Petiushko, and Li]{zhang2025safeauto}
Jiawei Zhang, Xuan Yang, Taiqi Wang, Yu~Yao, Aleksandr Petiushko, and Bo~Li.
\newblock {SafeAuto}: Knowledge-enhanced safe autonomous driving with multimodal foundation models.
\newblock \emph{arXiv preprint arXiv:2503.00211}, 2025{\natexlab{d}}.

\bibitem[Zhang et~al.(2025{\natexlab{e}})Zhang, Tang, Hu, Pan, Guo, Liu, Huang, Yuan, Zhang, Long, Cao, and Yin]{zhang2025epona}
Kaiwen Zhang, Zhenyu Tang, Xiaotao Hu, Xingang Pan, Xiaoyang Guo, Yuan Liu, Jingwei Huang, Li~Yuan, Qian Zhang, Xiao-Xiao Long, Xun Cao, and Wei Yin.
\newblock Epona: Autoregressive diffusion world model for autonomous driving.
\newblock \emph{arXiv preprint arXiv:2506.24113}, 2025{\natexlab{e}}.

\bibitem[Zhang et~al.(2024{\natexlab{c}})Zhang, Zeng, Wang, and Lu]{zhang2024tinyllama}
Peiyuan Zhang, Guangtao Zeng, Tianduo Wang, and Wei Lu.
\newblock {TinyLLaMA}: An open-source small language model.
\newblock \emph{arXiv preprint arXiv:2401.02385}, 2024{\natexlab{c}}.

\bibitem[Zhang et~al.(2024{\natexlab{d}})Zhang, Huang, Gao, Chen, and Lv]{zhang2024wisead}
Songyan Zhang, Wenhui Huang, Zihui Gao, Hao Chen, and Chen Lv.
\newblock {WiseAD}: Knowledge augmented end-to-end autonomous driving with vision-language model.
\newblock \emph{arXiv preprint arXiv:2412.09951}, 2024{\natexlab{d}}.

\bibitem[Zhang et~al.(2025{\natexlab{f}})Zhang, Huang, Chen, Collister, Huang, and Lv]{zhang2025openread}
Songyan Zhang, Wenhui Huang, Zhan Chen, Chua~Jiahao Collister, Qihang Huang, and Chen Lv.
\newblock {OpenREAD}: Reinforced open-ended reasoing for end-to-end autonomous driving with {LLM-as-Critic}.
\newblock \emph{arXiv preprint arXiv:2512.01830}, 2025{\natexlab{f}}.

\bibitem[Zhang et~al.(2025{\natexlab{g}})Zhang, Haß, Chao, Petrovic, Song, Wu, and Knoll]{zhang2025unified}
Yi~Zhang, Erik~Leo Haß, Kuo-Yi Chao, Nenad Petrovic, Yinglei Song, Chengdong Wu, and Alois Knoll.
\newblock A unified perception-language-action framework for adaptive autonomous driving.
\newblock \emph{arXiv preprint arXiv:2507.23540}, 2025{\natexlab{g}}.

\bibitem[Zhang et~al.(2021)Zhang, Liniger, Dai, Yu, and Van~Gool]{zhang2021end}
Zhejun Zhang, Alexander Liniger, Dengxin Dai, Fisher Yu, and Luc Van~Gool.
\newblock End-to-end urban driving by imitating a reinforcement learning coach.
\newblock In \emph{IEEE/CVF Int. Conf. Comput. Vis.}, pages 15222--15232, 2021.

\bibitem[Zhao et~al.(2025{\natexlab{a}})Zhao, Yuan, Li, Hu, Li, Gao, and Gao]{zhao2025sce2drivex}
Rui Zhao, Qirui Yuan, Jinyu Li, Haofeng Hu, Yun Li, Zhenhai Gao, and Fei Gao.
\newblock {Sce2DriveX}: A generalized {MLLM} framework for scene-to-drive learning.
\newblock \emph{IEEE Robot. Autom. Lett.}, 10\penalty0 (12):\penalty0 12580--12587, 2025{\natexlab{a}}.

\bibitem[Zhao et~al.(2025{\natexlab{b}})Zhao, Fu, Liang, Zhou, Zhang, Xie, Wang, and Bai]{zhao2025extending}
Zongchuang Zhao, Haoyu Fu, Dingkang Liang, Xin Zhou, Dingyuan Zhang, Hongwei Xie, Bing Wang, and Xiang Bai.
\newblock Extending large vision-language model for diverse interactive tasks in autonomous driving.
\newblock \emph{arXiv preprint arXiv:2505.08725}, 2025{\natexlab{b}}.

\bibitem[Zheng and Vedaldi(2023)]{zheng2023online}
Chuanxia Zheng and Andrea Vedaldi.
\newblock Online clustered codebook.
\newblock In \emph{IEEE/CVF Int. Conf. Comput. Vis.}, pages 22798--22807, 2023.

\bibitem[Zheng et~al.(2025{\natexlab{a}})Zheng, Mao, Ye, Li, Zhan, Lang, and Zhao]{zheng2025driveagent}
Weicheng Zheng, Xiaofei Mao, Nanfei Ye, Pengxiang Li, Kun Zhan, Xianpeng Lang, and Hang Zhao.
\newblock {DriveAgent-R1}: Advancing {VLM}-based autonomous driving with hybrid thinking and active perception.
\newblock \emph{arXiv preprint arXiv:2507.20879}, 2025{\natexlab{a}}.

\bibitem[Zheng et~al.(2024{\natexlab{a}})Zheng, Chen, Huang, Zhang, Duan, and Lu]{zheng2024occworld}
Wenzhao Zheng, Weiliang Chen, Yuanhui Huang, Borui Zhang, Yueqi Duan, and Jiwen Lu.
\newblock {OccWorld}: Learning a {3D} occupancy world model for autonomous driving.
\newblock In \emph{Eur. Conf. Comput. Vis.}, pages 55--72. Springer, 2024{\natexlab{a}}.

\bibitem[Zheng et~al.(2024{\natexlab{b}})Zheng, Song, Guo, Zhang, and Chen]{zheng2024genad}
Wenzhao Zheng, Ruiqi Song, Xianda Guo, Chenming Zhang, and Long Chen.
\newblock {GenAD}: Generative end-to-end autonomous driving.
\newblock In \emph{Eur. Conf. Comput. Vis.}, pages 87--104. Springer, 2024{\natexlab{b}}.

\bibitem[Zheng et~al.(2024{\natexlab{c}})Zheng, Wu, Zheng, Zuo, Xie, Yang, Pan, Hao, Jia, Lang, and Zhang]{zheng2024gaussianad}
Wenzhao Zheng, Junjie Wu, Yao Zheng, Sicheng Zuo, Zixun Xie, Longchao Yang, Yong Pan, Zhihui Hao, Peng Jia, Xianpeng Lang, and Shanghang Zhang.
\newblock {GaussianAD}: Gaussian-centric end-to-end autonomous driving.
\newblock \emph{arXiv preprint arXiv:2412.10371}, 2024{\natexlab{c}}.

\bibitem[Zheng et~al.(2024{\natexlab{d}})Zheng, Xia, Huang, Zuo, Zhou, and Lu]{zheng2024doe}
Wenzhao Zheng, Zetian Xia, Yuanhui Huang, Sicheng Zuo, Jie Zhou, and Jiwen Lu.
\newblock Doe-1: Closed-loop autonomous driving with large world model.
\newblock \emph{arXiv preprint arXiv:2412.09627}, 2024{\natexlab{d}}.

\bibitem[Zheng et~al.(2025{\natexlab{b}})Zheng, Yang, Xing, Zhang, Zheng, Gao, Li, Zhang, Xia, Jia, Lang, and Zhao]{zheng2025world4drive}
Yupeng Zheng, Pengxuan Yang, Zebin Xing, Qichao Zhang, Yuhang Zheng, Yinfeng Gao, Pengfei Li, Teng Zhang, Zhongpu Xia, Peng Jia, XianPeng Lang, and Dongbin Zhao.
\newblock {World4Drive}: End-to-end autonomous driving via intention-aware physical latent world model.
\newblock In \emph{IEEE/CVF Int. Conf. Comput. Vis.}, pages 28632--28642, 2025{\natexlab{b}}.

\bibitem[Zhong et~al.(2025)]{zhong2025survey}
Yifan Zhong et~al.
\newblock Yifan zhong and fengshuo bai and shaofei cai and xuchuan huang and zhang chen and xiaowei zhang and yuanfei wang and shaoyang guo and tianrui guan and ka nam lui and zhiquan qi and yitao liang and yuanpei chen and yaodong yang.
\newblock \emph{arXiv preprint arXiv:2507.01925}, 2025.

\bibitem[Zhou et~al.(2025{\natexlab{a}})Zhou, Pan, LeCun, and Pinto]{zhou2024dino}
Gaoyue Zhou, Hengkai Pan, Yann LeCun, and Lerrel Pinto.
\newblock {DINO-WM}: World models on pre-trained visual features enable zero-shot planning.
\newblock In \emph{Int. Conf. Mach. Learn.} PMLR, 2025{\natexlab{a}}.

\bibitem[Zhou et~al.(2024)Zhou, Liu, Yurtsever, Zagar, Zimmer, Cao, and Knoll]{zhou2024vision}
Xingcheng Zhou, Mingyu Liu, Ekim Yurtsever, Bare~Luka Zagar, Walter Zimmer, Hu~Cao, and Alois~C. Knoll.
\newblock Vision language models in autonomous driving: A survey and outlook.
\newblock \emph{IEEE Trans. Intell. Veh.}, pages 1--20, 2024.

\bibitem[Zhou et~al.(2025{\natexlab{b}})Zhou, Han, Yang, Ma, Tresp, and Knoll]{zhou2025opendrivevla}
Xingcheng Zhou, Xuyuan Han, Feng Yang, Yunpu Ma, Volker Tresp, and Alois Knoll.
\newblock {OpenDriveVLA}: Towards end-to-end autonomous driving with large vision language action model.
\newblock \emph{arXiv preprint arXiv:2503.23463}, 2025{\natexlab{b}}.

\bibitem[Zhou et~al.(2025{\natexlab{c}})Zhou, Cai, Zhao, Zhang, Huang, Zhou, and Ma]{zhou2025autovla}
Zewei Zhou, Tianhui Cai, Seth~Z. Zhao, Yun Zhang, Zhiyu Huang, Bolei Zhou, and Jiaqi Ma.
\newblock {AutoVLA}: A vision-language-action model for end-to-end autonomous driving with adaptive reasoning and reinforcement fine-tuning.
\newblock \emph{arXiv preprint arXiv:2506.13757}, 2025{\natexlab{c}}.

\bibitem[Zhu et~al.(2024)Zhu, Lin, Ning, Yan, Cui, Wang, Pang, Jiang, Zhang, Li, Zhang, Li, Liu, and Yuan]{zhulanguagebind}
Bin Zhu, Bin Lin, Munan Ning, Yang Yan, Jiaxi Cui, HongFa Wang, Yatian Pang, Wenhao Jiang, Junwu Zhang, Zongwei Li, Wancai Zhang, Zhifeng Li, Wei Liu, and Li~Yuan.
\newblock {LanguageBind}: Extending video-language pretraining to {N}-modality by language-based semantic alignment.
\newblock In \emph{Int. Conf. Learn. Represent.}, 2024.

\bibitem[Zhu et~al.(2025{\natexlab{a}})Zhu, Hu, Liu, Lu, Kong, and Ilic]{zhu2025spiral}
Dekai Zhu, Yixuan Hu, Youquan Liu, Dongyue Lu, Lingdong Kong, and Slobodan Ilic.
\newblock {SPIRAL}: Semantic-aware progressive {LiDAR} scene generation and understanding.
\newblock In \emph{Adv. Neural Inf. Process. Syst.}, volume~38, 2025{\natexlab{a}}.

\bibitem[Zhu et~al.(2025{\natexlab{b}})Zhu, Wang, Chen, Liu, Ye, Gu, Tian, Duan, Su, Shao, Gao, Cui, Wang, Cao, Liu, Wei, Zhang, Wang, Xu, Li, Wang, Deng, Li, He, Jiang, Luo, Wang, He, Shi, Zhang, Shao, He, Xiong, Qu, Sun, Jiao, Lv, Wu, Zhang, Deng, Ge, Chen, Wang, Dou, Lu, Zhu, Lu, Lin, Qiao, Dai, and Wang]{zhu2025internvl3}
Jinguo Zhu, Weiyun Wang, Zhe Chen, Zhaoyang Liu, Shenglong Ye, Lixin Gu, Hao Tian, Yuchen Duan, Weijie Su, Jie Shao, Zhangwei Gao, Erfei Cui, Xuehui Wang, Yue Cao, Yangzhou Liu, Xingguang Wei, Hongjie Zhang, Haomin Wang, Weiye Xu, Hao Li, Jiahao Wang, Nianchen Deng, Songze Li, Yinan He, Tan Jiang, Jiapeng Luo, Yi~Wang, Conghui He, Botian Shi, Xingcheng Zhang, Wenqi Shao, Junjun He, Yingtong Xiong, Wenwen Qu, Peng Sun, Penglong Jiao, Han Lv, Lijun Wu, Kaipeng Zhang, Huipeng Deng, Jiaye Ge, Kai Chen, Limin Wang, Min Dou, Lewei Lu, Xizhou Zhu, Tong Lu, Dahua Lin, Yu~Qiao, Jifeng Dai, and Wenhai Wang.
\newblock {InternVL3}: Exploring advanced training and test-time recipes for open-source multimodal models.
\newblock \emph{arXiv preprint arXiv:2504.10479}, 2025{\natexlab{b}}.

\bibitem[Zhuang et~al.(2024)Zhuang, Fang, Tong, Liu, Zeng, Zhou, and Chen]{zhuang2024online}
Huiping Zhuang, Di~Fang, Kai Tong, Yuchen Liu, Ziqian Zeng, Xu~Zhou, and Cen Chen.
\newblock Online analytic exemplar-free continual learning with large models for imbalanced autonomous driving task.
\newblock \emph{IEEE Trans. Veh. Tech.}, 74\penalty0 (2):\penalty0 1949--1958, 2024.

\bibitem[Zou et~al.(2025)Zou, Chen, Liao, Zheng, Song, Zhang, Zhang, Liu, and Wang]{zou2025diffusiondrivev2}
Jialv Zou, Shaoyu Chen, Bencheng Liao, Zhiyu Zheng, Yuehao Song, Lefei Zhang, Qian Zhang, Wenyu Liu, and Xinggang Wang.
\newblock {DiffusionDriveV2}: Reinforcement learning-constrained truncated diffusion modeling in end-to-end autonomous driving.
\newblock \emph{arXiv preprint arXiv:2512.07745}, 2025.

\end{thebibliography}

\end{document}